**THÈSE**

Présentée à la Faculté des Sciences

Département d'Informatique

Pour l'Obtention du Diplôme de

**Doctorat en sciences**

Option : Informatique

Par

**M. Sadik BESSOU**

**Thème**

# Contribution au Niveau de l'Approche Indirecte à Base de Transfert dans la Traduction Automatique

Soutenu le : 11/06/2015 devant la commission d'examen :

| | | | |
|---|---|---|---|
| Président | M Khireddine Kholladi | Professeur | Université El Oued |
| Rapporteur | Mohamed Touahria | Professeur | Université Sétif -1- |
| Examinateur | Allaoua Refoufi | MCA | Université Sétif -1- |
| Examinateur | Abdelkrim Amirat | Professeur | Université Souk Ahras |
| Examinateur | Abdelhak Boubetra | Professeur | Université BBA |

بسم الله الرحمن الرحيم

والصلاة والسلام على رسول الله صلى الله عليه وآله وسلم

# Résumé

Dans cette thèse, on aborde plusieurs points importants concernant l'analyse morphologique de la langue arabe appliquée à l'informatique documentaire et à la traduction automatique. Tout d'abord, on a dressé un aperçu sur la traduction automatique, son histoire et son développement, puis on a exposé les techniques de la traduction humaine pour une éventuelle inspiration dans la traduction automatique, ainsi on a exposé les approches linguistiques et particulièrement les approches indirectes de transfert. Enfin nous avons présenté nos contributions pour la résolution des problèmes morphosyntaxiques dans l'informatique documentaire comme la recherche d'information multilingue et dans la traduction automatique.

Dans une première contribution, nous avons développé un analyseur morphologique pour la langue arabe, et on l'a exploité dans la recherche d'information bilingue comme application de l'informatique documentaire multilingue. La validation des résultats montre une performance statistiquement signifiante.

Dans une seconde contribution, nous avons proposé une liste de règles de transfert morphosyntaxique de l'anglais vers l'arabe, pour une traduction en trois phases : analyse, transfert, génération. On a insisté sur la phase transfert, destinée à plonger sans distorsion sémantique une abstraction de l'anglais dans un sous-ensemble suffisant de l'arabe.

# Mots-clés

Traitement automatique du langage naturel, Traduction automatique, Recherche d'information, Transfert morphologique, Transfert syntaxique, Approche à base de règles, Langue arabe.

# ملخص

نستعرض في هذه الأطروحة عدة نقاط مهمة حول التحليل الصرفي للغة العربية في ميداني المعطيات النصية والترجمة الآلية. أولا، قمنا بعرض موجز حول الترجمة الآلية، تاريخها وتطورها، ثم عرضنا تقنيات الترجمة الإنسانية لاستنتاج أفكار محتملة في الترجمة الآلية. كما عرضنا المقاربات اللسانية، وخصوصا المقاربات غير المباشرة للتحويل. وفي الأخير عرضنا مساهماتنا لحل المشاكل الصرفية والنحوية في مسائل اللسانيات الحاسوبية كالبحث المعلوماتي متعدد اللغة والترجمة الآلية.

كمساهمة أولى، قمنا بتطوير محلل صرفي للغة العربية، يتم استغلاله في البحث المعلوماتي متعدد اللغة كتطبيق من تطبيقات الترجمة الآلية. حيث أثبتت النتائج فعالية احصائية معتبرة.

وكمساهمة ثانية، اقترحنا مجموعة من قواعد التحويل الصرفي النحوي من الإنجليزية الى العربية، لترجمة عبر ثلاث مراحل: تحليل، تحويل، توليد، حيث ركزنا في هذا البحث على مرحلة التحويل لترجمة دون تحريف دلالي بجعل الإنجليزية تمثل ضمن مجموعة فرعية في العربية.

# الكلمات المفاتيح

المعالجة الآلية للغات الطبيعية، الترجمة الآلية، البحث المعلوماتي، التحويل الصرفي، التحويل النحوي، مقاربة القواعد، اللغة العربية.

# Abstract

In this thesis, we address several important issues concerning the morphological analysis of Arabic language applied to textual data and machine translation. First, we provided an overview on machine translation, its history and its development, then we exposed human translation techniques for eventual inspiration in machine translation, and we exposed linguistic approaches and particularly indirect transfer approaches. Finally, we presented our contributions to the resolution of morphosyntactic problems in computer linguistics as multilingual information retrieval and machine translation.

As a first contribution, we developed a morphological analyzer for Arabic, and we have exploited it in the bilingual information retrieval such as a computer application of multilingual documentary. Results validation showed a statistically significant performance.

In a second contribution, we proposed a list of morphosyntactic transfer rules from English to Arabic for translation in three phases: analysis, transfer, generation. We focused on the transfer phase without semantic distortion for an abstraction of English in a sufficient subset of Arabic.

# Remerciements

<p dir="rtl" lang="ar">الحمد لله الذي بنعمته تتم الصالحات</p>

# Dédicaces

Je dédie ce travail à mon père, ma mère, ma femme, mes frères et sœurs.

A tous mes enseignants et tous mes étudiants.

A tous les amis.

# Table des matières

# Table des figures

# Liste des tableaux

# Introduction

*Traduction,* mot emprunté au latin *(traductio),* provient du verbe *(tradūcĕre)* : « faire passer », composé de *(trans)* « à travers » et *(dūcō)* « mener ». *(Traducere)* signifiait en latin 'transporter', signification largement attestée dans le latin médiéval. Jusqu'au XVe siècle, le mot le plus employé pour *traduire* était *(translatare)* ; or, l'extension de *(traducere)* avec la nouvelle signification fut si forte qu'il gagna toute l'Europe Occidentale ainsi que le roumain. Le sens le plus courant est : « faire passer un texte d'une langue à une autre ». Dans l'anglais *(translate)* et l'allemand *(übersetzen)*, c'est à la notion de déplacement que renvoie l'étymologie. Le verbe traduire apparaît pour la première fois en français en 1528, et le nom *traduction* en 1530.[1]

Traduire un texte suppose la compréhension du sens pour une restitution dans la langue cible. Ce processus apparemment simple est en réalité complexe. La traduction ne se limite pas à une simple substitution mot à mot. Le traducteur doit analyser et interpréter le texte et comprendre les relations entre les mots qui peuvent en influencer le sens. Ceci requiert une connaissance de la *grammaire*, de la *syntaxe* et de la *sémantique*, à la fois dans la langue source et dans la langue cible, dans notre cas l'anglais et l'arabe.

L'arabe est une langue littéraire très riche : il possède un vocabulaire fort étendu et une rare souplesse de formes. Il y aurait 80 termes différents pour exprimer le miel, 200 pour le serpent, 500 pour le lion, 1 000 pour le chameau, autant pour l'épée, et jusqu'à 4 000 pour rendre l'idée de malheur. C'est qu'une foule de nuances d'idées, dont la subtilité fait qu'elles sont traduites par des termes spéciaux, et que, dans le grand nombre d'expressions employées pour une même idée, il y a une foule de figures et de tropes. L'arabe comprend plus de 12.300.000 mots, alors que le dictionnaire Oxford comprend 1.000.000 mots, et le dictionnaire français 100.000 mots [2]

Lancée pendant la guerre froide conjointement aux Etats-Unis et en Union Soviétique, la traduction automatique a connu un progrès foudroyant. En seulement cinquante ans, la traduction automatique passa du stade de la science-fiction à la réalité. Les informaticiens n'ont plus à se désespérer devant un problème sans solution et les traducteurs, de leur côté, ne se sentent plus menacés par le développement de la traduction automatique. Même si les traducteurs humains pourront être difficilement remplacés par un ordinateur, la traduction automatique est déjà utilisée de manière effective, le plus souvent sous forme de traduction assistée par ordinateur (TAO), le traducteur humain limitant son concours à la levée des ambiguïtés irrésolues.

---

[1] Cf. Le Grand Robert, 1966.
[2] IBM France comptait en français un vocabulaire de 180 000 mots ayant 800 000 formes… ; la survie suppose de connaître 1500 mots, le baccalauréat suppose d'en connaître environ 5 000 à 5 500 ; seul Winston Churchill et Charles de Gaulle étaient censés utiliser potentiellement 75 000 mots.

Cette spécialité est un sous-domaine de la linguistique informatique qui travaille à la théorie et pratique de l'utilisation de l'outil informatique pour la transcription des textes d'une langue naturelle vers une autre.

La traduction automatique s'est développée autour de trois approches essentielles : les *approches symboliques*, les *approches à base de règles*, et les *approches statistiques*.

Dans cette thèse nous nous intéressons par les approches linguistiques à base de règles. Ces dernières formalisent des règles de réécritures et proposent une série de transformations des arbres morphologiques, syntaxiques et sémantiques entre langue source et langue cible.

**Position du problème**

Les avancées en traduction automatique sont, en grande partie, basées sur les développements de plus en plus importants des recherches en traitement automatique du langage naturel. A l'heure actuelle, les recherches en traduction automatique se concentrent sur les approches statistiques, mais beaucoup de problèmes d'ordre linguistiques n'ont pas été résolus, ce qui exige un recours aux méthodes linguistiques comme solution principale ou solution hybride.

L'ambiguïté est un problème omniprésent dans les recherches en traitement automatique du langage naturel, elle constitue l'un des défis du processus de traduction automatique. Cette ambiguïté est principalement d'ordre syntaxique et d'ordre sémantique.

L'analyse d'un texte en vue de sa traduction suppose l'élucidation complète des ambiguïtés, donc la recherche de l'acception exacte des vocables ambigus, au-delà de la simple analyse syntaxique.

S'intéressant au rapport signifiant/signifié, Umberto Eco (professeur de sémiologie) indique qu'un mot renvoie plutôt à une encyclopédie qu'à un dictionnaire : en effet, la levée de l'ambiguïté est un problème sémantique à résoudre dans un contexte précis, donc dans un cadre pragmatique donné.

Pour la langue arabe, l'ambigüité morphologique dans l'informatique documentaire est présente davantage par rapport à d'autres langues qui se limitent à des lemmatisations superficielles. Comment résoudre le problème d'ambiguïté lexicale due à l'absence des voyelles ? Comment à partir d'un mot déduire sa forme canonique différente du mot de départ ? Comment extraire la racine d'un mot qui ne partage pas toutes les lettres originales avec sa racine ?

Dans la traduction de l'anglais, l'analyse est difficile (détermination du genre, souvent indéfini ou défini tardivement[3], « *that* » facultatif d'où subordination implicite, ambiguïté de « but », verbalisation de noms…, doublons roman/germanique fluctuations dialectales entre British English, Australian English, American English, Canadian English…).

Dans la traduction vers la langue arabe, le problème ne se limite pas à l'analyse mais aussi à la génération. L'analyse (≠ synthèse) morphologique des mots est le cœur de tout travail en traitement automatique de la langue arabe, du fait que la langue arabe est une langue fortement flexionnelle et dérivationnelle.

Au niveau syntaxique, comment représenter une phrase en arabe par une représentation permettant le transfert depuis une autre représentation d'une phrase en anglais ? Comment assurer le transfert entre deux langues structurellement différentes ?

La langue arabe est caractérisée par la possibilité d'inclure toute une phrase dans un seul mot. Ce qui motive la nécessité du traitement morphologique. Des vas et viens entre le niveau morphologique et le niveau syntaxique sont indispensables.

**Contribution**

Dans cette optique, notre objectif est de résoudre les problèmes d'ambigüité morphologique et de proposer une approche basée sur les modèles linguistiques appliquée à l'analyse morphosyntaxique de la langue arabe, à la recherche d'information multilingue, et au transfert morphosyntaxique de/vers la langue arabe.

Cette « Contribution au niveau de l'approche indirecte à base de transfert dans la Traduction automatique » constitue la première pierre d'un projet de traduction automatique de l'anglais vers l'arabe, qui met en lumière les problèmes posés par l'anglais, et les solutions apportées en arabe.

---

[3] The baker opened the shop. *Her…*

Il s'agit d'une étude préalable, la réalisation et le test d'un module de transfert en vue de la traduction automatique supposant de le raccorder à deux modules complémentaires : l'*analyseur* de la langue à traduire en amont, et le *générateur* dans la langue visée en aval.

Cette étude se concentre sur la traduction phrase à phrase, en supposant au plus un verbe par phrase traduite, et en se limitant au sens propre. Cette relative faiblesse (l'unité de sens étant plutôt le paragraphe) est compensée par le traitement complet des mécanismes de subordination.

Une première réalisation *analyseur + transfert + générateur* devrait permettre une première campagne de tests. Le réalisme du corpus serait alors source d'indications sur les extensions et refontes nécessaires pour un traducteur toujours plus pertinent.

**Organisation du mémoire**

Ce mémoire est organisé en sept chapitres :

Le chapitre 1 introduit le contexte dans lequel se situent nos travaux de recherche. Il présente dans un premier temps la traduction automatique, son histoire, son développement, son importance et ses difficultés. Il aborde dans un deuxième temps les méthodes d'évaluation et les approches de traduction automatique.

Les techniques de la traduction humaine sont présentées au chapitre 2. Dans un premier temps, ce chapitre rappelle les origines, les définitions et les spécialisations de la traduction humaine. Les principales approches sont ensuite étudiées.

Conformément à notre contribution, les approches linguistiques sont détaillées au chapitre 3. Les différentes méthodes ainsi que leurs niveaux de représentation y sont expliqués. Commençant par l'approche directe, puis l'approche de transfert (objet de notre étude), et enfin l'approche à langue pivot.

Le chapitre 4 présente l'aspect multilingue, la recherche d'information multi–lingue comme application de la traduction automatique, et la taxonomie des modèles de recherche d'information et leurs approches. Nous montrons dans ce chapitre que les mêmes algorithmes proposés dans cette thèse sont exploitables pour la recherche d'information et pour la traduction automatique. Nous y expliquons aussi les frontières entre la traduction automatique et la recherche d'information multilingue. Nous y verrons également comment l'algorithme proposé aboutit à une exactitude statistiquement significative dans l'analyse morphologique et amé-liore les résultats de la traduction pendant la recherche multilingue, en diminuant le silence pendant la phase de recherche. Enfin, les résultats des différentes expériences, y sont exposés.

Nous proposons dans le chapitre 5, les règles morphologiques finalisant le transfert de l'anglais vers l'arabe, nous commençons par un survol sur les dernières propositions sur le transfert morphosyntaxique de la langue arabe. Les règles proposées, y sont ensuite détaillées pour les différentes classes grammaticales de mots.

En amont, les règles métataxiques, sont mises au point au chapitre 6, pour le transfert au niveau de l'entité phrase. Les aspects hypotaxiques / parataxiques sont discutés. Les interdépendances entre chapitre 5 et chapitre 6 sont chaque fois mentionnées pour montrer la non linéarité du processus de transfert. Pour une bonne compréhension chaque règle est dotée d'exemples représentés par des figures illustratives

Le chapitre 7 montre l'aspect réalisation du système avec un ensemble d'interfaces, des tests et des comparaisons avec des systemes existants. Un échantillon de résultats est présenté pour concrétiser les étapes de transferts expliquées dans les chapitres de transfert morphologique et syntaxique.

Ce document s'achève par un ensemble de conclusions générales et de perspectives des travaux en cours et avenir inchaa Allah.

# Chapitre 1. La Traduction Automatique

## *Introduction*

Depuis les origines, l'homme a eu besoin de communiquer que ce soit à l'aide de signaux visuels ou de signaux sonores. Il a toujours tenté de vaincre les distances et de mettre en place une transmission rapide de l'information. Il a voyagé à travers le monde, il a pris conscience de la nécessité de comprendre la langue de l'autre. Ce besoin est devenu plus urgent avec l'élargissement des contacts et des échanges entre les différents peuples. L'évolution intellectuelle qui a eu lieu dans les différentes civilisations a rendu ce besoin encore plus important. Depuis la révolution industrielle, cette tendance à la communication s'est développée à une vitesse vertigineuse, particulièrement avec l'émergence des technologies de l'information. En effet, l'utilisation généralisée des ordinateurs dans tous les domaines de la vie, l'avènement d'Internet et du Web et le phénomène de la mondialisation ont créé plus de besoins en matière de communication.

Informatique et langue ont des préoccupations communes : dans un programme informatique, l'ordinateur analyse les instructions et interprète leurs structures en fonction du langage utilisé. D'ailleurs, la recherche linguistique s'est beaucoup inspirée de la recherche informatique, et réciproquement. Aussi les informaticiens tentent-ils depuis une cinquantaine d'années d'automatiser le processus de traduction.

Aujourd'hui, des millions d'internautes effectuent quotidiennement la traduction de millions de pages à l'aide de « serveurs de traduction automatique » gratuits, comme ceux de Systran, Reverso, Google, ou au Japon Fujitsu, Toshiba, Nec, Oki… Toutefois, ce que ces serveurs fournissent peut être considéré comme étant des « pré traductions » utilisables par des réviseurs experts pour produire des traductions fiables.

Par ailleurs, diverses institutions s'intéressent à la question pour leurs besoins propres, souvent assumés par une centrale de traduction : veille technologique (devant ramener les documents collectés vers une même langue), diffusion de documentations (traduisant en di-verses langues[4] des textes rédigés dans une seule), ou échanges internes (cas de l'Union Européenne, avec ses 24 langues officielles et ses 3 langues de travail).

De nos jours, il paraît inimaginable qu'une découverte, une invention d'ordre scientifique ou artistique puisse rester cloîtré dans une seule langue. Ce serait priver le monde entier de telles connaissances, d'où la nécessité de traduire et de mettre ce savoir à la portée de tous.

C'est pour cette raison que la traduction a pris et prend de plus en plus d'importance, et ce dans tous les domaines, aussi bien en ce qui concerne *la littérature*, *l'industrie*, *les normes et les législations*, ainsi que *les sciences naturelles et humaines*.

## *1. Définition*

La Traduction Automatique (TA) est un processus par lequel des programmes informatiques sont utilisés pour traduire un texte d'une langue naturelle (comme l'anglais) vers une autre langue naturelle (comme l'arabe). La première langue est dite langue-source ou langue de départ et la seconde, langue-cible ou langue d'arrivée.

Ce sous-domaine de la linguistique computationnelle pousse à l'extrême l'utilisation des logiciels pour traduire un texte ou une parole d'une langue naturelle à une autre.

Pour traiter toute traduction, humaine ou automatisée, le sens du texte source doit être entièrement transcrit dans le texte cible – ce qui n'est simple qu'en surface. La traduction n'est pas une simple substitution mot-à-mot. Le traducteur doit interpréter et analyser tous les éléments dans le texte et savoir comment chaque mot peut influencer un autre. Cela nécessite une grande expertise dans *la grammaire*, *la syntaxe* (structure des phrases), *la sémantique* (sens), *la pragmatique* (non-dits du contexte) etc., dans les deux langues : source et cible [1].

La traduction humaine et la traduction automatique ont chacune leur part de défis. Par exemple, on ne peut pas trouver deux traducteurs qui produiraient des traductions identiques d'un même texte, et cela peut prendre plusieurs cycles de révisions pour répondre à la satisfaction du client. Mais le plus grand défi réside dans la manière dont la TA peut produire des traductions de qualité publiable [1].

La TA est un outil pratique, qui devrait à terme remplacer le traducteur humain. Moins ambitieuse mais plus fréquente, la Traduction Assistée par Ordinateur (TAO) est une version semi-automatique qui accroît grandement la productivité des traducteurs-experts, dont l'intervention est réduite à une certaine interaction pour mieux venir à bout des passages délicats.

---

[4] L'Inde possède 2 langues officielles nationales, 21 langues officielles régionales avec 12 systèmes d'écriture.

Un logiciel de TA analyse le texte dans la langue source et génère automatiquement le texte correspondant dans la langue cible, par le recours à des règles précises pour le transfert de la structure grammaticale. « Il existe aujourd'hui un certain nombre de systèmes produisant un résultat de qualité suffisante pour être utile dans certaines applications spécifiques, en général dans le domaine de la documentation technique. En outre, les logiciels de traduction essentiellement destinés à aider le traducteur humain à produire des traductions, jouissent d'une popularité croissante auprès des organismes professionnels de traduction » [2].

Les logiciels actuels de TA permettent souvent des personnalisations par domaine ou profession pour l'amélioration du rendement en limitant la portée des substitutions permises. Cette technique est particulièrement efficace dans les domaines où une langue formelle ou stéréotypée est utilisée.

Des résultats de qualité peuvent également être obtenus par l'intervention humaine : par exemple, certains systèmes sont capables de traduire plus précisément si l'utilisateur identifie clairement les différentes classes de mots. Il arrive que l'on soit tenu de préciser les classes de mots pour avoir un résultat escompté, parfois sans donner de précisions sur les classes de mots, le résultat peut être satisfaisant et fiable [3].

À l'aide de ces techniques, la TA s'est révélée utile pour assister les traducteurs humains.

## 2. Historique de la traduction automatique

L'idée d'un système de traduction d'une langue à une autre est aussi vieille que l'idée des systèmes informatiques. Warren Weaver écrit à propos de la TA dès 1949 – alors qu'on dispose à peine des premiers assembleurs.

Enthousiasmant, l'optimisme de Weaver a poussé les chercheurs à aller de l'avant, mais après plusieurs décennies de labeur, les résultats étaient encore embryonnaires.

Brown et al. (1988) [4] ont suggéré qu'il était possible de construire des systèmes de traduction automatiquement.

Au lieu de codifier l'introspection du processus de traduction humaine, Brown et al. proposent des techniques d'apprentissage automatique pour induire des modèles à partir des exemples.[5]

Dans les paragraphes qui suivent, on récapitule ce qu'a écrit John Hutchins à propos de l'histoire de la traduction automatique [5].

### 2.1. Avant l'ordinateur

En 1933 on note deux brevets délivrés en France et en Russie à Georges Artsrouni et Petr Trojanskij respectivement. Le brevet d'Artsrouni était pour une machine polyvalente qui peut également fonctionner comme un dictionnaire multilingue mécanique. Le brevet de Trojanskij était aussi essentiellement un dictionnaire mécanique, allant plus loin avec des propositions pour le codage et l'interprétation des fonctions grammaticales en utilisant « des symboles universels » (basés sur l'Espéranto[6]) dans un dispositif de traduction multilingue.

### 2.2. Après l'avènement de l'ordinateur

Aucun de ces précurseurs n'était connu d'Andrew Booth (cristallographe britannique) et Warren Weaver lors de leurs rencontres en 1946 et 1947, et leur mise en avant des premières tentatives pour utiliser les ordinateurs nouvellement inventés pour traduire les langues naturelles.

### 2.3. La période 1950 - 1955

En mai 1951, Yehoshua Bar-Hillel écrit un rapport sur l'état de l'art, dans lequel il expose quelques-unes des approches de base de TA, et en Juin 1952, il organisa la première conférence en TA au M.I.T. (Massachussets Institute of Technology).

En conséquence, il collabora avec IBM sur un projet qui aboutit à la première démonstration d'un système de TA le 7 Janvier 1954. Un échantillon de 49 phrases russes fut traduit en anglais, en utilisant un vocabulaire très restreint de 250 mots et seulement 6 règles de grammaire. La démonstration fut montée en épingle par la presse et fit grande impression sur le public et certains scientifiques. Elle attira beaucoup l'attention des médias aux États-Unis, et souleva d'immenses espoirs, marquant le début d'importants efforts de recherche.

---

[5] Cette voie est très inefficace, sauf à disposer de millions d'essais,… et de stratégies d'induction judicieuse.

[6] Langue artificielle créée par le Dr Zamenhof, 5 millions de locuteurs environ.

En 1953 la première thèse de doctorat en TA est présentée par Anthony G. Oettinger qui a réalisé un dictionnaire russe mécanique. Et en 1955, le premier livre de TA est apparu, contenant une collection éditée par Locke et Booth, comprenant aussi des notes de Weaver en 1949, les expériences de Booth et Richens, certaines communications présentées lors de la conférence de 1952, et d'autres contributions de Bar-Hillel, Dostert, Oettinger, Reifler et Yngve.

## 2.4. La période 1956 - 1966

La décennie 1956 - 1966 a vu les débuts des trois approches de base de TA. La première était le modèle de « *traduction directe* », la seconde approche est le modèle à langue-pivot, La troisième approche est moins ambitieuse : le modèle à base de *transfert*.

Toute évaluation de cette période ne doit pas oublier que les équipements informatiques étaient souvent insuffisants ; beaucoup d'efforts furent consacrés à l'amélioration du matériel de base. Pour des raisons politiques et militaires, presque toutes les recherches américaines étaient pour la traduction Russe-Anglais, et la plupart des recherches soviétiques étaient axées sur la traduction Anglais-Russe.

De nombreux pays s'engagent dans ces recherches : le Japon en 1956, la Tchécoslovaquie en 1957, la Chine en 1958 - 1959, l'Italie et la France en 1959, le Mexique en 1960, la Belgique en 1961. Les recherches restent peu développées en République fédérale d'Allemagne, en Suède et en Finlande.

C'est en 1959 - 1960 que se concrétise l'intérêt des Français pour la traduction automatique. Le CNRS y participe pleinement en créant en décembre 1959 le CETA (Centre d'études pour la traduction automatique) au sein de l'Institut Blaise-Pascal, avec deux sections, l'une à Paris, le CETAP, dirigé par Aimé Sestier, l'autre à Grenoble, le CETAG, dirigé par Bernard Vauquois. La création du CETA est précédée par celle de l'ATALA (l'Association pour l'étude et le développement de la traduction automatique et de la linguistique appliquée) en septembre 1959.[7]

En 1960, Bar-Hillel a critiqué l'hypothèse disant que l'objectif de la recherche en TA devrait être la création de systèmes de traduction entièrement automatique et de haute qualité (FAHQT : Fully Automatic High Quality Translation) produisant des résultats impossibles à distinguer de ceux des traducteurs humains. Son argument avait beaucoup de poids à l'époque, bien que les développements ultérieurs de l'intelligence artificielle et au sein de la TA et les systèmes à base de connaissances, aient démontré que son pessimisme n'était pas tout à fait justifié.

---

[7] Cf. *La Revue pour l'histoire du CNRS*, n°1, 1999, Le CNRS et les débuts de la traduction automatique en France.

En 1964, les commanditaires gouvernementaux de la TA aux États-Unis (principalement armée et renseignement) ont demandé à la NSF (National Science Foundation) de mettre en place le Traitement Automatique du Langage par le Comité consultatif (ALPAC) pour examiner la situation. Dans son célèbre rapport de 1966, ALPAC a conclu que la TA était très lente, imprécise et deux fois plus chère que la traduction humaine et qu' « il n'y a pas de perspective immédiate ou prévisible de la TA ». ALPAC n'a pas jugé nécessaire d'investir davantage dans la recherche en TA, qu'il rejetait parce qu'elle exigeait une post-édition.

### 2.5. La période post ALPAC : 1966 - 1980

Cependant, la recherche ne s'est pas complètement arrêtée – même dans les groupes de recherche aux États-Unis qui continuèrent pendant quelques années, à l'Université du Texas et à l'Université de Wayne State. La « deuxième génération » post-ALPAC devait être dominé par « les modèles indirects », *langue-pivot* et *transfert*.

Aux États-Unis, la principale activité était concentrée sur des traductions anglaises du russe portant sur des thèmes scientifiques et techniques

Les fonds alloués à la traduction chinois-anglais furent entièrement absorbés par les tentatives de réalisation des périphériques de lecture de textes chinois manuscrits !

Au Canada et en Europe les besoins étaient entièrement différents. La politique biculturelle du gouvernement du Canada a créé une demande pour le couple anglais-français. Les problèmes de traduction étaient également aigus dans la Communauté européenne avec la demande croissante de traductions de la documentation scientifique, technique, administrative et juridique de/vers toutes les langues de la communauté. Pendant qu'aux Etats-Unis la TA n'a pas été relancé durant de nombreuses années, au Canada et en Europe (et plus tard au Japon et ailleurs) sa nécessité n'a pas cessé d'être reconnu*e*, et le développement continu. À Montréal, la recherche a commencé en 1970 sur un système de *transfert syntaxique* pour une Traduction anglais-français. Le projet TAUM (Traduction Automatique de l'Université de Montréal) a entraîné deux réalisations majeures sous l'impulsion d'Alain Colmerauer : premièrement, le formalisme « Q-system » de manipulation de chaînes de caractères et des arbres linguistiques (origine du langage de programmation Prolog), et deuxièmement, le système Météo pour traduire les prévisions météorologiques. Météo fonctionne avec succès depuis 1976.

Les principales expériences innovantes ensuite portaient essentiellement sur les approches à base de *transfert*. Entre 1960 et 1971, le groupe créé par Bernard Vauquois à l'Université de Grenoble a développé un système pour la traduction de textes mathématiques et physiques du russe au français. Leur système devint performant quand sa mémoire centrale put enfin accueillir le dictionnaire, résident jusque-là sur bande magnétique.

Un modèle similaire a été adopté dans les années 1970 à l'Université de Texas dans son système METAL pour l'allemand et l'anglais. Il semblait à beaucoup à l'époque que l'approche la moins ambitieuse « à base de *transfert* » offrait de meilleures perspectives.

En 1970 l'ITF (Institut Textile de France) a présenté TITUS, un système multilingue pour traduire des résumés écrits dans un langage contrôlé, et en 1972 est venu CULT (Chinese University of Hong Kong) spécialement conçu pour la traduction de textes mathématiques du chinois vers l'anglais. Plus important, cependant, étaient les premières installations de Systran. Développé par Peter Toma, sa version la plus ancienne est le système Russe-Anglais à l'USAF Foreign Technology Division (Dayton, Ohio) installé en 1970.

Systran a été installé dans de nombreuses institutions intergouvernementales par exemple l'OTAN et l'International Atomic Energy Authority (Vienne), et dans de nombreuses grandes entreprises, comme General Motors, Dornier et Aérospatiale.

La relance de la recherche en TA durant la deuxième moitié des années 1970 et au début des années 1980 a été marquée par l'adoption quasi universelle de l'approche à base de *transfert* (principalement un *transfert syntaxique*) fondée sur la formalisation des règles lexicales et grammaticales influencés par les théories linguistiques de l'époque.

### 2.6. Les années 1980

Durant les années 1980, la plus grande activité commerciale était au Japon, où la plupart des sociétés informatiques (Fujitsu, Hitachi, NEC, Sharp, Toshiba) ont développé des logiciels pour la traduction assistée par ordinateur, principalement pour les marchés Japonais-Anglais et Anglais-Japonais, bien qu'ils n'aient pas ignoré les besoins de traduction de/vers le coréen, le chinois et autres.

Le groupe de Grenoble (GETA, Groupe d'Etudes pour la traduction automatique) a commencé le développement de son système Ariane. Considéré comme le paradigme de la « deuxième génération » des systèmes à base de *transfert* linguistique, Ariane a influencé les projets à travers le monde dans les années 1980. On note en particulier sa flexibilité et sa modularité, ses algorithmes pour manipuler des représentations d'arbres, et sa conception de grammaires statiques et dynamiques.

Un autre système similaire dans sa conception à GETA-Ariane était le système (Mu) développé à l'Université de Kyoto sous la direction de Makoto Nagao.

L'un des projets les plus connus des années 1980 était le projet Eurotra de la communauté européenne. Son but était la construction d'un système de *transfert* de pointe pour la traduction multilingue parmi toutes les langues de la communauté.[8]

Au cours de la deuxième moitié des années 1980, il y eut un regain d'intérêt général pour les systèmes *à langue-pivot* motivée en partie par la recherche contemporaine en intelligence artificielle et en linguistique cognitive. Le système DLT (Distributed Language Translation) de la société BSO à Utrecht (Pays-Bas), sous la direction de Toon Witkam, a été conçu comme un système d'exploitation multilingue et interactif sur des réseaux informatiques.

A la fin des années 1980 est apparue la version commerciale METAL Allemand-Anglais, provenant d'un groupe de recherche à l'Université de Texas. Après ses expériences en *langue-pivot* au milieu des années 1970, ce groupe a adopté une approche à base de *transfert*, pour des recherches financées depuis 1978 par la société Siemens à Munich (Allemagne). D'autres paires de langues ont ensuite été commercialisé pour le néerlandais, le français et l'espagnol ainsi qu'en anglais et en allemand.

### 2.7. Les années 1990

Depuis 1989, la domination de l'approche à base de règles a été rompue par l'émergence de nouvelles méthodes et stratégies qui sont maintenant appelés méthodes « à base de corpus ».

Les années 1990 voient le renouveau de l'intérêt pour la traduction automatique aux États-Unis. Celui-ci est d'abord d'ordre économique : nécessité de traduire la documentation des concurrents japonais, baisse des coûts de l'informatique et hausse des coûts de la traduction surtout pour les langues à alphabets non latins comme le japonais.

Mais surtout, une des caractéristiques des années 1990, c'est le retour des approches empiriques et du traitement statistique de grands corpus. Ce renouveau, suscité par le succès des méthodes stochastiques dans le traitement du signal et la reconnaissance de la parole, a bénéficié de la mise à disposition de corpus de données textuelles importants grâce aux nouvelles possibilités de traitement des ordinateurs et aux efforts de normalisation des textes.[9]

---

[8] Il leur a fallu modifier largement Systran, nettement insuffisant.

Depuis ce temps, la TA statistique (SMT pour Statistical Machine Translation) est devenue le principal objet de nombreux groupes de recherche, basée principalement sur le modèle IBM.

Les principaux centres de recherche en traduction statistique sont les universités d'Aix et Southern California, et ils ont récemment été rejoints par la société Google.

La deuxième grande approche qui a bénéficié également d'un accès rapide à l'amélioration de grandes banques de données de corpus de textes – est connue comme l'approche à « base d'exemples ».

L'utilisation de systèmes de TA s'est accélérée dans les années 1990. Le développement était plus marqué dans les agences commerciales, les services gouvernementaux et les entreprises multinationales, où les traductions sont produites à grande échelle, principalement de la documentation technique. Ce fut le principal marché pour les systèmes mainframe : Systran, Logos, METAL, et ATLAS. Toutes disposent d'installations où les traductions sont produites en grandes quantités ; d'ailleurs en 1995, on estimait que plus de 300 millions de mots par an avaient été traduites par ces sociétés.

Deux des premiers systèmes à être vendus largement sur des ordinateurs personnels étaient PC-translator (par Linguistic Products, Texas) et Power Translator (par Globalink).

### 2.8. Après l'avènement d'internet

Depuis le milieu des années 1990, Internet a exercé une grande influence sur le développement de la TA. Au début, il y a eu les logiciels de TA spécifiquement pour la traduction des pages Web et des messages de courrier électronique hors ligne.

Le service le mieux connu, Babelfish, est apparu sur les versions AltaVista, où les sites offrant Systran pour traduire en français, allemand et espagnol de/vers l'anglais (et plus tard dans de nombreux autres paires de langues). Elle a été suivie par de nombreux autres services en ligne (la plupart d'entre eux gratuits), par exemple Softissimo avec les versions en ligne de ses systèmes de Reverso, Logomedia avec les versions en ligne de LogoVista et PARS.

---

[9] Cf. *La Revue pour l'histoire du CNRS*, n°1, 1999, Le CNRS et les débuts de la traduction automatique en France.

Les services en ligne de TA ont été largement négligés par la plupart des chercheurs de TA. Un défi particulier pour les chercheurs de TA est l'utilisation de systèmes en ligne pour la traduction dans des langues que les utilisateurs ne connaissent pas bien. Une grande partie du langage utilisé sur Internet est familier, incohérent, non grammatical, plein d'acronymes et d'abréviations, d'allusions, et de jeux de mots, etc. – ceci est particulièrement vrai pour les courriers électroniques, chats et SMS. Ces types d'utilisation de la langue sont très différents de la langue des textes scientifiques et techniques pour qui les systèmes de TA ont été conçus.

L'emploi de la TA - pour la traduction du courrier électronique et des pages Web, des systèmes unilingues pour traduire les messages standards dans des langues peu connues, les systèmes de traduction de la parole dans des domaines restreints - est très utile. Certains de ces besoins sont satisfaits ou font l'objet de recherches actuelles, mais il y a beaucoup d'autres possibilités, en particulier en combinant la TA avec d'autres applications de technologie de langage (la recherche d'information, l'extraction d'information, le résumé automatique, etc.) Comme des systèmes de TA variés deviennent plus largement connu et utilisé, la gamme des besoins de traductions possibles et des types possibles de systèmes de TA apparaîtront plus clairement et stimuleront la recherche et le développement dans des directions non encore envisagées.

La fragilité du marché de la TA peut expliquer pourquoi après l'an 2000 des systèmes commerciaux basés sur des méthodes statistiques ont fait leur apparition. Contrairement aux approches à base de règles, les approches statistiques ont peut-être été considérées comme trop risquées ou peut-être prématurées, ou facilement en échec si « l'information c'est l'improbable ».

### 2.9. Les ressources linguistiques

Une grande base de données et des ressources linguistiques sont disponibles à travers le LDC (Linguistic Data Consortium) aux États-Unis et ELRA (European Language Resources Association), une organisation qui a également inauguré une série importante de conférences consacrées au sujet des ressources linguistiques et de leur évaluation (LREC).

### 2.10.    La traduction automatique et les langues dans le monde

Il existe actuellement un certain nombre de systèmes pour l'arabe (notamment Sakhr, Cimos et AppTek), et un nombre croissant de systèmes pour le chinois (par exemple Transtar, Logomedia, Systran, Transphere).

D'autres Systèmes de paires de langues où l'anglais n'est ni source ni cible sont moins importants. Cependant, la plupart des entreprises américaines et européennes mentionnées ci-dessus proposent des systèmes pour les paires telles que le français, l'allemand, l'italien, l'espagnol, le portugais ; et il y a un certain nombre de systèmes offrant le japonais-chinois, le japonais-coréen, etc.

Malgré toute cette activité commerciale, de nombreuses autres langues sont encore mal servies. On manque de systèmes commerciaux pour la plupart des langues de l'Afrique, l'Inde et l'Asie et ceux qui existent ne sont pas facilement accessibles [5].

Le Burkina-Faso compte 69 langues, et l'Inde 461, dont 23 officielles, l'Afrique du Sud 31 et la Bolivie 44, la Chine 299 langues, soient environ 12.000 langues sur la planète – pour 220 états[10] ; d'où le fréquent succès de « la langue du colonisateur », qui présente souvent l'avantage d'une langue véhiculaire sur une large étendue. Cependant, l'anglais recule en Inde et aux USA….

On pourrait envisager d'abord toute paire formable entre langues officielles de l'ONU : l'anglais, l'arabe, le chinois, l'espagnol, le français et le russe, avec peut-être l'adjonction prochaine de l'hindi, de l'indonésien ou du portugais (Portugal, Brésil, Angola, Mozambique : plus de locuteurs, tandis que le russe décline).

## 3. Le Traitement automatique du langage naturel

### 3.1. Définition

Le Traitement Automatique du Langage Naturel (TALN) est un ensemble de techniques pour l'analyse et la représentation des textes sur plusieurs niveaux d'analyse linguistique.

Le TALN est considéré comme une branche de l'intelligence artificielle.

### 3.2. Les objectifs

Les objectifs du TALN sont nombreux, et ils sont liés à l'application. S'il s'agit de la recherche d'information, l'objectif est de répondre aux besoins de l'utilisateur pour fournir les informations exactes et précises. Pour d'autres applications, ce peut être une traduction de texte dans une autre langue, ou la réponse aux questions sur le contenu d'un texte ou les conclusions à en tirer.

---

[10] Voir languages of the world: http://www.ethnologue.com/

### 3.3. Les origines de TALN

Le TALN est l'intersection de plusieurs disciplines comme la linguistique, l'intelligence artificielle, les sciences de l'information. D'abord appelé linguistique computationnelle, alors que l'ensemble du domaine est désigné comme traitement automatique du langage naturel, il y a en fait deux aspects majeurs : *traitement* du langage et *génération* du langage. Le premier se réfère à l'analyse du langage à des fins de production d'une représentation significative, tandis que le second se réfère à la production du langage à partir d'une représentation. La tâche du TALN est équivalente au rôle de lecteur/auditeur, tandis que la tâche de génération du langage naturel est celle de l'écrivain/conférencier – bien qu'une grande partie de la théorie et des technologies soient partagées par ces deux aspects.

La génération du langage naturel exige également une capacité de planification. Autrement dit, le système de génération exige un plan ou un modèle de l'objectif de l'interaction afin de décider ce que le système devrait générer à chaque point dans une interaction. Dans ce qui suit, nous allons nous concentré sur la tâche d'analyse du langage naturel.

### 3.4. Les Niveaux de traitement du langage naturel

Les systèmes de TALN sont souvent exprimés par des niveaux de langue qui diffèrent du modèle séquentiel avec l'hypothèse que les niveaux de TALN se suivent d'une manière strictement séquentielle et linéaire.

Les recherches psycholinguistiques suggèrent que le traitement de langage est plus dynamique, puisque les niveaux peuvent interagir de façon non linéaire.

**Le niveau phonologique (pour les entrées orales)**

Ce niveau traite l'interprétation des sons en mots, il existe trois types de règles employées dans l'analyse phonologique :

1. Les règles phonétiques pour le traitement des sons pour les mots ;
2. Les règles phonétiques pour les variations de prononciation quand les mots sont agglutinés ;
3. Les règles prosodiques pour la fluctuation de stress et l'intonation à travers une phrase.

Dans un système de traitement de la parole, les ondes sonores sont analysées et codées en un signal numérique, pour l'interprétation par diverses règles ou par comparaison avec le modèle de langage utilisé [6].

**Le niveau morphologique**

Ce niveau traite la forme des mots et leurs compositions. L'homme segmente un mot en ses constituants pour comprendre son sens. La même chose pour un système de TALN, il peut reconnaître chaque morphème pour aboutir à la reconnaissance du mot. Ce niveau est plus critique pour les langues flexionnelles que pour les langues analytiques [6].

**Le niveau lexical**

A ce niveau, l'homme ou la machine interprète les mots, plusieurs types de traitements contribuent dans la compréhension du niveau mot comme le *Part Of Speech* qui affecte la catégorie grammaticale la plus probable en se basant sur le contexte [6].

**Le niveau syntaxique**

Ce niveau se préoccupe de la structure de la phrase, l'objectif est de reconnaître la structure de la phrase ce qui nécessite une grammaire et un analyseur. La sortie de ce niveau est une représentation de la phrase qui révèle les relations structurelles entre les mots d'une phrase.

Tandis que ce niveau est critique pour les langues analytiques, il ne présente parfois qu'un intérêt rhétorique pour les langues flexionnelles [6].

**Le niveau sémantique**

C'est le niveau où le sens est en principe déterminé, bien que tous les niveaux participent à la détermination du sens.

Le traitement sémantique détermine les sens possibles d'une phrase, en se basant sur les interactions entre les mots au niveau de leurs significations. Il peut inclure la désambigüisation sémantique des mots polysémiques, ou le repérage des locutions figées [6].

**Le niveau discours**

Le niveau discours travaille sur des unités de texte supérieur à une phrase, comme le paragraphe. Il s'intéresse aux propriétés du texte. Plusieurs traitements à ce niveau sont possibles comme le traitement des anaphores [6].

**Le niveau pragmatique**

Ce niveau utilise le contexte pour la compréhension du texte, par exemple le fait que ce texte est relatif à un métier, émane d'une certaine école de pensée… ce qui amène à compléter ou redéfinir le sens apparent. L'objectif est d'expliciter comment un sens supplémentaire doit être lu dans le texte sans être codé. Certaines applications utilisent les bases de connaissances et des modules d'inférence [6].

### 3.5. Les approches de TALN

#### 3.5.1. Approche symbolique

Elle est caractérisée par une analyse profonde des phénomènes linguistiques, basée sur les niveaux cités auparavant.

La principale source dans les systèmes symboliques est le développement des règles et des lexiques.

Un bon exemple de l'approche *symbolique* est vu dans des systèmes logiques à base de règles. La structure symbolique est généralement sous forme de propositions logiques. Les manipulations de ces structures sont définies par des procédures d'inférence ; le moteur d'inférence sélectionne à plusieurs reprises la règle dont la condition est remplie. Un autre exemple d'approches symbolique est les réseaux sémantiques. L'approche symbolique a été employée dans plusieurs domaines de recherche comme *l'extraction d'information*, *la catégorisation de textes*, *la résolution d'ambigüité*, etc.

Les techniques typiques sont : l'apprentissage à base de règles, les arbres de décision, l'algorithme de K plus proches voisins, etc [6].

#### 3.5.2. Approche statistique

Elle emploie des techniques mathématiques, ainsi elle utilise des corpus de textes pour le développement des modèles des phénomènes linguistiques. Un modèle fréquemment utilisé est le modèle caché de Markov (HMM). Le HMM est un automate d'états finis avec un ensemble d'états et des probabilités attachés aux transitions (automate flou).

L'approche *statistique* a été typiquement utilisée dans des tâches telles que la reconnaissance vocale, l'acquisition de lexique, le POS tagging, la traduction statistique, etc [6].

#### 3.5.3. Approche connexionniste

Comme l'approche statistique, l'approche connexionniste développe des modèles à partir des exemples de phénomènes linguistiques. Ce qui diffère l'approche *connexionniste* de l'approche *statistique* est que les modèles connexionnistes combinent l'apprentissage statistique avec différentes théories de représentation. Ainsi dans les systèmes connexionnistes, les modèles linguistiques sont difficiles à observer en raison des architectures qui sont moins limités qu'aux statistiques.

Le modèle connexionniste est un réseau interconnecté de simples unités de traitement avec des connaissances stockées dans le poids des connexions entre les unités.

Les interactions entre les unités peuvent entrainer un comportement dynamique global qui à son tour, conduit à un calcul.[11]

Les recherches actuelles tendent vers le développement par des techniques hybrides qui utilisent les points forts de chaque approche pour tenter de résoudre les problèmes [6].

### 3.6. Les applications de TALN

Les applications de TALN sont multiples et diversifiées, dont on peut citer :

- La recherche d'informations ;
- L'extraction d'information ;
- La réponse aux questions ;
- Les ontologies et la terminologie ;
- Le résumé automatique ;
- La correction orthographique ;
- La reconnaissance de l'écriture manuscrite ;
- La reconnaissance de la parole ;
- La synthèse de la parole ;
- L'annotation sémantique ;
- La résolution des coréférences ;
- L'annotation des ressources ;
- La fouille de textes ;
- La génération de texte ;
- La classification de documents ;
- La traduction automatique.

## *4. La traduction automatique, mythes et réalités*

Maintes idées fausses circulent sur la TA, chez les utilisateurs des systèmes de TA et même chez les professionnels de traduction, les écrivains et les éditeurs.

---

[11] Ces approches sont limitées dans les niveaux d'analyse.

### 4.1. Idée 1 : la traduction automatique est un processus direct

C'est-à-dire qu'il suffit de remplacer chaque mot par son équivalent dans la langue cible ou même chaque phrase pour avoir une traduction, or la qualité de la traduction est décrite comme une réécriture dans une autre langue.

Dans le niveau sémantique, un mot ou une phrase peuvent être traduits par plusieurs possibilités. Dans le niveau syntaxique, une construction en langue source est traduite différemment dans la langue cible, selon les structures de cette dernière [7].

### 4.2. Idée 2 : la traduction est automatisable

En réalité, la TA a des difficultés avec l'ambiguïté linguistique, sous ses différents niveaux. Ce qui nécessite une réécriture par l'humain, or rendre une traduction avec plus de qualité prend (encore actuellement) plus de temps et d'effort que de créer directement une traduction de qualité à partir du texte original [7].

### 4.3. Idée 3 : la traduction est hors de mes mains

C'est problématique quand l'utilisateur ne parle pas la langue cible, alors il délègue tout le processus au système automatique, et il se trouve obligé de faire confiance aux résultats. Or il fallait que cet utilisateur donne des instructions avant de commencer la traduction, faire des vérifications sur les résultats par un natif de la langue cible [7].

### 4.4. Idée 4 : la TA est une perte de temps car elle ne traduit pas les textes de Shakespeare

La critique « la TA ne peut jamais produire une traduction des textes littéraires » est probablement correcte, mais cela ne veut pas dire que la TA est impossible ou inutile. Premièrement, la traduction de la littérature demande des compétences spéciales en littérature, i.e. seuls les traducteurs qui ont des compétences littéraires peuvent intervenir dans ce type de traductions et non pas tous les traducteurs.[12] Deuxièmement, la TA de la littérature constitue une petite partie de la traduction avec ses différents domaines.[13] Finalement, quand on veut traduire les textes de Shakespeare par la machine, il faut penser d'abord que les traducteurs humains trouvent dans cette traduction un défi. Donc les systèmes de TA ne sont pas conçus pour ce type de traduction [8].

---

[12] En France, Edgar Allan Poe est toujours publié dans la traduction de Baudelaire, qu'on a pu comparer à une traduction de Mérimée. La traduction de Baudelaire est supérieure à celle de Mérimée, *et même parfois à l'original…*

[13] Ainsi, on compare souvent les tirages littéraires entre eux. Mais les grands tirages de l'édition sont des utilitaires comme le Petit Larousse (environ 1.000.000 par an) ou La Cuisine de Tante Marie, en réédition permanente depuis près de 90 ans). Le manuel « PDP8S Handbook », en 1968, commençait par un mot du PDG « heureux et fier de préfacer le 400.000$^{ème}$ exemplaire… ».

### 4.5. Idée 5 : la qualité des traducteurs automatiques est mauvaise, alors la TA est inutile

Jour après jour, les systèmes de TA sont plus utilisés à travers le monde. Citons : *METEO, SYSTRAN, LOGOS, ALPS, ENGSPAN, METAL, GLOBAL LINK, …* L'intérêt pour les institutions est majeur.

Certes la TA n'est pas encore arrivée à un stade de haute qualité mais cela ne peut nier l'utilité de la TA. Imaginez que vous ayez devant vous un journal en chinois et que vous attendiez des informations importantes pour vous. Une traduction de faible qualité peut cependant être utile pour vous.

Deuxièmement, quand le traducteur humain fait sa traduction, il produit une première version sans s'intéresser au style d'écriture et à l'organisation, puis une deuxième réécriture pour réviser la première. En ce sens, la TA assure la première étape de traduction [8].

### 4.6. Idée 6 : la traduction automatique menace le métier des traducteurs

La TA ne menace les traducteurs humains que dans leur activité la plus routinière. Actuellement, la qualité des traducteurs automatiques ne peut pas concurrencer celle des traducteurs. Au futur, elle ne peut influencer le métier des traducteurs, car le volume des documents à traduire est énorme et en augmentation continue, ce qui rend la seule intervention des traducteurs humains impossible pour répondre aux besoins.

Ainsi, le facteur de temps à une importance, par exemple il y a des traductions momentanées, comme la traduction d'un journal. Combien, un traducteur prend-il de temps pour le traduire ? Alors que son utilité peut disparaître avant qu'il achève sa traduction [8].

## 5. L'importance de la traduction automatique

### 5.1. Des sociétés utilisant la TA

Parmi les sociétés internationales qui sont actuellement intéressées par le marché de TA des sociétés géantes comme *Microsoft, Siemens, Fujitsu, Hitachi, Toshiba, Oki, NEC,*

*Mitsubishi, Xerox* et *Sharp*. Des entreprises comme *IBM, Intel, Symantec, Cisco, Autodesk, Adobe, Fortis Banque* et *Siemens* ont déjà lancé des pilotes de TA, et ont commencé à intégrer la TA dans leur travail.

### 5.2. Des pays et des institutions utilisant la TA

La liste des utilisateurs de logiciels de TA est encore plus impressionnante. Elle comprend *les Nations Unies*, *le gouvernement américain*, y compris un grand nombre de ses agences civiles et militaires, l'Union européenne. Elle comprend également *le bureau américain des brevets et des marques de commerce*, la *National Science Foundation*, *l'Office Européen des Brevets de La Haye*[14], et de nombreuses institutions internationales importantes dans le monde entier. En outre, l'Europe, le Canada, la Russie, la Chine, la Corée, la Malaisie, l'Indonésie et la Thaïlande ne sont que quelques-uns parmi les nombreux pays s'intéressant à la TA. Pratiquement toute entreprise non-gouvernementale ou organisation ou organisme gouvernemental, qui a besoin de communiquer dans plus d'une langue serait intéressé à utiliser un logiciel de TA.

### 5.3. Le marché mondial de la TA

En ce qui concerne les coûts de traduction, il y a maintenant de nombreuses entreprises sur Internet qui offrent des traductions humaines de pages Web qui coute au moins 0,15 $ US par mot, avec une période d'attente de plus de 24 heures. La TA devrait réduire ces nombres de manière drastique. Actuellement, les logiciels de TA existants peuvent faire baisser le coût de la traduction d'un document tout en minimisant considérablement le temps de traduction.

Dans un avenir proche, un système de TA aurait encore besoin d'intervention humaine pour être en mesure de produire des traductions de qualité acceptable.

*Common Sense Advisory*, une société d'étude et de conseil américaine spécialisée dans le secteur de la traduction et de la localisation, qui suit le marché avec beaucoup de précision depuis plusieurs années, estime que le marché mondial de la traduction représentait en 2009 un poids global de 15 milliards de dollars, et qu'il est en croissance régulière de 15% par an. En 2005, la croissance annuelle était estimée à seulement 12% par la société d'étude IDC. L'Europe constitue la zone la plus importante de ce marché, avec 43%. Elle est suivie des Etats-Unis (40%), de l'Asie (12%), et du reste du monde (5%).

---

[14] Notamment pour la classification automatique des demandes de brevets.

Le marché de la traduction est en croissance continue depuis une vingtaine d'années. Il existe plusieurs raisons principales pour cette croissance :

- La traduction répond à un besoin fondamental : communiquer en s'affranchissant de l'obstacle de la langue. Alors que l'acte de communication n'a jamais été rendu aussi facile par les nouvelles technologies (Internet, téléphonie portable, satellites, etc.), et que les frontières ne cessent de disparaître avec la constitution de grandes zones de libre-échange (USA, UE, Alena, etc.), les langues constituent encore un obstacle incontournable au dialogue interculturel. Plus les technologies se développeront, plus le libre échange des biens et des personnes sera encouragé, plus le marché de la traduction se développera ;

- La mondialisation pousse les entreprises à gagner sans cesse de nouveaux territoires. A chaque fois, elles doivent réapprendre à communiquer dans la langue de leurs nouveaux clients si elles veulent leur vendre ;

- Conjointement à cette expansion économique des entreprises, les échanges sont de plus en plus réglementés, « judiciarisés », ce qui est la source de nouveaux textes, règlements, contrats, etc. Avec une très forte tendance à imposer les concepts à la source du droit anglo-saxon un peu partout dans le monde ;

- Pour proposer les nouvelles technologies à un public toujours plus vaste, les industriels simplifient sans cesse les interfaces d'utilisation, ce qui, a contrario, encourage la création de documentations toujours plus étoffées, plus complètes : il existe une véritable inflation des textes d'aide (sites Web, documentations, fichiers d'aide, livres, forums, blogs...), dont une très grande partie doit être traduite [9].

Alors, l'investissement en TA est une priorité absolue. Bien que seulement 8% de la population mondiale connaisse l'anglais (comme langue maternelle ou seconde), près de 50% de littérature scientifique et technologique est publié en anglais[15], et plus de 28% des utilisateurs Internet utilisent l'anglais (figure 1.1). Cela signifie qu'une personne qui ignore l'anglais serait privée de la moitié de la littérature disponible en matière de technologie.

La figure 1.1 présente la répartition des 10 premieres langues utilisées sur Internet.

---

[15] Des États comme l'Iran ou la Chine développent des Internet clos, tandis que l'Inde vient d'autoriser les 3 premiers masters n'utilisant pas l'anglais… afin de garder les langues indiennes plus vivantes.

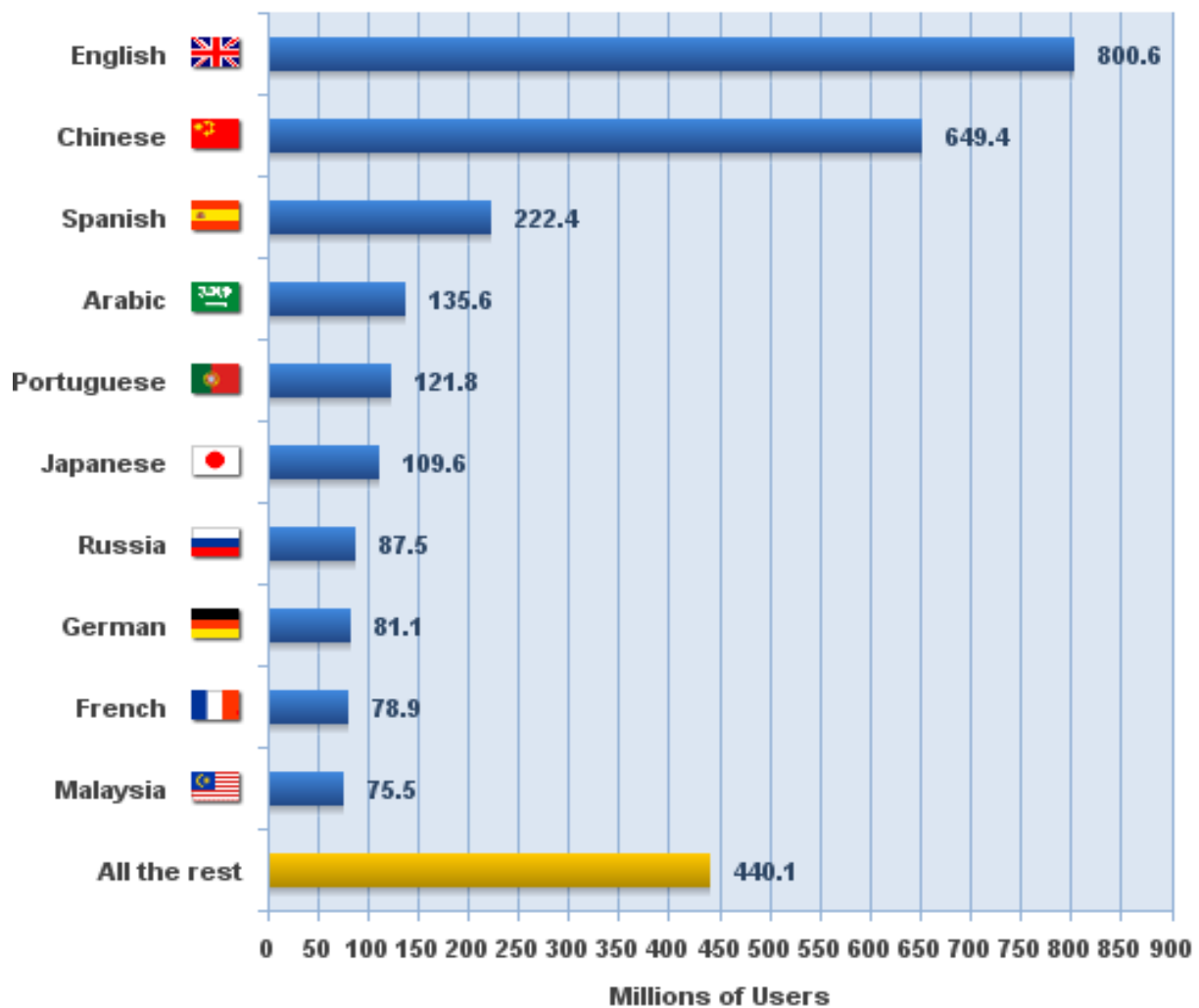

**Figure 1.1 Les dix premières langues sur Internet 2013 – par millions d'utilisateurs [10]**

## 6. Les raisons d'utiliser des systèmes de traduction automatique

Les technologies de TA sont développées au prix de millions de dollars pour répondre à un besoin réel et exprimé. Pourquoi les utilisateurs porteraient-ils autant d'intérêt à la TA s'ils n'avaient pas avant tout un problème avec la traduction humaine ?

La vraie question donc est de comprendre pourquoi cette dernière n'est plus capable de répondre aux besoins des utilisateurs. Deux aspects principaux : Avant tout, la traduction humaine traditionnelle peut être trop lente. Malgré le professionnalisme et l'expérience, les efforts sont insuffisants pour répondre aux attentes des clients. Ensuite, la qualité de la traduction humaine est trop incertaine. Le jugement diffère trop souvent d'un traducteur à un autre, et sous la pression de délais de plus en plus courts et d'une gestion manuelle, il devient encore plus difficile de pouvoir compter sur une qualité constante.

C'est pour ces deux raisons, notamment, que la TA intéresse tant. Ce qui peut être automatique est entrepris plus rapidement ; et une activité automatique est censée livrer les mêmes résultats tout le temps [11].

Les systèmes de TA peuvent être utilisés pour plusieurs raisons :

- Comme moyen de traduction de documents scientifiques et techniques, des procédures commerciales et administratives, des manuels, des modes d'emplois, des dépliants publicitaires, des rapports d'informations,…

- Comme une opportunité commerciale, pour exporter des produits locaux à l'étranger. Donc la TA est un outil qui facilite le commerce mondial qui est devenu global avec l'anglais comme langue officielle. Afin de permettre aux non anglophones de bénéficier du commerce mondial, il faut trouver un moyen pour traduire les documents produits dans leurs langues aux langues de leurs partenaires et vice versa. En fait, pour que le commerce soit totalement globalisé, soit le monde entier devrait parler une seule langue (ce qui serait absurde et nuirait à la créativité), soit de trouver un moyen de traduction entre les différentes langues du monde.

- Comme une technologie qui permet aux non-anglophones de faire partie d'Internet qui est devenu une partie vitale de commerce et d'échange d'informations, partout dans le monde. Alors que les non-anglophones sont désavantagés du fait que la plupart des informations sur internet sont en anglais.[16]

- Comme un outil qui aide chaque communauté à préserver sa culture et à la présenter aux autres peuples. Toute culture qui s'efforce de survivre dans le village planétaire doit investir beaucoup d'efforts et de ressources dans la TA.

---

[16] Cependant le mouvement « web sémantique » essaie de représenter les connaissances en confinant les langues au nommage des concepts des ontologies.

- Comme plateforme commune pour le développement de la recherche en Intelligence artificielle, sciences de l'information, la linguistique computationnelle, la reconnaissance vocale, la recherche multilingue, la génération de textes, la gestion de connaissances, … [12].

## *7. Les difficultés de la traduction automatique*

La TA, généralement est une tâche fastidieuse et difficile comme tout domaine d'activité humaine. Considérant le passage suivant du roman « *L'histoire de la pierre* » appelé aussi « *rêve de la chambre rouge* » de Cao Xueqin en 1792 transcrite en mandarin : (*dai yu zi zai chuang shang gan nian bao chai. . . you ting jian chuang wai zhu shao xiang ye zhe shang, yu sheng xi li, qing han tou mu, bu jue you di xia lei lai*).

La figure 1.2 montre la traduction anglaise de ce passage par *David Hawkes* dans les phrases E1-E4. Pour faciliter la lecture au lieu de donner la phrase en chinois, on la présente en anglais. Les mots en bleu sont des mots chinois, non traduits en anglais ou des mots anglais sans équivalents en chinois. Les lignes d'alignement présentent les correspondances des mots dans les deux langues [13].

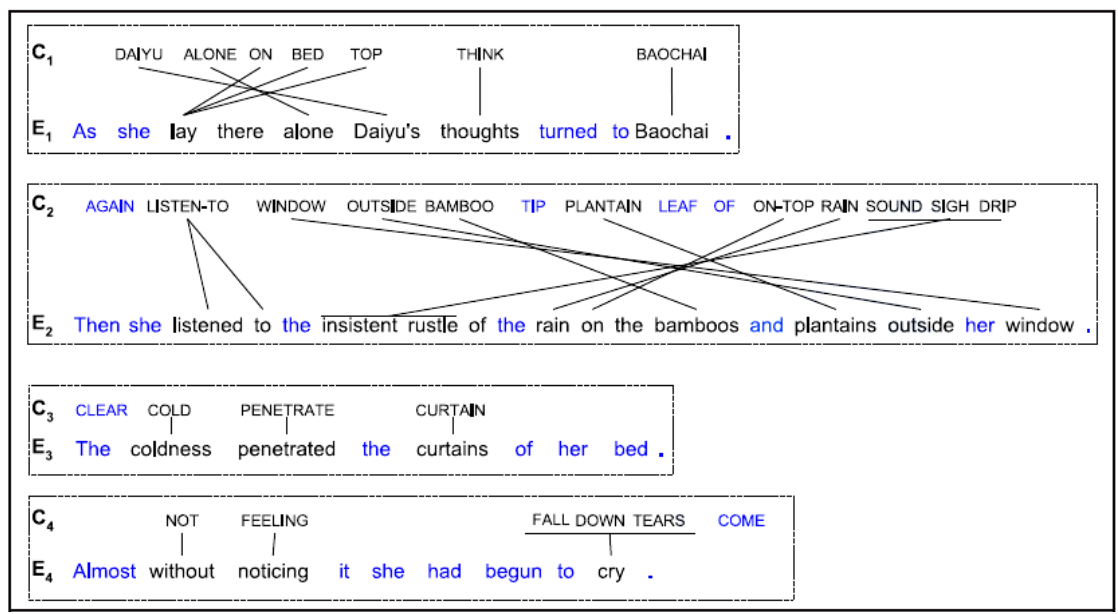

**Figure 1.2 Traduction d'un passage chinois vers l'anglais [13]**

Considérons certaines questions soulevées dans cette traduction. Tout d'abord, les textes en anglais et en chinois sont très différents structurellement et lexicalement. Les quatre phrases en anglais correspondent à une longue phrase chinoise. L'ordre des mots des deux textes est très différent, comme nous pouvons le voir par le grand nombre de mots anglais marqués en bleu. Beaucoup de ces différences sont dues aux différences structurelles entre les deux langues. Par exemple, comme le chinois marque rarement l'aspect verbal ou temporel, la traduction en anglais emploie des mots supplémentaires comme (*as*), (*turned to*) et (*had begun*). *Hawkes* a du décider de traduire le mot chinois (*tou*) comme (*penetrated*) plutôt que de dire (*was penetrating*) ou (*had penetrated*).

Le chinois a moins d'articles et aussi moins de pronoms que l'anglais, d'où le grand nombre de (*the*) en bleu.

*Hawkes* dut insérer (*she*) et (*her*) dans de nombreuses positions dans la traduction anglaise.

Les différences stylistiques et culturelles sont une autre source de difficulté pour le traducteur. Contrairement aux noms anglais, les noms chinois sont constitués de mots avec des significations. *Hawkes* a choisi d'utiliser la translitération (*Daiyu*) pour les noms avec les principaux caractères, mais à traduire les noms de fonction par leur signification (*Aroma, Seky bright*).

La phrase (*bamboo tip plantain leaf*) bien élégante en chinois où ses quatre parties sont une caractéristique de la prose serait gênant si traduit mot à mot en anglais. *Hawkes* utilise donc simplement (*bamboos*) (*plantains*), la traduction de ce type de phrases exige clairement une compréhension profonde et riche de la langue source et le texte d'entrée et un système sophistiqué et créatif de la langue cible.

Le problème d'effectuer automatiquement une traduction littéraire de haute qualité entre des langues différentes comme le chinois et l'anglais est donc beaucoup trop difficile à automatiser complètement.

Cependant, même les traductions non littéraires entre des langues comme l'anglais et le français peuvent être difficile. Voici une phrase en anglais à partir du corpus *Hansard* des débats du parlement canadien, avec sa traduction en français :

**Anglais :** *Following a two-year transitional period, the new Foodstuffs Ordinance for Mineral Water came into effect on April 1, 1988. Specifically, it contains more stringent requirements regarding quality consistency and purity guarantees.*

**Français :** *La nouvelle ordonnance fédérale sur les denrées alimentaires concernant entre autres les eaux minérales, entrée en vigueur le 1$^{er}$ avril 1988 après une période transitoire de deux ans, exige surtout une plus grande constance dans la qualité et une garantie de la pureté.*

French gloss : *THE NEW ORDINANCE FEDERAL ON THE STUFF FOOD CONCERNING AMONG OTHERS THE WATERS MINERAL CAME INTO EFFECT THE 1ST APRIL 1988 AFTER A TRANSITORY PERIOD OF TWO YEARS REQUIRES ABOVE ALL A LARGER CONSISTENCY IN THE QUALITY AND A GUARANTEE OF THE PURITY.*

Malgré les chevauchements structurels forts et les différences de vocabulaires entre l'anglais et le français, cette traduction, comme la traduction littéraire, doit encore faire face à des différences dans l'ordre des mots (par exemple, l'emplacement de la phrase *of the following a two-year transitional period*) et en structure (par exemple, l'anglais utilise **le nom** *requirements* alors que les Français utilise **le verbe** *exige*).[17]

Néanmoins, ces traductions sont beaucoup plus faciles, et un certain nombre de tâches de traduction non littéraire peut être abordé avec les modèles computationnels actuels de la TA, notamment : (1) des tâches pour lesquelles une traduction approximative est suffisante, (2) des tâches où une poste édition humaine est utilisée, et (3) des tâches limitées à des domaines restreints avec un sous-langage dans lesquelles la TA avec une qualité est toujours possible. L'acquisition d'information sur le web est un exemple de tâches où une traduction approximative peut toujours être utile [13].

## 8. *Evaluation des systèmes de traduction automatique*

Une question importante à poser lorsqu'il s'agit d'un système de TA est la possibilité de l'évaluer scientifiquement. C'est en vue d'établir son utilité et de le comparer avec les systèmes existants. Cette évaluation devrait couvrir à la fois les caractéristiques linguistiques et computationnelles, et doit être informative pour les clients et les utilisateurs du système. Toutefois, une partie importante de l'évaluation doit être effectuée par les développeurs pour vérifier que le système fonctionne comme prévu ; et les modifications sont possibles, sans qu'elles soient radicales. L'évaluation peut être effectuée en plusieurs phases :

- par les développeurs (éliminatoire)
- par les traducteurs (pour la mise au point, l'exactitude)
- par les utilisateurs finaux (pour la lisibilité, la clarté)

Plusieurs types d'évaluation ont été développés, on cite :

---

[17] Cf. L. Tesnière, *Éléments de syntaxe structurale,* Paris : Klincksiek, 1959.

### 8.1. Évaluation en boîte noire (Black box evaluation)

Le système est évalué en regardant les résultats et en les comparant avec ce qui est attendu. Ceci est réalisé souvent comme comparaison avec la traduction humaine ; autrement dit, les résultats sont comparés avec ce que les experts humains pourraient produire. Ce qui est subjectif comme évaluation d'un système de TA – mais donne quand même des indications sur les faiblesses générales du système [13].

### 8.2. Évaluation en boîte blanche (Glass box evaluation)

Il s'agit d'évaluer les éléments constitutifs du système. Ce type d'évaluation est effectué généralement par les développeurs, afin de mesurer les améliorations apportées au produit [13].

### 8.3. Évaluation selon les critères de la clientèle

C'est une évaluation basée sur les critères et les exigences de la clientèle. Il s'agit de déterminer l'utilité d'un système dans l'environnement réel dans lequel il fonctionnerait.

L'évaluation des systèmes de TA a suscité l'intérêt des agences financières, depuis les étapes précoces du développement du système. Des efforts ont été consacrés à l'élaboration des critères, des indicateurs et des méthodologies d'évaluation [12].

Lors du développement d'un système de TA, l'évaluation est une tâche essentielle mais elle est subjective. Elle peut être humaine ou automatique.

### 8.4. Evaluation humaine

On évalue à travers deux dimensions : la fidélité et la clarté :

**Clarté :** on peut chercher combien la traduction est intelligible, claire, lisible et combien les résultats sont naturels.

**Fidélité :** elle traite deux points : la pertinence et le caractère informatif.

1. **Pertinence :** combien d'information originale du texte source est préservée dans le texte cible.
2. **Caractère informatif** : y a-t-il suffisamment d'informations dans le texte résultant ? (on peut lire le texte source, et poser des questions à une personne qui n'a que le texte résultat).[18]

---

[18] Il est assez rare mais parfois pragmatiquement nécessaire que la traduction soit plus riche que la source.

On peut évaluer par post-édition, quel est le nombre de mots et le temps nécessaire à une personne pour corriger les résultats [13].

### 8.5. Evaluation automatique

La tâche d'évaluation humaine consomme beaucoup de temps, prenant des jours ou même des semaines. Il est donc utile d'avoir des métriques automatiques plus rapides à évaluer, mais il faut accepter qu'elles soient pires que l'évaluation humaine, tant elle est corrélée avec les jugements humains.

En fait, il y a plusieurs méthodes heuristiques, comme Bleu, TER, Précision et Rappel et METEOR. L'intuition de ces métriques dérive du *Miller* et *Beebe Center* (1958) qui a souligné que les bons résultats d'un système de TA sont celles qui sont très similaires à la traduction humaine.

Pour chacune de ces métriques, nous supposons que nous avons déjà une ou plusieurs traductions humaines de phrases pertinentes.

Etant donné une phrase résultat d'un système de TA nous calculons la distance de traduction entre les résultats automatiques et les phrases humaines.

Un résultat est classé comme meilleur s'il est en moyenne proche de la traduction humaine. Les métriques diffèrent dans le calcul de la distance de traduction (proximité de la traduction).

Dans le domaine de la reconnaissance vocale, la métrique de proximité de transcription est le taux d'erreur Mot, qui est la distance minimale d'édition à une transcription humaine.

Par contre dans la TA, nous ne pouvons pas utiliser le même mot comme paramètre d'erreur, car il y a plusieurs traductions possibles d'une phrase source.

Un bon résultat pourrait ressembler à une traduction humaine, mais très différente d'une autre. Pour cette raison, la plupart des métriques jugent un résultat en le comparant à plusieurs traductions humaines [13].

### 8.6. La métrique Bleu (BiLingual Evaluation Understudy)

 [14]

Bleu est un algorithme d'évaluation de qualité des textes traduits par des machines. La qualité est considérée comme la correspondance entre le résultat d'une machine et celui d'un homme. La meilleure TA est celle qui est très proche de la traduction humaine.

Pour le principe de travail, on classe les résultats par une moyenne pondérée du nombre de N-gramme qui chevauchent avec la traduction humaine.

La figure 1.3 montre une intuition, à partir de deux traductions candidates d'une phrase chinoise, représenté avec 3 traductions de référence humaine. La phrase candidate 1 partage beaucoup de mots avec les traductions de référence.

**Figure 1.3 Intuition de Bleu sur deux traductions candidates [13]**

Observons maintenant comment le score Bleu est calculé. Bleu est basé sur le calcul de la précision. Une précision 1 gramme compte le nombre de mots dans la traduction candidate (résultat du système) qui apparaissent dans une des traductions références et le divise par le nombre total des mots dans la traduction candidate. Si une traduction candidate avait 10 mots, et 6 entre eux apparaissent dans au moins une des traductions références, nous aurions une précision de 6 /10 = 0,6.

Le problème est qu'il y a des failles dans l'utilisation de la simple précision ; elle récompense les phrases candidates avec des mots répétés.

La figure 1.4 montre un exemple d'une phrase candidate pathologique composée de plusieurs instances du mot (*the*), et puisque chaque mot des sept identiques apparaît dans une des traductions références, la précision serait 7/7= 1 ! [13].

**Figure 1.4 Un exemple pathologique dans la précision 1 gramme de Bleu [13]**

Afin d'éviter ce problème, Bleu utilise une précision de N grammes modifiée. Nous comptons d'abord le nombre maximal de fois qu'un mot est employé dans chaque traduction référence. Le nombre pour chaque mot est ensuite ajusté par cette référence maximale.

Alors, la précision modifiée dans l'exemple précédent serait 2/7, selon la référence avec un maximum de 2 (*the*).

Revenant à l'exemple chinois, la première phrase candidate a une précision 1 gramme modifiée de 17/18, tandis que la deuxième possède 8/14.

Nous calculons de la même façon, la précision modifiée pour N>1, la précision bi-gramme modifiée pour la phrase candidate 1 est 10/17 et pour la candidate 2 est 1/13.

Pour calculer le score total, Bleu calcule le N gramme pour chaque phrase, puis elle ajoute les calculs ajustés sur toutes les phrases candidates et le divise par le nombre total des N grammes. Le score de décision modifié sera :

$$p_n = \frac{\sum_{C \in \{Candidates\}} \sum_{n\text{-}gram \in C} \text{Count}_{clip}(n\text{-}gram)}{\sum_{C' \in \{Candidates\}} \sum_{n\text{-}gram' \in C'} \text{Count}(n\text{-}gram')} \qquad (1.1)$$

Bleu utilise les 1gramme, bigrammes, trigrammes et même les quadri grammes. Elle combine ces précisions modifiées par la moyenne géométrique.

En outre, Bleu ajoute une pénalité supplémentaire pour pénaliser les traductions trop courtes.

Considérons la traduction (*of the*) comparée avec les référence 1-3 (figure 1.3), comme cette phrase est très courte, et tous ces mots apparaissent dans certaines traductions, sa précision 1 gramme modifiée est gonflée à 2/2.

Normalement, nous traitons ce type de problèmes en combinant la précision avec le rappel, mais comme il a été expliqué avant, nous ne pouvons pas utiliser le rappel sur plusieurs traductions humaines, puisque le rappel exige (incorrectement) qu'une bonne traduction doit contenir beaucoup de N grammes de chaque traduction. Au lieu de cela, Bleu comprend une brève pénalité sur l'ensemble du corpus. Soit *C* la longueur totale du corpus des traductions candidates. On calcule la référence *r* de ce corpus, avec un additionnement pour chaque phrase candidate des longueurs des meilleures combinaisons alors la pénalité est une exponentielle de *r/c* :

$$BP = \begin{cases} 1 & \text{if } c > r \\ e^{(1-r/c)} & \text{if } c \leq r \end{cases}$$
$$\text{Bleu} = BP \times \exp\left(\frac{1}{N}\sum_{n=1}^{N}\log p_n\right) \quad (1.2)$$

Tant que des métriques comme Bleu (NIST, METEOR,…) sont très utiles pour le développement des systèmes de traduction et correspondent souvent aux jugements humains, elles ont certaines limites qui sont importantes à considérer. Premièrement, beaucoup de méthodes se concentrent sur des informations très locales.

Considérons l'exemple de la figure 1.3, pour produire une phrase candidate comme « *Ensures that the military it is a guide to action which always obeys the commands of the party* ».

Cette phrase aurait un score inférieur. Par ailleurs, les métriques automatiques comparent mal des systèmes avec des architectures différentes.

Ainsi Bleu par exemple n'est pas toujours d'accord avec les jugements humains sur la qualité des traductions. Pour évaluer le rendement des systèmes commerciaux comme Systran contre les systèmes N-grammes statistiques, on conclut que les métriques automatiques sont les plus appropriés lors de l'évaluation des changements progressifs d'un seul système, ou des systèmes différents mais avec des architectures similaires [13].

## *9. Les approches de la traduction automatique*

Cette section résume plusieurs approches qui ont jalonné la recherche sur la TA, de 1950 à nos jours.

Les premiers programmes d'ordinateurs relatifs à la traduction étaient destinés à servir d'aide à la traduction. Quelques règles et surtout un dictionnaire bilingue composaient le cœur du système. Les années suivantes voient les dictionnaires grandir ; ce qui engendre une augmentation du nombre de règles régissant le ré-ordonnancement des mots. La nécessité d'automatiser l'acquisition des règles et de progresser leur généricité participe au développement de la linguistique informatique.

### 9.1. L'approche à base de règles

Le triangle présenté à la figure 1.5 est attribué à Vauquois. Il présente de manière synthétique une analyse du processus de traduction encore pleinement pertinente et employée de nos jours.

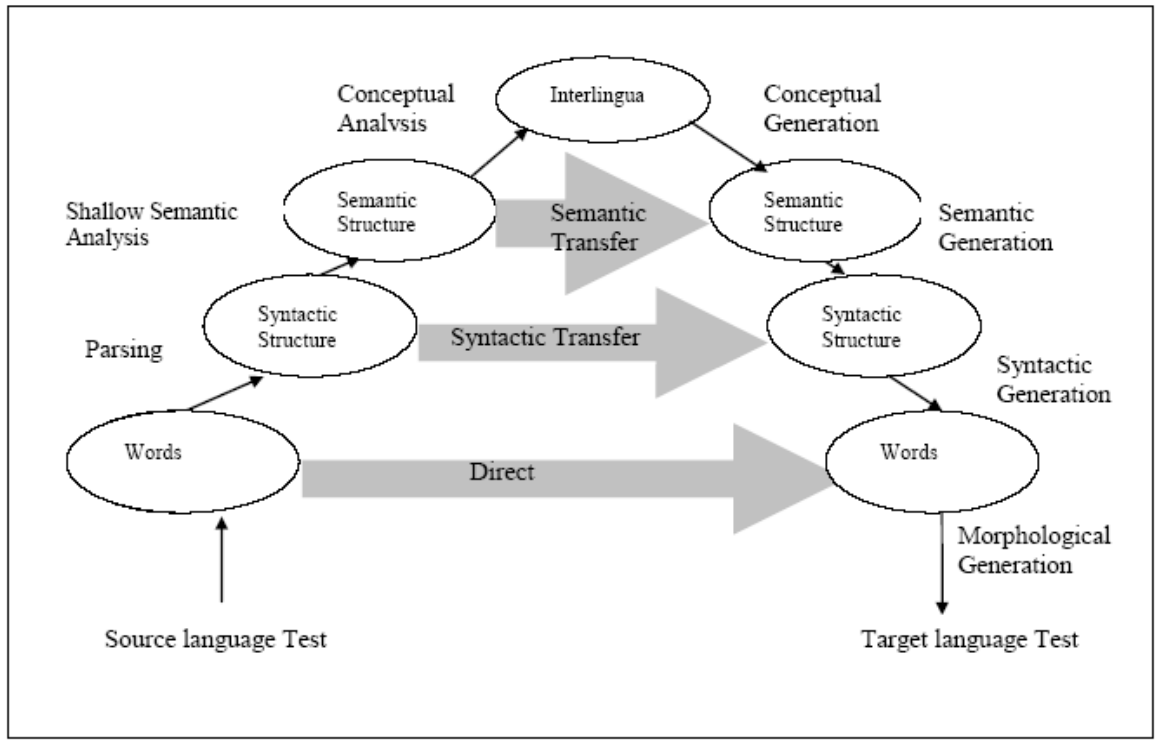

**Figure 1.5 Triangle de Vauquois**

La traduction peut s'opérer à plusieurs niveaux.

Au niveau le plus bas, on retrouve la *traduction directe*, qui passe directement des mots de la langue source aux mots de la langue cible.

Les *systèmes semi-directs* ont une phase de segmentation ou d'analyse morphologique, voire morphosyntaxique, et une phase de génération morphologique. Si l'on effectue une analyse syntaxique de la phrase source, le transfert à la langue cible devrait être simplifié. À ce niveau, les détails spécifiques à la constitution des groupes nominaux, par exemple, n'ont pas besoin d'être connus des règles régissant le transfert.

Avec une analyse plus approfondie de la phrase source, au niveau sémantique, le transfert devient uniquement sémantique. En revanche, la génération des mots après le transfert est plus complexe qu'au niveau inférieur.

Enfin, une analyse totale de la phrase source peut aboutir à une représentation de son sens dans une « inter-langue » artificielle, de laquelle on dérive ensuite les mots cible.

Le pivot est un type de représentation utilisant des attributs et relations interlingues, et des unités lexicales de chacune des langues. Ces systèmes sont à transfert simple, alors qu'on a un double transfert en « pivot ».

L'approche reposant sur une inter-langue est attractive car elle remplace le problème de la traduction par deux problèmes unilingues, d'analyse et de synthèse. L'avantage est que les modules unilingues sont à priori réutilisables. Pour couvrir tous les sens de traduction entre n langues, il suffit de n modules d'analyse et de n modules de synthèse, contre n*(n−1) systèmes de transfert [15].

Des systèmes à véritable *langue-pivot*, on peut citer : ATLAS-II de Fujitsu ou IVOT/Crossroads de NEC, ou KANT/CATALYST de CMU/Caterpillar, ou UNL, ou MASTOR-1 d'IBM.

Le début des années 1990 voit le développement d'autres types d'approches. Les ordinateurs se répandent et gagnent en puissance, ce qui permet l'émergence de stratégies qui se fondent sur de grandes quantités de données « corpus-based approaches ». On distingue en particulier deux grands types d'approches : la TA à base d'exemples et la TA par méthodes statistiques.

### 9.2. L'approche statistique

Elle doit ses origines aux travaux de Brown et al. en 1990 [16] et en particulier au prototype Candide de Berger et al. en 1994 [17], un système de traduction construit à partir de discours disponibles en français et en anglais de parlementaires canadiens. En effet, comme la traduction à base d'exemples, la traduction par méthodes statistiques repose sur un corpus parallèle.

Un modèle statistique de traduction est défini, comprenant une ou plusieurs lois de probabilités. Le corpus est traité afin d'estimer ces lois qui sont souvent constituées de plusieurs milliers, voire millions de paramètres.

### 9.3. L'approche à base d'exemples

L'approche à base d'exemples (« Example-based machine translation », ou EBMT) repose sur un ensemble « d'exemples » préalablement traduits : un corpus parallèle de phrases traduites l'une de l'autre. Lorsqu'on lui présente une phrase à traduire, le système parcourt sa base d'exemples et produit trivialement une traduction si la phrase s'y trouve. Dans le cas général, la phrase n'apparaît pas dans la base et le système s'emploie alors à rassembler des exemples qui contiennent des fragments communs (des groupes de mots) avec la phrase à traduire. Pour chaque fragment d'exemple dans la langue source, il s'agit ensuite de retrouver sa traduction dans la langue cible : c'est la phase d'alignement. Enfin, la phase de génération assemble les fragments dans la langue cible et produit la traduction. À chacune des trois étapes, il est possible d'utiliser des sources externes de connaissances, telles que des lexiques bilingues, des listes de synonymes, des étiquettes ou des arbres syntaxiques, etc [18].

### 9.4. Les approches hybrides

Leur idée directrice est qu'une approche unique du problème de la traduction, aussi perfectionnée soit-elle, ne parviendra pas à produire une traduction satisfaisante dans tous les cas. Au contraire, une approche par règle peut s'avérer particulièrement adaptée à certaines phrases, tandis que d'autres phénomènes linguistiques sont correctement traités par une approche reposant sur des corpus.

Les systèmes hybrides sont actuellement envisagés comme des systèmes combinant les méthodes statistiques ou méthodes à base d'exemples avec des méthodes linguistiques (à base de règles), en particulier pour l'analyse morphologique et syntaxique.

Un système hybride pourrait parvenir à tirer profit des forces de chaque approche. Une première stratégie pour mettre en œuvre un système hybride est d'utiliser les différentes approches en parallèle. Enfin, dans un système statistique, il est courant de traiter par un système de règles spécialisées certains fragments de phrases, typiquement les nombres, les dates, etc. Les morceaux de phrases ainsi identifiés et traduits en isolation par le système à base de règles peuvent être transmis au système statistique [18].

## *Conclusion*

Nous avons vu dans ce chapitre la traduction automatique, son histoire, son développement et ses progrès. Les mythes et les réalités, les fausses idées sur la TA et les différentes discussions et polémiques entre traduction humaine et TA, son importance et les raisons qui interpellent son utilisation. Les difficultés de la TA sont les mêmes que celles du traitement du langage naturel et plus, car on a affaire à deux langues et à un double traitement : *analyse* et *génération*. Donc la TA est une tâche fastidieuse mais fructueuse, les systèmes actuels donnent de l'espoir et affirment que la TA est possible.

Les approches de la TA tirent leurs idées de la traduction humaine notamment les approches linguistiques, les différents modèles, procédés et approches de la traduction humaine seront l'objet du prochain chapitre.

# Chapitre 2. La Traduction Humaine

## *Introduction*

Le traitement automatique de l'information diffère selon le type d'information traitée : texte, image, son ou vidéo. Pour maîtriser la traduction automatique, il faut comprendre la traduction humaine, ses procédés, ses modèles et ses principes ; comment cela se passe dans la tête du traducteur, quels sont les procédés mentaux qui permettent à l'homme de passer d'une langue source vers une langue cible, quelles sont les différentes approches qui ont été proposée jusqu'à présent. Ce chapitre donne un survol sur la traduction et ses approches.

### *1. Origine de la traduction*

L'histoire de la traduction est celle du monde et des civilisations. Son origine n'a jamais été déterminée avec exactitude. Van Hoof affirme en effet que la traduction remonte au moins à 3000 ans avant J.C, si l'on s'en tient au témoignage le plus ancien dont on dispose, et selon lequel « les égyptiens disposaient d'interprètes et usaient d'un hiéroglyphe spécifique pour exprimer la fonction d'interprétation » [19].

La traduction trouve son fondement dans *La Genèse* (XI, 1:9) avec le mythe de la tour de Babel (Hébreu : מגדל בבל Migdal Bavel, Arabe : برج بابل Burj Babil). Après le Déluge, les premiers hommes, qui parlent une seule langue[19], entreprennent de bâtir une ville et une tour dont le sommet touche le ciel, pour atteindre Dieu, car ils veulent tout le pouvoir. Dieu les voit, et pour les punir de cette mauvaise pensée Il brise la tour et brouille leur langue afin qu'ils ne se comprennent plus. Ils sont ensuite dispersés sur toute la surface de la terre. La ville est alors nommée Babel (terme proche du mot hébreu traduit par « brouiller »).

---

[19] Des travaux récents sont consacrés à la caractérisation d'une hypothétique langue-mère d'où procéderaient toutes les langues, la filiation pouvant s'aider de considérations génétiques.

Après la confusion des langues et la chute de la tour de Babel, les hommes deviennent comme sourds au discours d'autrui, et ne se comprennent plus. Ils auront désormais besoin de la traduction pour se comprendre.

En un certain sens, la traduction est au langage, ce que la lumière est à la vue.

Depuis la destruction de la tour de Babel, symbole de la rupture d'une unité linguistique universelle, les hommes ont dû recourir à la traduction chaque fois que des communautés de langues différentes éprouvaient le besoin de communiquer[20] [19].

Une autre théorie [20] explique que la première civilisation humaine après le Déluge était en Irak, et la langue du peuple était le Babylonien, le nombre de la population se multiplie, les gens se trouvent dispersés sur la Terre dans des migrations massives, à la recherche de moyens de subsistance ; l'une des premières migrations devait atterrir sur le Nil, une autre en Chine ; elles y ont formé des civilisations célèbres. Les gens ont commencé à trouver de nouvelles choses, ils ont inventé des noms à ces objets, et ces mots inventés étaient différents avec ceux d'autres civilisations. Les ajouts dépendaient de ce qu'ils voyaient et entendaient, et de jour en jour, ils ajoutaient de nouveaux mots selon le besoin. La différence s'intensifia avec le passage des années, ainsi d'autres langues apparaissent après d'autres migrations. La traduction est alors devenue un moyen de rétablir la communication.

Dans le Coran, on trouve la diversité des langues des hommes dans la sourate d'Ar rum (les Byzantins) :

۩ وَمِنْ آيَاتِهِ خَلْقُ السَّمَاوَاتِ وَالْأَرْضِ وَاخْتِلَافُ أَلْسِنَتِكُمْ وَأَلْوَانِكُمْ ۚ إِنَّ فِي ذَٰلِكَ لَآيَاتٍ لِلْعَالِمِينَ ۩

۩ *parmi ses signes, la création des cieux et de la terre et la différence de vos langues et de vos sortes, en quoi résident des signes pour ceux qui savent* ۩

Ibn Kathir dans son exégèse de ce verset dit : « Allah dit que (*parmi Ses signes*) prouvant Son immense omnipotence ; il y a la création des cieux dans leur hauteur et leur immensité, et de la terre avec ses dépressions et ses montagnes, ses continents et ses mers, etc. Le segment (*la différence de vos langues et de vos sortes*) donne une indication sur les multiples langues et les multiples aspects distinguant les uns des autres. Il y a les Arabes, les Tartares, les Byzantins, les Européens, les Berbères, les Abyssiniens, les Indiens, les Kurdes, etc., le Blanc aux lèvres retroussées, le Blanc aux lèvres épaisses, le Noir au front bombé, le Noir au front large, etc. En tout cela donc (*résident des signes pour ceux qui savent*) » [21].

---

[20] Généralement à des fins commerciales « Niemetz » (les muets) est le nom donné par les Russes aux Allemands, les premiers contacts ayant eu lieu lors de trocs effectués sans mot dire.

Dans l'exégèse de Jalalayne : « Parmi Ses signes encore : Il a créé les cieux et la terre et la diversité de vos langues : arabe et non arabes, et de couleurs : blanche, noire et autre alors que toute l'humanité est née d'un seul couple : Adam et Eve. Il y a là une preuve de Son omnipotence pour des hommes sensés qui raisonnent » [22].

Dans l'exégèse de Tabari : « Et la différence de logique de vos langues, (*de vos sortes*) et la différence de couleurs de vos corps (*des signes pour ceux qui savent*) en faisant cela, il donne des leçons et des preuves pour ses créatures qui comprennent qu'il ne se fatigue pas de les retourner à leur état dont ils étaient avant de mourir après leur dissolution» [23].

Concluons notre discussion sur les langues avec la figure 2.1 qui présente la répartition des langues dans le monde. D'où, on peut voir l'aménagement linguistique, l'inégalité géographique des langues et la cohabitation des langues.

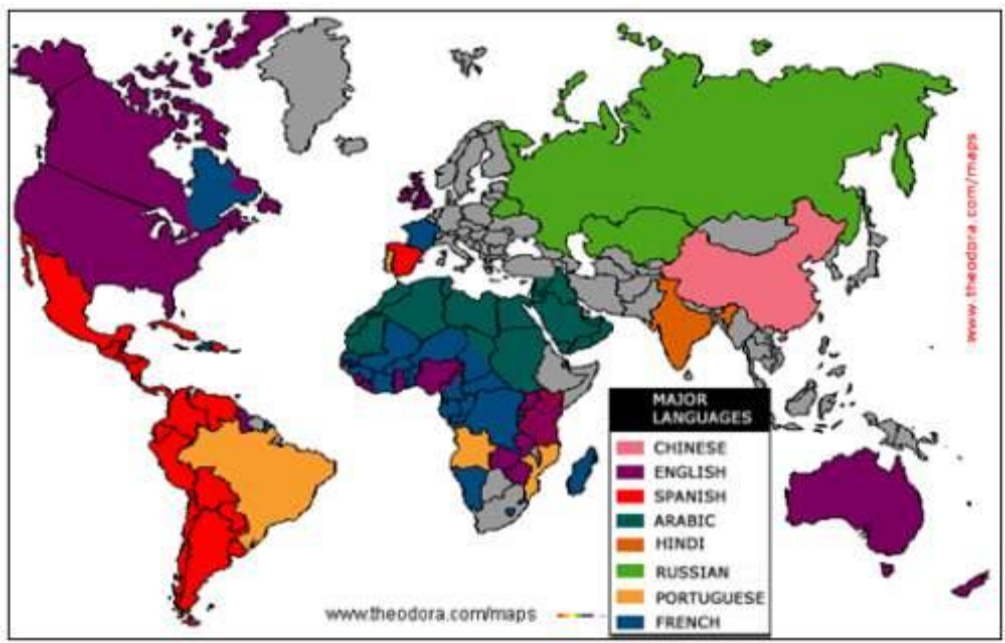

**Figure 2.1 Les langues dans le monde [24]**

## *2. Définitions*

### 2.1. Traduction

La traduction écrite est la réécriture, dans une langue, d'un énoncé écrit dans une autre ; au sens large la traduction englobe aussi bien l'activité qui part d'un texte écrit pour aboutir à un autre texte écrit que celle qui va d'un discours improvisé à une interprétation orale et d'un texte écrit à une traduction orale [25].

Au sens du dictionnaire, traduire consiste à faire passer un texte ou un discours d'une langue à une autre. Autrement dit, pour traduire un texte, deux éléments sont indispensables : *la parfaite compréhension* du texte source, et *la connaissance de la formulation équivalente* dans la langue cible, qui doit être *de préférence* la langue maternelle du traducteur – car la règle d'or en traduction est que l'on ne traduit *bien* que vers sa langue maternelle [26].

### 2.2.     Interprétariat

*L'interprétariat – interprétation* diront les puristes – est la traduction orale d'une intervention orale. Alors que *l'usage* veut que les traducteurs ne travaillent que vers leur langue maternelle, les interprètes quant à eux peuvent traduire dans les deux sens. On distingue l'interprétariat de conférence (ou simultané) de l'interprétariat consécutif. Comme son nom l'indique, l'interprétariat simultané consiste à traduire à voix haute et en même temps qu'il s'exprime, les propos d'un orateur. Ce type d'interprétariat nécessite un entraînement et des facultés de concentration exceptionnelles[21], ainsi que du matériel dédié. L'interprétariat consécutif, qui consiste à traduire les propos de l'orateur à chacune de ses interruptions, l'interprète utilise un système de prise de notes spécifique pendant que le locuteur s'exprime, puis restitue le propos dans une autre langue à la fin de l'intervention. L'interprétariat consécutif présente moins de contraintes, mais rallonge la durée de la présentation. C'est une prestation adaptée aux réunions en petit comité ou à l'accompagnement d'un petit groupe [26].

### 2.3. Traduction et interprétariat

Les traducteurs manient l'écrit. Ils ont peu de contacts avec les auteurs des textes qu'ils traduisent et avec les utilisateurs de leur travail, et ils disposent d'un certain temps pour produire leur traduction. Leur texte est appelé à durer ; il peut être publié, relu plusieurs fois, utilisé à plusieurs reprises, voire devenir une référence[22].

---

[21] Souvent par phases limitées à ¼ d'heure (ONU… ).

Quant aux interprètes, leur matière première est la parole : ils doivent trouver instantanément la formule juste car ils participent à un processus de communication destiné à un public présent [27].

### 2.4. La traduction est-elle indispensable ?

On ne peut y répondre que par un « oui » franc ! On ne saurait imaginer notre monde sans la traduction... Comment comprendre une autre culture sans la traduction ? Comment faire du commerce sans la traduction ? La traduction est indispensable aussi bien pour la littérature où, d'après Goethe lui-même, elle s'intègre dans la cadre de la « Weltliteratur », que pour le commerce et l'industrie, où elle permet échanges, progrès, innovation et développement [28].

Nul ne peut nier l'importance de la traduction dans le monde entier et dans tous les domaines du savoir : développement des langues et littératures nationales, dissémination du savoir, progrès scientifiques et techniques, développement et expansion des religions [29].

## 3. *Spécialisations possibles dans la Traduction*

### 3.1. Traducteur audiovisuel

Le traducteur audiovisuel intervient au cinéma ou à la télévision aussi bien pour le sous-titrage que pour le doublage. Il commence par visionner le film pour en traduire les dialogues ou les commentaires. Pour les sous-titres, il doit tenir compte des contraintes de temps et d'espace à l'écran. Pour le doublage, il doit faire « coller » le dialogue traduit avec le mouvement de lèvres originel de l'acteur. Un exercice délicat : la synchronisation labiale demande près d'une journée de travail pour dix minutes de film !

Les documentaires constituent toutefois la plus grande part du marché de la traduction audiovisuelle, notamment pour les voix off qui ne nécessitent pas de synchronisation [27].

---

[22] D'où des problèmes sans fin quand les diverses traductions d'une résolution de l'ONU ne sont pas strictement équivalentes, alors qu'elles sont censées l'être…

Voir aussi le cas de la chaîne d'information continue Euronews. Lors du lancement en 1993, Euronews diffuse en cinq langues : *français, allemand, anglais, espagnol et italien*

- en 1999, le *portugais* s'ajouté.
- En septembre 2001, le *russe* permet de répondre à l'extension de la chaîne à l'Est.
- Au 12 juillet 2008, Euronews diffuse également en *arabe* afin de couvrir une large population en Afrique du Nord, Moyen-Orient et en Europe.
- Le 30 janvier 2010, le *turc* s'ajoute aux huit premiers canaux audio.
- Le 27 octobre 2010, le *persan* permet de satisfaire les publics d'origine iranienne.
- Le 24 août 2011, l'*ukrainien* est lancé, elle devient la 11e langue offerte.
- Le 16 décembre 2011, le *polonais* est accessible en langue sélectionnable
- Depuis le 18 décembre 2012, la chaîne diffuse ses programmes en *grec* à destination des téléspectateurs en Grèce et Chypre.
- Le *hongrois* est lancé le 30 mai 2013.

### 3.2. Traducteur expert ou assermenté

Agréé par la Justice, le *traducteur expert ou assermenté* est chargé de la transcription textes administratifs (diplôme d'université, acte de mariage ou de naissance…) rédigés dans une langue étrangère. On le sollicite également pour traduire oralement les propos de personnes ne parlant pas les langues du pays et confrontées à l'appareil judiciaire [27].

*Toute erreur de traduction peut mener à une erreur judiciaire.*

### 3.3. Traducteur littéraire

La traduction littéraire est un exercice ardu qui exige de réels talents d'écriture pour faire ressortir tout le « sel » d'un auteur étranger, la musicalité de son écriture, le rythme de son phrasé, etc. Et malgré tout, le traducteur doit disparaître derrière l'œuvre et faire oublier au lecteur qu'il lit une traduction. Quelle exigence… [27].

Longtemps, l'école française a privilégié l'élégance de la traduction plutôt que la fidélité[23]. A partir de 1800, l'école allemande a insisté sur l'exactitude, au prix de lourdeurs considérables.

---

[23] Fausses traductions : Gil Blas de Santillane, de Lesage, faux roman espagnol du XVIII[ème] siècle, les romans policiers « traductions de Vernon Sullivan », directement écrits en français par Boris Vian.

### 3.4. Traducteur technique

80 % des traductions concernent des documents commerciaux, techniques ou scientifiques. Mais traduire en français un ouvrage anglais de médecine en ignorant tout du vocabulaire médical anglais occasionnerait des erreurs lourdes de conséquences. D'où l'importance du rôle du traducteur technique.

Spécialisé dans un ou plusieurs domaines techniques, il apporte toute sa connaissance du sujet qui lui permet des ajustements pragmatiques au-delà de la simple traduction littérale[24]. Les traducteurs techniques travaillent surtout pour l'édition et les entreprises [27].

Des hôpitaux ont administré des radiothérapies excessives (brûlures etc…), les notices des appareils n'ayant pas été traduites.[25]

## *4. La traductologie*

La traductologie est la discipline scientifique qui s'intéresse à la traduction. Assimilée à la linguistique dans les années 50 à 60, elle est devenue une discipline académique à part entière dans les années 80 et draine beaucoup de chercheurs. L'œuvre de Holmes[26] a été un tournant dans la reconnaissance et l'autonomie de cette discipline. Par le terme « *translation studies* », il désigne toutes les activités de recherche qui portent sur la traduction et sur le processus de traduction. Cette discipline lui paraît empirique, aussi tente-il une description du cadre général. Holmes distingue entre la traductologie pure et la traductologie appliquée.

La traductologie pure a deux objectifs : décrire le processus de traduction et les traductions elles-mêmes et établir des principes généraux permettant d'expliquer et de prévoir ces phénomènes, ce qui revient à la diviser en deux principales branches, à savoir la descriptive et la théorique.

Outre la traductologie pure, Holmes distingue la traductologie appliquée qui comprend l'enseignement de la traduction, la formation du traducteur et les critiques en traductologie.

---

[24] Cf. *La traduction scientifique et technique,* Jean Maillot, Paris : Eyrolles, 1969, aussi : *Manuel de russe à l'usage des scientifiques*, Gentilhomme, Paris : Dunod, 1963.

[25] L'Université de Sétif a été équipée très tôt d'un bon système Unix (1989) mais toute la documentation était en allemand…

[26] Cf. *Translated! :* Papers on Literary Translation and Translation Studies, J. S. Holmes, Rodopi B.V., Amesterdam, Atlanta, GA, 1988.

Les sous-branches de la traductologie pure (théorique et descriptive) et la traductologie appliquée sont intimement liées, bien qu'elles soient présentées comme des branches distinctes [30].

## 5. Les approches de la traduction humaine
[31]

Chaque approche explicative de la traduction se caractérise par une terminologie propre, des catégories spécifiques et une méthodologie distincte. L'application d'une approche particulière à la traduction peut être qualifiée en fonction du trait dominant : par exemple, l'approche *linguistique* ou *sémiotique* de la traduction, l'approche *sociologique* ou *sociolinguistique*, l'approche *philosophique*, *culturelle* ou encore *idéologique* du phénomène traductionnel.

On peut faire des distinctions au sein d'une même approche. Par exemple, l'approche linguistique se caractérise par le fait qu'elle envisage la traduction avant tout comme une opération d'essence verbale. Au sein de cette approche, il est possible de discerner le *modèle structuraliste* qui étudie les relations entre systèmes linguistiques, le *modèle textuel* qui s'intéresse aux situations communicatives dans les textes, le *modèle psycholinguistique* ou cognitiviste qui étudie le processus mental de la traduction, etc. Ces *modèles* délimitent le domaine de la traduction de façon différente, et chacun met en relief un aspect particulier de l'activité générale. Malgré leur divergences théoriques et méthodologiques, ces modèles doivent être perçus comme également pertinents et tout à fait complémentaires. Outre l'avantage de l'interdisciplinarité la conjonction de leur acquis ne peut qu'enrichir la traductologie [31].

Ce qui suit détaille chacune de ces approches, mais on donne davantage d'explications à l'approche linguistique, car elle inspire notre modèle de transfert tiré de cette approche linguistique de la traduction automatique.

### 5.1. Les approches linguistiques

Le développement de la traductologie au cours du XX$^{\text{ème}}$ siècle se dissocie lentement de celui de la linguistique. La traduction a beaucoup intéressé les linguistes qui ont appliqué diverses approches théoriques qui se sont succédé au cours du siècle : structuralisme, générativisme, fonctionnalisme, linguistique formelle, énonciative, textuelle, cognitive, sociolinguistique, psycholinguistique. Chaque courant est parti de ses propres postulats, employant des concepts différents pour étudier le phénomène de la traduction, sans jamais parvenir à l'appréhender dans sa complexité ni même dans sa globalité. Mais certaines approches ont été plus convaincantes que d'autres parce qu'elles ont capté des aspects essentiels de l'activité traductionnelle [31].

La linguistique s'intéresse aux langues et au langage, tandis que la traductologie s'occupe des traducteurs et des traductions, reproduisant le clivage entre savoir et savoir-faire.

### 5.1.1. L'approche fonctionnelle

Les approches fonctionnelles de la traduction sont essentiellement inspirées des travaux du linguiste britannique G.R. Firth.

Firth rejetait la conception du langage comme un simple code servant à transmettre l'information – c'est le cadre de la théorie de la communication à l'époque – et définissait plutôt le sens en terme de fonction relative à un contexte particulier.

Dans la perspective fonctionnaliste, le contexte revêt une importance cruciale et renvoie à un certain nombre d'éléments tels que *les actants*, *l'action*, *l'espace* et *le temps*, qui doivent être pris en considération pour saisir le sens du message[27].

L'un des premiers ouvrages à adopter une approche proprement linguistique de la traduction est (introduction à la théorie de la traduction) d'Andrei Fedorov [32]. Celui-ci cherche à mener une étude systématique de la traduction suivant un paradigme linguistique parce qu'il est convaincu que « toute théorie de la traduction doit être incorporé dans l'ensemble des disciplines linguistiques » [33].

D'autres auteurs ont la même conviction et s'évertuent à faire de la traduction un domaine parmi d'autre en recherche en linguistique, Vinay et Darbelnet publient leur fameuse *Stylistique comparée du français et de l'anglais* (Paris, Didier, 1958), que l'on tient pour la « première vraie méthode de la traduction fondée explicitement sur les apports de la linguistique » [33].

D'autres « méthodes » du même genre suivront, dans *Stylistique comparée du français et du l'allemand (méthode de traduction)* de Malblanc, et *Traité de stylistique comparée, analyse comparative de l'italien et du français* de P. Scavé et P. Intravaia [31].

### 5.1.2. L'approche stylistique comparée

La *Stylistique comparée du français et de l'anglais* de Vinay et Darbelnet est l'un des ouvrages qui « a le plus marqué les études de traduction » [33]. Dans cet ouvrage, les deux auteurs revendiquent le rattachement de la traductologie à la linguistique, mais ils ne se privent pas de faire appel à d'autres disciplines pour compléter leur approche de traduction stylistique comme la rhétorique et la psychologie.

---

[27] Cf. Tesnière.

L'objectif est de dégager « une théorie de la traduction reposant à la fois sur la structure linguistique et sur la psychologie des sujets parlants » [34]. Pour ce faire, ils s'efforcent de « reconnaître les voix que suit l'esprit, consciemment ou inconsciemment, quand il passe d'une langue à l'autre, et d'en dresser la carte ». A partir d'exemples, ils procèdent à l'étude des attitudes mentales, sociales et culturelles qui donnent lieu à des procèdes de traduction [31].

### 5.1.3. Les unités de traductions

Vinay et Darbelnet définissent *l'unité de traduction* comme le « plus petit segment de l'énoncé dont la cohésion des signes est tel qu'ils ne doivent pas être traduite séparément ».

A partir de cette définition les auteurs distinguent quatre types d'unité de traduction :

- les *unités fonctionnelles*, qui ont les mêmes fonctions grammaticales dans les deux langues ;
- les *unités sémantiques* qui possèdent le même sens ;
- les *unités dialectiques*, qui procèdent du même raisonnement ;
- les *unités prosodiques* qui impliquent la même intonation [34].

### 5.1.4. Procèdes de traduction [35]

**1. L'emprunt**

Procédé le plus simple, consistant à ne pas traduire et à laisser tel quel un mot ou une expression de la langue de départ dans la langue d'arrivée.

- pour des raisons d'usage : *the spectators said **'encore'** → bravo*
- ou d'absence d'équivalent : *let's go to the **pub** → allons au **pub***
- ou pour créer un effet rhétorique (couleur locale, humour etc.)

Particulièrement pratiqué lorsqu'il n'existe pas de terme équivalent dans la langue cible. Cela permet également de situer clairement un texte dans son contexte culturel par l'intermédiaire du registre de vocabulaire utilisé.

Exemples :
- *Weight Watchers*
- *une rave [rêve]*

- *une after*
- *the Bibliothèque Nationale*
- *the gendarmes*

**2. Le calque**

Le calque traduit littéralement le mot ou l'expression de la langue de départ. C'est une « copie » de l'original, un emprunt qui a été traduit.

Exemples :
- The United States of America : *Les États-Unis d'Amérique*
- the Cold War : *la Guerre Froide*
- AIDS : *SIDA*
- World Health Organization : Organisation Mondiale de la Santé

Voir aussi certaines expressions courantes au Québec telles que : *tomber en amour, chiens chauds, chars usagés…* tous directement calqués de l'anglais.

Certains calques à partir de l'anglais sont acceptés en français :
- *ce n'est pas ma tasse de thé*
- *développer un procédé*
- *être dans le rouge*
- *contrôler la situation.*

D'autres peuvent être considérés comme fautifs[28]
- *there are no other alternatives* → *il n'y a pas d'autres choix/ possibilités.*

Le calque ne doit être utilisé qu'avec précaution car il conduit très facilement à des contresens ou même des non-sens, fautes très graves en traduction.

**3. La traduction littérale**

Procédé qui consiste à traduire la langue source mot à mot, sans effectuer de changement dans l'ordre des mots ou au niveau des structures grammaticales et tout en restant correct et idiomatique.

Exemples :
- avaler la pilule : *to swallow the pill*

---

[28] Cf. Etiemble, *Parlez-vous franglais ?*

- avoir un mot sur le bout de la langue : *to have a word on the tip of the tongue*
- tirer à sa fin : *to draw to an end*
- voir rouge : *to see red*

Les obstacles liés à la traduction littérale sont nombreux et elle n'est pas recommandée dans la traduction académique. Elle ne fonctionne parfaitement que très rarement !

## 4. La transposition

Procédé qui entraîne un changement de catégorie grammaticale d'un mot en passant d'une langue à l'autre. La table 2.1 expose quelques exemples de traductions adoptant le procédé de transposition.

Table 2.1 Exemples de traduction utilisant le procédé de transposition

| Langue de départ | Langue d'arrivée |
|---|---|
| **Nom** *the hour of indulgence…* | **Verbe** *le moment de se faire plaisir* |
| **Nom** *at some level of consciousness* | **Adverbe** *plus ou moins consciemment* |
| **Adjectif** *festival paper* *medical students* | **Nom** *papier-cadeau* *étudiants en médecine* |
| **Adjectif** *endless* | **Verbe** *qui ne s'arrête jamais, qui n'en finit pas* |
| **Verbe** *to bleep* *for sale* | **Nom** *le beep* *à vendre* |
| **Adverbe** *with a certain testy reluctance* | **Nom** *avec une certaine réticence et quelque irritation* |
| **Préposition** *into a shallow rippled expanse* *driving through the city* *hoping…* | **Verbe** *pour former une étendue peu profonde et ridée* *traversant la ville en voiture* *dans l'espoir…* |

La transposition doit être utilisée lorsque la traduction littérale n'a aucun sens, entraîne une erreur de traduction, ou est incompréhensible (problème de structure). Si la traduction n'est ni authentique ni idiomatique, on doit avoir recours à la transposition.

**5. La modulation**

Procédé impliquant un changement de point de vue afin d'éviter l'emploi d'un mot ou d'une expression qui passe mal dans la langue d'arrivée. Il permet aussi de tenir compte des différences d'expression entre les deux langues : passage de l'abstrait au concret, de la partie au tout, de l'affirmation à la négation, évacuation des formes passives…

Exemples :

1. les occupations auxquelles il passe la plus grande partie de ses **heures** → *the occupations that take up most of his **day***
2. le **milieu** avec lequel il est **en contact** → *the **circles** in which he **moves***
3. vu son **attitude** → *in view of his **behavior***
4. *café **soluble*** → ***instant** coffee*
5. avoir du pain sur la **planche** → *to have a lot on one's **hands***

**6. L'équivalence**

Procédé consistant à traduire un *message* dans sa globalité (surtout utilisé pour les exclamations, les expressions figées ou les expressions idiomatiques). Le traducteur doit comprendre la situation dans la langue de départ et doit trouver l'expression équivalente appropriée et qui s'utilise dans la même situation dans la langue d'arrivée. C'est une rédaction du message entièrement différente d'une langue à l'autre.

Exemples

- What's up? → *Quoi de neuf ?*
- Mind your own business. → *Occupe-toi de tes oignons.*
- Aïe ! → *Ouch !*
- Formidable ! → *Great!*
- C'est pas vrai ? → *No kidding?*
- Attention à la peinture. → *Wet paint.*
- Fermeture pour cause de travaux → *Closed for renovation.*
- L'Hexagone → *France.*
- Les personnes du troisième âge. → *Senior citizens.*

**7. Les collocations**

Les mots se marient et forment des couples indissociables, l'un entraînant *automatiquement* l'autre. Le moindre changement risque de provoquer une gêne à la lecture d'un texte traduit, qui manque alors de naturel et d'authenticité. Il faut au contraire profiter de cette dynamique qui relie les mots d'une langue selon des relations privilégiées toujours identiques.

Être conscient de l'existence de ces collocations et savoir les manipuler avec habileté permet de bien mieux traduire un texte et de le rendre bien plus authentique dans la langue d'arrivée.

- …bottles that were kept for Occasions → *les bouteilles qu'on réservait pour les grandes occasions*
- il n'a pour tout bagage… → *all he has in the way of baggage*
- il n'en a pas la tête → *he doesn't look like one*
- il se mettait en quatre → *he bent over backward.*

**8. L'étoffement**

Généralement consiste à traduire une préposition, un pronom ou un adverbe interrogatif anglais par un syntagme verbal ou nominal en français. L'anglais est en général plus abstrait que le français qui lui nécessite l'utilisation de ce procédé plus systématiquement.

- ***off** the motorway, problems arise for the motorist* → **lorsqu'il quitte** l'autoroute...
- *the wreck **off** Land's End* → *l'épave **au large de** Land's End*

Il est souvent utile et même parfois indispensable d'ajouter une précision en traduisant afin d'obtenir le même effet que dans la langue de départ. L'étoffement permet également de parvenir à une formulation plus authentique que la simple traduction littérale.

- to sit to her meal → *s'asseoir pour prendre son repas :* la phrase complète serait *to sit and have her meal*, l'étoffement obligatoire redonne le verbe sous-entendu dans une expression très usuelle.

### 5.1.5. L'approche linguistique théorique

Dans *Les problèmes théoriques de la traduction,* Georges Mounin consacre la linguistique comme cadre conceptuel de référence pour l'étude de la traduction. Le point de départ de sa réflexion est que la traduction est un « contact de langues, un fait de bilinguisme » [36].

Son souci premier est la scientificité de la discipline, ce qui le conduit à poser une question lancinante pour l'époque, « l'étude scientifique de l'opération traduisant doit-elle être une branche de la linguistique ? » [36].

En réalité, l'objectif de Mounin est de faire accéder la traductologie au rang de « science » mais il ne voit pas d'autres possibilités que de passer par la linguistique. C'est pourquoi « il revendique pour l'étude scientifique de la traduction le droit de devenir une branche de la linguistique » [37].

### 5.1.6. L'approche linguistique appliquée

La linguistique appliquée est une branche de la linguistique qui s'intéresse davantage aux applications pratiques de la langue qu'aux théories générales sur le langage. Pendant longtemps, la traduction a été perçue comme une chasse gardée de la linguistique appliquée. L'exemple type de cette approche est le livre de Catford intitulé (*A Linguistic theory of translation*), dont le sous-titre est sans ambiguïté quant à la nature de l'approche : (*essay in applied linguistics*).

Catford affirme son intention de se focaliser sur l'analyse de ce que la traduction est afin de mettre en place une théorie qui soit suffisamment générale pour être applicable à tous les types de traductions [31].

Pour Catford, la traduction est une opération réalisée sur les langues : un processus de substitution d'un texte dans une langue par un texte dans une autre langue [38].

### 5.1.7. L'approche sociolinguistique

La sociolinguistique étudie la langue dans son contexte social à partir du langage concret. Apparue dans les années 1960 aux États Unis sous l'impulsion de Labov, Gumperz et Hymes, elle a bénéficié de l'apport de la sociologie. Parmi ses centres d'intérêts, on trouve les différences socioculturelles et l'analyse des interactions, mais aussi les politiques linguistiques et l'économie de la traduction [31].

Dans *Les fondements sociolinguistiques de la traduction,* Maurice Pergnier s'interroge sur la nature de la traduction en mettant exergue le caractère ambigu du terme même : « le phénomène recouvert par le terme de traduction ne comporte pas, en dépit des apparences, de frontières nettes et bien définies » [39].

### 5.2. L'approche Herméneutique

L'herméneutique est un mot forgé à partir du grec « *Hermêneuein* » qui signifie à l'origine « comprendre, expliquer », mais qui a fini par désigner un courant et une méthode d'interprétation initiée par les auteurs romantiques allemands. Le principal promoteur de cette méthode dans le domaine de la traduction est Friedrich Schleiermacher.

Pour lui la traduction doit être fondée sur un processus de compréhension de type empathique, dans lequel l'interprétant se projette dans le contexte concerné et s'imagine à la place de l'auteur pour essayer de ressentir ce qu'il a senti et réfléchir comme lui [31].

### 5.3. Les approches idéologiques

L'idéologie est un ensemble d'idées orientées vers l'action politique. L'approche idéologique a connu un essor important dans le sillage du courant culturaliste, qui a mis les études sur les rapports de pouvoir au centre de ses préoccupations. Le domaine de la traduction a été maintes fois analysé suivant ce paradigme particulier. Plusieurs questions ont été posées à ce sujet : la traduction est-elle motivée idéologiquement ? Comment séparer notre vision du monde de l'idéologie qui peut entacher la traduction ? La traduction est-elle toujours idéologique ?

Berman fait une distinction entre les *traductions ethnocentriques* qui mettent en avant le point de vue de la cible (langue d'arrivée), et les *traductions hypertextuelles*, qui privilégient les liens implicites entre les textes des différentes cultures.

De son coté, Penrod distingue deux grandes tendances idéologiques : la *naturalisation* des éléments contenus dans la traduction et *l'exotisation* qui préserve les éléments originaux tels quels.

En réalité, derrière l'approche idéologique profile le vieux débat sur la *fidélité* à la source, lequel débat oppose la *traduction littérale* à la *traduction libre* [31].

### 5.4. L'approche poétologique

La poétique est l'étude de l'art littéraire en tant que création verbale. Ainsi, Tzvetan Todorov distingue trois grandes familles de théories de la poésie dans la traduction occidentale :

- le premier courant développe une conception rhétorique qui considère la poésie comme un ornement du discours, un *plus* ajouté au langage ordinaire ;
- le deuxième courant conçoit la poésie comme l'inverse du langage ordinaire, un moyen de communiquer ce que celui-ci ne saurait traduire ;
- le troisième met l'accent sur le jeu du langage poétique qui attire l'attention sur lui-même en tant que création davantage que sur le sens qu'il véhicule.

Certains traductologues ont mis cette problématique au centre de leur réflexion. Ainsi, dans (*Un art en crise*), Efim Etkind estime que la traduction poétique passe par une crise profonde dont il essaie de comprendre les causes.

Il existe, en effet, en matière de traduction poétique, deux grands courants représentés par deux poètes majeurs de la littérature française : Charles Baudelaire et Paul Valéry.

Pour Baudelaire, il n'est pas possible de traduire la poésie autrement que par de la prose rimée. A l'inverse, pour Valéry, il ne suffit pas de traduire le sens poétique ; il faut tenter de rendre la forme jusque dans la prosodie : « s'agissant de poésie, la fidélité restreinte au sens est une manière de trahison[29] » [31].

Etkind propose de ne pas se focaliser sur un aspect en particulier du poème, ni sur le sens, ni sur les sons, ni sur les images. Il faut simplement prendre conscience que « le texte forme un tout et le traducteur doit absolument redonner à ce tout, dans sa propre langue, sa fonction, en respectant la forme et la pensée » [40].

### 5.5. L'approche textuelle

L'approche textuelle postule que tout discours peut être « mis en texte ». Qu'il s'agisse d'une interaction orale ou écrite, le résultat est le même : c'est un texte qui possède des caractéristiques propres et un sens précis. Il en découle que toute traduction est censée être précédée d'une analyse textuelle, au moins au niveau typologique, pour assurer la validité de la compréhension – et donc de l'interprétation – qui s'ensuit. Mais il existe plusieurs perspectives d'étude du texte, ce qui rend l'analyse traductologique compliquée : *le type, la fonction envisagée, la finalité, le sens, le contexte, l'idéologie* du texte déterminent la traduction.

Dans les domaines de spécialités, l'analyse du discours sert notamment à montrer le marquage culturel de la terminologie. Ainsi, la traduction d'un ouvrage ou d'un article de médecine du français vers l'arabe nécessitera, par exemple, le passage d'une manière abstraite de penser et d'écrire à une manière plus concrète et plus pratique, une variété de modalités et de registres différents, un choix de concepts et de métaphores médicales plus adaptés à la culture cible.

Les métaphores apparaissent comme des marqueurs de visions culturelles et de points de vue idéologiques, marqueurs qui forment un réseau de signification incontournable lors de la traduction. Car il ne s'agit pas simplement de procédés décoratifs du texte, mais de véritables déclencheurs d'effets chez le récepteur. Donc, on peut redéfinir le rôle du traducteur, comme un médiateur culturel avant tout.

Le linguiste canadien Robert Larose a analysé les éléments constitutifs des discours sur la traduction, en particulier ceux de Vinay et Darbelnet, Mounin, Nida, Catford, Steiner, Delisle, Ladmiral et Newmark. Cette étude comparative met en évidence à la fois les qualités et les limites des titres qu'elle passe en revue, mais il s'agit d'une synthèse orientée vers la conceptualisation, en ce sens que Larose vise à proposer, à travers cet exposé, son propre modèle explicatif de la traduction [31].

---

[29] « Tradittore, trattore » (tout traducteur est un traître – proverbe italien).

### 5.6. Les approches sémiotiques

La sémiotique[30] est l'étude des signes et des systèmes de signification. Elle s'intéresse aux traits généraux qui caractérisent ces systèmes quelle que soit leur nature : verbale, picturale, plastique, musicale, etc.

Le principe est qu'une comparaison des systèmes de signification peut contribuer à une meilleure compréhension du sens en général.

Pour Peirce, le processus de signification est le résultat de la coopération de trois éléments : un *signe*, son *objet* et son *interprétant*. Aussi, d'un point de vue sémiotique, toute traduction est envisagée comme une forme d'interprétation qui porte sur des textes ayant un contenu encyclopédique différent et un contexte socioculturel particulier.

La *Sémio-traductologie* analyse les traductions portant sur des signes verbaux et non verbaux.

Gorlée insiste sur le rôle capital de l'interprétant-traducteur. Celui-ci doit être à la fois l'interprète du *signifiant* dans le texte source et l'énonciateur du *signifié* en langue cible.

Dans cette perspective, la notion d'équivalence occupe une place centrale. Elle est définie comme une identité à travers des codes : ainsi, deux signes sont équivalents dans la mesure où ils déterminent un même signifié[31] [31].

### 5.7. Les approches communicationnelles

Les approches communicationnelles sont nées de la focalisation des linguistes sur la fonction du langage humain. Dès le début du XX[ème] siècle, Ferdinand de Saussure (en son *Cours de Linguistique Générale*, Genève : Payot, 1913/1995) distingue la parole que nous produisons pour communiquer, de la langue qui est un ensemble de mots présents dans le cerveau des locuteurs. Conçu en ces termes, le langage n'a dans la communication humaine qu'une fonction utilitaire : par exemple, dans la théorie de Shannon et Weaver, il est un code parmi d'autres qui sert à transmettre l'information entre deux individus.[32]

---

[30] Cf. Umberto Eco, *Sémiotique et Philosophie du Langage,* PUF 1988, trad. de *Semiotica e filosofia del linguaggio*, 1984, Turin : Einaudi.

[31] Ces questions sont notamment critiques en matière de signalisation s'adressant à une population multiculturelle : signalisation routière, signalisation dans les stations multi-modales, rencontres olympiques….

Dans cette optique, la communication est analysée en termes d'encodage et de décodage portant sur un message particulier. L'encodage renvoie aux informations que le locuteur met dans son message et le décodage renvoie à la compréhension du récepteur de ce même message ; l'un encode, l'autre décode, de façon quasi mécanique pour ainsi dire.

Cette conception simpliste et binaire fait que le traducteur est perçu comme un simple décodeur du message original et un réencodeur du message final. Il doit se contenter de relayer le message en apportant le minimum de modifications, i.e. qui servent uniquement à prédire le sens dans la langue cible.[33]

Cette idée de la communication est appliquée à la traduction pour la première fois par Nida. Celui-ci propose de concentrer le travail du traducteur sur les informations prédictibles entre deux langues. Le traducteur aurait ainsi pour tâche principale de compenser le bas niveau de prédictibilité de certains messages [41]. Cette compensation peut être requise pour des raisons linguistiques telles que l'existence d'un ordre des mots inhabituel ou d'une expression peu familière. Elle peut l'être également pour des raisons culturelles telles que l'absence de certaines notions, genres textuels ou mêmes objets de la vie courante [31].

**L'approche pragmatique**

La pragmatique est l'étude du langage du point de vue de sa *praxis*, c'est-à-dire des finalités et des conditions de son utilisation.

Son champ d'investigation privilégié concerne les actes de langage, c'est-à-dire les expressions impliquant une action telles que les ordres, les requêtes, les excuses ou encore les compliments ; ou toute expression langagière qui produit un effet.

Pour décrire ce type d'expressions, Austin a défini trois catégories d'actes de langage (*locution*, *illocution*, *perlocution*) qui ont été mises à profit pour l'étude du processus de traduction et d'interprétation. Baker a exploité cette approche qui vise à produire dans la langue cible des actes « locutoires », ayant la même force « perlocutoire » que ceux de la langue source. Hickey a également appliqué cette approche à la traduction, mais de façon plus systématique et sur une échelle plus large.

---

[32] Un langage est plus qu'un code : c'est un ensemble de suites structurées de codes.
[33] Cette idée est d'autant plus simpliste qu'on doit passer d'une langue VSO (à verbe préfixé) à une langue SOV (à verbe postfixé…)

L'intérêt principal de l'approche pragmatique pour la traductologie est qu'elle permet de mettre en relief les éléments les plus saillants de la communication dans un texte ou dans un discours particulier [31].

### 5.8. Les approches cognitives

Les sciences cognitives s'intéressent aux processus mentaux qui sont mis en œuvre dans les différentes activités humaines. De ce point de vue, la traduction est envisagée comme un processus de compréhension et de reformulation du sens entre deux langues, intégrant un traitement particulier de l'information.

Il fallait recourir à une discipline qui puisse aborder à la fois la psychologie de l'humain et le fonctionnement du langage. C'est pourquoi la discipline phare qui illustre aujourd'hui l'approche cognitive est la psycholinguistique. Celle-ci étudie la manière de communiquer et de gérer les informations par un être humain au sein d'une langue, et postule que la traduction est une forme de communication bilingue.

D'un point de vue psycholinguistique, ces formes de traduction engagent quelques activités mentales de base (*lire*, *écouter*, *écrire*, *parler*) qui sont soumises à des contraintes spécifiques et qui utilisent des ressources cognitives particulières lors de la traduction. Ainsi par exemple, l'interprète de conférence doit écouter et parler en *temps réel* pour ainsi dire, mais cette contrainte temporelle ne pèse pas de la même façon sur le traducteur de l'écrit [31].

## *Conclusion*

Ce chapitre a donné un aperçu sur la traduction humaine, son histoire, ses spécialités et ses différentes approches, en insistant davantage sur l'approche linguistique, cadre de notre contribution à la traduction automatique.

La traduction automatique a suscité l'ambition des chercheurs de différents domaines commençant par la traduction elle-même, l'intelligence artificielle, la linguistique computationnelle, le TALN et les statistiques. Différentes approches sont issues de ces efforts, détaillées dans le prochain chapitre.

# Chapitre 3. Les approches linguistiques de la traduction automatique

## *Introduction*

Trois grandes tendances de la recherche actuelle dans la TA peuvent être identifiées. La première est l'exploitation des techniques actuelles de la linguistique computationnelle pour éclairer la relation traductionnelle entre deux textes. La seconde, l'utilisation des ressources existantes de toutes sortes, que ce soit pour en extraire des informations utiles ou directement en tant que composants dans les systèmes. La troisième, la tendance vers des modèles statistiques ou empiriques de la traduction.

Bien que nous insistions au long de cette thèse sur les méthodes linguistiques, beaucoup de travaux récents préconisent une combinaison de techniques, avec des méthodes statistiques.

## *1. Les approches linguistiques de la TA*

L'approche linguistique de la TA regroupe trois familles : *l'approche directe*, *l'approche de transfert*, *l'approche à langue-pivot*.

Dans les approches linguistiques, on a recours à des dictionnaires bilingues, avec des structures spécifiques [42].

Dans la traduction *directe*, on procède mot-à-mot dans le texte source. La traduction *directe* utilise un grand dictionnaire bilingue. Le programme traduit mot-à-mot.

Dans les approches de *transfert*, d'abord on analyse le texte d'entrée, puis on applique des règles pour transformer la structure syntaxique de la phrase source vers une structure syntaxique de la langue cible. Ensuite à partir de cette structure on génère la phrase en langue cible.

Dans les approches *à langue-pivot*, on analyse le texte en langue source en une représentation abstraite, appelée *interlingua*[34] ou *langue-pivot*. A partir de cette représentation on génère ensuite un texte dans la langue cible.

Une façon courante de visualiser ces trois approches est le *triangle de Vauquois* représenté dans la figure 3.1. Le triangle montre la profondeur croissante de l'analyse requise (à la fois analyse et génération) pour passer de l'approche *directe* par les approches de *transfert*, à l'approche *à langue-pivot*.

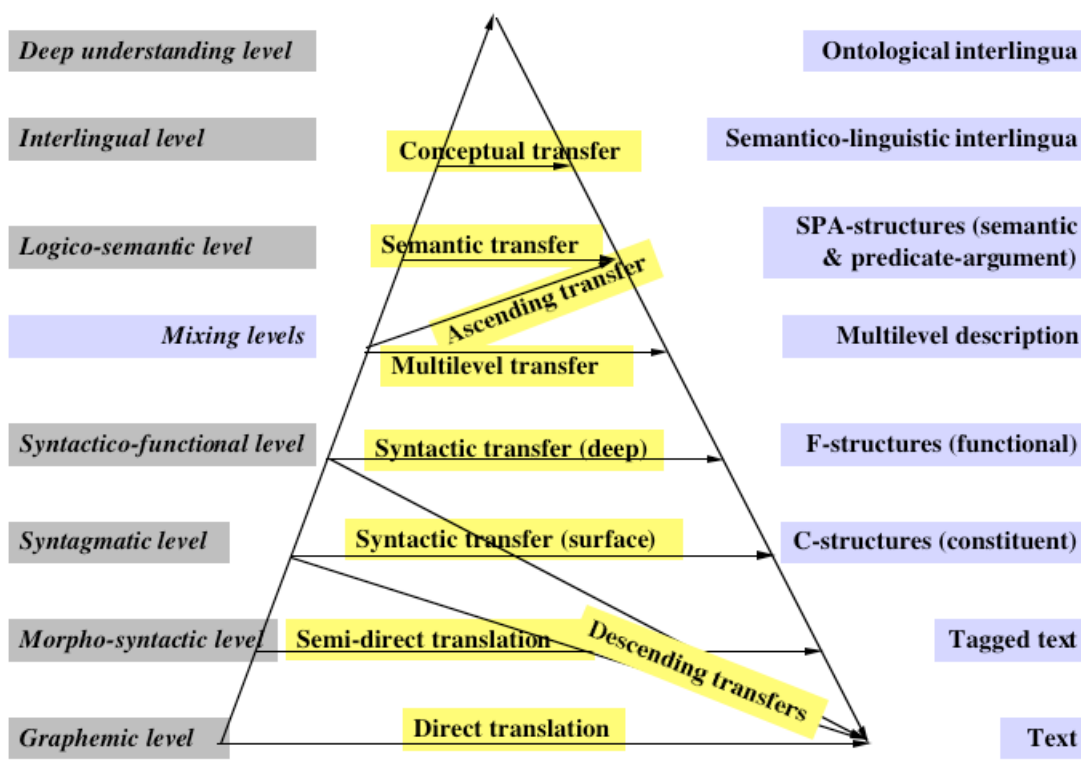

**Figure 3.1 Triangle de Vauquois [43]**

En outre, le triangle indique la quantité décroissante de transfert de connaissances nécessaire quand nous nous dirigeons vers le haut de triangle :

- niveau *direct* : toute connaissance est le transfert de connaissances pour chaque mot.
- via le *transfert* : règles de transfert par des arbres d'analyse.
- à travers *la langue-pivot* : pas de transfert de connaissances spécifiques.

---

[34] Interlingua peut désigner 3 langues artificielles :
- langue élaborée par l'International Auxiliary Language Association (IALA)
- langue construite en 1922 par Edgar de Wahl
- le *latin sans flexions*, langue inventée en 1903 par le mathématicien italien Giuseppe Peano.

La plupart des systèmes à base de règles de *transfert* ou *à langue-pivot* sont basés sur l'idée que la réussite de la TA consiste à définir un niveau de représentation des textes suffisamment abstrait pour simplifier la traduction, mais en même temps suffisamment étendu pour permettre aux différentes phrases de la langue source ou cible d'être mise en correspondance avec ce niveau de représentation i.e. une TA réussie implique un compromis entre la profondeur de l'analyse (compréhension) du texte source et le besoin de calculer une représentation abstraite [8].

Donc, l'idée est d'avoir un formalisme pivot très régulier, avec des atomes en nombre suffisants pour une bonne couverture sémantique/pragmatique/glossématique. Alors, en prenant la précaution d'une paire initiale de langues très éloignées pour un bon formalisme, on peut ensuite greffer d'autres analyseurs et d'autres générateurs, et passer d'un traducteur 1 : 1 à un système traducteur k : n, au profit d'un effort en k+n plutôt qu'en k*n.

### 1.1 L'approche directe

L'approche *directe* est la forme la plus simple de la TA, Elle consiste à retrouver la forme de base de chaque mot par le biais d'une analyse morphologique (lemmatisation). Cette forme de base du mot est recherchée dans un dictionnaire. Ensuite le mot correspondant dans la langue cible est inséré dans le texte.

Les systèmes *directs* donc, n'utilisent que deux représentations, le texte d'entrée et le texte de sortie.

Pour les langues ayant des systèmes d'écriture à séparateurs de mots ou de syllabes, le texte d'entrée n'est souvent pas strictement le flot de caractères tel quel, mais une suite de « mots typographiques » séparés grâce à des règles simples. Les systèmes *semi-directs* ont une phase de segmentation ou d'analyse morphologique, voire morphosyntaxique, et une phase de génération morphologique [44].

Dans la traduction *directe* (figure 3.2), on procède mot-à-mot dans le texte source. Nous n'utilisons pas de structures intermédiaires, sauf pour l'analyse morphologique peu profonde ; chaque mot source est directement mis en correspondance avec un mot cible. La traduction *directe* est donc basée sur de gros dictionnaires bilingues comprenant des règles de correspondance ; Après la traduction des mots et des expressions figées, des règles simples de ré-ordonnancement peuvent s'appliquer, par exemple déplacer les adjectifs après les noms lors de la traduction de l'anglais vers le français.

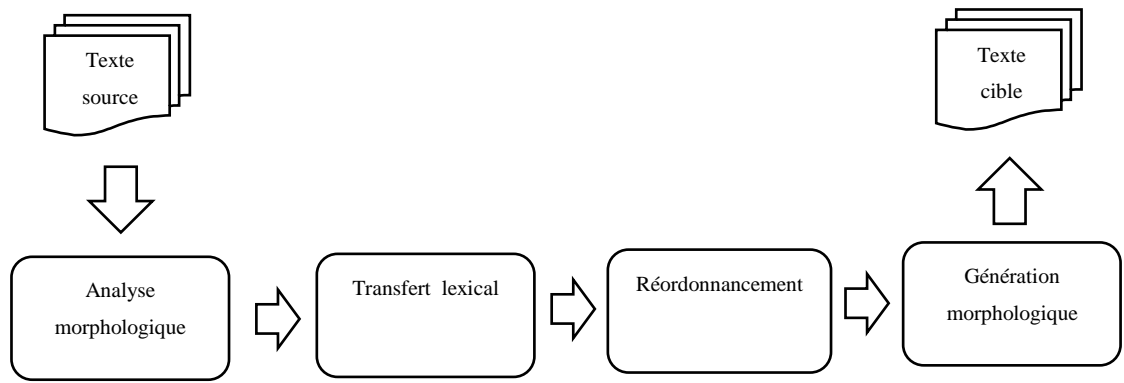

**Figure 3.2 Traduction automatique directe**

Examinons un système direct simplifié sur un exemple[35], traduisant de l'anglais vers l'espagnol dans la figure 3.3.

> Mary didn't slap the green witch
>
> *Maria no di´o una bofetada a la bruja verde*
>
> Mary not gave a slap to the witch green

**Figure 3.3 Exemple de traduction directe de l'anglais vers l'espagnol [13]**

Les quatre étapes décrites dans la Figure 3.2 procéderont comme indiqué sur la figure 3.4.

> Entrée: Mary didn't slap the green witch
>
> Étape 1: Morphologie Mary DO-PAST not slap the green witch
>
> Étape 2: Transfert lexical: Maria PAST no dar una bofetada a la verde bruja
>
> Étape 3: Ré-ordonnancement local: Maria no dar PAST una bofetada a la bruja verde
>
> Étape 4: Morphologie: Maria no abofeteó a la bruja verde

**Figure 3.4 Un exemple de traitement dans un système direct [13]**

L'étape 2 suppose que le dictionnaire bilingue ait l'expression (*dar una bofetada a*) comme traduction en espagnol du mot (*slap*). Le ré-ordonnancement local de l'étape 3 aurait besoin de changer l'ordre adjectif-nom de (*green witch*) à (*bruja verde*) avec une combinaison de règles d'ordonnancement. Le dictionnaire traitera de la négation et du passé en anglais du mot (*didn't*). Ces entrées de dictionnaire peuvent être complexes [13].

---

[35] La majorité des exemples de ce chapitre sont tirés de (Daniel Jurafsky & James H, *Speech and Language Processing*, chapter 24: machine translation).

Un échantillon d'entrée du dictionnaire d'un des premiers systèmes directs Anglais-Russe est représenté sur la Figure 3.5.

---

**Function** DIRECT TRANSLATE MUCH/MANY (word) **returns** Russian translation

    **if** preceding word is *how* **return** *skol'ko*

    **else if** preceding word is *as* **return** *stol'ko zhe*

    **else if** word is *much*

    **if** preceding word is *very* **return** nil

    **else if** following word is a noun **return** *mnogo*

    **else** /* word is many */

    **if** preceding word is a preposition and following word is a noun **return** *mnogii*

    **else return** *mnogo*

---

**Figure 3.5 Une procédure pour traduire (much) et (many) en russe [45]**

Bien que l'approche *directe* puisse traiter cet exemple simple, et gérer un ré-ordonnancement par mot, elle n'a pas de composante d'analyse syntaxique, voire aucune connaissance sur la structure grammaticale ni de la langue source ni de la langue cible. Elle ne peut pas donc gérer de manière fiable un ré-ordonnancement complexe, ou des phrases avec des structures complexes comme l'exemple de la figure 3.6 [13].

> « Il rajusta son col et son gilet de velours noir *sur lequel* se croisait plusieurs fois une de ces grosses chaînes d'or fabriquées à Gênes ; puis, après avoir jeté par un seul mouvement sur son épaule gauche son manteau doublé de velours en le drapant avec élégance, il reprit sa promenade sans se laisser distraire par les œillades bourgeoises qu'il recevait. »
>
> — Balzac, Gambara

**Figure 3.6 Exemple de texte avec des structures complexes**

Cela peut se produire même dans des langues similaires à l'anglais, comme l'allemand, où des adverbes comme (*heute*) (aujourd'hui) apparaissent dans des positions différentes. Le sujet par exemple, (*Hexe*) peut apparaître après le verbe principal, comme le montre la figure 3.7.

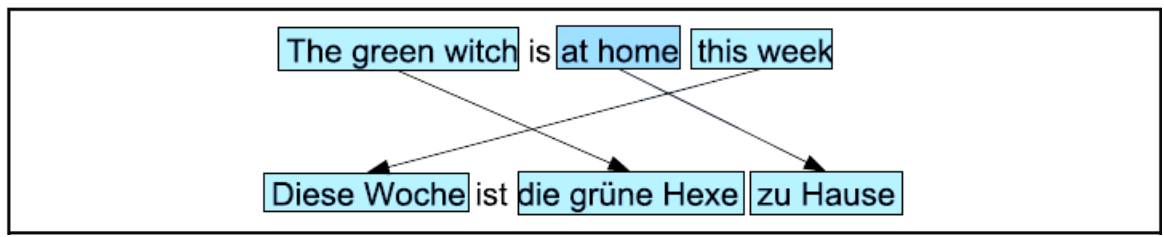

**Figure 3.7 Ré-ordonnancement complexe lors de la traduction de l'anglais vers l'allemand [13]**

L'Allemand met souvent l'adverbe en position initiale alors que dans l'anglais, il serait plus naturellement mis après. Les verbes conjugués apparaissent souvent en deuxième position dans la phrase, ce qui provoque l'inversion du sujet et du verbe. De plus, les temps composés sont éclatés : l'auxiliaire reste central, le participe est rejeté en fin de phrase [13] comme illustré dans l'exemple de la figure 3.8.

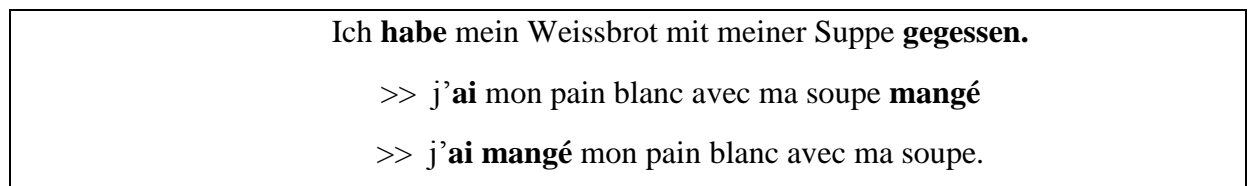

**Figure 3.8 Exemple d'éclatement des temps composés dans l'allemand [13]**

Enfin, des ré-ordonnancements plus complexes se produisent lorsqu'on traduit des langues SVO à des langues SOV [13], comme nous le voyons dans l'exemple de la figure 3.9.

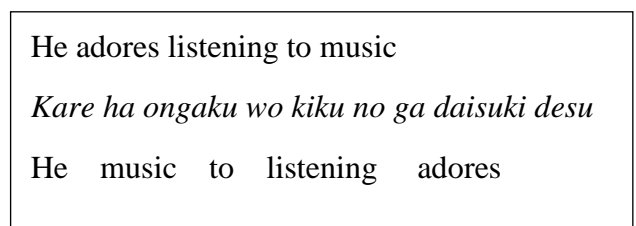

**Figure 3.9 Exemple de traduction de l'anglais vers le japonais [46]**

Ce genre de problème surgit au moins chaque fois que la traduction doit changer de type syntaxique, les langues se répartissant en 6 types principaux.

Soient V, S, O trois symboles représentant Verbe, Sujet et Objet, pouvant former 6 types syntaxiques. D'après Wikipédia, on a la typologie présenté dans la table 3.1.

**Table 3.1 Typologie syntaxique**

| Type | phrase-type | langues (ex.) | % langues |
|---|---|---|---|
| **SOV** | le chat la souris mange | japonais, turc, coréen… | 41% |
| **SVO** | **le chat mange la souris** | anglais, français, swahili, langues chinoises… | 39% |
| **VSO** | mange le chat la souris | arabe classique, langues celtiques, hawaïen... | 15% |
| **VOS** | mange la souris le chat | fidjien, malgache... | 5% |
| **OSV** | la souris le chat mange | Xavánte… | |
| **OVS** | la souris mange le chat | Hixkaryana… | |

Les exemples montrent que l'approche *directe* est trop axée sur des unités individuelles (mots). Ce qui la rend très utile pour des applications où on a affaire juste à des mots et non pas à des phrases comme pour la recherche d'information multilingue (voir chapitre 4). Pour faire face à des phrases cohérentes, nous aurons besoin d'ajouter des connaissances structurelles dans nos modèles de TA.

### 1.2 L'approche de transfert

Comme illustré dans la Section 1, les langues diffèrent structurellement. Une stratégie pour réussir la TA est de traduire par un processus qui surmonte ces différences, En modifiant la structure de l'entrée pour la rendre conforme aux règles de la langue cible. Cela peut être fait en appliquant des connaissances sur les différences structurelles entre les deux langues. Les systèmes qui utilisent cette stratégie sont appelés modèle à base de *transfert*. Le modèle de *transfert* présuppose une analyse de la langue source, suivie par une phase de génération pour générer la phrase de sortie. Ainsi, sur ce modèle, la TA comporte trois phases : l'*analyse,* le *transfert* et la *génération*, où le transfert comble l'écart entre la sortie de l'analyseur de la langue source et l'entrée du générateur de la langue cible (figure 3.10).[36]

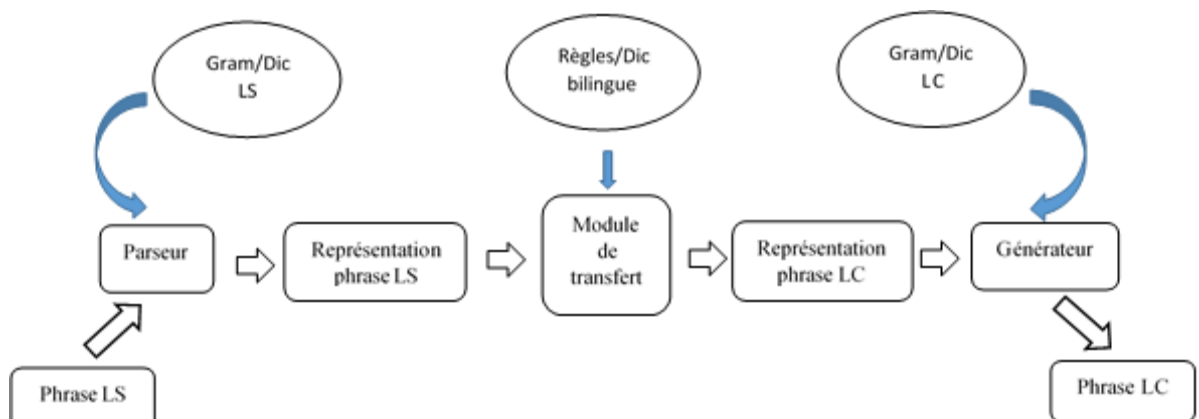

**Figure 3.10 Approche de transfert**

---

[36] Modèle proposé par Victor Yngve, MIT, 1957 ; implanté après 1970.

Le contenu de la phrase analysée est enregistré sous une forme intermédiaire symbolique la plus neutre possible sous forme d'un arbre syntaxique à partir de laquelle la phrase cible va être générée selon des règles linguistiques. Le résultat de la traduction peut être satisfaisant pour les textes ayant une structure syntaxique simple et un dictionnaire personnalisé.

Dans la mesure où une langue possède une morphologie assez riche, c'est un modèle qui s'impose pour toute étude visant le traitement automatique des langues. Faute de quoi, on serait obligé de construire des dictionnaires gigantesques pour reconnaître toutes les formes de mots.

Le niveau syntaxique est utilisé comme niveau de *transfert* pour établir une correspondance entre des sous-arbres, correspondants à des groupes syntaxiques dans la langue source et dans la langue cible.

On substitue ainsi, par morceaux, des sous- arbres (éléments de formule de la structure syntaxique en langue cible) à ceux qui ont été obtenus par l'analyse syntaxique de la langue source jusqu'à la création de la structure syntaxique complète en langue cible.

Il reste alors à remplacer les mots du texte initial par leurs équivalents (appartenant à la classe syntaxique exigée par la nouvelle structure) dans la langue cible et à réaliser la construction du texte traduit au moyen d'un modèle de synthèse [47].

Il existe au moins sept variantes des systèmes de *transfert*. La structure obtenue en fin d'analyse peut être syntagmatique, ou bien dépendancielle, et dans ce cas surfacique (fonctions syntaxiques : sujet, objet direct, épithète, attribut…) ou profonde (relations sémantiques : agent, patient, cause, concession…). Les systèmes de *transfert profond* fondés sur les théories de Tesnière, puis de l'Ecole de Prague et de celle de Moscou, utilisent des représentations logico-sémantiques distinguant les arguments des circonstants. Il y a seulement « *transfert lexical* » quand on passe directement de « l'espace lexical » de la langue source à celui de la langue cible [44].

Il est à noter qu'une analyse pour la TA peut différer de l'analyse requise à d'autres applications de TALN. Par exemple, supposons qu'on ait besoin de traduire la phrase *(John saw the girl with the binoculars)* vers le français. L'analyseur n'a pas besoin de se soucier de savoir où le syntagme prépositionnel doit s'attacher, parce que les deux possibilités conduisent à la même phrase en français. C'est une caractéristique de la traduction par rapport aux autres applications de TALN — l'ambigüité syntaxique n'influe pas sur le résultat.

Une fois le texte source analysé, nous aurons besoin de règles pour le *transfert syntaxique* et le *transfert lexical*. Les règles de *transfert syntaxique* nous indiquent comment modifier l'arbre d'analyse du texte source à ressembler à l'arbre d'analyse du texte cible.[37]

La figure 3.11 donne une intuition sur le ré-ordonnancement des cas simples comme adjectif-nom ; on transforme un arbre d'analyse, permettant de décrire une expression en anglais, en un autre arbre d'analyse, permettant de décrire une phrase en espagnol. Ces transformations sont des opérations qui correspondent une structure d'arborescence à une autre [13].

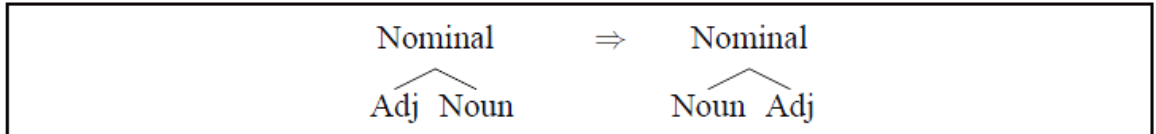

**Figure 3.11 Transformation simple qui réorganise les adjectifs et les noms [13]**

L'approche de *transfert* et cette règle peuvent être appliquées à l'exemple *(Mary did not slap the green witch)*. En plus de cette règle de transformation, on doit supposer que l'analyse morphologique précise que *(didn't)* est composé de *(do-PAST)* plus (*not*), l'analyseur fixe la caractéristique (PAST) sur le GV. Ensuite, le *Transfert lexical*, par le biais de recherche dans le dictionnaire bilingue, retire *(do)*, change *(not)* à (*no*), et transforme *slap en* une phrase (*dar una bofetada a*), avec un léger réarrangement de l'arbre syntaxique, comme le suggère la figure 3.12.

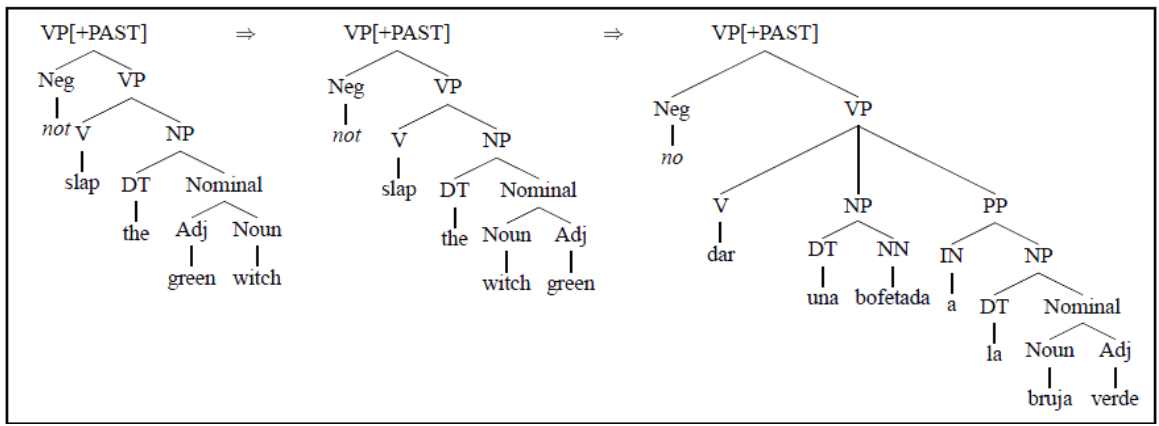

**Figure 3.12 Transformations dans l'approche à base de transfert [13]**

Pour la traduction des langues SVO comme l'anglais aux langues SOV comme le japonais, nous aurons besoin de transformations encore plus complexes, en déplaçant le verbe à la fin, en changeant les prépositions, et ainsi de suite. Un exemple du résultat de ces règles est montré dans la figure 3.13.

---

[37] Tesnière parle de transformation métataxique.

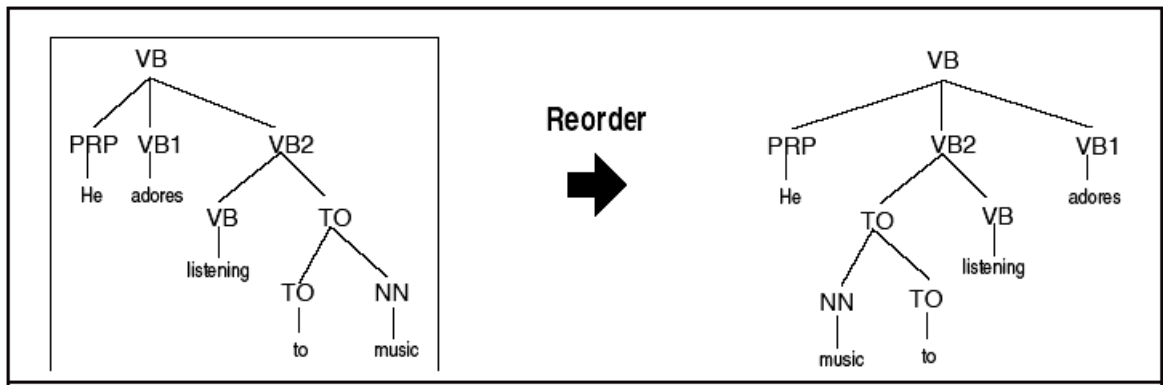

**Figure 3.13 Transformations syntaxiques de l'anglais (SVO) au japonais (SOV) [13]**

Les systèmes de *transfert* peuvent être basés sur des structures plus riches qu'une simple analyse syntaxique pure [13].

Par exemple, un système à base de *transfert* pour la traduction du chinois vers l'anglais pourrait avoir des règles pour traiter le fait qu'en chinois les locutions qui marquent un BUT comme (*to the store* dans la phrase (*I went to the store*)) ont tendance à précéder le verbe, alors qu'en anglais, ces locutions doivent apparaître après le verbe. Afin de traiter ce type de phrases et d'autres, l'analyse du texte Chinois doit percevoir la structure thématique, de façon à distinguer les phrases prépositionnelles qui doivent apparaître avant le verbe / de préférence apparaître avant le verbe / apparaître après le verbe [48].

En plus des transformations syntaxiques, les systèmes de *transfert* doivent avoir des règles de *transfert lexical*. Le *transfert lexical* est généralement basé sur un dictionnaire bilingue, tout comme pour la traduction *directe*. Le dictionnaire lui-même peut également être utilisé pour traiter les problèmes d'ambiguïté lexicale. Par exemple, le mot anglais *home* a plusieurs traductions possibles en allemand :

- *nach Hause* : dans le sens de rentrer à la maison ;
- *Heim* : dans le sens de jeu à domicile ;
- *Heimat :* dans le sens de la patrie, pays d'origine, ou maison spirituelle ;
- *zu Hause* : dans le sens d'être à la maison.

Dans ce cas, l'expression (*at home*) est très susceptible d'être traduite en (*zu Hause*), et donc le dictionnaire bilingue peut lister cette traduction idiomatique.

De nombreux cas de *transfert lexical* sont trop complexes pour les traiter via un dictionnaire phrastique. Dans ces cas, les systèmes de *transfert* peuvent appliquer les techniques de désambiguïsation lors de l'analyse du texte source, [13].

### 1.3 Combinaison des approches directe et de transfert

Bien que l'approche de *transfert* offre la possibilité de traiter des phénomènes linguistiques plus complexes par rapport à l'approche *directe*, il s'avère que les règles entre langues SVO et SOV ne sont pas suffisantes. Dans la pratique, on a besoin de règles qui allient la richesse des connaissances lexicales des deux langues avec des caractéristiques syntaxiques et sémantiques [13].

Pour cette raison, les systèmes commerciaux de la TA ont tendance à faire des combinaisons de l'approche *directe* et les approches de *transfert* et en utilisant des dictionnaires bilingues riches, mais aussi en utilisant des étiqueteurs et des analyseurs. Le système Systran, par exemple, comme décrit dans Hutchins et Somers [49], Senellart et al. [50], comporte trois composantes. La première est une phase d'analyse, elle comprend :

- analyse morphologique et une partie d'étiquetage
- segmentation des groupes nominaux et prépositionnelles
- analyse des dépendances superficielles (sujets, formes passives,…)

Vient ensuite la phase de *transfert*, qui comprend :

- La traduction des expressions idiomatiques ;
- la désambiguïsation sémantique ;
- l'attribution des prépositions.

Enfin, dans l'étape de synthèse, le système :

- applique un dictionnaire bilingue riche pour faire la traduction lexicale ;
- traite le ré-ordonnancement ;
- effectue la génération morphologique.

Comme dans un système direct, dans le système Systran une grande partie du traitement s'appuie sur le dictionnaire bilingue, qui comporte des connaissances lexicales, syntaxiques et sémantiques ; puis Systran ré-ordonne dans une étape de post-traitement.

Mais comme dans un système de *transfert*, de nombreuses étapes sont informées par un traitement syntaxique et sémantique [13].

### 1.4 L'approche à langue-pivot

Les systèmes de *transfert* sont moins ambitieux que les systèmes *à langue-pivot*, parce qu'ils reconnaissent la nécessité des règles de correspondance (souvent très complexes) entre les représentations les plus abstraites de phrases source et cible [8].

Un autre problème avec le modèle de *transfert* est qu'il nécessite un ensemble précis de règles de *transfert* pour chaque paire de langues. Ceci est clairement insuffisant pour les systèmes de traduction utilisés dans plusieurs-à-plusieurs contextes multilingues comme l'Union européenne. Les situations possibles sont :

- 1 : 1 (source unique, cible unique)
- 1 : n (source unique, cibles multiples)
- n : 1 (sources multiples, cible unique)
- k : n : (multi-sources/multi-cibles)

Ceci suggère une perspective différente sur la nature de la traduction.

Au lieu de transformer directement les mots de la phrase en langue source vers la langue cible, l'intuition de la *langue-pivot* est de traiter la traduction comme un processus d'extraction de sens du texte d'entrée, puis d'exprimer ce sens via l'interlangue dans la langue cible. Un système de TA pourrait alors se passer de la connaissance contrastive, simplement en s'appuyant sur les mêmes règles syntaxiques et sémantiques utilisées par un interpréteur standard et générateur de langue. La quantité de connaissances nécessaires serait alors proportionnelle au nombre de langues dont le système traite plutôt qu'à leur produit.

Cela suppose l'existence d'une représentation du sens sous une forme indépendante de la langue canonique, comme les représentations sémantiques. L'idée est que l'approche à *langue-pivot* représente toutes les phrases qui veulent dire la même chose de la même manière, quelle que soit la langue d'origine. La traduction dans ce modèle procède en effectuant une analyse sémantique profonde sur un texte dans la langue *X* vers une représentation *à langue-pivot* et génère à partir de la *langue-pivot* un texte dans la langue *Y* [13].

Les systèmes à véritable *langue-pivot* utilisent trois espaces lexicaux, car une véritable *langue-pivot* possède son propre vocabulaire, même si ce vocabulaire est construit comme union des acceptions d'un certain nombre de langues. Dans les systèmes de TA, il existe des *langue-pivot* « linguistico-sémantiques» dont les « lexèmes » sont construits à partir des lemmes et des lexies de dictionnaires d'une ou plusieurs langues naturelles, et des *langue-pivot* « sémantiques » ou « sémantico-pragmatiques », dont les lexèmes sont construits à partir des entités, propriétés, actions et processus d'un domaine précis et d'un ensemble de tâches bien identifiées (par exemple, la réservation touristique) [44].

La question qui se pose quel système de représentation à utiliser comme *langue-pivot* ? Le calcul des prédicats est une possibilité, ou une variante comme la sémantique à récursion minimale. Nous illustrons une troisième approche commune, une simple représentation basée sur des événements, dans laquelle les événements sont liés à leurs arguments via un petit ensemble fixe de rôles thématiques. Que nous utilisions logique ou autres représentations d'évènements, nous aurons besoin de spécifier les propriétés temporelles et aspectuelles des évènements, et nous devrons aussi représenter les relations non évènementielles entre les entités, telles que la relation (*has-color*) entre (*green*) et (*witch*). La Figure 3.14 montre une représentation *à langue-pivot* pour *(Mary did not slap the green witch)* comme une structure de traits.

On peut créer ces représentations *à langue-pivot* du texte de la langue source à l'aide des techniques de l'analyse sémantique ; à l'aide d'un étiqueteur de rôles sémantiques[38] pour découvrir la relation AGENT entre (*Mary*) et l'évènement (*slap*), ou la relation THEME entre (*witch*) et l'événement (*slap*). Nous aurions aussi besoin de faire une désambiguïsation de la relation nom-modificateur pour reconnaître que la relation entre (*the green*) et (*witch*) est la relation (HAS-COLOR), et nous aurons besoin de découvrir que cet événement a une polarité négative du mot (*didn't*). La *langue-pivot* nécessite donc une analyse plus approfondie que le modèle de *transfert*, qui ne nécessite que l'analyse syntaxique (ou un étiquetage de rôles thématiques superficielles). Mais la génération peut maintenant passer directement de la *langue-pivot* sans avoir besoin de transformations syntaxiques [13].

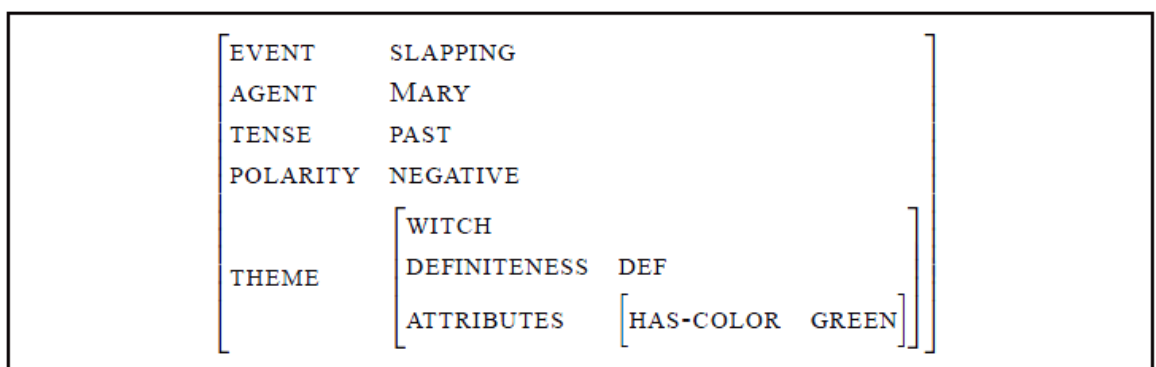

**Figure 3.14 Représentation à langue-pivot de la phrase (Mary did not slap the green witch) [13]**

En plus de l'absence de transformations syntaxiques, le système *à langue-pivot* n'a pas besoin de règles lexicales.

---

[38] L'utilisation des rôles sémantiques est généralement appelé *transfert* sémantique.

Par contre, le modèle *à langue-pivot* a ses propres problèmes. Par exemple, pour traduire du japonais vers le chinois, la *langue-pivot* universelle doit inclure des concepts tels le *grand-frère* (= aîné) et le *petit-frère* (=cadet ou benjamin) ! L'utilisation de ces concepts pour la traduction de l'allemand vers l'anglais nécessiterait alors de grandes quantités de désambiguïsation inutiles. La même chose pour le concept de nombre, où on trouve le singulier et le pluriel alors que dans l'arabe existe un autre nombre qui est le *Duel*, peu fréquent dans les langues modernes.[39]

En outre, le modèle *à langue-pivot* nécessite une analyse exhaustive de la sémantique du domaine et de la formaliser en une ontologie[40]. En général, cela n'est possible que dans des domaines relativement simples basés sur un modèle de base de données, comme dans le transport aérien, la réservation d'hôtel, ou domaines de recommandation de restaurants, où la définition de la base de données détermine les entités et les relations possibles. Pour ces raisons, les systèmes *à langue-pivot* sont généralement utilisés dans des domaines restreints [13].

## 2. Niveaux de représentations

Trouver un niveau suffisamment abstrait pour une représentation de la TA est un objectif réalisable. Cependant, ce n'est pas toujours vrai que le niveau le plus profond de la représentation soit le meilleur niveau pour la traduction.

Ceci peut être illustré facilement en réfléchissant à la traduction entre langues apparentées telles que le Norvégien et le Suédois (figure 3.15 et figure 3.16).

> a. Min nya bil ar bla  (Suédois)
>    'my new car is blue'
> b. Den nye bilen min er bla  (Norvégien)
>    'the new car mine is blue'

**Figure 3.15 Exemple 1 : comparaison de traduction entre le suédois et le norvégien [8]**

---

[39] Tend à disparaître, par exemple dans les langues slaves ou indiennes ; usuel en hébreu et dans les langues polynésiennes.

[40] Umberto Eco dit qu'on ne traduit pas à l'aide de dictionnaires, mais à l'aide d'encyclopédies.

> a. Var har du hittat en saful slips? (Suédois)
>
>    'Where did you find a such ugly tie'
>
> b. Hvor har du funnet et sastygt slips? (Norvégien)
>
>    'Where did you find a such ugly tie'

**Figure 3.16 Exemple 2 : comparaison de traduction entre le suédois et le norvégien [8]**

Dans la figure 3.16, les deux langues ont exactement le même ordre des mots, bien que les mots et leurs caractéristiques grammaticales diffèrent. Dans la figure 3.15, on voit que le suédois (comme l'anglais) ne permet pas l'utilisation d'un article avec un pronom possessif, alors que le norvégien (comme l'italien) le fait.

Pour faire face à cette différence structurelle on exprime une correspondance entre les structures décrites par les règles syntaxiques présentées dans la figure 3.17 [8].

> (Suédois)    GN → Poss Adj N
>
> (Norvégien)  GN → Det Adj N Poss

**Figure 3.17 Correspondance entre des structures du suédois et du norvégien [8]**

## 3. Flexibilités des systèmes de TA

Il serait simple de concevoir un système de TA équipé uniquement d'un seul type de règles linguistiques pour effectuer des manipulations des structures syntaxiques. Mais un certain nombre de considérations, notamment économiques, s'y opposent. On peut conclure que ce qui est nécessaire est une approche à base de règles de traduction suffisamment souples pour n'effectuer une analyse en profondeur qu'en cas de nécessité, de sorte que le même moteur de TA puisse être utilisé pour traiter des paires de langues proches et des paires de langues différant fortement. Ces idées sont à l'origine des tentatives pour concevoir des systèmes flexibles pouvant fonctionner dans une variété de modes, en fonction de la profondeur de l'analyse requise pour la paire de langues.

Il y a d'autres raisons à l'intérêt actuel pour les systèmes flexibles. L'exemple ci-dessus, montre que ce qui est le « niveau approprié » de l'analyse pour une paire de langues peut être tout à fait inapproprié pour une autre paire. La Traduction dépend des informations sur les différents niveaux d'information linguistique. Une représentation en termes de relations sémantiques (agent, patient, etc.) est attrayante. Cependant, une telle représentation ne sera pas capable de distinguer entre (a), (b) et (c) de la figure 3.18. Cela signifie qu'ils seront traduits de la même façon, si c'est la représentation qui est produite par l'analyse. Mais dans de nombreux cas, cela ne produit pas une très bonne traduction [8].

> a. Sam broke the printer.
> b. It was the printer that Sam broke.
> c. It was Sam that broke the printer

**Figure 3.18 Trois formulations différentes de la même signification [8]**

## *Conclusion*

Au long de ce chapitre, nous avons examiné les approches linguistiques – dites à base de règles. Ces approches sont basées sur les techniques de la linguistique computationnelle.

Nous avons décrit l'approche *directe*, où le passage de la langue source à la langue cible s'effectue par traduction mot à mot.

La deuxième approche est l'approche de *transfert*, qui suppose une représentation intermédiaire entre les structures de la langue source et les structures de la langue cible. Cette représentation assure un *transfert lexical*, *syntaxique* voire même *sémantique*. Les approches par *transfert* visent donc à formuler des équivalences langue à langue à un niveau d'abstraction suffisant pour tenir compte du contexte grammatical et de contraintes sémantiques.

La troisième approche est l'approche *à langue-pivot*. A l'inverse de l'approche de *transfert* qui travaille avec deux langues, l'approche à *langue-pivot* travaille avec plusieurs langues en même temps. Dans un schéma 1 : n, elle transfère le contenu d'un texte source correctement et sans perte de sens dans une langue pivot à partir de laquelle seront générées des phrases dans toutes les langues cibles possibles, correctement et sans perte de sens.

Le prochain chapitre étudiera un autre domaine de TALN, lié à la traduction automatique, celui de *la recherche d'information*, plus précisément la *recherche d'information multilingue.*

# Chapitre 4. Analyse morphologique : de la Recherche d'Information à la Traduction automatique de la langue arabe

## *Introduction*

Le rapide développement des technologies de l'information a eu pour conséquence un accroissement de sources d'informations textuelles. Il devient de plus en plus difficile pour les utilisateurs de retrouver exactement ce qu'ils recherchent dans cette masse de données, ce qui engendre ce qu'on appelle le Web caché. Le problème qui se pose actuellement donc n'est plus tant la disponibilité de l'information mais la capacité d'accès et de sélection de l'information répondant aux besoins précis d'un utilisateur, à partir des représentations qu'il perçoit.

La qualité de cette recherche pourra s'exprimer de façon traditionnelle en informatique documentaire, à l'aide de deux concepts :

- le *silence*, ou absence de documents qui auraient dû être trouvés ;
- le *bruit*, ou sélection de documents inappropriés.

Silence et bruit ne s'annulent pas, mais remettent tous deux en cause la qualité de la recherche effectuée.

**Aspect multilingue**

Alors qu'Internet rend de nombreux fonds (ou gisements) documentaires accessibles, le multilinguisme s'impose comme une réalité sociétale incontournable. Après que la langue anglaise eut largement dominé Internet il y a une dizaine d'années, la représentation du chinois, de l'espagnol ou du portugais tend à refléter l'importance réelle de ces langues. Internet reflète maintenant mieux la diversité des langues nationales et régionales [10]. [41]

---

[41] Google propose des moteurs de recherche qui fonctionnent dans 136 langues (nationales et régionales), et des outils de TA et de recherche d'information interlingue : en novembre 2010, 52 langues et 2550 paires de langues étaient disponibles sur Internet, et la bibliothèque *Google Book Search* contenait 7 millions de documents en 44 langues. Microsoft propose le correcteur orthographique de *Word* en 126 langues et un correcteur grammatical dans 6 langues (61 si l'on considère les variantes).

Les applications qui nécessitent la prise en compte d'aspect multilingue sont diverses :

- Veille stratégique, commerciale, scientifique, brevets.
- Surveillance d'opinion au niveau international.
- Partage d'information dans des sociétés multinationales.
- Partage d'information dans des pays pluri-linguistiques.
- Sécurité nationale (terrorisme, prolifération nucléaire).
- Lutte contre les trafics (drogue, blanchiment d'argent,…) [51].

Logiquement, les applications informatiques développées – que ce soit en traduction, en recherche d'information[42](RI), en aide à la communication écrite ou en apprentissage des langues,… – doivent s'adapter à cette réalité en prenant en compte des langues de plus en plus diverses avec parfois une grande distance linguistique entre elles.

Si les institutions internationales et les gouvernements de pays « multilingues », par exemple, ont toujours été des utilisateurs de systèmes de RI multilingue, le besoin de tels systèmes pour la vie de tous les jours se développe avec l'ensemble des activités liées au tourisme et au commerce électronique [52].

Les outils de RI sont aujourd'hui indispensables pour consulter des informations sur Internet. Ainsi en 2013, environ 45 % des utilisateurs emploient anglais ou chinois alors que moins de 11 % utilisent le français, l'allemand ou l'arabe comme langue principale [10]. D'autre part, les pages rédigées en anglais étaient encore largement dominantes avant 2000[43], mais les langues utilisées se sont beaucoup diversifiées ces dernières années. Il est donc devenu indispensable de considérer la RI dans plusieurs langues. Si les tentatives pour prendre en compte une dimension multilingue en RI datent de la fin des années 60, un renouveau de cette problématique a surgi dans les années 90 avec l'émergence du web et la disponibilité d'un grand nombre de pages écrites dans différentes langues [52]. Les recherches dans ce domaine ont débuté en 1996 lors du premier atelier CLIR (Cross-Lingual Information Retrieval) à la conférence SIGIR. Ces ateliers ont lieu annuellement depuis 2000.

---

[42] Ce nom fut donné par Calvin N. Mooers en 1948 pour la 1ère fois quand il travaillait sur son mémoire de maîtrise.
[43] Supérieures à 70 % selon http ://www.clickz.com/clickz/stats/1697080/web-pages-language.

## *1. Recherche d'information multilingue*

Le but d'un système de recherche d'information est de retrouver, parmi une collection de documents préalablement stockés, les documents qui répondent au besoin de l'utilisateur exprimé sous forme de requête.

Dans cette optique, la recherche d'information traite de la représentation, du stockage, de l'organisation de l'information, et des procédures d'accès.

*La recherche d'information multilingue* a pour but de repérer l'information lorsque la langue des requêtes est différente de la langue (ou des langues) des documents à repérer [52].

Cette recherche d'information translinguistique (*Cross-Language*), nécessite donc, soit la traduction des documents vers la langue de la requête, soit la traduction de la requête vers la langue des documents, soit de trouver des représentations des documents et des requêtes indépendantes de la langue [53].

## *2. Motivation des Systèmes de RI multilingues*

La RI multilingue tire son importance des cas où on ne peut pas satisfaire des besoins d'information par des systèmes de RI unilingue. Par exemple :

- Pour une collection contenant des documents écrits en plusieurs langues, il est peu pratique de formuler une requête dans chaque langue pour la recherche.
- Un même document peut être écrit dans plus d'une langue. A titre d'exemple : des documents dans lesquels des passages en anglais apparaissent confondus avec le texte de récit dans une autre langue ; actes de congrès multilingues….
- Un utilisateur qui ne maîtrise pas suffisamment la langue de la collection pour exprimer sa requête dans cette langue, mais est capable de se servir des documents qui sont identifiés [54].

## *3. Taxonomie des modèles de recherche d'information*
[55]

### 3.1. Le modèle booléen ou ensembliste

Il est basé sur la théorie des ensembles et sur l'algèbre de Boole. La requête est représentée sous forme d'une équation logique. Les termes sont reliés par des connecteurs logiques ET, OU et NON [56].

### 3.2. Le modèle vectoriel

Dans ce modèle, les requêtes et les documents sont vus comme des vecteurs dans un espace Euclidien de dimension élevée. Cet espace est celui engendré par tous les termes d'indexation. Le mécanisme de recherche consiste à retrouver les vecteurs documents qui se rapprochent le plus du vecteur requête. Cela implique que la pertinence d'un document relativement à une requête est reliée à la mesure de similarité des vecteurs associés [57].

### 3.3. Le modèle probabiliste

La similarité entre un document et une requête est mesurée par le rapport entre la probabilité qu'un document $d$ donné soit pertinent pour une requête $Q$, notée $p(d/Q)$, et la probabilité qu'il soit non pertinent et $p'(d,Q)$ [58].

### 3.4. Les réseaux inférentiels bayésiens

Un réseau inférentiel bayésien est un graphe de dépendances, orienté et acyclique. Dans ce graphe les nœuds représentent des variables propositionnelles (des concepts, des groupes de termes ou des documents) et les arcs des liens de dépendances entre les nœuds (les dépendances entre termes et entre termes et documents). Ainsi, si la proposition représentée par le nœud $p$ cause ou implique la proposition représentée par le nœud $q$, on trace alors un arc de $p$ vers $q$ [59].

### 3.5. Le modèle connexionniste

L'idée de base est que la RI est un processus associatif bien représenté par les mécanismes de propagation et d'activation des réseaux de neurones. Par ailleurs, les capacités d'apprentissage de ces modèles peuvent permettre d'obtenir des SRI adaptatifs [60].

### 3.6. Les modèles de langage

Ils sont basés sur l'hypothèse qu'un utilisateur en interaction avec un SRI fournit une requête en pensant à un ou plusieurs documents qu'il souhaite retrouver. La requête est alors inférée par l'utilisateur à partir de ce(s) document(s) [61].

### 3.7. Latent Semantic Indexing : LSI

L'objectif de LSI est d'aboutir à une représentation conceptuelle des documents. Dans ces documents les effets dus à la variation d'usage des termes dans la collection sont nettement atténués. Ainsi, des documents partageant des termes co-occurents ont des représentations proches dans l'espace défini par le modèle. Ceci permet de sélectionner des documents pertinents même s'ils ne contiennent aucun mot de la requête [62].

## *4. Approches de recherche d'information multilingue*

Les approches proposées sont basées soit sur des bases de connaissances, soit sur des textes parallèles. La première catégorie comporte trois techniques : *la traduction automatique*, *les dictionnaires bilingues* et *les vocabulaires contrôlés*. Ces approches peuvent être classées en trois catégories :

- une catégorie basée sur la traduction ; cette traduction peut être effectuée, soit en utilisant les traducteurs automatiques, ou des dictionnaires ou encore les corpus alignés, cette approche est celle qui nous intéresse dans ce chapitre ;
- une catégorie basée sur l'utilisation d'un vocabulaire prédéfini (thesaurus) comme un référentiel pour représenter les documents et les requêtes ;
- une catégorie basée sur le Croisement de Langues ou Latent Semantic Indexing (CL-LSI), représentant les documents multilingues et les termes d'indexation dans un même espace vectoriel [55].

### 4.1. Approches basées sur la TA

Ces approches nécessitent l'intégration d'un logiciel de TA dans le SRI. Les systèmes basés sur la TA sont utilisés pour obtenir un même texte dans plusieurs langues [55]. On utilise donc un système de recherche unilingue pour rechercher les documents. Deux approches dans la TA peuvent être appliquées : la traduction des documents ou la traduction des requêtes.

### 4.1.1. Traduction de la requête

Le système traduit *la requête* vers la langue des documents. Il s'agit de présenter au moteur de recherche les traductions de cette requête dans les différentes langues souhaitées. Le système récupérera alors les différents documents correspondants à chaque traduction [55].

La traduction de la requête présente moins de précision que celle de la collection de documents [63], qui contiennent un contexte d'information plus important, ce qui réduit les risques de mauvaise traduction. Toutefois, la traduction de tous les documents dans toutes les langues souhaités est trop compliquée à réaliser pour des corpus de taille importante.

### 4.1.2. Traduction des documents

Le système traduit *les documents* vers la langue de la requête. Les documents sont traduits dans la langue de la requête à l'aide d'outils de traduction. Le SRI procède ensuite à une simple interrogation unilingue. Son principal inconvénient est lié à la taille du fond documentaire. Il n'est pas concevable de traduire une collection de documents dans toutes les langues souhaitées pour l'interrogation [55].

### 4.1.3. Traduction de la requête et des documents

Le système traduit *la requête et les documents*. Dans ce cas, il s'agit de représenter la requête et les documents dans un même référentiel. Ce référentiel est souvent un vocabulaire multilingue prédéfini qui peut être par exemple un thesaurus[44] [55]. Cependant l'inconvénient de ce type de vocabulaire est qu'il n'est pas toujours disponible et qu'il doit vivre avec le fond.

Actuellement, la plupart des travaux dans ce domaine se focalisent sur la traduction de la requête. Cette traduction est moins coûteuse que celle de tous les documents de la collection [55]. Cependant, la traduction des requêtes n'est pas sans engendrer des problèmes ; Yamabana et al. observent que les techniques adoptées avec succès par la communauté de la TA sont peu adaptées à traduire des requêtes, puisque les requêtes sont souvent une suite de termes, situation que l'on rencontre couramment dans les moteurs de recherche. Donc, ce sont rarement des phrases complètes et plus souvent juste une séquence de mots [64]. Alors que les systèmes de TA sont prétendus à traduire des phrases syntaxiquement correctes.

Le problème d'expressivité de la requête traduite est posé aussi quand les termes issus de la traduction ne sont pas suffisants pour représenter la requête initiale. D'où la nécessité d'expansion pour enrichir la requête avec des termes plus courants. Mais le problème le plus crucial à résoudre est sans doute le problème d'ambiguïté, notamment quand plusieurs traductions pour un ou plusieurs termes de la requête sont possibles [65].

---

[44] Exemple Euro WordNet.

Contrairement aux systèmes de TA qui sur la base d'une phrase, restituent une phrase traduite, les approches basées sur les dictionnaires et sur les corpus alignés proposent une traduction mot à mot [55].

## 4.2. Approches basées sur les dictionnaires

Les dictionnaires, utilisés dans ce domaine, sont généralement des listes de termes donnés dans la langue source alignés avec d'autres termes de la langue cible. L'approche de base est de prendre chaque terme dans la requête et de le remplacer par une liste de toutes ses traductions possibles, ceci par l'entremise d'une recherche des termes de la requête dans un dictionnaire compréhensible par une machine sans se préoccuper de la syntaxe. Ainsi, les termes (*mad cow*) seront traduits (*fou vache*) et non (*vache folle*) [66]. Les techniques basées sur les dictionnaires ne sont pas donc totalement satisfaisantes à cause de la difficulté de la TA et des imperfections des dictionnaires bilingues [55], qui posent souvent des problèmes, en effet :

- le dictionnaire ne contient pas tous les mots possibles retrouvés dans un texte. Certains termes sont explicites. Ils ne sont pas nécessairement dans un dictionnaire, car l'utilisateur humain est capable de dériver automatiquement ses formes [55] ;
- la traduction par dictionnaire produit habituellement une expansion importante de la requête qui devient bruitée. En effet, les termes ont beaucoup de traductions possibles. Parmi ces traductions, non seulement on retrouve des synonymes, mais également des termes correspondant à des sens différents [54] ;
- le dictionnaire contient la plupart des termes de la langue, cependant les termes techniques existent rarement. C'est ce qu'on appelle le problème de couverture [55];
- le dictionnaire ne contient pas les noms propres. Les noms des pays et des personnes ont besoin aussi d'être traduits [55] ;
- D'autre part, la traduction basée sur des mots ne réussit pas toujours à traduire correctement des mots composés ou des expressions qui contiennent plus d'un mot. Ce sont les dictionnaires idiomatiques[45] ou terminologiques qui peuvent résoudre ce problème. Malheureusement, il est difficile de trouver ce genre de dictionnaires à large couver-ture. Beaucoup de paires de langues telles que anglais-arabe ne bénéficient pas de ce type de dictionnaires [54].

---

[45] Les expressions figées sont fréquentes en français, plus rares en allemand, turc ou hongrois...

Les études faites par Ballesteros [67] et Grefenstette [68] montrent que la traduction des requêtes par les dictionnaires bilingues peut mener à une baisse de 40 à 60 % de performance de la RI translinguistique par rapport à la performance unilingue.

L'utilisation des dictionnaires pour la traduction de requêtes pose des problèmes liés à l'absence des termes spécifiques à un domaine ou l'absence de certaines formes d'un terme, et dans la plupart des cas, ces dictionnaires proposent pour un terme donné différentes traductions. C'est la raison pour laquelle la communauté de la RI multilingue s'est orientée vers les méthodes basées sur les corpus alignés. Ces derniers tentent d'y répondre par extraction automatique de l'information manquante [55].

### 4.3. Approches basées sur les corpus alignés

L'approche basée sur les corpus alignés analyse les collections de textes en se basant sur les statistiques. Elle extrait automatiquement l'information requise pour construire des techniques spécifiques à la traduction de requêtes. Un corpus aligné est constitué d'un ensemble de documents exprimés dans une langue et alignés avec des documents dans une autre langue. L'alignement entre ces documents consiste à mettre en correspondance les documents de langues différentes selon un critère donné. Il peut être parallèle ou comparable [55].

#### 4.3.1. L'alignement parallèle

Il consiste à mettre en correspondance chaque document d'une langue source $L1$ avec le document représentant sa traduction dans la langue cible $L2$. Dans ce cas, l'alignement peut être fait sur : *le document*, *les paragraphes*, *les phrases* ou *les termes*. Les corpus basés sur ce type d'alignement sont appelés les corpus parallèles [55]. En général, ces méthodes procèdent en alignant les phrases des corpus phrase par phrase. Ensuite, le système crée une représentation globale permettant de traduire un terme en un ensemble de termes possibles selon plusieurs paramètres tels que la position des mots dans les phrases [54].

#### 4.3.2. L'alignement comparable

Plus délicat à réaliser, revient à mettre en correspondance des documents en se basant sur des critères comme par exemple la présence de même dates, de même noms de personnes dans des documents de langues différentes [68], [63]. Les corpus basés sur ce type d'alignement sont appelés les corpus comparables [55].

## 4.4. Approches basées sur un vocabulaire prédéfini (thésaurus)

Un thésaurus est un dictionnaire hiérarchisé de concepts. Dans un thésaurus, les mots sont classés dans des catégories appelées concepts et les concepts sont reliés selon leurs relations sémantiques. Pour chaque concept, sont indiqués ses synonymes. Dans un vocabulaire contrôlé (thésaurus), chaque concept est étiqueté par un terme descriptif unique dans le sens où l'usager peut facilement spécifier les concepts appropriés dans sa requête [54].

Cette approche consiste d'une façon générale à utiliser un vocabulaire contrôlé, représenté sous forme d'un thésaurus multilingue. Les correspondances entre termes de différentes langues étant prédéfinies par le vocabulaire et regroupées dans des classes. Une classe représente une entrée du vocabulaire. Ces approches sont utilisées pour la représentation des documents et des requêtes. L'indexation des documents est guidée par le vocabulaire. Ainsi, chaque document est représenté par une liste de classes de termes. La RI revient donc à représenter la requête dans ce référentiel (liste de classes) et à récupérer les documents exprimés dans les différentes langues et indexés par cette liste [55]. La traduction des concepts se fait par une simple consultation d'un thésaurus incluant pour chaque concept les termes correspondants dans le langage cible. Un des avantages de l'approche des thésaurus est le contrôle des synonymes et de la polysémie par l'utilisation d'informations syntaxiques et sémantiques [54].

## 4.5. Approches basées sur Croisement de Langue - LSI

L'idée de base consiste à considérer un ensemble de documents dans une langue, les traduire dans le but de construire un ensemble de documents duaux. Un document dual ou virtuel est la concaténation d'un document exprimé dans une langue et sa traduction intégrale dans une autre langue. Lors de la phase d'analyse, le document dual est considéré comme un seul document indépendamment de la langue. L'ensemble de documents duaux est analysé en utilisant la LSI. Le résultat est représenté par l'espace sémantique réduit où les termes reliés sont regroupés dans la même classe. Du fait qu'un document dual contient des termes en Français et en Anglais, l'espace LSI va automatiquement contenir les termes dans les deux langues. Les termes identiques, auront une représentation identique dans l'espace, cependant les termes qui apparaissent fréquemment, sont représentés de façon similaire. Puis il faut représenter les documents dans chaque langue autour des termes de l'espace. Dans ce cas, l'utilisateur peut poser sa requête soit en Français ou en Anglais et récupérer les documents les plus semblables[46] dans ces deux langues [55].

## 5. Recherche d'information translinguistique avec l'arabe

Les premières expériences en RI translinguistique avec l'arabe ont été abordées en 2001 et 2002 dans la conférence TREC[47]. Plusieurs travaux y présentaient une problématique de recherche de documents pertinents dans une large collection de documents en arabe, en utilisant des requêtes en anglais. La majorité des systèmes a utilisé une approche de traduction des requêtes basée sur des dictionnaires bilingues, d'autres systèmes ont exploité un modèle de traduction statistique entraîné sur des corpus parallèles. La troisième catégorie a opéré une combinaison d'un modèle de traduction statistique avec des dictionnaires bilingues. Cette combinaison a donné de meilleures performances par rapport aux autres approches utilisant des ressources individuelles [54].

## 6. Contribution dans la recherche d'information en langue arabe

Dans ce qui suit nous décrivons un système que nous avons conçu et réalisé appelé (ESAIR) : *Enhanced Stemmer for Arabic Information Retrieval*. Les résultats obtenus ont été publiés dans le journal *Neural Network World Journal*[48] [69]. Le système propose une méthode d'indexation et de recherche pour les textes en langue arabe basée sur les techniques de traitement du langage naturel.

### 6.1. Problématique

Les documents électroniques accessibles dans les sites web constituent un champ de recherches documentaires et de veille technologique vaste et en pleine expansion [70]. Mais ces documents sont, selon l'inventeur du web, « destinés aux humains plutôt que des données et informations qui peuvent être analysées automatiquement » [71], le défi est justement d'extraire automatiquement de l'information de ces documents écrits en langage naturel.

---

[46] À contrario, http://www.linguee.fr/francais-anglais permet de vérifier que la traduction d'une même expression varie avec le contexte…

[47] Text REtrieval Conference.

[48] http://www.nnw.cz/ : International journal on non standard computing and artificial intelligence, Czech Technical University in Prague, Faculty of Transportation Sciences, Czech Republic.

L'utilisation de la langue arabe, dans le domaine d'indexation constitue un grand pas vers son intégration dans la technologie de l'information. Vue sa puissance et sa richesse.

« La puissance de la langue naturelle crée un obstacle à son utilisation pour le traitement de l'information » [72].

L'application de l'indexation sur la langue arabe pose des problèmes majeurs [73, 74] :

- Le problème d'ambiguïté issue de l'absence des voyelles, ceci exige des règles morphologiques complexes [75] ;
- Le problème de la reconnaissance des formes fléchies, car l'arabe est une langue fortement flexionnelle [76].

Les objectifs fixés pour ce traitement sont :

- La reconnaissance intelligente des différentes formes que peut prendre un mot *sans l'utilisation d'un dictionnaire des formes fléchies* c'est à dire sans la connaissance à priori de ces mots ;
- La réduction du taux d'ambiguïté issu de l'absence des voyelles ;
- L'identification d'information contenue dans tout texte et sa représentation au moyen d'index, dans un espace restreint relativement à l'espace où ces documents sont stockés [77].
- La réduction du temps de réponse d'une RI.

Les SRI doivent non seulement représenter, stocker et organiser les documents mais surtout restituer un ensemble de documents pertinents satisfaisant les besoins d'un utilisateur exprimés par sa requête [78, 79, 80].

## 6.2. Démarche d'analyse

Pour comprendre un texte, il est nécessaire d'en faire une analyse linguistique complète. Les techniques linguistiques sont théoriquement les meilleures car elles permettent d'avoir une bonne compréhension des documents indexés. Cepen-

---

Mot-clé : mot qui a une importance particulière. ex : cheval.

Racine ou stemme : mot privé de ses derniers caractères s'ils ne marquent que des différences de forme — n'est pas forcément un mot ; *cheval* est le radical de *cheval, chevalier* et *chevalin*, mais *cheval* et *chevaux* ne partagent que *cheva-*.

Lemme : mot ni conjugué ni accordé représentant tous les mots d'une même famille, indépendamment de leur forme ; *cheval* est le lemme de *cheval* et de *chevaux* ; de chevalin de chevalier.

dant, la mise en œuvre de telles méthodes nécessite des algorithmes de traitement du langage naturel très puissants, et un temps de traitement considérable.

On s'était d'abord contenté d'évaluer la pertinence d'un document pour une requête sur la base des *mots-clés* partagés. Par ce mécanisme fondé sur une simple comparaison de chaines de caractères, les SRI se retrouvaient confrontés à un problème lié à la complexité du langage naturel, qui peut manifester de différentes manières un même concept.

Un document pertinent peut ainsi contenir des termes différents de ceux de la requête mais sémantiquement proches. Alors, les synonymies engendrent des silences, et les polysémies du bruit !

On a ensuite évalué la pertinence d'un document pour une requête sur la base des *racines* ou stemmes. La *racinisation* consiste alors à enlever des mots les derniers caractères (considérés comme décrivant les flexions de mots) pour ne retenir que le radical – certaines racinisations utilisant des connaissances morphologiques plus complètes (suffixes, préfixes…). La racinisation réduit le silence mais augmente le bruit.

Nous nous sommes donc attachés à une troisième stratégie, la lemmatisation[49], qui désigne l'analyse lexicale du contenu d'un texte, regroupant les mots d'une même famille en une entité appelée lemme (forme canonique), dénotant un mot indépendamment de sa forme.

Le processus de lemmatisation convertit les documents préalablement pour extraire des mots de référence ou lemmes.

Au lieu d'indexer un texte par des mots on l'indexera par des lemmes correspondants aux mots. On espère ainsi améliorer l'appariement des mots utilisés dans la requête avec ceux représentant le contenu des documents (figure 4.1).

Pour faire face à ce problème, nous nous tournons vers le traitement automatique du langage naturel (TALN), afin de disposer en RI de descripteurs plus robustes et plus pertinents que de simples chaines de caractères [81].

---

[49] « La Lemmatisation des grandes bases de textes », Dominique Labbé, CERAT, Institut d'études politiques (Grenoble, France), sur
http://www.uottawa.ca/academic/arts/astrolabe/articles/art0030.htm/Lemmatisation.htm

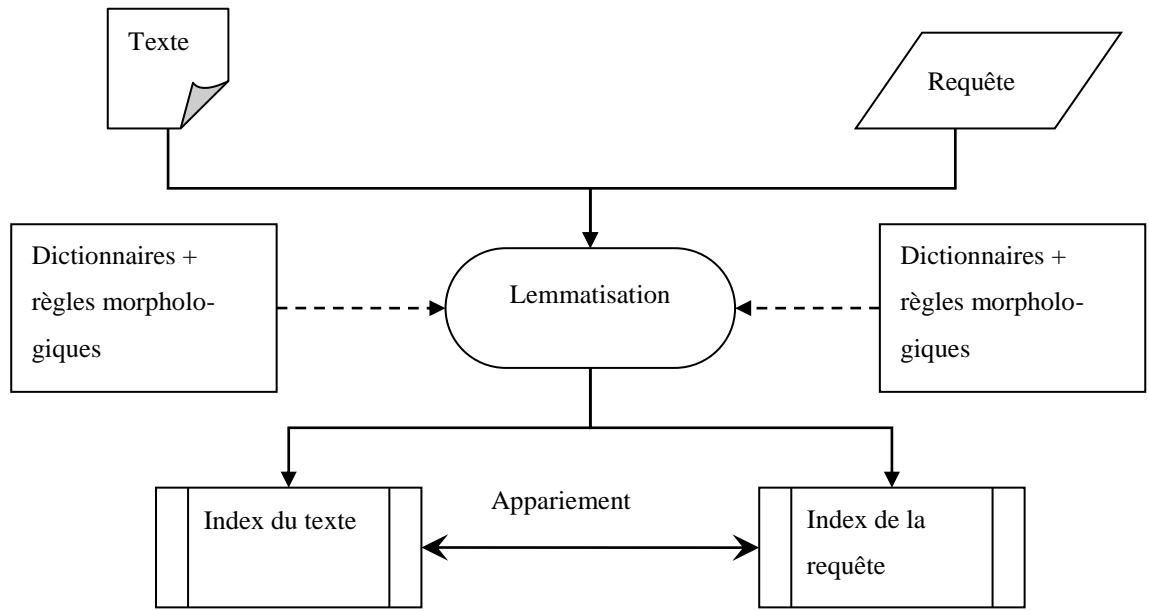

**Figure 4.1 Lemmatisation des mots du texte et de la requête pour l'appariement.**

L'algorithme ESAIR consiste à décomposer les mots en morphèmes [82, 83], sans tenir compte des liens grammaticaux entre ces derniers (lemmatisation hors contexte). Il procède en premier lieu par :

- *Une Etape de normalisation*, qui transforme le document dans un format plus facile à manipuler [84]. Cette étape est délicate car l'arabe est une langue flexionnelle et fortement productive [85].
- *Une Etape d'analyse lexicale* : l'analyse morphologique ne peut fonctionner sans l'aide des dictionnaires contenants les unités lexicales (dictionnaire des racines, des schèmes, des mots vides, des mots outils, des mots spécifiques) ; cette étape permet de vérifier si l'unité lexicale appartient bien à la langue, comme la compatibilité entre les différents constituants du mot.
- *Etape d'indexation* : qui engendre l'index en regroupant les mots par leurs lemmes.

Pour vérifier qu'un mot appartient aux lexèmes arabes à l'exception des noms propres, de quelques noms communs et des mots-outils, il suffit de trouver sa racine et le schème correspondant.

Suivant cette méthode le travail comprend trois étapes essentielles :

- Le découpage ;
- La recherche des schèmes et des racines ;
- La génération d'index.

### 6.2.1. Découpage

Pour résoudre l'ambiguïté, Aljlayl et Frieder montrent que la lemmatisation légère (approche basée sur suppression de suffixe et de préfixe) surpasse significativement celle basée sur la détection de racine dans le domaine de RI [86].

Découper un mot ou le segmenter consiste à extraire ses différentes parties (préfixe, racine, suffixe,…).

Le principe du découpage est le suivant :

> Clitique : élément à mi-chemin entre un mot indépendant et un morphème lié, toujours prosodiquement lié à un mot adjacent, dénommé hôte (très souvent un verbe), de telle sorte qu'il ne forme avec celui-ci une seule unité phonétique complexe. Les clitiques peuvent être : des mots séparés, des mots distincts de leur hôte, attachés à leur hôte par un élément de ponctuation (trait d'union, apostrophe).
> Proclitique : clitique s'appuyant sur le mot qui le suit.
> Enclitique : clitique s'appuyant sur le mot qui le précède
> Source: http://fr.wikipedia.org/wiki/Clitique

- Découpage du mot en *proclitique +base1+enclitique* qui consiste à repérer tous les proclitiques et les enclitiques qui apparaissent dans le mot. *Base1* est en général une base (racine + des infixes) munie de préfixes et de suffixes.

- Découpage de la *base1* (résultat de la phase précédente) en *préfixes + base + suffixes* ; le principe de cette phase est le même que celui de la phase précédente.

- Découpage de la *base* en *racine* et *schème* i.e. trouver un schème parmi ceux stockés dans le dictionnaire des schèmes qui corresponde à la base. La méthode de reconnaissance du schème sera décrite en 6.2.2.1.

#### 6.2.1.1. Reconnaissance des proclitiques et des enclitiques

**Recherche des proclitiques et enclitiques**

La liste des proclitiques et des enclitiques de la langue arabe est limitée. On peut utiliser la liste proposé par Darwish [87], plusieurs d'entre eux ont été utilisés pour la lemmatisation par Chen et Habash [88, 89]. La liste des proclitiques et des enclitiques utilisé est représentée dans la table 4.1.

**Table 4.1 Liste des proclitiques et des enclitiques**

| Proclitiques | ' | ب | ك | ل | ف | س | أ | ال | كال | لل | فب | فس | فال | فك | فل | فلل | أف | أس | فبال | فكال | بال |
|---|---|---|---|---|---|---|---|---|---|---|---|---|---|---|---|---|---|---|---|---|---|
| Enclitiques | ' | ه | ي | ك | هم | هن | ها | هما | كم | كن | كما | ني | نا | | | | | | | | |

Le découpage du mot en " proclitiques + base1+ enclitiques" ne se limite pas à la recherche d'un proclitique (resp. un enclitique) parmi la liste au début (resp. la fin) du mot, mais aussi à vérifier une certaine compatibilité entre les proclitiques et les enclitiques repérés dans le mot à décomposer.

### Test de compatibilité

Après l'extraction du proclitique *P* et de l'enclitique *E* du mot à analyser, ces deux sous-chaînes sont fusionnées en une chaîne *C* pour les tester dans une table des incompatibilités. Si la chaine *C* n'y est pas trouvée alors ce proclitique *P* est dit compatible avec cet enclitique *E*.

### Principe d'analyse

Lors du découpage du mot en *proclitique+ base1+enclitique* le processus identifie le plus long *proclitique* (respectivement *enclitique*) du mot, puis il accède à la table pour vérifier la compatibilité (*proclitique et enclitique*). Si cette décomposition est acceptée, elle sera stockée dans la table des résultats de cette phase, puis on continue avec une nouvelle décomposition pour traiter tous les cas possibles. Si la décomposition est rejetée, on passe à une autre décomposition.

#### 6.2.1.2. Reconnaissance des préfixes et des suffixes

Le principe de cette étape est pratiquement le même que la précédente sauf que la liste utilisée est celle des préfixes et des suffixes et la table de compatibilité utilisée est celle des préfixes et des suffixes.

La table 4.2 représente la liste des préfixes et des suffixes.

**Table 4.2 Liste des préfixes et des suffixes**

| Prefixes | ا | ت | ن | ي | إ | '' | | | | | | | | | |
|---|---|---|---|---|---|---|---|---|---|---|---|---|---|---|---|
| Suffixes | '' | ات | ية | ة | يات | نا | ت | تما | تم | تن | ن | ين | ان | ون | وا | ا | ي |

### 6.2.2. Recherche de schème et de racine

Après avoir découpé le mot en base et constituants, cette base est l'objet de traitement de l'étape de recherche de schème et de racine.

#### 6.2.2.1. Recherche de schème

Pour un *mot X*, un *schème I* du dictionnaire des schèmes correspond au *mot X* si la taille du *schème* est égale à la taille du *mot X*, et si toutes les lettres correspondantes aux positions dans le champ (liste des infixes) du dictionnaire des schèmes se trouvent dans le *mot X* aux mêmes positions révélées par ce champ (liste des infixes) (figure 4.2).

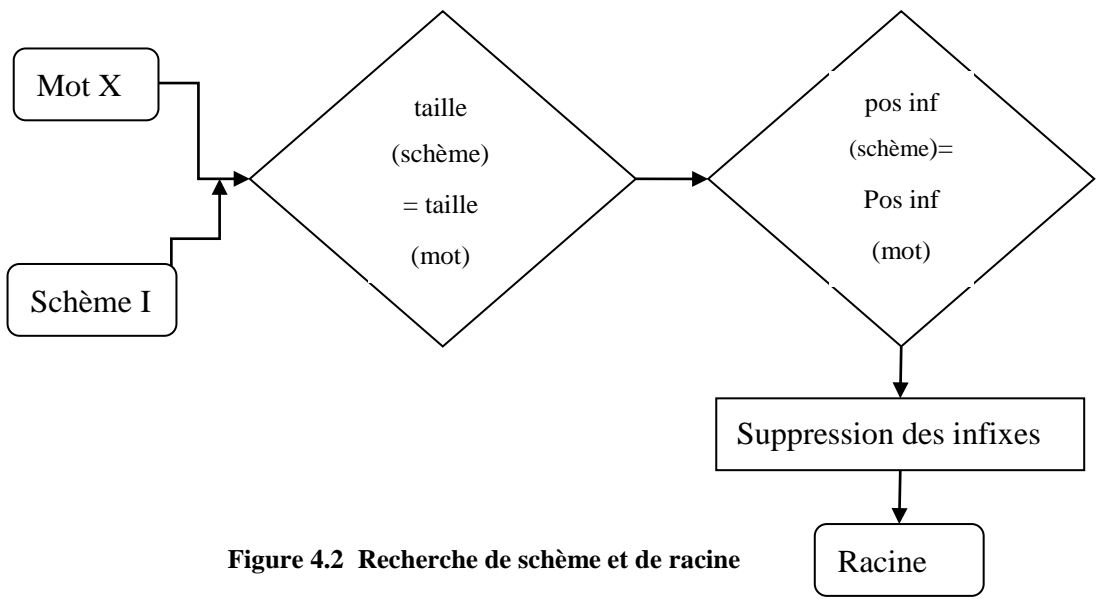

**Figure 4.2  Recherche de schème et de racine**

*Exemple*

Soit un Mot = 'صالح'. Le processus de recherche de schème parcourt tous les enregistrements qui ont la même taille que le mot jusqu'à rencontrer le schème 'فاعل'. Le champ liste des infixes correspondant est '2' la lettre 'ا' se trouve à la position 2 du mot 'صالح' donc c'est probablement le bon schème.

#### 6.2.2.2.    Recherche de la racine

Après détermination du schème, l'extraction de la racine se limite à la suppression de toutes les lettres correspondantes aux positions de champs (liste des infixes) dans le mot à décomposer.

*Exemple*

Le mot 'مفاتيح' [MAFATIH] a pour schème 'مفاعيل' [MAFAIL], le champ liste des infixes est '135'. (figure 4.3).

L'élimination de la lettre 'م' de la position 1 du mot 'مفاتيح' [MAFAIL] qui est la même du champ liste des infixes, la lettre 'ا' de la position 3 et la lettre 'ي' de la position 5 donne 'فتح' [FTH]. Ainsi on a retrouvé la racine correcte du mot 'مفاتيح' [MAFATIH] qui est la racine 'فتح' [FTH].

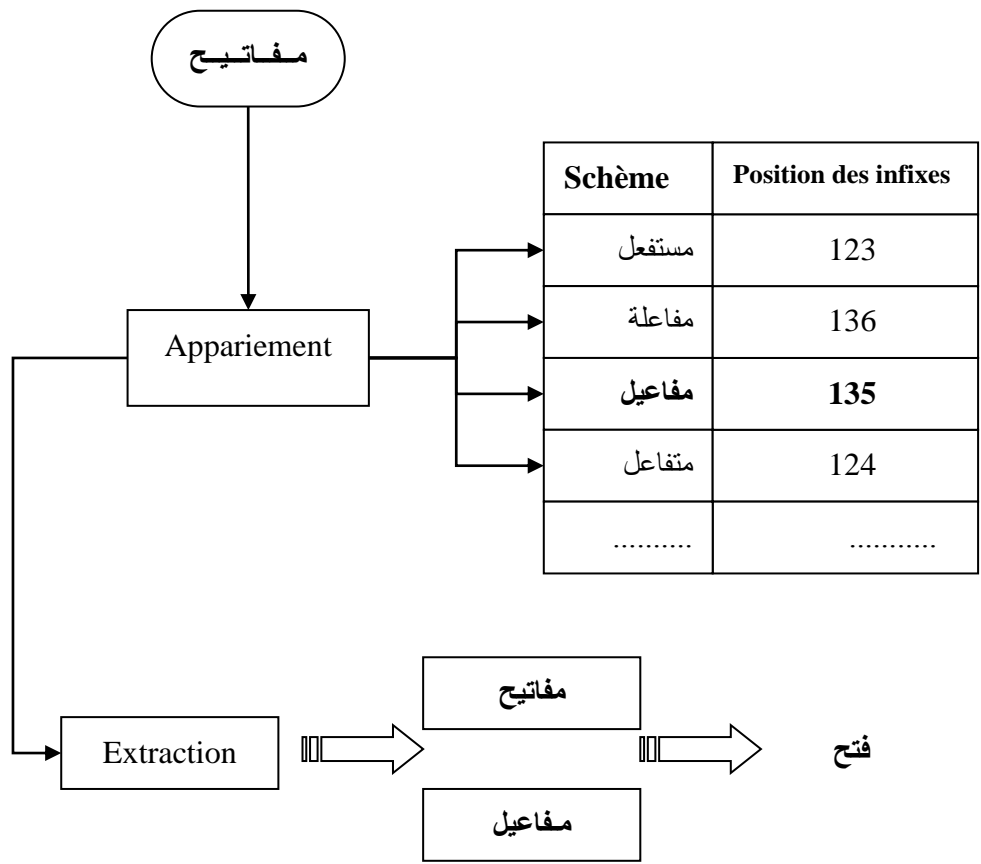

**Figure 4.3 Recherche de schème et de racine et extraction de racine**

**Remarque**

Ces étapes ne sont pas les seules, et des sous-étapes peuvent être introduites en cas de besoin (traitement de compatibilité entre proclitiques/préfixes, enclitiques/suffixes, et désambiguïsation dans le cas où il y a plusieurs interprétations).

### 6.2.3. Génération d'index

Les descripteurs peuvent être les mots du texte, les lemmes, les concepts et, plus rarement, les *N*-grammes, ou encore les contextes (cas du « Latent Semantic Indexing » ou des méthodes basées sur l'Analyse Factorielle des Correspondances). Les modèles utilisant les mots peuvent fonctionner avec la langue anglaise (peu de flexions, peu d'homographies), mais se révèlent nettement insuffisants pour les autres langues (particulièrement pour les langues agglutinantes) ; on peut alors utiliser les lemmes, mais, pour avoir de bonnes performances, il faut recourir à une analyse linguistique pour lever certaines ambiguïtés [90].

L'objectif du niveau morphologique est de diminuer fortement le nombre de mots analysés (par lemmatisation) [91], puis d'éliminer les mots vides en utilisant un anti dictionnaire, pour alléger les index et diminuer le silence à la recherche [92], donné par (équation 4.1).

$$\text{Silence} = \frac{\text{documents pertinents non trouvés}}{\text{documents pertinents}} \qquad (4.1)$$

Le bruit et le silence sont expliqués dans la figure 4.4.

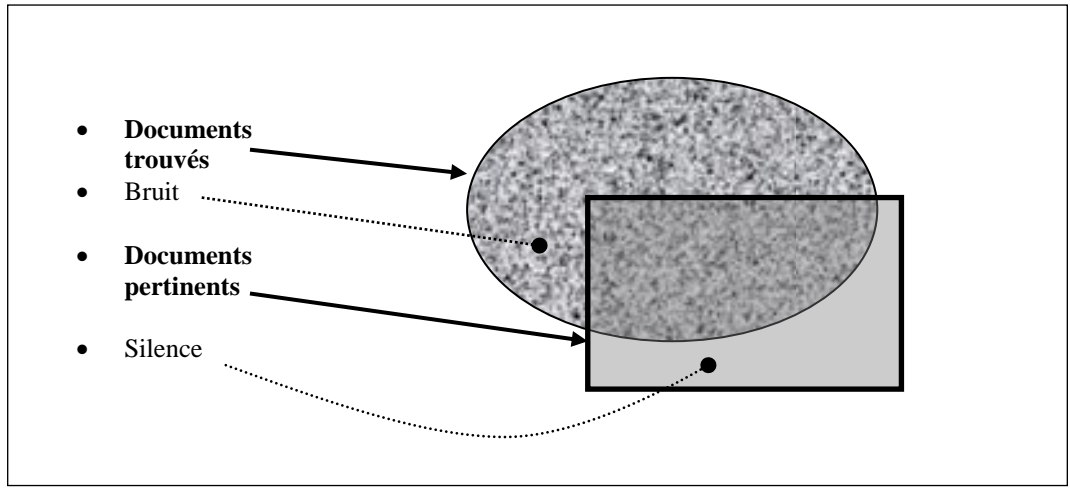

- **Documents trouvés**
- Bruit
- **Documents pertinents**
- Silence

**Figure 4.4 Silence et bruit dans la recherche d'informations**

### 6.3. Illustration

Dans cette section, on discute les résultats obtenus en utilisant notre algorithme. Nous avons appliqué les étapes susmentionnées sur des corpus. L'utilisateur formule sa requête en langue arabe, cette requête passe par toutes les étapes d'indexation y compris les étapes d'analyse morphologique. Puis les termes de la requête sont mis dans une liste pour qu'elle soit comparée avec les index des documents.

Le système analyse le contenu de la requête et le convertit en éléments du langage d'indexation. Les documents étant représentés par des éléments de ce même langage d'indexation, le système, après comparaison des éléments de la requête avec ceux des documents, détermine les degrés de ressemblance de ces derniers avec la requête et sélectionne ceux qui ont un degré de ressemblance supérieur à un seuil donné.

L'avantage d'ESAIR est la diminution de silence lors de la recherche ; des mots non mentionnés dans la requête peuvent figurer dans les résultats.

Considérons ce texte avec 162 mots :

( مما صح عنه عليه الصلاة و السلام قوله : " عليكم بالصدق، فان الصدق يهدي إلى البر. و إن البر يهدي إلى التقوى. و لا يزال الرجل يصدق و يتحرى الصدق، حتى يكتب عند الله صديقا. و إياكم و الكذب، فان الكذب يهدي إلى الفجور. و إن الفجور يهدي إلى النار. و لا يزال الرجل يكذب و يتحرى الكذب، حتى يكتب عند الله كذابا ". في هذا الحديث الشريف، دعوة قوية من النبي صلى الله عليه و سلم إلى إتباع الصدق و التحلي به قولا و فعلا. بل على الرجل الفطن الكيس أن يفتش و ينقب و يتحرى أثاره في كل زمان و مكان. لأن ذلك يهديه و يقوده إلى البر و الخير، فان فعل ذلك كان مع الصديقين و الشهداء و حسن أولئك رفيقا أي جوارا و صحبة في الجنة. و من معاني

الصدق كذلك التصديق بكل ما أمر الله تعالى الإيمان به. و من معانيه كذلك الصدقة و الإنفاق في سبل و أوجه الخير الكثيرة، مصداقا لقوله تعالى " إنما الصدقات للفقراء و المساكين ".)

La table 4.3 montre l'index après la suppression des mots vides, on obtient un index avec 72 mots (44% du texte original).

**Table 4.3 Index après suppression des mots vides**

| بالصدق | يصدق | يهدي | الكذب | الصدق | أثاره | الصديقين | التصديق | الخير |
|---|---|---|---|---|---|---|---|---|
| الصدق | يتحرى | الفجور | يكتب | التحلي | زمان | الشهداء | الله | الكثيرة |
| يهدي | الصدق | الفجور | الله | قولا | مكان | حسن | الإيمان | الصدقات |
| البر | يكتب | يهدي | كذابا | فعلا | يهديه | رفيقا | معانيه | للفقراء |
| البر | الله | النار | دعوة | الرجل | يقوده | جوارا | الصدقة | المساكين |
| يهدي | صديقا | الرجل | النبي | يفتش | البر | صحبة | الإنفاق | قوية |
| التقوى | الكذب | يكذب | الله | ينقب | الخير | الجنة | أوجه | الفطن |
| الرجل | الكذب | يتحرى | إتباع | يتحرى | فعل | الصدق | سبل | الكيس |

La table 4.4 montre l'index après l'application des étapes précédentes sur chaque mot, on obtient un index avec 43 mots (26% du texte original).

**Table 4.4 Index après l'application d'ESAIR**

| صدق | رجل | صديق | جنة | حلي | أثار | شهداء | إنفاق | فطن |
|---|---|---|---|---|---|---|---|---|
| هدي | حرى | كذب | إيمان | قول | زمان | حسن | أوجه | كيس |
| بر | كتب | فجور | دعو | فعل | مكان | رفيق | سبل | كثير |
| تقوى | اله | نار | نبي | فتش | قاد | جوار | فقراء | |
| قوي | تصديق | كذاب | إتباع | نقب | خير | صحب | مساكين | |

La diminution de la taille d'index est relative au texte ; si le texte contient beaucoup de termes dérivés d'une même racine, comme dans le texte 1, la réduction tend vers 98%, comme illustré en figure 4.5.

Texte 1 : كتب الكاتب في مكتبه بالكاتبة كل الكتب غير المكتوبة في المكتبة [KATABA ALKATIBO FI MAKTABIHI BILKATIBATI KOULLA ALKOUTOUBI GHAIRA ALMAKTOUBATI FI ALMAKTABATI]

Index 1 : كتب [KTB].

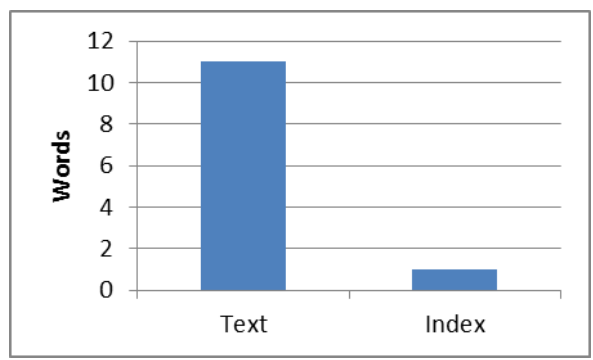

**Figure 4.5 Text1 vs. Index1**

Si le texte ne contient jamais deux mots communs à une racine ou même base, comme dans le texte 2, la réduction sera de l'ordre de 0%, comme illustré en figure 4.6.

Texte 2: تقدم الدولة حاليا تحفيزات كبيرة ومعتبرة للنهوض بالبحث العلمي في الجزائر [TOKADIMO ADAWLATO HALIEN TAHFIZET KABIRA WAMOETABRA LINOHODI BILBAHTI ALILMI FI ALJAZEIR]

Index 2 : قدم، دول، حفز، كبر،عبر، نهض، بحث، علم، الجزائر [KDM, DWL, HFZ, KBR, ABR, NHD, BHT, ALM, aljazeir]

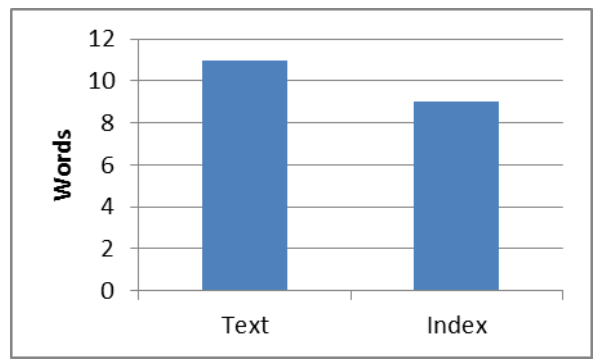

**Figure 4.6 Text2 vs. Index2**

La requête passe par toutes les étapes d'indexation, puis les termes de la requête sont comparés avec les termes des indexes des documents comme dans l'exemple suivant :

Texte 1 : (يساهم المعلم بشكل مباشر في بناء الأجيال) [YOSSAHIMO ALMOALIMO BICH-KLIN MOBACHIR FI BINAI ALAJIALI]

Texte 2 : (أقام المدير حفلا على شرف المعلمات المتقاعدات) [AKAMA ALMODIRO HAFLEN ALA CHARAFI ALMOALIMATI ALMOTAKAIDATI]

Texte 3 : (زار التلاميذ و المعلمون مركزا للبحث العلمي) [ZARA ATALAMIDO WA ALMOA-LIMONA MARKAZEN LILBAHTI ALILMI]

Requête : (معلم) [MOALIM]

Si l'appariement s'effectue mot à mot aucun texte ne sera trouvé (0/3). Avec ESAIR, les trois textes (3/3) sont retenus, car la lemmatisation des mots (المعلم ، المعلمات، المعلمون) [AL-MOALIMO, ALMOALIMATO, ALMOALIMONA] donne le même mot (معلم) [MOALIM].

Considérant le texte de la table 4.3, et un ensemble de 49 autres textes, et la requête "الصدق في القول" [ASSIDKO FI ALKAOULI], on en déduit : le nombre de documents pertinents trouvés est de 9 sur 12, le nombre de documents donné comme réponse est de 14 documents. Par conséquent la précision = (9/14) = 0.64, le rappel = (9/12) = 0.75 et le silence = (3/12) = 0.25. L'exactitude calculé sur ce texte est (69/72) = 0.958.

## 6.4. Expérimentations et résultats

Les expériences[50] ont été effectuées en exécutant ESAIR sur un ensemble aléatoire de documents du corpus d'ESSEX (Essex Arabic Summaries Corpus (EASC)), EASC est une ressource linguistique développée à l'université d'ESSEX, Royaume-Uni. Le corpus contient 153 articles tirés du journal Alwatan et le journal Alrai, qui couvrent plusieurs sujets : éducation, science et technologies, finance, santé, politique, religion et sport. Chaque document comporte en moyenne 389 mots, avec un total de 59'548 mots dans le corpus.

On extrait de façon manuelle les racines des mots pour des raisons de comparaison avec les résultats d'ESAIR. Un ensemble de 25 requêtes avec leurs jugements pertinents, créés pour chercher des informations particulières, sont utilisées pour évaluer la méthode proposée. Rappelons les formules de rappel et de précision : (4.4) et (4.5).

Le rappel mesure la capacité du système à restituer l'ensemble des documents pertinents (en lien avec le *silence documentaire* (formule 4.3).

$$Silence = 1 - rappel \qquad (4.3)$$

$$Rappel = \frac{Nb\ de\ documents\ pertinents\ retrouvés}{Nb\ de\ documents\ pertinents} \qquad (4.4)$$

La précision mesure la capacité du système à ne restituer que des documents pertinents (en lien avec le *bruit documentaire* (formule 4.6).

$$Précision = \frac{Nb\ de\ documents\ pertinents\ retrouvés}{Nb\ de\ documents\ retrouvés} \qquad (4.5)$$

$$Bruit = 1 - précision \qquad (4.6)$$

Dans la table 4.5, nous résumons une comparaison entre ESAIR et NoStem (recherche sans lemmatisation) en matière de précision et de rappel.

**Table 4.5 Moyenne de précision et de rappel**

| Algorithme | Précision | Rappel |
|---|---|---|
| ESAIR | 0.5732 | 0.6916 |
| NoStem | 0.4328 | 0.4152 |

---

[50] Je remercie Dr. Guy Tremblay de l'Université du Québec à Montréal (UQAM), Canada pour son soutien pendant la réalisation des expériences dans le laboratoire LATECE, et Dr. Mahmoud El haj de l'Université d' Essex, Royaume Uni pour fournir le corpus de test.

Les résultats confirment clairement que l'algorithme proposé dépasse l'approche minimaliste[51] de recherche mot par mot. Cela indique que la lemmatisation a un effet crucial sur la RI pour les langues fortement fléchies comme la langue arabe.

La figure 4.7 montre *la précision en 11 points*[52] pour ESAIR et NoStem.

La *précision en 11 points* (Moyenne des précisions obtenues chaque fois qu'un document pertinent est retrouvé.) est la moyenne des 11 précisions interpolées obtenues pour les points de rappels fixes, de 0 %, à 100 % (0-1-2-……10) par pas de 10 %. La règle d'interpolation est : la valeur interpolée de la précision pour un niveau de rappel i est la précision maximale obtenue pour tous les rappels supérieurs ou égaux à i.

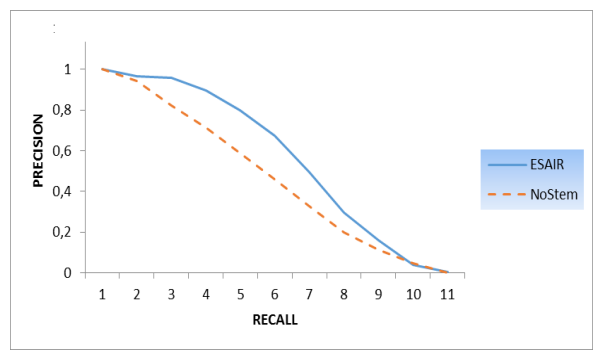

**Figure 4.7 11 points de précision pour ESAIR et NoStem**

Après la consultation de tous les lemmes, l'exactitude pour chaque document est calculée par la formule 4.7 :

$$Exactitude = \frac{\sum \text{mots lemmatisés correctement}}{\text{total des mots valides}} \quad (4.7)$$

Les résultats obtenus indiquent que l'algorithme extrait les lemmes corrects avec une exactitude plus de 96%, qui permet d'améliorer la RI.

### 6.5. Recherche multilingue

Ce genre de système, s'apparente à une recherche unilingue, cependant le processus de recherche est capable de traiter des requêtes dans différentes langues. Le corpus est découpé en bases documentaires unilingues, indépendantes les unes des autres. Les documents de chacune des bases ne peuvent être retrouvés que par une requête dans leur langue.

---

[51] On cherche tels quels les mots de la requête dans les documents.
[52] Cette mesure a été introduite dans TREC2 pour sa capacité à résumer les mesures de précision aux 11 points de rappel.

Avant de procéder à la recherche dans le corpus en langue arabe, on traduit d'abord le(s) mot(s) de la requête par traduction *directe* (mot à mot) car on n'a pas d'exigences syntaxiques ; la requête est constituée juste d'un ensemble de mots clés non reliés entre eux syntaxiquement et ne constituent pas une phrase grammaticalement cohérente. Cela facilite la tâche de traduction ; aucun *transfert syntaxique* ni *sémantique* n'est sollicité. Avant la traduction, une phase de prétraitement est nécessaire, il s'agit d'analyse morphologique des mots clés de la requête pour extraire leurs lemmes à partir des formes fléchies. La traduction s'effectue par une simple consultation d'un dictionnaire bilingue. On peut prendre juste une seule entrée du dictionnaire ou plusieurs équivalents lexicaux pour enrichir la requête. Le mot en langue cible passera ensuite par les étapes cités en dessus. La liste des termes est puis comparée avec les entrés d'index de la base de documents (figure 4.8).

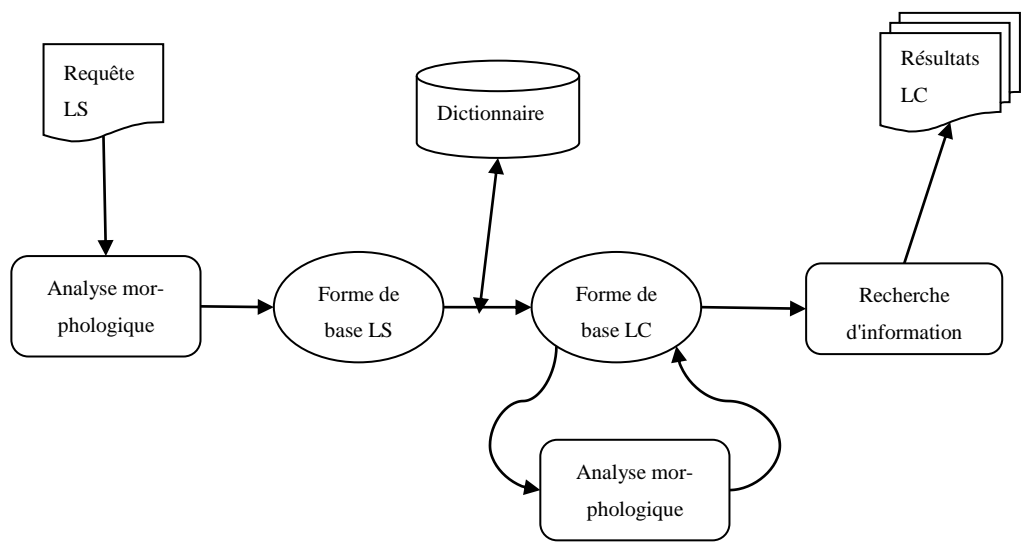

**Figure 4.8 Recherche multilingue**

**Exemple**

Soit la requête suivante : (*teachers*) et les textes suivants :

Texte 1 : (يساهم المعلم بشكل مباشر في بناء الأجيال) [YOSSAHIMO ALMOALIMO BICHKLIN MOBACHIR FI BINAI ALAJIALI]

Texte 2 : (أقام المدير حفلا على شرف المعلمات المتقاعدات) [AKAMA ALMODIRO HAFLEN ALA CHARAFI ALMOALIMATI ALMOTAKAIDATI]

Texte 3 : (زار التلاميذ و المعلمون مركزا للبحث العلمي) [ZARA ATALAMIDO WA ALMOALIMONA MARKAZEN LILBAHTI ALILMI].

Le terme de la requête passe par toutes les étapes citées dans la section précédente.

**Phase 1 :** *analyse morphologique* : on ramène les formes fléchies à des formes canoniques. Le mot (*teachers*) est transformé en (*teacher*) : [nom singulier].

**Phase 2 :** *traduction directe* : on cherche dans un dictionnaire bilingue l'équivalent du mot qui est (معلم), on peut enrichir la requête par d'autres équivalents comme (مدرس), (أستاذ), …etc.

**Phase 3 :** *recherche d'information* : si le mot cible n'est pas dans sa forme canonique, on peut procéder à une analyse morphologique avant de lancer la recherche.

**Résultats** : les trois textes seront retenus car les mots (المعلم، المعلمات، المعلمون) ont comme forme canonique dans l'index le terme (معلم).

## *Conclusion*

Dans ce chapitre, nous avons défini la recherche d'information *multilingue*, et les différentes approches proposées basées sur la linguistique comme les approches basées sur la TA, les dictionnaires, et les corpus. Puis, nous avons proposé un Algorithme ESAIR qui assure une recherche unilingue arabe, et qui fournit des composantes nécessaires pour une RI *multilingue*.

Nous avons appliqué une méthode basée sur les techniques linguistiques (notion de schème), qui nous a permis :

(1) d'adjoindre aux textes des index de tailles minimes représentant les informations pertinentes,

(2) d'analyser la requête et de la comparer avec l'index.

Cette approche n'exige pas une connaissance préalable des mots et peut trouver des relations avec les mots sans la spécification de l'utilisateur.

Compte tenu de l'intérêt de la recherche linguistique, nous avons construit un moteur de recherche qui permet une recherche efficace grâce à l'analyse du contenu des requêtes, assurant ainsi une diminution de silence.

L'algorithme a été testé sur des mots tirés du corpus EASC constituié de 153 articles du journal Alwatan et du journal Alrai. Des jugements d'experts ont été utilisés pour évaluer les résultats. L'algorithme a pu extraire correctement les lemmes avec une exactitude de plus de 96%. On a démontré que la lemmatisation entraîne des améliorations considérables dans la recherche d'information en langue arabe avec une précision de 0,5732. *La différence avec les approches à base de mots est statistiquement significative.*

Pour la *recherche multilingue*, on appliquerait une analyse morphologique sur des termes de requêtes, puis une traduction *directe* des termes, ces derniers seraient analysés morphologiquement si besoin. Puis ils rentreraient dans le processus de recherche pour aboutir à des résultats en langue cible qui correspondent aux mots clés écrits en langue source.

# Chapitre 5. Règles morphologiques pour le transfert de l'anglais vers l'arabe

## *Introduction*

Ce chapitre s'intéresse à l'entité mot, le chapitre 6 à l'entité phrase. Notre travail est divisé en deux parties : la première concerne la traduction directe, qui est très importante pour la langue arabe car le mot peut inclure toute une phrase. Les différents constituants du mot sont ajoutés au lemme dans cette phase selon les traits du mot source. La deuxième partie est la traduction au niveau phrase. A ce niveau on effectue des transferts abstraits entre l'anglais et l'arabe avec des arbres syntaxiques sans prendre en compte les mots qui représentent les entités de l'arbre, la construction des mots est faite dans la première partie. L'interaction entre les deux étapes s'effectue de façon non linéaire.

L'arabe a une morphologie plus complexe que l'anglais. En arabe, le nom et l'adjectif sont fléchis par les marques de genre et de nombre. Le verbe est fléchi en plus par les marques de temps, de mode et de personne. En outre des clitiques peuvent être attachées aux mots (conjonctions, prépositions, pronoms possessifs, …)

## *1. Les recherches morpho-syntaxiques de l'arabe pour la TA*

Dans cette section on présente quelques exemples de travaux effectués sur l'analyse morphologique et l'analyse syntaxique de la langue arabe pour la traduction automatique, publiés de 2010 à 2013.

Bouzit et al. [93] présentent une contribution en vue de développer un système de traduction de l'arabe vers n'importe quelle autre langue en utilisant des méthodes purement linguistiques.

Bies et al. [94] décrivent les ressources linguistiques notamment les plus adaptées avec la traduction automatique pour l'arabe. Ces ressources ont été publiées par The Linguistic Data Consortium, LDC.

Les travaux de Habash [95] étudient l'effet de prétraitement des différents niveaux du mot arabe sur la qualité de la TA statistique à base de phrase.

Zbib et al. [96] décrivent l'application de deux techniques de prétraitement de la traduction statistique de l'anglais vers l'arabe : la segmentation morphologique et le réordonnancement syntaxique. Ces techniques peuvent être adaptées pour les appliquer à la traduction vers l'arabe en fournissant des améliorations signifiantes pour les systèmes à base de phrase.

Le travail de Violetta et al. [97] décrit comment l'information morphologique est utilisée dans les approches à base d'exemples pour la traduction de l'arabe vers l'anglais pour produire des améliorations signifiantes dans la qualité de la traduction que ce soit pour de petits ou de gros corpus.

Le travail de Zbib et al. [98] présente plusieurs techniques pour intégrer des informations à partir des systèmes de traduction à base de règles à des systèmes de traduction statistique. Ces informations sont regroupées en trois niveaux (morphologique, lexical, système), le travail montre comment des informations spécifiques à une langue obtenues à partir des méthodes à base de règles peuvent améliorer la traduction statistique qui est souvent indépendante de la langue.

Marton et al. [99] explorent la contribution des caractéristiques lexicales et morphologiques pour l'analyse de dépendance de l'arabe. Le travail montre que les traits fonctionnels de genre et de nombre et le trait humain sont plus utiles dans l'analyse que les méthodes basées sur la forme de genre et de nombre.

Arwa et al. [100] ont proposé une approche de transfert en utilisant l'analyse morphologique pour induire à une symétrie morphologique et syntaxique qui a pour objectif d'améliorer la TA entre l'arabe et l'anglais.

Dans le travail d'El Kholy et al. [101], les auteurs étudient l'effet de la segmentation morphologique et la normalisation orthographique sur la traduction statistique vers la langue arabe. Les résultats montrent que la meilleure segmentation est celle de Penn Arabic Treebank.

Dans le travail de Hasan [102], les auteurs ont examinés les différentes méthodologies de la segmentation dans la langue arabe pour la traduction statistique par une comparaison entre la segmentation dans la traduction à base de règles et les segmentations dans la traduction statistique.

Alkuhlani et al. [103] présentent une version enrichie de Penn Arabic Treebank où les caractéristiques nécessaires pour la modélisation d'accord morpho-syntaxique en arabe sont annotées manuellement. Ainsi, ils présentent une analyse quantitative du phénomène morpho-syntaxique de l'arabe.

Hassan [104] a réalisé une extension de la traduction statistique à base de phrase avec des descriptions lexicales syntaxiques qui localisent l'information globale syntaxique dans le mot sans introduire l'ambigüité syntaxique redondante.

Le travail de Shaalan et al. [105] utilise les techniques d'apprentissage automatique supervisé pour induire les règles de transfert de l'arabe à l'anglais à partir des ressources linguistiques parallèlement alignées. Les règles de transfert structurel assurent le passage d'une structure syntaxique de l'arabe vers une structure syntaxique de l'anglais. Ainsi l'induction des règles morphologiques pour assurer l'apprentissage des règles morphologiques de l'arabe.

Alawneh et al. [106] présentent une approche de traduction des phrases structurées de l'anglais vers l'arabe en utilisant les méthodes basées sur la grammaire et les techniques de traduction à base d'exemples pour résoudre les problèmes d'accord et d'ordonnancement.

Abu Shquier et al. [107] ont étudié les problèmes d'accord et d'ordonnancement dans les systèmes de TA. Ils exposent une approche à base de règles pour la traduction des phrases en anglais bien structurées vers des phrases en arabe structurés.

Le travail de Bisazza et al. [108] propose une méthode de segmentation pour le réordonnancement pour déplacer le verbe de la première position dans la langue arabe vers la deuxième position en anglais. La méthode est utilisée pour le prétraitement des données d'apprentissage pour avoir des informations statistiques sur le mouvement du verbe. Les résultats montrent des améliorations au niveau de l'ordonnancement ainsi dans la qualité de traduction.

Le travail de Carpuat et al. [109] étudie les défis d'ordre du verbe arabe et de son sujet dans la traduction statistique. Les auteurs proposent une nouvelle méthode pour intégrer les informations de verbe et de son sujet dans la traduction statistique.

Marton et al. [110] intègrent à un modèle de traduction issu des textes parallèles alignés des informations syntaxiques de la langue source. Les résultats montrent des améliorations importantes dans la performance pour la traduction de l'arabe vers l'anglais.

Alawneh et al. [111] présentent une approche de traduction des phrases en anglais bien structurées vers des phrases en arabe bien structurées aussi en utilisant les techniques à base de grammaires et les techniques à base d'exemples pour traiter les problèmes d'accord et de ré-ordonnancement. Cette technique est hybride car elle combine les techniques de traduction à base de règles et les techniques à base d'exemples.

Ture et al. [112] comparent les modèles de traduction linéaires et hiérarchiques à base de phrase pour la traduction des requêtes. Les deux méthodes donnent de meilleurs résultats par rapport aux méthodes à base de jetons ou à base d'une traduction référence.

## *2. Algorithme de TA pour l'arabe*

### 2.1. Problématique

Il s'agit ici d'un algorithme de traduction directe de l'anglais vers l'arabe et de l'arabe vers l'anglais. Ce travail résulte d'une étude morphosyntaxique de la langue arabe dont, on a besoin de combiner entre l'approche directe et l'approche de transfert. Ce chapitre s'intéresse à l'aspect morphologique, en ce qui concerne l'aspect syntaxique, il sera détaillé dans le chapitre 6 avec des règles de transfert.

Pour la traduction vers l'arabe, il s'agit d'une synthèse de mots arabes. Pour la traduction de l'arabe vers l'anglais, il s'agit d'analyse des mots arabes. Cette génération ou cette analyse est basée sur les notions de schème avec proclitiques, enclitiques, préfixes et suffixes. Comment à partir d'informations grammaticales des mots en anglais engendrer les constituants du mot en arabe, sachant qu'un mot en arabe peut inclure toute une phrase ? Et comment à partir des constituants du mot en arabe engendrer le résultat en anglais ?

### 2.2. Méthode proposée

Cette section est publiée en partie dans le journal : *International Journal of Computer Applications*[53] [91].

#### 2.2.1. Traduction de l'arabe vers l'anglais

C'est une analyse morphologique des mots arabes basée sur la notion de schème, en suivant les étapes décrites dans la section 6 du chapitre 4. Cette fois–ci, le travail est orienté vers les structures de traits i.e. chaque constituant est lié avec une catégorie grammaticale ou une information morphosyntaxique. Cette catégorie ou information est interprété en mots en langue cible ou en information de ré-ordonnancement des mots dans la phrase cible.

---

[53] http://www.ijcaonline.org/ scholarly peer-reviewed research publishing journal, Foundation of Computer Science, New York, USA.

La reconnaissance des proclitiques et des enclitiques, en tant que parties du mot analysé sans ambiguïté après l'analyse finale et après les tests de compatibilité, est très utile pour représenter les traits du mot et pour préserver les caractéristiques morphologiques du mot. La même chose pour les préfixes et les suffixes.

Le mot (أستخرجانها) (*est-ce que vous allez la sortir ?*) est analysé en lemme et d'autres constituants (figure 5.1), qui sont interprétés comme traits utilisés dans la traduction vers la langue cible.

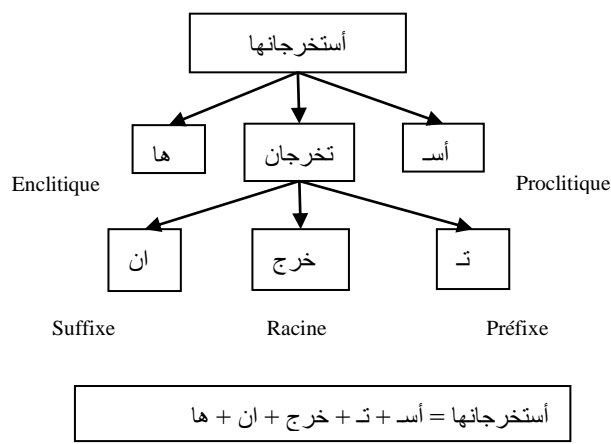

**Figure 5.1 Analyse morphologique du mot 'أستخرجانها'**

Les différents constituants ont les traits suivants représentés dans la table 5.1.

**Table 5.1 Les traits des constituants du mot (أستخرجانها)**

| Constituant | Trait |
|---|---|
| (أ) | Interrogative pronoun |
| (سـ) | Tense = future |
| (تـ) | Subject = 2nd person singular |
| (ان) | Subject = dual |
| (ها) | object = feminine singular |

Ces traits sont utilisés dans la traduction comme illustré dans la table 5.2.

**Table 5.2 Exploitation des traits des mots**

| Trait | Mot cible |
|---|---|
| `Interrogative pronoun` | (be) you |
| `Tense = future` | Will |
| Verb = see dictionary | Take out |
| `Subject = 2ⁿᵈ person singular` | You[54] |
| `Subject = dual` → plural | You |
| `object = feminine singular` | She, it |

Alors, on obtient la phrase (*will you take out it*), puis avec des règles de ré-ordonnancement et des règles d'accord, la phrase sera réécrite selon les règles de la grammaire de la langue cible comme suit (*will you take it out*) ou (*will you take her out*), et d'autres possibilités. Les traits en tant qu'informations contextuelles éliminent plusieurs possibilités et ne laissent que les cas les plus probables.

**Remarque**

Les traits de nombre peuvent être appliqués systématiquement, par contre les traits de genre ne peuvent pas être appliqués toujours car le genre des mots relève d'aspects culturels de chaque langue.

### 2.2.2. Traduction de l'anglais vers l'arabe

Dans la traduction vers la langue arabe, on utilise la génération au lieu de l'analyse. Le principe est le même qu'avec l'analyse mais cette fois ci le processus est inversé i.e. à partir d'un mot source avec ses traits, on génère un mot en arabe (figure 5.2).

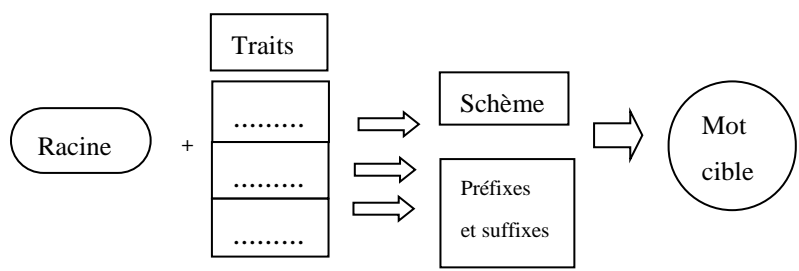

**Figure 5.2 Génération de mots à partir des traits et de la racine**

Au niveau du mot, on commence par la traduction de la racine en arabe, puis, on effectue des transformations sur la racine par le biais du schème. Enfin selon les traits du mot en anglais, on rajoute les préfixes et les suffixes correspondants. Pour l'entité phrase, des règles de ré-ordonnancement et des règles d'accord sont appliquées. Rappelons que ces règles s'ins-crivent dans une génération morphosyntaxique et non pas une génération purement syntaxique qui sera l'objet du prochain chapitre.

---

[54] « Thou » ne survit que dans les textes religieux, poétiques ou pour des effets littéraires.

**Exemple**

Le mot (*the workers*) a les traits suivants : (*noun, definite, masculine, plural*).

L'équivalent de la racine (*work*) est (عمل).

On applique la caractéristique (*agent*) par l'application du schème (فاعل), pour le trait : (*definite*), on rajoute au lemme l'article de définition (الـ). Pour le trait *gender : (masculine)*, on ne rajoute pas le (ة). Enfin le trait *number : (plural),* on rajoute (ون), pour aboutir en fin au mot cible (العاملون).

## *3. Règles de transfert*

L'étiqueteur des parties de discours utilisé dans ces deux derniers chapitres est le Stanford Log-linear Part-Of-Speech Tagger (POS Tager) [113] crée par (The Stanford Natural Language Processing Group) [114]. La liste des parties de discours utilisées par Stanford POS tager est celle utilisée par Penn Treebank POS tager exprimée dans la table 5.3.

**Table 5.3 Liste des parties de discours de Stanford POS tagger [115]**

| N°= | Tag | Description |
|---|---|---|
| 1. | CC | Coordinating conjunction |
| 2. | CD | Cardinal number |
| 3. | DT | Determiner |
| 4. | EX | Existential there |
| 5. | FW | Foreign word |
| 6. | IN | Preposition or subordinating conjunction |
| 7. | JJ | Adjective |
| 8. | JJR | Adjective, comparative |
| 9. | JJS | Adjective, superlative |

| 10. | LS | List item marker |
|---|---|---|
| 11. | MD | Modal |
| 12. | NN | Noun, singular or mass |
| 13. | NNS | Noun, plural |
| 14. | NNP | Proper noun, singular |
| 15. | NNPS | Proper noun, plural |
| 16. | PDT | Predeterminer |
| 17. | POS | Possessive ending |
| 18. | PRP | Personal pronoun |
| 19. | PRP$ | Possessive pronoun |
| 20. | RB | Adverb |
| 21. | RBR | Adverb, comparative |
| 22. | RBS | Adverb, superlative |
| 23. | RP | Particle |
| 24. | SYM | Symbol |
| 25. | TO | to |
| 26. | UH | Interjection |
| 27. | VB | Verb, base form |
| 28. | VBD | Verb, past tense |
| 29. | VBG | Verb, gerund or present participle |
| 30. | VBN | Verb, past participle |
| 31. | VBP | Verb, non-3rd person singular present |
| 32. | VBZ | Verb, 3rd person singular present |
| 33. | WDT | Wh-determiner |

| 34. | WP   | Wh-pronoun            |
|-----|------|-----------------------|
| 35. | WP$  | Possessive wh-pronoun |
| 36. | WRB  | Wh-adverb             |

Ainsi, d'autres traits sont nécessaires pour les particularités de la langue arabe :

[M] : Masculin

[F] : Féminin

[hmn]= "vrai" : Humain, [hmn]= "faux" : Non humain.

[S] : singulier

[B] : Both

[P] : pluriel

Pré : préfixe

Suf : suffixe

Pro : proclitique

Enc : enclitique

"+" : concaténation.

La table 5.4 présente les différentes traductions du mot (*beliver*) selon les traits associés au mot. On remarque bien que le même mot aura une traduction différente avec chaque changement de traits.

**Table 5.4 Traduction du mot (beliver)**

| Mot source  | Traits                       | Mot cible |
|-------------|------------------------------|-----------|
| A believer  | (S/M):  Singular masculine   | مؤمن      |
| A believer  | (S/F):  Singular feminine    | مؤمنة     |
| A believers | (B/M): Both masculine        | مؤمنان    |
| A believers | (B/F):  Both feminine        | مؤمنتان   |
| A believers | (P/M): Plural masculine      | مؤمنون    |
| A believers | (P/F):  Plural feminine      | مؤمنات    |

Un autre exemple, avec des verbes cette fois-ci, est présenté dans la table 5.5 pour la phrase (*you fear*). On remarque que la traduction du verbe (*fear*) change selon les traits du mot qui le précède (*you*).

**Table 5.5 Traduction du verbe (fear)**

| Mot source | Traits (the doer of the action) | Mot cible |
|---|---|---|
| You fear | (S/M):  Singular masculine | تخاف |
| You fear | (S/F):  Singular feminine | تخافين |
| You fear | (B/M):  Both masculine | تخافان |
| You fear | (B/F):  Both feminine | تخافان |
| You fear | (P/M):  Plural masculine | تخافون |
| You fear | (P/F):  Plural feminine | تخفن |

**Remarque**

La seule présence d'un masculin avec un ou plusieurs féminins, rend tout le groupe masculin.

### *3.1.   Le verbe*

On entend par accord sujet-verbe/ verbe-objet, l'accord intrinsèque dans le mot puisque on est au niveau morphologique.

#### 3.1.1.  L'accord verbe/sujet

Si l'ordre des mots dans la phrase est SVO, le verbe s'accorde avec le sujet en genre et en nombre. Si l'ordre est VSO, le verbe s'accorde uniquement en genre.

**Exemple**

(The boy writes the lesson) → (الولد يكتب الدرس) or (يكتب الولد الدرس)

(The boys write the lesson) → (الأولاد يكتبون الدرس) or (يكتب الأولاد الدرس)

La forme la plus courante en arabe est la forme VSO. L'accord du verbe avec son sujet est expliqué ci-dessous

##### 3.1.1.1.  Le présent

Si $NN_c(M)$: [$DT_s$ $NN_s$ $VBZ_s$ $DT_s$ $NN_s$]   → ["يـ"+$VB_c$ …………..]

Si $NN_c(F)$: [$DT_s$ $NN_s$ $VBZ_s$ $DT_s$ $NN_s$]   → ["تـ"+$VB_c$ …………..]

**Exemple**

(The boy writes the lesson) → (يكتب الولد الدرس)

(The girl writes the lesson) → (تكتب البنت الدرس)

### 3.1.1.2. Le passé

Si $NN_c(M)$: [$DT_s$ $NN_s$ $VBD_s$ $DT_s$ $NN_s$]  → [$VB_c$ …………..]

Si $NN_c(F)$: [$DT_s$ $NN_s$ $VBD_s$ $DT_s$ $NN_s$]  → [$VB_c$ +"ت"…………..]

**Exemple**

(The boy wrote the lesson) → (كتب الولد الدرس)

(The girl wrote the lesson) → (كتبت البنت الدرس)

### 3.1.1.3. Le futur

Si $NN_c(M)$: [$DT_s$ $NN_s$ $MD_s$ $VB_s$ $DT_s$ $NN_s$]  → ["ﺳ"+ "ﻳ"+$VB_c$ …………..]

Si $NN_c(F)$: [$DT_s$ $NN_s$ $MD_s$ $VB_s$ $DT_s$ $NN_s$]  → ["ﺳ"+ "ﺗ"+$VB_c$ …………..]

**Exemple**

(The boy will write the lesson) → (سيكتب الولد الدرس)

(The girl will write the lesson) → (ستكتب البنت الدرس)

L'accord sujet-verbe et objet-verbe quand la phrase est réduite en un seul mot est effectué par l'utilisation des pronoms pour remplacer le sujet et l'objet par le biais des préfixes, suffixes et proclitiques, enclitiques.[55]

**Exemple**

(He writes the lesson) → (هو يكتب الدرس) ou (يكتب الدرس)

[$PRP_s$ $VBZ_s$ $DT_s$ $NN_s$] → ["pré"+$VB_c$+"suf" $DT_c$+ $NN_c$]

(He writes it) → (يكتبه)

[$PRP_s$ $VBZ_s$ $PRP_s$] → ["pré"+$VB_c$+"suf"+"Enc"]

On constate que dans le deuxième exemple la traduction des trois entités est réduite à un seul mot.

Alors que Systran nous donne une traduction mot à mot (هو يكتب) en trois entités, et Google donne (انه يكتب عليه) en trois entités aussi. On constate (dans ces produits en 2013) l'absence d'intégration des pronoms dans les verbes et les noms par rapport à notre approche.

Donc la traduction mot à mot est valable pour les entités nom (NN) et verbe (VB) et non pas pour les pronoms (PRP) qui plutôt s'agglutinent aux noms et aux verbes.

---

[55] Les préfixes, suffixes possibles et les proclitiques, enclitiques possibles sont représentés dans les tables 4.1, 4.2 respectivement.

### 3.1.2. Le pronom sujet

#### 3.1.2.1. Le pronom sujet lié à un verbe conjugué au présent

Le pronom (sujet) est collé au verbe conjugué au **présent** implicitement par un préfixe (أ - نـ - يـ - تـ -).

Notons qu'en arabe, le sujet est inclus dans le verbe conjugué (comme trait). Il n'est donc pas nécessaire (comme c'est le cas en anglais) de précéder le verbe conjugué de son pronom.

[PRP$_s$ VBP$_s$] → ["Pré" +VB$_c$] ou ["Pré" +VB$_c$ +"Suf"]

Si PRP$_s$ = {I, YOU(S, M), WE}: [PRP$_s$ VBP$_s$] → ["Pré" +VB$_c$].

Si PRP$_s$ ={I}: Pré={أ}:[PRP$_s$ VBP$_s$] → ["أ" +VB$_c$]

Si PRP$_s$ ={YOU(S,M)}: Pré={تـ}:[PRP$_s$ VBP$_s$] → ["تـ" +VB$_c$]

Si PRP$_s$ ={WE}: Pré={نـ}:[PRP$_s$ VBP$_s$] → ["نـ" +VB$_c$]

Si PRP$_s$ = {HE, SHE}: [PRP$_s$ VBZ$_s$] → ["Pré" +VB$_c$].

Si PRP$_s$ ={HE}: Pré={يـ}:[PRP$_s$ VBZ$_s$] → ["يـ" +VB$_c$]

Si PRP$_s$ ={SHE}: Pré={تـ}:[PRP$_s$ VBZ$_s$] → ["تـ" +VB$_c$]

Si PRP$_s$ = {YOU(S,F), YOU(P, M), YOU(B), YOU(P, F), THEY(B,M), THEY(B,F), THEY(M), THEY(F)}: [PRP$_s$ VBP$_s$] → ["Pré" +VB$_c$ +"Suf"]

Si PRP$_s$ ={YOU(S,F)}: pré ={تـ} & suf ={ين}:[PRP$_s$ VBP$_s$] → ["تـ" +VB$_c$ +"ين"]

Si PRP$_s$ ={YOU(B)}: pré ={تـ} & suf ={ان}:[PRP$_s$ VBP$_s$] → ["تـ" +VB$_c$ +"ان"]

Si PRP$_s$ ={YOU(P,M)}: Pré={تـ} & suf ={ون}:[PRP$_s$ VBP$_s$] → ["تـ" +VB$_c$ +"ون"]

Si PRP$_s$ ={YOU(P,F)}: Pré={تـ} & suf ={ن}:[PRP$_s$ VBP$_s$] → ["تـ" +VB$_c$ +"ن"]

Si PRP$_s$ ={THEY(B,M)}: pré ={يـ} & suf ={ان}:[PRP$_s$ VBP$_s$] → ["يـ" +VB$_c$ +"ان"]

Si PRP$_s$ ={THEY(B,F)}: pré ={تـ} & suf ={ان}:[PRP$_s$ VBP$_s$] → ["تـ" +VB$_c$ +"ان"]

Si PRP$_s$ ={THEY(M)}: pré ={يـ} & suf ={ون}:[PRP$_s$ VBP$_s$] → ["يـ" +VB$_c$ +"ون"]

Si PRP$_s$ ={THEY(F)}: pré ={تـ} & suf ={ن}:[PRP$_s$ VBP$_s$] → ["تـ" +VB$_c$ +"ن"]

Sachant que les traits masculin (M) et féminin (F) duel ou both (B) n'existe pas en anglais, mais ils sont déduits selon le nom cible.

Le pronom sujet lié à un verbe conjugué au **présent** est traduit comme illustré dans la table 5.6.

**Table 5.6 Traduction des pronoms liés à un verbe conjugué au présent**

| Mot source | Mot cible |
|---|---|
| I  VBPs | "أَ"+ VBc |
| You(S, M)  VBPs | "تَ"+ VBc |
| You(S, F)  VBPs | "تَ"+VBc+"ين" |
| You(B)  VBPs | "تَ"+VBc+"ان" |
| You(P, M)  VBPs | "تَ"+VBc+"ون" |
| You(P, F)  VBPs | "تَ"+VBc+"ن" |
| He  VBZs | "يَ"+ VBc |
| She  VBZs | "تَ"+ VBc |
| We  VBPs | "نَ"+ VBc |
| They(B,M)  VBPs | "يَ"+ VBc+"ان" |
| They(B,F)  VBPs | "تَ"+ VBc+"ان" |
| They(M)  VBPs | "يَ"+ VBc+"ون" |
| They(F)  VBPs | "يَ"+VBc+"ن" |

**Exemple**

La table 5.7 illustre la règle précédente.

**Table 5.7 Traduction implicite des pronoms en arabe**

| Verbe source | Verbe cible |
|---|---|
| I write | أكتبُ |
| You write | تكتبُ/تكتبين/تكتبان/تكتبون/تكتبن |
| He writes | يكتبُ |
| She writes | تكتبُ |
| We write | نكتبُ |
| They write | يكتبان/تكتبان/يكتبون/يكتبن |

Alors que sur Systran, on constate que les pronoms sont réécrits explicitement avant le verbe comme illustré dans la figure 5.3.

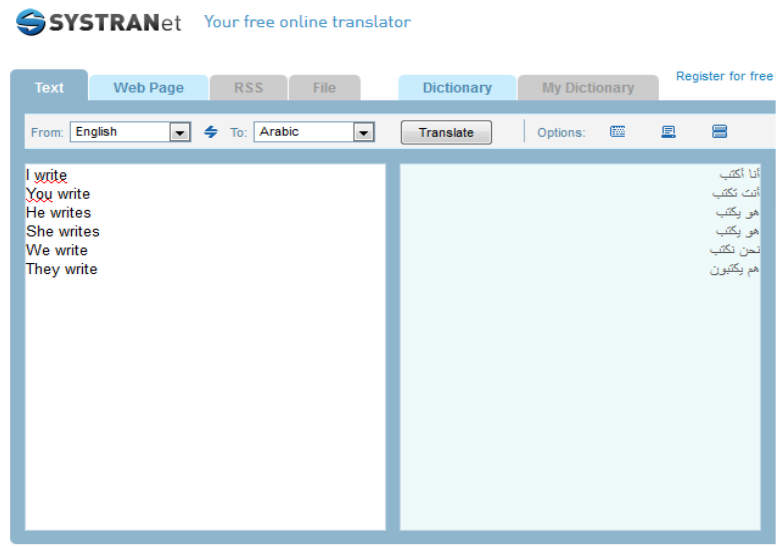

**Figure 5.3 Traduction Systran du pronom-verbe de l'anglais vers l'arabe (07/12/2013)**

### 3.1.2.2. Le pronom sujet lié à un verbe conjugué au passé

Si le verbe est conjugué à un temps **passé (VBD)** : Le pronom (sujet) est collé au verbe implicitement par un suffixe (تـ،نا،وا،ن، تم، تن) :

[PRP$_s$ VBD$_s$] → [VB$_c$ +"Suf"]

Si PRP$_s$ = {I, YOU(S), SHE}: suf= { ت }: [PRP$_s$ VBD$_s$] → [VB$_c$ +" ت "]

Si PRP$_s$ = {YOU (B)}: suf= { تما }: [PRP$_s$ VBD$_s$] → [VB$_c$ +"تما"]

Si PRP$_s$ = {YOU (P,M)}: suf= { تم }: [PRP$_s$ VBD$_s$] → [VB$_c$ +"تم "]

Si PRP$_s$ = {YOU (P,F)}: suf= { تن }: [PRP$_s$ VBD$_s$] → [VB$_c$ +"تن"]

Si PRP$_s$ = {HE}: suf=φ: [PRP$_s$ VBD$_s$] → [VB$_c$]

Si PRP$_s$ = {WE}: suf= { نا }: [PRP$_s$ VBD$_s$] → [VB$_c$ +"نا "]

Si PRP$_s$ = {THEY(B,M)}: suf= { ا }: [PRP$_s$ VBD$_s$] → [VB$_c$ +"ا"]

Si PRP$_s$ = {THEY(B,F)}: suf= { تا }: [PRP$_s$ VBD$_s$] → [VB$_c$ +"تا"]

Si PRP$_s$ = {THEY(M)}: suf= {وا}: [PRP$_s$ VBD$_s$] → [VB$_c$ +"وا "]

Si PRP$_s$ = {THEY(F)}: suf= {ن}: [PRP$_s$ VBD$_s$] → [VB$_c$ +" ن"]

Le pronom sujet lié à un verbe conjugué au passé est traduit comme illustré dans la table 5.8.

**Table 5.8 Traduction des pronoms liés à un verbe conjugué au passé**

| Mot source | Mot cible |
|---|---|
| I  VBD$_s$ | VB$_c$+ "ت" |
| You(S)  VBD$_s$ | VB$_c$+ "ت" |
| You(B)  VBD$_s$ | VB$_c$+ "تما" |
| You(P,M)  VBD$_s$ | VB$_c$+ "تم" |
| You(P,F)  VBD$_s$ | VB$_c$+ "تن" |
| He  VBD$_s$ | VB$_c$ |
| She  VBD$_s$ | VB$_c$+ "ت" |
| We  VBD$_s$ | VB$_c$+ "نا" |
| They(B,M)  VBD$_s$ | VB$_c$+ "ا" |
| They(B,F)  VBD$_s$ | VB$_c$+ "تا" |
| They(M)  VBD$_s$ | VB$_c$+ "وا" |
| They(F)  VBD$_s$ | VB$_c$+ "ن" |

### 3.1.2.3.  Le pronom sujet lié à un verbe conjugué au futur

Le pronom sujet lié à un verbe conjugué au **futur** est traduit comme dans le cas des verbes conjugués au présent précédé du préfixe "ســ" ou du mot "سوف".

[PRP$_s$ MD$_s$ VB$_s$] → ["ســ" +"Pré" +VB$_c$] ou ["ســ" +"Préf" +VB$_c$ +"Suf"]

Si PRP$_s$ = {I, YOU (S,M), HE,SHE, WE}: [PRP$_s$ MD$_s$ VB$_s$] → ["ســ" +"Pré" +VB$_c$]

Si PRP$_s$ = {I}: pré= {أ}: [PRP$_s$ MD$_s$ VB$_s$]→ ["ســ" +"أ " +VB$_c$]

Si PRP$_s$ = {YOU(S,M)}: pré= {تـ}: [PRP$_s$ MD$_s$ VB$_s$]→ ["ســ" +"تـ " +VB$_c$]

Si PRP$_s$ = {HE}: pré= {يـ}: [PRP$_s$ MD$_s$ VB$_s$]→ ["ســ" +"يـ " +VB$_c$]

Si PRP$_s$ = {SHE}: pré= {تـ}: [PRP$_s$ MD$_s$ VB$_s$]→ ["ســ" +"تـ " +VB$_c$]

Si PRP$_s$ = {WE}: pré= {نـ}: [PRP$_s$ MD$_s$ VB$_s$]→ ["ســ" +"نـ " +VB$_c$]

Si PRP$_s$ = {YOU (S, F), YOU(P,M),YOU(P,F), THEY(M), THEY(F)}: [PRP$_s$ MD$_s$ VB$_s$] → ["ســ" +"Pré" +VB$_c$ +"Suf"]

Si PRP$_s$ ={YOU(S,F)}:pré= {تـ} & suf={ين}:[PRP$_s$ MD$_s$ VB$_s$]→["ســ" +"تـ" +VB$_c$+"ين"]

Si PRP$_s$ ={YOU(B)}:pré= {تـ} & suf={ان}:[PRP$_s$ MD$_s$ VB$_s$] ➔ ["سـ" +"تـ" +VB$_c$ +"ان"]

Si PRP$_s$ ={YOU(P,M)}:pré={تـ}& suf={ون}:[PRP$_s$ MD$_s$ VB$_s$]➔["سـ" +"تـ"+VB$_c$+"ون"]

Si PRP$_s$ = {YOU(P,F)}:pré= {تـ} & suf={ن}:[PRP$_s$ MD$_s$ VB$_s$] ➔ ["سـ" +"تـ"+VB$_c$+"ن"]

Si PRP$_s$={THEY(B,M)}:pré={يـ}& suf={ان}:[PRP$_s$ MD$_s$ VB$_s$]➔["سـ" +"يـ"+VB$_c$+"ان"]

Si PRP$_s$={THEY(B,F)}:pré={تـ}& suf={ان}:[PRP$_s$ MD$_s$ VB$_s$]➔["سـ" +"تـ"+VB$_c$+"ان"]

Si PRP$_s$ ={THEY(M)}:pré={يـ} & suf={ون}:[PRP$_s$ MD$_s$ VB$_s$]➔["سـ" +"يـ"+VB$_c$+"ون"]

Si PRP$_s$ = {THEY(F)}:pré= {يـ} & suf={ن}:[PRP$_s$ MD$_s$ VB$_s$] ➔ ["سـ" +"يـ" +VB$_c$ +"ن"]

La table 5.9 résume tous les cas avec tous les pronoms.

**Table 5.9 Traduction des pronoms liés à un verbe conjugué au futur**

| Mot source | Mot cible |
|---|---|
| I will VB$_s$ | "سـ"+ "أ"+ VB$_c$ |
| You(S,M) will VB$_s$ | "سـ"+ "تـ"+ VB$_c$ |
| You(S,F) will VB$_s$ | "سـ"+ "تـ"+VB$_c$+"ين" |
| You(B) will VB$_s$ | "سـ"+ "تـ"+VB$_c$+"ان" |
| You(P,M) will VB$_s$ | "سـ"+ "تـ"+VB$_c$+"ون" |
| You(P,F) will VB$_s$ | "سـ"+ "تـ"+VB$_c$+"ن" |
| He will VB$_s$ | "سـ"+ "يـ"+ VB$_c$ |
| She will VB$_s$ | "سـ"+ "تـ"+ VB$_c$ |
| We will VB$_s$ | "سـ"+ "نـ"+ VB$_c$ |
| They(B,M) will VB$_s$ | "سـ"+ "يـ"+VB$_c$+"ان" |
| They(B,F) will VB$_s$ | "سـ"+ "تـ"+VB$_c$+"ان" |
| They(M) will VB$_s$ | "سـ"+ "يـ"+VB$_c$+"ون" |
| They(F) will VB$_s$ | "سـ"+ "يـ"+VB$_c$+"ن" |

### 3.1.3. Le pronom objet

Pour le pronom objet le temps du verbe n'a pas d'influence sur la terminaison du verbe cible. Toutes les formes des verbes sont traitées comme verbes simples :

{[PRP VBP PRP], [PRP VBZ PRP], [PRP VBD PRP], [PRP MD VB PRP], [PRP VBZ VBG PRP], [PRP VBZ VBN PRP]} ➔ {[PRP VB PRP]}

Le pronom (objet) : (complément d'objet direct) est collé au verbe s'il est représenté dans la phrase source par un pronom.

La traduction pour les pronoms objet s'effectue comme suit :

[$VB_s$ $PRP_s$] → ["Pré"+$VB_c$+"suf" +"Enc"]

Si $PRP_s$= {me}→ Enc={ني}: [$VB_s$ $PRP_s$] → ["Pré"+$VB_c$+"suf"+ "ني"]

Si $PRP_s$= {you(S)}→ Enc={ك}: [$VB_s$ $PRP_s$] → ["Pré"+$VB_c$+"suf"+ "ك"]

Si $PRP_s$= {you(B)}→ Enc={كما}: [$VB_s$ $PRP_s$] → ["Pré"+$VB_c$+"suf"+ "كما"]

Si $PRP_s$= {you(P,M)}→ Enc={كم}: [$VB_s$ $PRP_s$] → ["Pré"+$VB_c$+"suf"+ "كم"]

Si $PRP_s$= {you(P,F)}→ Enc={كن}: [$VB_s$ $PRP_s$] → ["Pré"+$VB_c$+"suf"+ "كن"]

Si $PRP_s$= {him, it}→ Enc={ـه}: [$VB_s$ $PRP_s$] → ["Pré"+$VB_c$+"suf"+ "ـه"]

Si $PRP_s$= {her}→ Enc={ها}: [$VB_s$ $PRP_s$] → ["Pré"+$VB_c$+"suf"+ "ها"]

Si $PRP_s$= {us}→ Enc={نا}: [$VB_s$ $PRP_s$] → Enc={نا} ["Pré"+$VB_c$+"suf"+ "نا"]

Si $PRP_s$= {them(B)}→ Enc={هما} : [$VB_s$ $PRP_s$] → ["Pré"+$VB_c$+"suf"+ "هما"]

Si $PRP_s$={them(M)}&(hmn="vrai")→Enc={هم}: [$VB_s$ $PRP_s$] → ["Pré"+$VB_c$+"suf"+"هم"]

Si $PRP_s$={them(F)}&(hmn="vrai")→Enc={هن}:[$VB_s$$PRP_s$]→ ["Pré"+$VB_c$+"suf"+"هن"]

Si $PRP_s$={them} & (hmn= "faux")→ Enc={ها}:[$VB_s$ $PRP_s$]→["Pré"+$VB_c$+"suf"+ "ها"]

La table 5.10 résume les différents cas des traductions des pronoms objets que ce soit le temps du verbe (**présent, passé, futur**).

**Table 5.10 Traduction des pronoms objets**

| Pronom source | Pronom cible |
|---|---|
| $VB_s$ me | "Pré"+$VB_c$+"suf"+"**ني**" |
| $VB_s$ **you**(S) | "Pré"+$VB_c$+"suf"+"**ك**" |
| $VB_s$ **you**(B) | "Pré"+$VB_c$+"suf"+"**كما**" |
| $VB_s$ **you**(P,M) | "Pré"+$VB_c$+"suf"+"**كم**" |
| $VB_s$ **you**(P,F) | "Pré"+$VB_c$+"suf"+"**كن**" |
| $VB_s$ **him** | "Pré"+$VB_c$+"suf"+"**ـه**" |
| $VB_s$ **her** | "Pré"+$VB_c$+"suf"+"**ها**" |



| | |
|---|---|
| VB$_s$ **it** | "Pré"+VB$_c$+"suf"+"ـه" |
| VB$_s$ **us** | "Pré"+VB$_c$+"suf"+"نا" |
| VB$_s$ **them**(B) | "Pré"+VB$_c$+"suf"+"هما" |
| VB$_s$ **them**(M) | "Pré"+VB$_c$+"suf"+"هم" |
| VB$_s$ **them**(F) | "Pré"+VB$_c$+"suf"+"هن" |
| VB$_s$ **them** (hmn= "faux") | "Pré"+VB$_c$+"suf"+"ها" |

La liste des préfixes et des suffixes qui correspond aux différents cas des pronoms est la même expliquée auparavant, pour les verbes conjugués au présent : (la table 5.6, les verbes conjugués au passé : la table 5.8, les verbes conjugués au futur : la table 5.9).

**Remarque**

Il y a des pronoms personnels qui ne peuvent pas subir l'action réfléchie. Ces cas ne figurent ni en langue source ni en langue cible.

**Exemple**

(I give me, We give me, I give us, You give you, …).

Mais plutôt ce type d'expression est exprimé par des pronoms réfléchis (myself, yourself, himself,…).

### 3.1.4. La forme négative

La traduction des phrases négatives est différente de celles affirmatives. La négation engendre des changements morphologiques selon le temps du verbe comme suit :

#### 3.1.4.1. Accord verbe/sujet

**Le présent**

Si NN$_c$(M): [DT$_s$ NN$_s$ VBZ$_s$ RB$_s$ VB$_s$ DT$_s$ NN$_s$] → ["لا" "يـ"+VB$_c$ …………..]

Si NN$_c$(F): [DT$_s$ NN$_s$ VBZ$_s$ RB$_s$ VB$_s$ DT$_s$ NN$_s$] → ["لا" "تـ"+VB$_c$ …………..]

Tel que VBZ$_s$="does", RB$_s$= "not".

**Le passé**

Si NN$_c$(M): [DT$_s$ NN$_s$ VBD$_s$ RB$_s$ VB$_s$ DT$_s$ NN$_s$] → ["لم" "يـ"+VB$_c$ …………..]

Si NN$_c$(F): [DT$_s$ NN$_s$ VBD$_s$ RB$_s$ VB$_s$ DT$_s$ NN$_s$] → ["لم" "تـ"+VB$_c$ …………..]

Tel que VBD$_s$="did", RB$_s$="not".



**Le futur**

Si NN$_c$(M): [DT$_s$ NN$_s$ MD$_s$ RB$_s$ VB$_s$ DT$_s$ NN$_s$]  → ["لن" "يـ"+VB$_c$ …………..]

Si NN$_c$(F): [DT$_s$ NN$_s$ MD$_s$ RB$_s$ VB$_s$ DT$_s$ NN$_s$]  → ["لن" "تـ"+VB$_c$ …………..]

Tel que MD$_s$="will" ou "shall", RB$_s$="not".

### 3.1.4.2. Accord verbe /pronoms au présent

Si le verbe est conjugué au **présent** et précédé d'une marque de négation, le verbe en langue cible sera précédé de la particule (لا) et aucun changement au niveau du verbe :

[PRP$_s$ VBP$_s$ RB$_s$ VB$_s$] → ["لا"   pré+ VB$_c$] ou ["لا"   préf+ VB$_c$+ "suf"]

[PRP$_s$ VBZ$_s$ RB$_s$ VB$_s$] → ["لا"   pré+ VB$_c$]

Si PRP$_s$ = {I, YOU(S,M), WE}: [PRP$_s$ VBP$_s$ RB$_s$ VB$_s$] → ["لا"  "Pré" +VB$_c$]

Si PRP$_s$= {He, She}: [PRP$_s$ VBZ$_s$ RB$_s$ VB$_s$] → ["لا"   pré+ VB$_c$]

Si PRP$_s$ = {YOU(S,F), YOU(P, M), YOU(B), YOU(P, F), THEY(B,M),THEY(B,F), THEY(M), THEY(F)}: [PRP$_s$ VBP$_s$ RB$_s$ VB$_s$] → ["لا"  "Pré" +VB$_c$ +"Suf"]

Tel que VBP$_s$ ="do", RB$_s$= "not", VBZ$_s$="does".

La liste des préfixes et des suffixes en arabe qui correspond aux différents pronoms en anglais est la même qu'avec celle de conjugaison du verbe au présent dans la table 5.6.

La table 5.11 résume tous les cas des pronoms liés à des verbes conjugués au présent dans des phrases négatives

**Table 5.11 Traduction des pronoms liés à un verbe au présent dans la forme négative**

| Mot source | Mot cible |
|---|---|
| I  do  not  VB$_s$ | "لا" "أ"+VB$_c$ |
| You(S,M) do not VB$_s$ | "لا" "تـ"+VB$_c$ |
| You(S,F) do not VB$_s$ | "لا" "تـ"+VB$_c$+"ين" |
| You(B) do not VB$_s$ | "لا" "تـ"+VB$_c$+"ان" |
| You(P,M) do not VB$_s$ | "لا" "تـ"+VB$_c$+"ون" |
| You(P,F) do not VB$_s$ | "لا" "تـ"+VB$_c$+"ن" |
| He does not VB$_s$ | "لا" "يـ"+VB$_c$ |
| She does not VB$_s$ | "لا" "تـ"+VB$_c$ |
| We do not VB$_s$ | "لا" "نـ"+VB$_c$ |
| They(B,M) do not VB$_s$ | "لا" "يـ"+VB$_c$+"ان" |



| They(B,F) do not VB$_s$ | "ان"+VB$_c$+"تـ" "لا" |
| They(M) do not VB$_s$ | "ون"+VB$_c$+"يـ" "لا" |
| They(F) do not VB$_s$ | "ن"+VB$_c$+"يـ" "لا" |

**Exemple**

(I do not write) → (لا أكتب)

### 3.1.4.3. Accord verbe /pronoms au passé

Si le verbe est conjugué au **passé** et précédé d'une marque de négation, le verbe en langue cible sera précédé de la particule (لم) avec des changements au niveau du verbe :

[PRP$_s$ VBD$_s$ RB$_s$ VB$_s$] → ["لم" "pré"+ VB$_c$] ou ["لم" "pré"+ VB$_c$+ "suf"]

Tel que VBD$_s$="did", RB$_s$="not".

Si PRP$_s$= {I, You(S, M), He, She, We, They(M) } [PRP$_s$ VBD$_s$ RB$_s$ VB$_s$] → ["لم" "pré"+ VB$_c$]

Si PRP$_s$= {I}: Pré= {أ} : [PRP$_s$ VBD$_s$ RB$_s$ VB$_s$] → ["لم" "أ"+ VB$_c$]

Si PRP$_s$= {You(S, M)}: Pré={تـ}: [PRP$_s$ VBD$_s$ RB$_s$ VB$_s$] → ["لم" "تـ"+ VB$_c$]

Si PRP$_s$= {He}: Pré={يـ}: [PRP$_s$ VBD$_s$ RB$_s$ VB$_s$] → ["لم" "يـ"+ VB$_c$]

Si PRP$_s$= {She}: Pré={تـ}: [PRP$_s$ VBD$_s$ RB$_s$ VB$_s$] → ["لم" "تـ"+ VB$_c$]

Si PRP$_s$= {We}: Pré={نـ}: [PRP$_s$ VBD$_s$ RB$_s$ VB$_s$] → ["لم" "نـ"+ VB$_c$]

Si PRP$_s$= {You(S,F), You(P,M), You(P,F), You(B), They(F)} [PRP$_s$ VBD$_s$ RB$_s$ VB$_s$] → ["لم" "pré"+ VB$_c$+ "suf"]

Si PRP$_s$= {You(S,F)}: Pré={تـ} & suf ={ي}: [PRP$_s$ VBD$_s$ RB$_s$ VB$_s$] → ["لم" "تـ"+ VB$_c$+ "ي"]

Si PRP$_s$= {You(B)}: Pré={تـ} & suf ={ا}: [PRP$_s$ VBD$_s$ RB$_s$ VB$_s$] → ["لم" "تـ"+ VB$_c$+ "ا"]

Si PRP$_s$= {You(P,M)}:Pré={تـ} & suf ={وا}: [PRP$_s$ VBD$_s$ RB$_s$ VB$_s$]→["لم" "تـ"+ VB$_c$+ "وا"]

Si PRP$_s$= {You(P,F)}: Pré={تـ} & suf ={ن}: [PRP$_s$ VBD$_s$ RB$_s$ VB$_s$] → ["لم" "تـ"+ VB$_c$+"ن"]

Si PRP$_s$= {They(B,M)}: Pré={يـ} & suf ={ا} [PRP$_s$ VBD$_s$ RB$_s$ VB$_s$] → ["لم" "يـ"+ VB+ "ا"]

Si PRP$_s$= {They(B,F)}: Pré={تـ} & suf ={ا} [PRP$_s$ VBD$_s$ RB$_s$ VB$_s$] → ["لم" "تـ"+ VB+ "ا"]

Si PRP$_s$= {They(M)}: Pré={يـ} & suf ={وا} [PRP$_s$ VBD$_s$ RB$_s$ VB$_s$] → ["لم" "يـ"+ VB+ "وا"]

Si PRP$_s$={They(F)}: Pré={يـ} & suf ={ن}: [PRP$_s$ VBD$_s$ RB$_s$ VB$_s$] → ["لم" "تـ"+ VB$_c$+"ن"]



La table 5.12 résume tous les cas des pronoms liés à des verbes conjugués au passé dans des phrases négatives

**Table 5.12 Traduction des pronoms liés à un verbe au passé dans la forme négative**

| Mot source | Mot cible |
|---|---|
| I did not VB$_s$ | "لم" "أ"+VB$_c$ |
| You(S,M) did not VB$_s$ | "لم" "تَ"+VB$_c$ |
| You(S,F) did not VB$_s$ | "لم" "تَ"+VB$_c$+"ي" |
| You(B) did not VB$_s$ | "لم" "تَ"+VB$_c$+"ا" |
| You(P,M) did not VB$_s$ | "لم" "تَ"+VB$_c$+"وا" |
| You(P,F) did not VB$_s$ | "لم" "تَ"+VB$_c$+"ن" |
| He did not VB$_s$ | "لم" "يَ"+VB$_c$ |
| She did not VB$_s$ | "لم" "تَ"+VB$_c$ |
| We did not VB$_s$ | "لم" "نَ"+VB$_c$ |
| They(B,M) did not VB$_s$ | "لم" "يَ"+VB$_c$+"ا" |
| They(B,F) did not VB$_s$ | "لم" "تَ"+VB$_c$+"ا" |
| They(M) did not VB$_s$ | "لم" "يَ"+VB$_c$+"وا" |
| They(F) did not VB$_s$ | "لم" "يَ"+VB$_c$+"ن" |

**Exemple**

(I did not write) → (لم أكتب)

### 3.1.4.4. Accord verbe /pronoms au futur

Si le verbe est conjugué au futur et précédé d'une marque de négation, le verbe en langue cible sera précédé de la particule (لن) avec les mêmes changements au niveau du verbe qu'avec les verbes conjugués au passé :

[PRP$_s$ MD$_s$ RB$_s$ VB$_s$] → ["لن" "pré"+ VB] ou ["لن" "pré"+ VB$_c$+ "suf"]

Tel que MD$_s$="will" ou "shall", RB$_s$="not".

Si PRP$_s$= {I, You(S,M), He, She, We} [PRP$_s$ VBD$_s$ RB$_s$ VB$_s$] → ["لن" "pré"+ VB$_c$]

Si PRP$_s$= {You(S,F), You(B), You(P,M), You(P,F), They(B,M), They(B,F), They(M), They(F)} [PRP$_s$ MD$_s$ RB$_s$ VB$_s$] → ["لن" "pré"+ VB$_c$+"suf"]

La liste des préfixes et des suffixes en arabe qui correspond aux différents pronoms en anglais est la même qu'avec celle des pronoms liés à des verbes conjugués au passé dans des phrases négatives dans la table 5.12.



**Exemple**

(I will not write) → (لن أكتب)

### 3.1.4.5. Les pronoms d'objet

La traduction des pronoms d'objets dans des phrases négatives s'effectue de la même façon qu'avec les phrases affirmatives, tout en précédant la phrase par une particule de négation (لا، لن، لم). La liste des enclitiques est présentée dans la table 5.10.

## 3.1.5. La forme Interrogative

Les phrases interrogatives sont des phrases affirmatives précédées par les verbes (do, does, did, will). La traduction vers l'arabe rajoute à la phrase cible la particule "هل", ou le préfixe "أ", tout en traduisant les pronoms sources en préfixes et suffixes dans la langue cible comme suit :

### 3.1.5.1. Accord verbe/sujet

**Le présent**

Si $NN_c(M)$: [$VBZ_s$ $DT_s$ $NN_s$ $VB_s$ $DT_s$ $NN_s$?]   → ["هل" "يـ"+$VB_c$ …………..]

Si $NN_c(F)$: [$VBZ_s$ $DT_s$ $NN_s$ $VB_s$ $DT_s$ $NN_s$?]   → ["هل" "تـ"+$VB_c$ …………..]

Tel que $VBZ_s$="Does".

**Le passé**

Si $NN_c(M)$: [$VBD_s$ $DT_s$ $NN_s$ $VB_s$ $DT_s$ $NN_s$?]   → ["هل" $VB_c$…………..]

Si $NN_c(F)$: [$VBD_s$ $DT_s$ $NN_s$ $VB_s$ $DT_s$ $NN_s$?]   → ["هل" $VB_c$ +"ت" …………..]

Tel que $VBD_s$ = "did"

**Le futur**

Si $NN_c(M)$: [$MD_s$ $DT_s$ $NN_s$ $VB_s$ $DT_s$ $NN_s$?]   → ["هل" "سـ"+"يـ"+$VB_c$ …………..]

Si $NN_c(F)$: [$MD_s$ $DT_s$ $NN_s$ $VB_s$ $DT_s$ $NN_s$?]   → ["هل" "سـ"+"تـ"+$VB_c$ …………..]

Tel que $MD_s$="will"ou "shall".

### 3.1.5.2. Accord verbe /pronoms au Présent

Si le verbe de la phrase est conjugué au **présent**, la traduction sera comme suit :

[$VBP_s$ $PRP_s$ $VB_s$?] → ["هل"  "pré"+$VB_c$ ؟] ou ["هل"  "pré"+$VB_c$+"suf"  ؟]

[$VBZ_s$ $PRP_s$ $VB_s$?] → ["هل"  "pré"+$VB_c$  ؟]

Tel que $VBP_s$="Do",  $VBZ_s$="Does".



Si PRP$_s$={ I, YOU(S,M),WE}:[VBP$_s$ PRP$_s$ VB$_s$?]→["هل" "pré"+VB$_c$ ؟]

Si PRP$_s$={He, She}:[VBZ$_s$ PRP$_s$ VB$_s$?] → ["هل" "pré"+VB$_c$ ؟]

Si PRP$_s$ = {YOU(S,F), YOU(P, M), YOU(B), YOU(P, F), THEY(B,M), THEY(B,F), THEY(M), THEY(F)}: [VBP$_s$ PRP$_s$ VB$_s$?] → ["هل" "pré"+VB$_c$+"suf" ؟]

La liste des préfixes et des suffixes en arabe qui correspond aux différents pronoms en anglais est la même qu'avec celle de conjugaison du verbe au présent dans la table 5.6.

**Exemple**

(Do you write?) → (هل تكتب ؟)

### 3.1.5.3. Accord verbe /pronoms au passé

Si le verbe de la phrase est conjugué au **passé**, la traduction sera comme suit :

[VBD$_s$ PRP$_s$ VB$_s$?] → ["هل" VB$_c$+"suf" ؟]

Tel que VBD$_s$ = "did"

$\forall$ PRPi ∈ PRPs : [VBD$_s$ PRP$_s$ VB$_s$?] → ["هل" VB$_c$+"suf" ؟]

La liste des suffixes en arabe qui correspond aux différents pronoms en anglais est la même qu'avec celle de conjugaison du verbe au passé dans la table 5.8.

**Exemple**

(Did you write?) → (هل كتبت ؟)

### 3.1.5.4. Accord verbe /pronoms au futur

Si le verbe de la phrase est conjugué au **futur**, la traduction sera comme suit :

[MD$_s$ PRP$_s$ VB$_s$] → ["هل" "سـ" + "pré"+VB] ou ["هل" "سـ" + "pré"+VB$_c$+"suf"]

Tel que MD$_s$="will"ou "shall".

Si PRP$_s$={I, YOU(S,M), HE,SHE, WE}:[MD$_s$ PRP$_s$ VB$_s$?]→["هل" "سـ"+"pré"+VB$_c$ ؟]

Si PRP$_s$={YOU (S,F), YOU(P,M),YOU(P,F), THEY(B,M), THEY(B,F), THEY(M), THEY(F)}: [MD$_s$ PRP$_s$ VB$_s$?] → ["هل" "سـ" + "pré"+ VB$_c$+ "suf" ؟]

La liste des préfixes et des suffixes en arabe qui correspond aux différents pronoms en anglais est la même qu'avec celle de conjugaison du verbe au futur dans la table 5.9.



**Exemple**

(Will you write?) → (هل ستكتب ؟) ou (أستكتب ؟)

### 3.1.5.5. Les pronoms d'objet

La traduction des pronoms d'objets dans des phrases interrogatives s'effectue de la même façon qu'avec les phrases affirmatives, tout en précédant la phrase par une particule d'interrogation (هل). La liste des enclitiques est présentée dans la table 5.10.

## 3.1.6. Les verbes irréguliers

Dans cette section on présente quelques[56] règles de transfert pour les verbes irréguliers (dits malades). Les règles citées auparavant ne s'appliquent qu'aux verbes réguliers.

Les racines malades sont des verbes trilitères qui comportent une des lettres {ا،و،ي}, on peut les catégoriser en trois catégories selon la position de la lettre malade dans le verbe.

1- Le verbe assimilé : est un verbe qui commence par une lettre malade, comme :
(وصل، وقف، وزن)

2- Le verbe concave : est un verbe qui contient une lettre malade au milieu, comme :
(قال، نام، باع)

3- Le verbe défectueux : est un verbe qui se termine par une lettre malade, comme :
(دعا، رمى، نهى)

Dans la génération des verbes malades lors du transfert il y a des lettres qui disparaissent et d'autres qui apparaissent. Pour montrer quelle lettre disparaisse chaque fois, on ne présente pas le verbe cible en VBc mais on expose plus de détails avec les lettres du verbe : F3L. Pour chaque lettre {ا،و،ي} il y a des tables de conjugaison, ainsi pour la même lettre il y a trois possibilités selon la prononciation du verbe (selon les signes diacritiques) : FA3ALA, FA3ILA, FA3OLA

### 3.1.6.1. Le verbe assimilé

On présente le cas des verbes de type FA3ALA : F= {و}

**Le pronom sujet lié à un verbe assimilé conjugué au présent**

Le pronom sujet lié à un verbe assimilé de type FA3ALA : F= {و} conjugué au **présent** est traduit comme illustré dans la table 5.13.

---

[56] Le même principe est applicable pour les autres verbes selon leurs structures. Il ne nous est pas possible de développer davantage ou de répéter les règles des verbes et des noms irréguliers dans le cadre restreint de ce manuscrit.



Table 5.13 Traduction des pronoms liés à un verbe assimilé FA3ALA : F= {و} au présent

| Mot source | Mot cible |
|---|---|
| I  VBP$_s$ | "أ"+ 3L |
| You(S, M)  VBP$_s$ | "تـ"+3L |
| You(S, F)  VBP$_s$ | "تـ"+3L+"ين" |
| You(B)  VBP$_s$ | "تـ"+3L+"ان" |
| You(P, M)  VBP$_s$ | "تـ"+3L+"ون" |
| You(P, F)  VBP$_s$ | "تـ"+3L+"ن" |
| He  VBZ$_s$ | "يـ"+3L |
| She  VBZ$_s$ | "تـ"+3L |
| We  VBP$_s$ | "نـ"+ 3L |
| They(B,M)  VBP$_s$ | "يـ"+3L +"ان" |
| They(B,F)  VBP$_s$ | "تـ"+ 3L+"ان" |
| They(M)  VBP$_s$ | "يـ"+3L +"ون" |
| They(F)  VBP$_s$ | "يـ"+ 3L+"ن" |

**Le pronom sujet lié à un verbe assimilé conjugué au passé**

Le pronom sujet lié à un verbe assimilé de type FA3ALA : F= {و} conjugué au **passé** est traduit de la même façon des verbes réguliers, comme illustré dans la table 5.8.

**Le pronom sujet lié à un verbe assimilé conjugué au futur**

Le pronom sujet lié à un verbe assimilé de type FA3ALA : F= {و} conjugué au **futur** est traduit de la même façon pour les verbes assimilés conjugués au présent on ajoutant le préfixe "سـ".

### 3.1.6.2. Le verbe concave

On présente le cas des verbes de type F3L : 3= {ا}, selon le modèle de (قام)[57]

**Le pronom sujet lié à un verbe concave conjugué au présent**

Le pronom sujet lié à un verbe concave de type F3L : 3= {ا}, selon le modèle de (قام) conjugué au **présent** est traduit comme illustré dans la table 5.14.

Table 5.14 Traduction des pronoms liés à un verbe concave F3L: 3= {ا}, modèle (قام) au présent

| Mot source | Mot cible |
|---|---|
| I  VBP$_s$ | "أ"+ F+"و"+L |
| You(S, M)  VBP$_s$ | "تـ"+ F+"و"+L |
| You(S, F)  VBP$_s$ | "تـ"+ F+"و"+L +"ين" |
| You(B)  VBP$_s$ | "تـ"+ F+"و"+L +"ان" |
| You(P, M)  VBP$_s$ | "تـ"+ F+"و"+L +"ون" |
| You(P, F)  VBP$_s$ | "تـ"+ F+L +"ن" |
| He  VBZ$_s$ | "يـ"+ F+"و"+L |
| She  VBZ$_s$ | "تـ"+ F+"و"+L |
| We  VBP$_s$ | "نـ"+ F+"و"+L |

---

[57] Les verbes concaves peuvent être de types (قام) qui donne (يقوم) ou de type (باع) qui donne (يبيع)



| They(B,M) VBP$_s$ | "ان"+ L +"و"+F +"يـ" |
| They(B,F) VBP$_s$ | "ان"+ L +"و"+F +"تـ" |
| They(M)  VBP$_s$ | "ون"+ L +"و"+F +"يـ" |
| They(F)  VBP$_s$ | "ن"+ F+L +"يـ" |

**Le pronom sujet lié à un verbe concave conjugué au passé**

Le pronom sujet lié à un verbe concave de type F3L : 3= {ا}, selon le modèle de (قام) conjugué au **passé** est traduit comme illustré dans la table 5.15.

**Table 5.15 Traduction des pronoms liés à un verbe concave F3L: 3= {ا}, modèle (قام) au passé**

| Mot source | Mot cible |
|---|---|
| I  VBD$_s$ | FL+ "ت" |
| You(S)  VBD$_s$ | FL + "ت" |
| You(B)  VBD$_s$ | FL + "تما" |
| You(P,M)  VBD$_s$ | FL + "تم" |
| You(P,F)  VBD$_s$ | FL + "تن" |
| He  VBD$_s$ | F3L |
| She  VBD$_s$ | F3L + "ت" |
| We  VBD$_s$ | FL + "نا" |
| They(B,M)  VBD$_s$ | F3L + "ا" |
| They(B,F)  VBD$_s$ | F3L + "تا" |
| They(M)  VBD$_s$ | F3L + "وا" |
| They(F)  VBD$_s$ | FL + "ن" |

**Le pronom sujet lié à un verbe concave conjugué au futur**

Le pronom sujet lié à un verbe concave de type F3L : 3= {ا}, selon le modèle de (قام) conjugué au **futur** est traduit de la même façon pour les verbes concaves conjugués au présent on ajoutant le préfixe "سـ".

### 3.1.6.3. Le verbe défectueux

On présente le cas des verbes de type F3L : L= {ي}, selon le modèle de (نسي).

**Le pronom sujet lié à un verbe défectueux conjugué au présent**

Le pronom sujet lié à un verbe défectueux de type F3L : L= {ي}, selon le modèle de (نسي) conjugué au **présent** est traduit comme illustré dans la table 5.16.

**Table 5.16 Traduction des pronoms liés à un verbe défectueux F3L: L={ي}, modèle (نسي) au présent**

| Mot source | Mot cible |
|---|---|
| I  VBP$_s$ | "أ"+ F3+"ى" |



| You(S, M)  VBP$_s$ | "تـ"+F3+"ى" |
|---|---|
| You(S, F)  VBP$_s$ | "تـ"+VB$_c$+"ن" |
| You(B)  VBP$_s$ | "تـ"+VB$_c$+"ان" |
| You(P, M)  VBP$_s$ | "تـ"+F3+"ون" |
| You(P, F)  VBP$_s$ | "تـ"+VB$_c$+"ن" |
| He  VBZ$_s$ | "يـ"+F3+"ى" |
| She  VBZ$_s$ | "تـ"+F3+"ى" |
| We  VBP$_s$ | "نـ"+ F3+"ى" |
| They(B,M)  VBP$_s$ | "يـ"+VB$_c$+"ان" |
| They(B,F)  VBP$_s$ | "تـ"+ VB$_c$ +"ان" |
| They(M)  VBP$_s$ | "يـ"+F3 +"ون" |
| They(F)  VBP$_s$ | "يـ"+ VB$_c$ +"ن" |

**Le pronom sujet lié à un verbe défectueux conjugué au passé**

Le pronom sujet lié à un verbe défectueux de type F3L : L= {ي}, selon le modèle de (نسي) conjugué au **passé** est traduit de la même façon des verbes réguliers, comme illustré dans la table 5.8. sauf pour (They) ; VBc est remplacé par F3:

Si PRP$_s$ = {THEY(M)}: suf= {وا}: [PRP$_s$ VBD$_s$] → [F3 +"وا "]

**Le pronom sujet lié à un verbe défectueux conjugué au futur**

Le pronom sujet lié à un verbe défectueux de type F3L : L= {ي}, selon le modèle de (نسي) conjugué au **futur** est traduit de la même façon pour les verbes défectueux conjugués au présent on ajoutant le préfixe "سـ".

Le même principe est appliqué pour les autres cas, mais avec des modifications différentes. De ce fait, On se contente de ces règles pour ne pas encombrer le manuscrit, et pour éviter la répétition.

## *3.2. Le Nom*

### 3.2.1. La définition (la détermination)

On dit qu'un mot est défini ou déterminé s'il est précédé d'une marque de définition (déterminant de définition) ou défini par annexion.

[DT$_s$ NN$_s$]→ [DT$_c$+ NN$_c$] ou [NN$_c$] tel que DT$_c$ ="الـ" toujours

Si DT$_s$= "The" : [DT$_s$ NN$_s$]→ ["الـ" + NN$_c$]

Si DT$_s$= "a" : [DT$_s$ NN$_s$]→ [NN$_c$]

Si DT$_s$= ϕ : [NN$_s$]→ [NN$_c$]

**Exemple**



(The book) →(الكتاب)

(A book)→(كتاب )

(Book) → (كتاب)

### 3.2.2. Le sujet

On a expliqué les variations du verbe dans la section verbe avec les différents cas des pronoms, dans cette section on s'intéresse aux variations morphologiques du sujet.

Ici on ne s'intéresse pas à l'ordre des mots qui sera détaillé dans le chapitre 6 dans le transfert syntaxique.

Le sujet est toujours dans la forme **nominative**, si le sujet est singulier aucun suffixe n'est ajouté, si le sujet est au duel ou au pluriel, des suffixes apparaissant à la fin du sujet.

#### 3.2.2.1. Le duel

Que ce soit le temps du verbe lié à un sujet duel, le sujet cible sera suffixé de "ان" ou de "تان".

[DT$_s$ **NNS**$_s$ VBP$_s$ DT$_s$ NN$_s$]

[DT$_s$ **NNS**$_s$ VBD$_s$ DT$_s$ NN$_s$]

[DT$_s$ **NNS**$_s$ MD$_s$ VB$_s$ DT$_s$ NN$_s$]

**Le masculin**

∀ NN$_c$ (M): [… NNS$_s$ …] →[… NN$_c$+"ان" …]

**Exemple**

(The two teachers explain the lesson) → (يشرح المعلمان الدرس)

**Le féminin**

∀ NN$_c$ (F): [… NNS$_s$ …] →[… NN$_c$+"تان" …]

**Exemple**

(The two teachers explain the lesson) → (تشرح المعلمتان الدرس)

#### 3.2.2.2. Le pluriel

Que ce soit le temps du verbe lié à un sujet pluriel, le sujet cible sera suffixé de "ون" ou de "ات" s'il appartient aux noms réguliers.



Si $NN_c \in \{NNS\ rég\}$ & $NN_c(M)$: [… $NNS_s$ …] → [… $NN_c$+"ون" …]

**Exemple**

(Teachers explain the lesson) → (يشرح المعلمون الدرس)

Si $NN_c \in \{NNS\ rég\}$ & $NN_c(F)$: [… $NNS_s$ …] → [… $NN_c$+"ات" …]

**Exemple**

(Teachers explain the lesson) → (تشرح المعلمات الدرس)

**Remarque**

Si le nom n'appartient pas aux noms réguliers, une forme de pluriel irrégulier est affectée à la traduction du nom.

### 3.2.3. Le complément d'agent

Les mêmes règles de sujet sont applicables sur le complément d'agent.
[$DT_s\ NNS_s\ VBD_s\ VBN_s$] → [$VB_c\ DT_c$+$NNS_c$], tel que $VBD_s$ ="was" ou "were".

**Exemple**

(The teachers were honored) → (كُرِّم المعلمون)

### 3.2.4. Le complément d'objet direct

L'objet est toujours dans la forme **accusative**, des suffixes apparaissent à la fin du complément d'objet selon le nombre de ce dernier.

#### 3.2.4.1. Le singulier indéfini

Si le complément d'objet est indéfini ($DT_s$ ="a" ou "ϕ"), le complément d'objet cible sera suffixé de "ا"
Si $DT_{s2}$ ="a" ou "ϕ" : [$DT_{s1}\ NN_s\ VB_s\ DT_{s2}\ NN_{s2}$] → [… $NN_{c2}$+"ا"]

**Exemple**

(The boss saw a teacher) → (رأى المدير معلما)

#### 3.2.4.2. Le duel

Que ce soit le temps du verbe lié à un complément d'objet duel, l'objet cible sera suffixé de "ين" ou de "تين".



[DT$_s$ NNS$_s$ VBP$_s$ DT$_s$ **NNS$_s$**]

[DT$_s$ NNS$_s$ VBD$_s$ DT$_s$ **NNS$_s$**]

[DT$_s$ NNS$_s$ MD$_s$ VB$_s$ DT$_s$ **NNS$_s$**]

**Le masculin**

$\forall$ NN$_c$ (M): [… NNS$_s$ …] $\rightarrow$ [… NN$_c$+"ين" …]

**Exemple**

(The boss saw the two teachers) $\rightarrow$ (رأى المدير المعلمين)

**Le féminin**

$\forall$ NN$_c$ (F): [… NNS$_s$ …] $\rightarrow$ [… NN$_c$+"تين" …]

**Exemple**

(The boss saw the two teachers) $\rightarrow$ (رأى المدير المعلمتين)

### 3.2.4.3. Le pluriel

Que ce soit le temps du verbe lié à un sujet pluriel, le sujet cible sera suffixé de "ين" ou de "ات" si 'il appartient aux noms réguliers.

**Le masculin**

Si NN$_c$ ∈ {NNS rég} & NN$_c$(M): [… NNS$_s$ …] $\rightarrow$ [… NN$_c$+"ين" …]

**Exemple**

(The boss saw the teachers) $\rightarrow$ (رأى المدير المعلمين)

**Le féminin**

Si NN$_c$ ∈ {NNS rég} & NN$_c$(F): [… NNS$_s$ …] $\rightarrow$ [… NN$_c$+"ات" …]

**Exemple**

(The boss saw the teachers) $\rightarrow$ (رأى المدير المعلمات)

### 3.2.5. Le nombre

Quand un mot est précédé d'un nombre des changements sur le mot apparaissent :



- Si $CD_c = 2$ : $[CD_s\ NNS_s] \rightarrow [NN_c(B)]$

Si $CD_c = 2$ : $[CD_s\ NNS_s] \rightarrow [NN_c + "ان"]$ ou $[NN_c + "ين"]$, selon la forme du nom (nominative, accusative, génitive).

- Si $CD_c \in [3-10]$ : $[CD_s\ NNS_s] \rightarrow [CD_c\ NNS_c]$

Si $CD_c \in [3-10]$ : si $NNS_c(M) \rightarrow CD_c(F)$

: si $NNS_c(F) \rightarrow CD_c(M)$

Si $NNS_c(M)\ \&\ NN_c \in \{NNS\ rég\}$: $[CD_s\ NNS_s] \rightarrow [CD_c + "ة"\ NN_c + "ين"]$

Si $NNS_c(F)\ \&\ NN_c \in \{NNS\ rég\}$: $[CD_s\ NNS_s] \rightarrow [CD_c\ NN_c + "ات"]$

- Si $CD_c \in [11-19]$ : $[CD_s\ NNS_s] \rightarrow [CD_{c1}\ CD_{c2}\ NN_c]$

Si $CD_c \in \{11,12\}$: si $NNS_c(M) \rightarrow CD_{c1}(M)\ CD_{c2}(M)$

: si $NNS_c(F) \rightarrow CD_{c1}(F)\ CD_{c2}(F)$

Si $NN_c(M)$: $[CD_s\ NNS_s] \rightarrow [CD_{c1}\ CD_{c2}\ NN_c + "ا"]$, $CD_{c1} = \{"أحد"\}$ ou $\{"اثنا"\}$

Si $NN_c(F)$: $[CD_s\ NNS_s] \rightarrow [CD_{c1}\ CD_{c2} + "ة"\ NN_c]$, $CD_{c1} = \{"احدى"\}$ ou $\{"اثنتا", "اثنتي"\}$ selon la forme du mot (nominative, accusative, génitive)

- Si $CD_c \in [13-19]$ : si $NN_c(M) \rightarrow CD_{c1}(F)\ \&\ CD_{c2}(M)$

: si $NN_c(F) \rightarrow CD_{c1}(M)\ \&\ CD_{c2}(F)$

Si $NN_c(M)$: $[CD_s\ NNS_s] \rightarrow [CD_{c1} + "ة"\ CD_{c2}\ NN_c + "ا"]$

Si $NN_c(F)$: $[CD_s\ NNS_s] \rightarrow [CD_{c1}\ CD_{c2} + "ة"\ NN_c]$

- Si $CD_c \in \{20, 30, 40, 50, 60, 70, 80, 90\}$ : $[CD_s\ NNS_s] \rightarrow [CD_c\ NN_c]$

Nominatif :

$[CD_s\ NNS_s] \rightarrow [CD_c + "ون"\ NN_c + "ا"]$

Accusatif et génitif :

$[CD_s\ NNS_s] \rightarrow [CD_c + "ين"\ NN_c + "ا"]$

- Si $CD_c$
  $\in \{21-29, 31-39, 41-49, 51-59, 61-69, 71-79, 81-89, 91-99\}$

  $[CD_{s1}\ CD_{s2}\ NNS_s] \rightarrow [CD_{c2}\ "و"\ CDS_{c1}\ NN_c]$



Si $CD_{c2} \in \{1,2\}$ : si $NN_c(M) \rightarrow CD_{c2}(M)$ & $CDS_{c1}$

: si $NN_c(F) \rightarrow CD_{c2}(F)$ & $CDS_{c1}$

Si $NN_c(M)$: $[CD_{s1}\ CD_{s2}\ NNS_s] \rightarrow [CD_{c2}\ "و"\ CDS_{c1}\ NN_c + "ا"]$

Si $NN_c(F)$: $[CD_{s1}\ CD_{s2}\ NNS_s] \rightarrow [CD_{c2} + "ة"\ "و"\ CDS_{c1}\ NN_c]$

Si $CD_{c2} \in \{3,9\}$ : si $NN_c(M) \rightarrow CD_{c2}(F)$ & $CDS_{c1}$

: si $NN_c(F) \rightarrow CD_{c2}(M)$ & $CDS_{c1}$

Si $NN_c(M)$: $[CD_{s1}\ CD_{s2}\ NNS_s] \rightarrow [CD_{c2} + "ة"\ "و"\ CDS_{c1}\ NN_c + "ا"]$

Si $NN_c(F)$: $[CD_{s1}\ CD_{s2}\ NNS_s] \rightarrow [CD_{c2}\ "و"\ CDS_{c1}\ NN_c]$

Nominatif : $CDS_{c1} \rightarrow CD_{c1} + "ون"$

Accusatif ou génitif : $CDS_{c1} \rightarrow CD_{c1} + "ين"$

### 3.2.6. Le nom de nombre

Le nom du nombre est considéré comme un adjectif, donc les mêmes règles d'accord nom-adjectif sont applicables. (Voir section 3.2.11)

$[DT_s\ JJ_s\ NN_s] \rightarrow [DT_c + NN_c\ DT_c + JJ_c]$

**Exemple**

(The first book) → (الكتاب الأول)

(The first school) → (المدرسة الأولى)

Si $CD_c \in [11 - 19]$

$[DT_s\ CD_s\ JJ_s\ NN_s] \rightarrow [DT_c + NN_c\ DT_c + JJ_c\ CD_c]$

Si $CD_c \in [21 - 99]$

$[DT_s\ CD_s\ JJ_s\ NN_s] \rightarrow [DT_c + NN_c\ DT_c + JJ_c\ "و"\ "الـ" + CD_c]$

**Exemple**

(The twenty first book) → (الكتاب الواحد و العشرون)

La morphologie du $CD_c$ suit celle du $NN_c$

Si $NN_c$ est nominatif $CD_c$ sera suffixé de "ون"



Si NN$_c$ est accusatif ou génitif CD$_c$ sera suffixé de "ين"

### 3.2.7. Les conjonctions de coordination

Les noms liés par des conjonctions de coordination sont toujours de la même forme (nominative, accusative, génitive).

[DT$_{s1}$ NN$_{s1}$ CC$_s$ DT$_{s2}$ NN$_{s2}$] → [DT$_{c1}$+NN$_{c1}$ CC$_c$ DT$_{c2}$+NN$_{c2}$]

Si NN$_{c1}$ est nominatif → {NN$_{c2}$, NN$_{c3}$, ....} sont nominatifs

Si NN$_{c1}$ est accusatif ou génitif → {NN$_{c2}$, NN$_{c3}$, ....} sont accusatifs ou génitifs

(Teachers and learners) →(المعلمون و المتعلمون)

(With the teachers and learners)→(مع المعلمين و المتعلمين)

### 3.2.8. Les prépositions

Les prépositions relient des noms, des pronoms et des expressions à d'autres mots dans une phrase. Le mot ou l'expression que la préposition introduit est appelé l'*objet* de la préposition.

Si le nom est précédé de la préposition (**with**), le nom cible sera préfixé de la lettre ("بـ").

Si IN$_s$ ="with": [IN$_s$ DT$_s$ NN$_s$] → ["بـ"+ DT$_c$+ NN$_c$]

Si IN$_s$ ="with" & NNS$_c$ ∈ {NNS rég} & NNS$_{c2}$(M): [IN$_s$ DT$_s$ NNS$_s$] → ["بـ"+ DT$_c$+ NN$_c$+"ين"]

Si IN$_s$ ="with" & NNS$_c$ ∈ {NNS rég} & NNS$_{c2}$(F): [IN$_s$ DT$_s$ NNS$_s$] → ["بـ"+ DT$_c$+ NN$_c$+"ات"]

Si IN$_s$ ="with" & NNS$_c$ ∈ {NNS irrég}: [IN$_s$ DT$_s$ NNS$_s$] → ["بـ"+ DT$_c$+ NNS$_c$]

**Exemple**

(With the book) →(بالكتاب)

Si le nom est précédé de la préposition (**as**), le nom cible sera préfixé de la lettre ("كـ").

Si IN$_s$ ="as": [IN$_s$ DT$_s$ NN$_s$] → ["كـ"+ DT$_c$+ NN$_c$]

Si IN$_s$ ="as" & NNS$_c$ ∈ {NNS rég} & NNS$_{c2}$(M): [IN$_s$ DT$_s$ NNS$_s$] → ["كـ"+ DT$_c$+ NN$_c$ +"ين"]

Si IN$_s$ ="as" & NNS$_c$ ∈ {NNS rég} & NNS$_{c2}$(F): [IN$_s$ DT$_s$ NNS$_s$] → ["كـ"+ DT$_c$+ NN$_c$ +"ات"]



Si $IN_s$ ="as" & $NNS_c \in \{\text{NNS irrég}\}$: $[IN_s \, DT_s \, NNS_s] \rightarrow ["\text{كـ}" + DT_c + NNS_c]$

**Exemple**

(As a book) →(ككتاب)

Si le nom est précédé de la préposition (**to**), le nom cible sera préfixé de la lettre ("لـ").
$[TO_s \, DT_s \, NN_s] \rightarrow ["\text{لـ}" + DT_c + NN_c]$

Si $NNS_c \in \{\text{NNS rég}\}$ & $NNS_{c2}(M)$: $[IN_s \, DT_s \, NNS_s] \rightarrow ["\text{لـ}" + DT_c + NN_c + "\text{ين}"]$

Si $NNS_c \in \{\text{NNS rég}\}$ & $NNS_{c2}(F)$: $[IN_s \, DT_s \, NNS_s] \rightarrow ["\text{لـ}" + DT_c + NN_c + "\text{ات}"]$

Si $NNS_c \in \{\text{NNS irrég}\}$: $[IN_s \, DT_s \, NNS_s] \rightarrow ["\text{لـ}" + DT_c + NNS_c]$

**Exemple**

(To the doctor) →(للطبيب)

Les noms pluriels réguliers précédés par une préposition (من، الى، عن، على، في، حتى، منذ، كي، عدا) sont traduits par des noms suffixés de "ين" pour le masculin et de "ات" pour le féminin.

Si $NNS_c \in \{\text{NNS rég}\}$ & $NNS_{c2}(M)$: $[IN_s \, DT_s \, NNS_s] \rightarrow [IN_c \, DT_c + NN_c + "\text{ين}"]$

Si $NNS_c \in \{\text{NNS rég}\}$ & $NNS_{c2}(F)$: $[IN_s \, DT_s \, NNS_s] \rightarrow [IN_c \, DT_c + NN_c + "\text{ات}"]$

Si $NNS_c \in \{\text{NNS irrég}\}$: $[IN_s \, DT_s \, NNS_s] \rightarrow [IN_c \, DT_c + NNS_c]$

### 3.2.9. L'annexion

L'annexion[58] peut être exprimée par deux noms liés par (of) ou ('s).

$[NN_{s1} \text{ 's } NN_{s2}]$ ou $[NN_{s2} \text{ of } NN_{s1}]$

Si POS= ( 's) : les noms sont inversés en matière d'ordre (voir chap.6) et traduits par deux noms : le premier indéfini et le deuxième est défini.

$[DT_s \, NN_{s1} \, POS_s \, NN_{s2}] \rightarrow [NN_{c2} \, DT_c + NN_{c1}]$

**Exemple**

---

[58] Le cas possessif est toujours génitif.



(Student's book) → (كتاب التلميذ)

Si IN= (of) : les noms dans la langue cible entretiennent le même ordre et sont traduits par deux noms : le premier indéfini et le deuxième est soit défini ou indéfini selon la détermination du mot source.

Si $DT_{s2}$ ="the" & $IN_s$ ="of" :{ $[DT_{s1}\ NN_{s1}\ IN_s\ DT_{s2}\ NN_{s2}]$→ $[NN_{c1}\ "الـ"+NN_{c2}]$, $[DT_{s1}\ NNS_{s1}\ IN_s\ DT_{s2}\ NNS_{s2}]$→ $[NNS_{c1}\ "الـ"+NNS_{c2}]$, $[DT_{s1}\ NN_{s1}\ IN_s\ DT_{s2}\ NNS_{s2}]$→ $[NN_{c1}\ "الـ"+NNS_{c2}]$, $[DT_{s1}\ NNS_{s1}\ IN_s\ DT_{s2}\ NN_{s2}]$ → $[NNS_{c1}\ "الـ"+NN_{c2}]$ }

Si $DT_{s2}$ ="a" & $IN_s$ ="of": { $[DT_{s1}\ NN_{s1}\ IN_s\ DT_{s2}\ NN_{s2}]$→ $[NN_{c1}\ NN_{c2}]$, $[DT_{s1}\ NNS_{s1}\ IN_s\ DT_{s2}\ NNS_{s2}]$→ $[NNS_{c1}\ NNS_{c2}]$, $[DT_{s1}\ NN_{s1}\ IN_s\ DT_{s2}\ NNS_{s2}]$→ $[NN_{c1}\ NNS_{c2}]$, $[DT_{s1}\ NNS_{s1}\ IN_s\ DT_{s2}\ NN_{s2}]$→ $[NNS_{c1}\ NN_{c2}]$}

**Exemple**

**The** book of **the** student → (كتاب التلميذ)

**The** book of **a** student →(كتاب تلميذ)

**A** book of **the** student →(كتاب التلميذ)

**A** book of **a** student →(كتاب تلميذ)

On remarque que le premier nom cible ($NN_{c1}$) est toujours défini par annexion, alors que le deuxième nom ($NN_{c2}$) est défini par (الـ) ou indéfini.

Les noms pluriels en première position qui appartiennent aux pluriels réguliers sont suffixés de "وا".

Les noms pluriels en deuxième position qui appartiennent aux pluriels réguliers et masculins sont suffixés de "ين".

Les noms pluriels en deuxième position qui appartiennent aux pluriels réguliers et féminins sont suffixés de "ات".

Les pluriels irréguliers restent invariables.

Si $NNS_{c1} \in \{NNS\ rég\}$ & $NNS_{c2}(M)$: $[DT_{s1}\ NNS_{s1}\ IN_s\ DT_{s2}\ NN_{s2}]$→ $[NN_{c1}+"وا"\ "الـ"+NN_{c2}]$

Si $NNS_{c1} \in \{NNS\ rég\}$ & $NNS_{c2}(F)$: $[DT_{s1}\ NNS_{s1}\ IN_s\ DT_{s2}\ NN_{s2}]$→ $[NN_{c1}+"ات"\ "الـ"+NN_{c2}]$

Si $NNS_{c1} \in \{NNS\ irrég\}$ : $[DT_{s1}\ NNS_{s1}\ IN_s\ DT_{s2}\ NN_{s2}]$→ $[NNS\ "الـ"+NN_{c2}]$



Si NNS$_{c2}$ ∈ {NNS rég} & NNS$_{c2}$(M): [DT$_{s1}$ NN$_{s1}$ IN$_s$ DT$_{s2}$ NNS$_{s2}$] → [NN$_{c1}$ "الـ"+NN$_{c2}$ +"ين"]

Si NNS$_{c2}$ ∈ {NNS rég} & NNS$_{c2}$(F): [DT$_{s1}$ NN$_{s1}$ IN$_s$ DT$_{s2}$ NNS$_{s2}$] → [NN$_{c1}$ "الـ"+NN$_{c2}$ +"ات"]

Si NNS$_{c2}$ ∈ {NNS irrég} : [DT$_{s1}$ NNS$_{s1}$ IN$_s$ DT$_{s2}$ NN$_{s2}$] → [NN$_{c1}$ "الـ"+NNS$_{c2}$]

### Annexion multiple

Si le nombre de mots dans l'annexion est supérieur à 1, il faut toujours garder la règle qu'en langue cible, on obtient des mots tous définis, les premiers sont définis par annexion et le dernier est défini par déterminant.

[DT$_{s1}$ NN$_{s1}$ IN$_s$ DT$_{s2}$ NN$_{s2}$ IN$_s$ DT$_{s3}$ NN$_{s3}$] → [NN$_{c1}$ NN$_{c2}$ DT$_c$+NN$_{c2}$]

Si DT$_{sn}$ ="the" & IN$_s$ ="of" : [DT$_{s1}$ NN$_{s1}$ IN$_s$ DT$_{s2}$ NN$_{s2}$ IN$_s$ … DT$_{sn}$ NN$_{sn}$] → [NN$_{c1}$ NN$_{c2}$ … "الـ"+NN$_{cn}$]

**Exemple**

(The key of the door of the house) → (مفتاح باب البيت)

Si DT$_{sn}$ ="a" & IN$_s$ ="of": [DT$_{s1}$ NN$_{s1}$ IN$_s$ DT$_{s2}$ NN$_{s2}$ IN$_s$ … DT$_{sn}$ NN$_{sn}$] → [NN$_{c1}$ NN$_{c1}$ NN$_{cn}$]

**Exemple**

(A key of a door of a house) → (مفتاح باب بيت)

Les autres cas du féminin singulier et du pluriel irrégulier masculin ou féminin restent les mêmes comme expliqué en dessus. Sauf pour le pluriel régulier, traité comme suit :

Si les noms sont masculins, le premier sera suffixé de "وا", et le deuxième jusqu'à l'avant dernier de "ي" et le dernier de "ين".

Si NNS$_{c1..n-1}$ ∈ {NNS rég} & NNS$_{c1..n-1}$(M): [DT$_{s1}$ NNS$_{s1}$ IN$_s$ DT$_{s2}$ NNS$_{s2}$ IN$_s$ DT$_{s3}$ NNS$_{s3}$ IN$_s$ … DT$_{sn}$ NN$_{sn}$] → [NN$_{c1}$+"وا" NN$_{c2}$+"ي" NN$_{c3}$+"ي" … "الـ"+NN$_{cn}$]

Si NNS$_{cn}$ ∈ {NNS rég} & NNS$_{cn}$(M): [… DT$_{sn}$ NNS$_{sn}$] → ["الـ"+NN$_{cn}$ +"ين"]

Si les noms sont féminins, tous les mots sont suffixés de "ات"

Si NNS$_{c1.. n-1}$ ∈ {NNS rég} & NNS$_{c1..n-1}$(F): [DT$_{s1}$ NNS$_{s1}$ IN$_s$ DT$_{s2}$ NNS$_{s2}$ IN$_s$ …DT$_{sn}$ NN$_{sn}$] → [NN$_{c1}$+"ات" NN$_{c2}$+"ات" … "الـ"+NN$_{cn}$]



Si les noms de la phrase sont confondus, entre masculin, féminin et singulier, pluriel, la traduction se fait selon chaque cas du mot.

### 3.2.10. Les pronoms possessifs

[PRP$$_s$ NN$_s$] → [NN$_{c+}$ "Enc"]

**Exemple**

(His book) → (كتابه)

[PRP$$_s$ NN$_s$] ou [PRP$$_s$ NNS$_s$] → [NN$_c$ +"Enc"]

Si PRP$_s$ ={MY}: [PRP$$_s$ NN$_s$] → [NN$_c$ +"ـي"]

Si PRP$_s$ ={YOUR(S)}: [PRP$$_s$ NN$_s$] → [NN$_c$ +"ك"]

Si PRP$_s$ ={YOUR(B)}: [PRP$$_s$ NN$_s$] → [NN$_c$ +"كما"]

Si PRP$_s$ ={YOUR(P,M)}: [PRP$$_s$ NN$_s$] → [NN$_c$ +"كم"]

Si PRP$_s$ ={YOUR(P,F)}: [PRP$$_s$ NN$_s$] → [NN$_c$ +"كن"]

Si PRP$_s$ ={HIS}: [PRP$$_s$ NN$_s$] → [NN$_c$ +"ـه"]

Si PRP$_s$ ={HER}: [PRP$$_s$ NN$_s$] → [NN$_c$ +"ها"]

Si PRP$_s$ ={ITS(M)}: [PRP$$_s$ NN$_s$] → [NN$_c$ +"ـه"]

Si PRP$_s$ ={ITS(F)}: [PRP$$_s$ NN$_s$] → [NN$_c$ +"ها"]

Si PRP$_s$ ={OUR}: [PRP$$_s$ NN$_s$] → [NN$_c$ +"نا"]

Si PRP$_s$ ={THEIR(B)}: [PRP$$_s$ NN$_s$] → [NN$_c$ +"هما"]

Si PRP$_s$ ={THEIR(M)}: [PRP$$_s$ NN$_s$] → [NN$_c$ +"هم"]

Si PRP$_s$ ={THEIR(F)}: [PRP$$_s$ NN$_s$] → [NN$_c$ +"هن"]

Si PRP$_s$ ={THEIR} & (hmn= "faux"): [PRP$$_s$ NN$_s$] → [NN$_c$ +"ها"]

La table 5.17 résume tous les cas des pronoms possessifs (PRP$) que ce soit liés à des noms singuliers ou pluriels (NN ou NNS) : Masculins ou féminins (M ou F).

**Table 5.17 Traduction des pronoms possessifs**

| Mot source | Mot cible |
|---|---|
| My NN$_s$ | NN$_c$ + "ـي" |
| Your(S)  NN$_s$ | NN$_c$ + "ك" |
| Your(B)  NN$_s$ | NN$_c$ + "كما" |
| Your(P,M) NN$_s$ | NN$_c$ + "كم" |
| Your(P,F) NN$_s$ | NN$_c$ + "كن" |
| His  NN$_s$ | NN$_c$ + "ـه" |
| Her  NN$_s$ | NN$_c$ + "ها" |
| Its  NN$_s$ (M) | NN$_c$ + "ـه" |
| Its  NN$_s$ (F) | NN$_c$ + "ها" |
| Our  NN$_s$ | NN$_c$ + "نا" |
| Their  NN$_s$ (B) | NN$_c$ + "هما" |
| Their  NN$_s$ (M) | NN$_c$ + "هم" |
| Their  NN$_s$ (F) | NN$_c$ + "هن" |
| Their  NN$_s$ ([hmn]="faux") | NN$_c$ + "ها" |
| Whose  NN$_s$ | NN$_c$ + "هم"  "الذين" |

### 3.2.11. Accord Nom-adjectif

Ce type d'accord est absent en anglais. Par contre la langue arabe exige que les adjectifs soient en accord avec les noms :

- en définition.
- en genre
- en nombre

Les traits de l'adjectif source sont appliqués sur l'adjectif cible et sur le nom cible.

#### 3.2.11.1. La définition

La traduction de la phrase (the big car) : [DT$_s$ JJ$_s$ NN$_s$] ne donne pas (السيارة كبيرة) avec le nom défini et l'adjectif indéfini, ni (سيارة الكبيرة) avec le nom indéfini et l'adjectif défini.

Après l'inversion du nom et de l'adjectif et l'application des règles de passage entre les deux langues, on obtient la traduction correcte suivante :

(The big car) → (السيارة الكبيرة)

[DT$_s$ JJ$_s$ NN$_s$] → [DT$_c$+JJ$_c$ DT$_c$+NN$_c$], toujours DT$_c$ ="الـ"

Si DT$_s$="The": [DT$_s$ JJ$_s$ NN$_s$] → ["الـ"+ NN$_c$  "الـ"+JJ$_c$]

Si DT$_s$ ="a": [DT$_s$ JJ$_s$ NN$_s$] → [NN$_c$  JJ$_c$]

Si DT$_s$ = $\phi$ : [DT$_s$ JJ$_s$ NN$_s$] → [NN$_c$  JJ$_c$]

Alors que sur Systran (figure 5.4) on constate que l'inversion est effectuée par contre la définition est erronée (l'adjectif n'a pas été marqué par le déterminant bien que la phrase commence par "The").

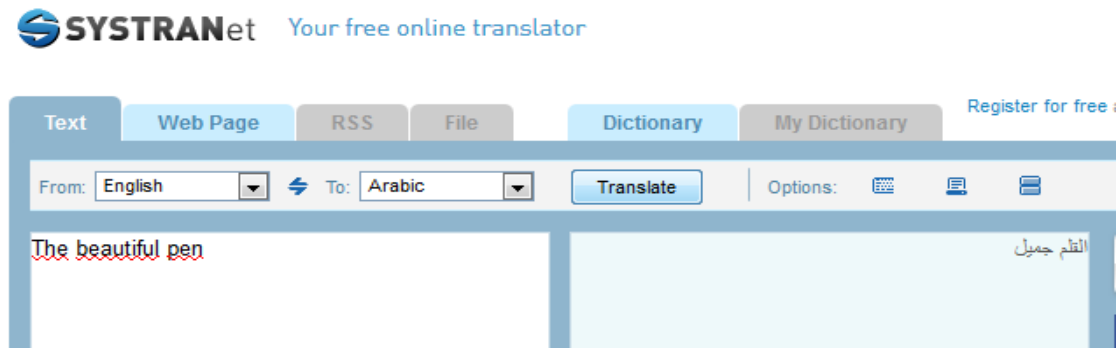

**Figure 5.4 Traduction erronée des adjectifs (07/12/2013)**

### 3.2.11.2. Le genre

Dans les exemples précédents on remarque aussi un accord en genre. Le mot cible (سيارة) a le trait (féminin), l'adjectif (*big*) est neutre, alors que pour une traduction vers l'arabe : le genre est appliqué sur le nom cible et sur l'adjectif cible. Ce qui donne (سيارة كبيرة) et non pas (سيارة كبير).

Si $NN_c(M)$: $[DT_s\ JJ_s\ NN_s] \rightarrow [DT_c+NN_c\ DT_c+JJ_c]$,

Si $NN_c(F)$: $[DT_s\ JJ_s\ NN_s] \rightarrow [DT_c+NN_c\ DT_c+JJ_c+"ة"]$

Si $DT_c = \{"الـ", \phi\}$, les règles expliquées ci-dessus restent valables (section : la définition).

### 3.2.11.3. Le nombre

Les adjectifs liés à des noms singuliers sont traduits par un nom singulier et un adjectif singulier.

Les adjectifs liés à des noms pluriels sont traduits par des noms pluriels et des adjectifs singuliers dotés de la marque de féminin.

$[DT_s\ JJ_s\ \mathbf{NN}_s] \rightarrow [DT_c+NN_c\ DT_c+JJ_c]$

$[DT_s\ JJ_s\ \mathbf{NNS}_s] \rightarrow [DT_c+NNS_c\ DT_c+JJ_c+"ة"]$

Si $DT_c = \{"الـ", \phi\}$, les règles expliquées en dessus restent valables (section : la définition).

**Exemple**

(Beautiful pen) → (قلم جميل)

(Beautiful pens) → (أقلام جميلة)

Si le nom est au pluriel et a le trait humain, l'adjectif aura le genre du nom.

Si $NN_c \in \{NNS\ rég\}$ & $NN_c(M)$ & $JJ_c \in \{NNS\ rég\}$: [$DT_s\ JJ_s\ \mathbf{NNS_s}$] →[$DT_c$+$NN_c$+"ون" $DT_c$+$JJ_c$+"ون"]

Si $NN_c \in \{NNS\ rég\}$ & $NN_c(M)$ & $JJ_c \in \{NNS\ irrég\}$: [$DT_s\ JJ_s\ \mathbf{NNS_s}$] →[$DT_c$+$NNS_c$+"ون" $DT_c$+$JJS_c$]

Si $NN_c \in \{NNS\ irrég\}$ & $NN_c(M)$ & $JJ_c \in \{NNS\ rég\}$: [$DT_s\ JJ_s\ \mathbf{NNS_s}$] →[$DT_c$+$NNS_c$ $DT_c$+$JJ_c$+"ون"]

Si $NN_c \in \{NNS\ irrég\}$ & $NN_c(M)$ & $JJ_c \in \{NNS\ irrég\}$: [$DT_s\ JJ_s\ \mathbf{NNS_s}$] →[$DT_c$+$NNS_c$ $DT_c$+$JJS_c$]

Si $NN_c \in \{NNS\ rég\}$ & $NN_c(F)$: [$DT_s\ JJ_s\ \mathbf{NNS_s}$] →[$DT_c$+$NNS_c$+"ات" $DT_c$+$JJ_c$+"ات"]

Si $NN_c \in \{NNS\ irrég\}$ & $NN_c(F)$: [$DT_s\ JJ_s\ \mathbf{NNS_s}$] →[$DT_c$+$NNS_c$ $DT_c$+$JJ_c$+"ات"]

**Exemple**

- (Strong men) → (رجال أقوياء)
- (Strong women) → (نساء قويات)

Si le nombre d'adjectifs est supérieur à 1, les adjectifs sont inversés et ils s'accordent avec le nom en genre, en nombre et en définition.

[$DT_s\ \mathbf{JJ_{s1}\ JJ_{s2}}\ NN_s$] → [$DT_c$+$NN_c$ $DT_c$+$\mathbf{JJ_{c2}}$ $DT_c$+$\mathbf{JJ_{c1}}$]

Si $DT_c = \{$"الـ", $\phi\}$, les règles expliquées en dessus restent valables (section : la définition).

**Exemple**

(A beutiful big car) → (سيارة كبيرة جميلة)

(He lives in **a** lovely, gigantic, ancient, house) → (يعيش في بيت قديم عملاق جميل), car DT="a".

(This is **the** lovely, gigantic, ancient, house) → (هذا هو البيت القديم العملاق الجميل), car DT="the".

Si $DT_s$="a" : [$DT_s\ JJ_{s1}\ JJ_{s2}\ JJ_{s3}\ ......JJ_{sn}\ NN_s$] → [$NN_c\ JJ_{cn}\ ......JJ_{c3}\ JJ_{c2}\ JJ_{c1}$]

Si $DT_s$= "the": [$DT_s\ JJ_{s1}\ JJ_{s2}\ JJ_{s3}\ ......JJ_{sn}\ NN_s$] →[$DT_c$+$NN_c$ $DT_c$+$JJ_n$......$DT_c$+$JJ_{c3}$ $DT_c$+$JJ_{c2}$ $DT_c$+$JJ_{c1}$] sachant que DT="الـ".

### 3.2.12. Accord sujet- prédicat

#### 3.2.12.1. Sujet-prédicat au présent

Le sujet-prédicat ou (sujet-attribut) est exprimé en anglais par le verbe (to be) au présent simple entre un nom et un adjectif.

**1. Le masculin**

Si le sujet source est masculin, le sujet-prédicat est traduit par deux noms successifs le premier défini et le deuxième indéfini.

Si $VBZ_s$="is" & $NN_c$(M): [$DT_s$ **$NN_s$** $VBZ_s$ $JJ_s$] → ["الـ"+ **$NN_c$** $JJ_c$]

**Exemple**

(The book is useful) → (الكتاب مفيد)

**2. Le féminin**

Si le sujet source est féminin, le sujet-prédicat est traduit par deux noms successifs le premier défini et le deuxième indéfini muni de la marque de féminin.

Si $VBZ_s$="is" & $NN_c$(F): [$DT_s$ **$NN_s$** $VBZ_s$ $JJ_s$] →["الـ"+ **$NN_c$** $JJ_c$ +"ة"]

**Exemple**

(The car is useful) → (السيارة مفيدة)

**3. Le duel**

**3.1.    Le duel masculin**

Si $CC_s$="and" & $VBP_s$="are" & ($NN_{c1}$(M) ou $NN_{c2}$(M)): [$DT_s$ $NN_{s1}$ $CC_s$ $DT_s$ $NN_{s2}$ $VBP_s$ $JJ_s$] →["الـ"+$NN_{c1}$ "و" "الـ"+ $NN_{c2}$ $JJ_c$+"ان"]

**Exemple**

(Book and pen are useful) → (الكتاب و القلم مفيدان)

**3.2.    Le duel féminin**

Si $CC_s$="and" & $VBP_s$="are" & ($NN_{c1}$(F) & $NN_{c2}$(F)): [$DT_s$ $NN_{s1}$ $CC_s$ $DT_s$ $NN_{s2}$ $VBP_s$ $JJ_s$] →["الـ"+$NN_{c1}$ "و" "الـ"+ $NN_{c2}$ $JJ_c$+"تان"]

**Exemple**

(Ruler and protractor are useful) → (المسطرة و المنقلة مفيدتان)

**4. Le pluriel**

**4.1.    Le pluriel entier**

Si le sujet source est au pluriel et est non humain, le prédicat cible aura la marque du féminin.

Si VBP$_s$="are" & NNS$_s$ (hmn="faux"): [DT$_s$ **NNS**$_s$ VBP$_s$ JJ$_s$] → ["الـ"+ **NNS**$_c$  JJ$_c$+"ة"]

**Exemple**

(Books are useful) → (الكتب مفيدة)

Si le sujet source est humain, l'adjectif cible sera au pluriel. (une des particularités de la langue arabe est que l'adjectif peut être singulier ou pluriel, masculin ou féminin).

Si NNS$_c$(M) & VBP$_s$="are" & **NNS**$_s$ (hmn="vrai"): [DT$_s$ **NNS**$_s$ VBP$_s$ JJ$_s$] → ["الـ"+ **NNS**$_c$ JJS$_c$]

**Exemple**

(Men are standing) → (الرجال واقفون)

**Remarque**

Si JJ$_c$ ∈ {NNS rég}: JJS$_c$ → JJ$_c$+"ون"

Si NNS$_c$(F) & VBP$_s$="are" & **NNS**$_s$ (hmn="vrai"): [DT$_s$ **NNS**$_s$ VBP$_s$ JJ$_s$] → ["الـ"+ **NNS**$_c$  JJ$_c$ +"ات"]

**Exemple**

(Women are standing) → (النساء واقفات)

### 4.2. Le pluriel divisé

$\forall\ NNc\ \in\ NN, \exists NNc(hmn = faux)$ :

Si CC$_{s1}$="and" & CC$_{s2}$="and" &VBP$_s$="are": [DT$_s$ NN$_{s1}$ CC$_{s1}$ DT$_s$ NN$_{s2}$ CC$_{s2}$ DT$_s$ NN$_{s3}$ VBP$_s$ JJ$_s$] →["الـ"+NN$_{c1}$ "و" "الـ"+ NN$_{c2}$ "و" "الـ"+ NN$_{c3}$ JJ$_c$+"ة"]

**Exemple**

(Ruler and protractor and compass are useful) → (المسطرة و المنقلة و الفرجار مفيدة)

$\forall\ NNc\ \in\ NN, NNc(hmn = vrai), \exists\ NNc(M)$ :

Si CC$_{s1}$="and" & CC$_{s2}$="and" &VBP$_s$="are": [DT$_s$ NN$_{s1}$ CC$_{s1}$ DT$_s$ NN$_{s2}$ CC$_{s2}$ DT$_s$ NN$_{s3}$ VBP$_s$ JJ$_s$] →["الـ"+NN$_{c1}$ "و" "الـ"+ NN$_{c2}$  "و" "الـ"+ NN$_{c2}$ JJS$_c$]

**Exemple**

(Father and mother and aunt are standing) → (الأب و الأم و العمة واقفون)

**Remarque**

∀ NNc ∈ NN, NNc(hmn = vrai), ∃ NNc(M), SI JJ$_c$ ∈ {NNS rég}: JJS$_c$ → JJ$_c$+"ون"

∀ NNc ∈ NN, NNc(hmn = vrai), NNc(F) :

Si CC$_{s1}$="and" & CC$_{s2}$="and" &VBP$_s$="are": [DT$_s$ NN$_{s1}$ CC$_s$ DT$_s$ NN$_{s2}$ CC$_{s3}$ DT$_s$ NN$_{s3}$ VBP$_s$ JJ$_s$] →["الـ"+NN$_{c1}$ "و" "الـ"+ NN$_{c2}$  "و" "الـ"+ NN$_{c2}$ JJ$_c$ +"ات"]

**Exemple**

(Aunt and mother and grandmother are standing) → (العمة و الأم و الجدة واقفات)

Les mêmes règles sont valables si le nombre des noms est supérieur à trois.

Les noms peuvent être séparés par des virgules et le dernier nom est précédé par "and": [NN$_1$, NN$_2$, NN$_3$ …..CC NN$_n$ VBP JJ]

5. **Les noms propres**

   5.1. **Le masculin**

   Si le sujet est un nom propre masculin, le sujet-prédicat est traduit par deux noms : un nom propre suivi d'un nom indéfini (adjectif).
   Si VBZ$_s$="is" & NNP$_s$(M): [**NNP**$_s$ VBZ$_s$ JJ$_s$]→[NNP$_c$ JJ$_c$]

**Exemple**

(Alper[59] is strong) →(ألبر قوي)

   5.2. **Le féminin**

   Si le sujet est un nom propre féminin, le sujet-prédicat est traduit par un nom propre suivi d'un nom indéfini (adjectif) muni de la marque de féminin.
   Si VBZ$_s$="is" & NNP$_s$(F): [**NNP**$_s$ VBZ$_s$ JJ$_s$]→[NNP$_c$ JJ$_c$ +"ة"]

**Exemple**

---

[59] Nom turc signifiant "courageux"

(مايا جميلة) ← (Maya is beautiful)

Les mêmes règles de duel et du pluriel sont applicables pour les noms propres.

### 3.2.12.2. Sujet-prédicat au passé

Si le verbe (to be) est conjugué au **passé (VBD)**, le sujet-attribut dans la langue cible sera précédé par (كان) ou (كانت). Si le sujet source est masculin singulier, le sujet cible sera précédé par (كان) et l'attribut aura le préfixe (ا). Si le sujet source est féminin ou pluriel (masculin ou féminin), le sujet cible sera précédé par (كانت) et l'attribut cible aura la marque du féminin.

**1. Le masculin**

Si $VBD_s$ ="was" & $NN_c$ (M) : [$DT_s$ **$NN_s$** $VBD_s$ $JJ_s$] → ["كان" "الـ"+ $NN_c$ $JJ_c$+"ا"]

**Exemple**

(كان الكتاب مفيدا) ← (The book was useful)

**2. Le féminin**

Si $VBD_s$ ="was" & $NN_c$ (F) : [$DT_s$ **$NN_s$** $VBD_s$ $JJ_s$] → ["كانت" "الـ"+ $NN_c$ $JJ_c$+"ة"]

**Exemple**

(كانت المكتبة مفيدة) ← (The Library was useful)

**3. Le duel**

**3.1. Le duel masculin**

Si $CC_s$="and" & $VBD_s$="were" & ($NN_{c1}$(M) & $NN_{c2}$(M): [$DT_s$ $NN_{s1}$ $CC_s$ $DT_s$ $NN_{s2}$ $VBD_s$ $JJ_s$] → ["كان" "الـ" + $NN_{c1}$ "و" "الـ"+$NN_{c2}$ $JJ_s$+"ين"]

(كان الكتاب و القلم مفيدين) ← (Book and pen were useful)

**3.2. Le duel confondu**

Si $CC_s$="and" & $VBD_s$="were" & ($NN_{c1}$(F) & $NN_{c2}$(M)): [$DT_s$ $NN_{s1}$ $CC_s$ $DT_s$ $NN_{s2}$ $VBD_s$ $JJ_s$] → ["كانت" "الـ" + $NN_{c1}$ "و" "الـ"+$NN_{c2}$ $JJ_s$+"ين"]

(كانت المسطرة و القلم مفيدين) ← (Ruler and pen were useful)

Si $CC_s$="and" & $VBD_s$="were" & ($NN_{c1}$(M) & $NN_{c2}$(F)): [$DT_s$ $NN_{s1}$ $CC_s$ $DT_s$ $NN_{s2}$ $VBD_s$ $JJ_s$] → ["كان" "الـ" + $NN_{c1}$ "و" "الـ"+$NN_{c2}$ $JJ_s$+"ين"]

(كان القلم و المسطرة مفيدين) ← (Pen and ruler were useful)

**3.3. Le duel féminin**

Si $CC_s$="and" & $VBD_s$="were" & ($NN_{c1}$(F) & $NN_{c2}$(F)): [$DT_s$ $NN_{s1}$ $CC_s$ $DT_s$ $NN_{s2}$ $VBD_s$ $JJ_s$] → ["كانت" "الـ" + $NN_{c2}$ "و" "الـ" $NN_{c1}$ + $JJ_c$+"تين"]

(كانت المسطرة و المنقلة مفيدتين) ← (Ruler and protractor are useful)

### 4. Le pluriel

#### 4.1. Le pluriel entier

Si $VBD_s$ ="were" & $NNS_c$ (hmn ="faux"): [$DT_s$ **$NNS_s$** $VBD_s$ $JJ_s$] → [$NNS_c$+"الـ" "كانت"] $JJ_c$+"ة"]

**Exemple**

(كانت الكتب مفيدة) ← (Books were useful)

Si $VBD_s$ ="were" & $NNS_c$ (hmn ="vrai") & $NNS_c$(M): [$DT_s$ **$NNS_s$** $VBD_s$ $JJ_s$] → ["كان" "الـ"+$NNS_c$ $JJS_c$]

**Exemple**

(كان الرجال واقفين) ← (Men were standing)

**Remarque**

Si $JJ_c$ ∈ {NNS rég}: $JJS_c$ → $JJ_c$+"ين"

Si $VBD_s$ ="were" & $NNS_c$ (hmn ="vrai") & $NNS_c$(F): [$DT_s$ **$NNS_s$** $VBD_s$ $JJ_s$] → ["كانت" "الـ"+$NNS_c$ $JJ_c$+"ات"]

**Exemple**

(كانت النساء واقفات) ← (Women were standing)

#### 4.2. Le pluriel divisé

∀ NNc ∈ NN, NNc(hmn = faux) :

Si $CC_{s1}$="and" & $CC_{s2}$="and" & $VBP_s$="were" & $NN_{s1}$(F): [$DT_s$ $NN_{s1}$ $CC_{s1}$ $DT_s$ $NN_{s2}$ $CC_{s2}$ $DT_s$ $NN_{s3}$ $VBP_s$ $JJ_s$] →["كانت" "الـ"+$NN_{c1}$ "و" "الـ" + $NN_{c2}$ "و" "الـ"+ $NN_{c2}$ $JJ_c$+"ة"]

**Exemple**

(كانت المسطرة و المنقلة و الفرجار مفيدة) ← (Ruler and protractor and compass were useful)

∀ NNc ∈ NN, NNc(hmn = faux) :

Si CC$_{s1}$="and" & CC$_{s2}$="and" &VBP$_s$="were" & NN$_{c1}$(M): [DT$_s$ NN$_{s1}$ CC$_{s1}$ DT$_s$ NN$_{s2}$ CC$_{s2}$ DT$_s$ NN$_{s3}$ VBP$_s$ JJ$_s$] → ["كان" "الـ"+NN$_{c1}$ "و" "الـ"+ NN$_{c2}$ "و" "الـ"+ NN$_{c2}$ JJS$_c$]

**Exemple**

(Protractor and ruler and compass were useful) → (كان الفرجار و المسطرة و المنقلة مفيدين)

**Remarque**

Si JJ$_c$ ∈ {NNS rég}: JJS$_c$ → JJ$_c$+"ين"

∀ NNc ∈ NN, NNc(hmn = vrai), ∃ NNc(M) :

Si CC$_{s1}$="and" & CC$_{s2}$="and" &VBP$_s$="were": [DT$_s$ NN$_{s1}$ CC$_{s1}$ DT$_s$ NN$_{s2}$ CC$_{s2}$ DT$_s$ NN$_{s3}$ VBP$_s$ JJ$_s$] → ["كان" "الـ"+NN$_{c1}$ "و" "الـ"+ NN$_{c2}$ "و" "الـ"+ NN$_{c2}$ JJS$_c$]

**Exemple**

(Father and mother and aunt were standing) → (كان الأب و الأم و العمة واقفين)

**Remarque**

∀ NNc ∈ NN, NNc(hmn = vrai), ∃ NNc(M), Si JJ$_c$ ∈ {NNS rég}: JJS$_c$ → JJ$_c$+"ين"

∀ NNc ∈ NN, NNc(hmn = vrai), NNc(F) :

Si CC$_{s1}$="and" & CC$_{s2}$="and" &VBP$_s$="were": [DT$_s$ NN$_{s1}$ CC$_s$ DT$_s$ NN$_{s2}$ DT$_s$ CC$_{s3}$ DT$_s$ NN$_{s3}$ VBP$_s$ JJ$_s$] → ["الـ"+ NN$_{c2}$ JJ$_c$ +"ات"] "و" "الـ"+ NN$_{c2}$ "و" "الـ"+NN$_{c1}$ "الـ" "كانت"]

**Exemple**

(Aunt and mother and grandmother were standing) → (كانت العمة و الأم و الجدة واقفات)

Les mêmes règles sont valables si le nombre des noms est supérieur à trois.

Les noms peuvent être séparés par des virgules et le dernier nom est précédé par "and": [NN$_1$, NN$_2$, NN$_3$ …..CC NN$_n$ VBP JJ]

### 5. Les noms propres

#### 5.1. Le masculin

Si le sujet est un nom propre masculin, le sujet-prédicat est traduit par deux noms un nom propre suivi d'un nom indéfini (adjectif) suffixé de "ا".

Si VBZ$_s$="was" & NNP$_s$(M): [**NNP$_s$** VBZ$_s$ JJ$_s$] → ["كان" NNP$_c$ JJ$_c$ +"ا"]

**Exemple**

(Albert was strong) →(كان ألبر قويا)

### 5.2. Le féminin

Si le sujet est un nom propre féminin, le sujet-prédicat est traduit par un nom propre suivi d'un nom indéfini (adjectif) muni de la marque de féminin.

Si VBZ$_s$="was" & NNP$_s$(F): [**NNP**$_s$ VBZ$_s$ JJ$_s$]→["كانت" NNP$_c$ JJ$_c$+ "ة"]

**Exemple**

(Maya was beautiful) →(كانت مايا جميلة)

Les mêmes règles de duel et du pluriel sont applicables pour les noms propres.

### 3.2.12.3. Sujet-prédicat au futur

Si le verbe (to be) est conjugué au **futur** (MD VB, MD= "will"), le sujet attribut dans la langue cible sera précédé par (سيصبح ou ستصبح). Si le sujet est masculin singulier le sujet cible sera précédé par (سيصبح) et l'attribut aura le préfixe (ا). Si le sujet est féminin ou pluriel non humain (masculin, féminin), le sujet cible sera précédé par (ستصبح) et l'attribut cible aura la marque du féminin. Si le sujet est pluriel humain et masculin le sujet cible sera précédé par (سيصبح) et l'attribut aura le préfixe (ين). Si le sujet est pluriel humain et féminin le sujet cible sera précédé par (ستصبح) et l'attribut aura le préfixe (ات)

1. **Le masculin**

Si MD$_s$ ="will" & VB$_s$ ="be" & NN$_c$ (M): [DT$_s$ **NN**$_s$ MD$_s$ VB$_s$ JJ$_s$] → ["سيصبح "+"الـ"+NN$_c$ JJ$_c$+"ا"]

**Exemple**

(The book will be useful) → (سيصبح الكتاب مفيدا)

2. **Le féminin**

Si MD$_s$ ="will" & VB$_s$ ="be" & NN$_c$ (F): [DT$_s$ **NN**$_s$ MD$_s$ VB$_s$ JJ$_s$] → ["ستصبح "+"الـ"+NN$_c$ JJ$_c$+"ة"]

**Exemple**

(The Library will be useful) → (ستصبح المكتبة مفيدة)

Les mêmes règles précédentes du passé (duel, pluriel entier, pluriel divisé, les noms propres) sont applicables pour le futur, tout en remplaçant (كان) par (سيصبح) et (كانت) par (ستصبح).

Les temps composés sont traités comme temps simples i.e.

Past perfect, past continuous, present perfect → simple past

Present continuous → simple future

$VBN_s$ → $VBD_s$

$VBG_s$ → $VB_s$

$VBP_s$, $VBZ_s$ → $VB_s$

$VBZ_s$ $VBN_s$ → $VBD_s$

### 3.2.13. L'adverbe

L'adverbe est toujours à l'accusatif.

[$DT_{s1}$ $NN_{s1}$ $VB_s$ $DT_{s1}$ $NN_{s1}$ $RB_s$] → [$VB_c$ $DT_{c1}$+$NN_{c1}$ $DT_{c2}$+$NN_{c2}$ $RB_c$]

**Exemple**

(The teacher explains the lesson quickly) → (يشرح الأستاذ الدرس سريعا)

## *Conclusion*

Dans ce chapitre, on a cité les différentes règles de passage morphologique pour la traduction de l'anglais, supposé parfaitement analysée, vers l'arabe, basées sur le modèle proposé dans le chapitre 4. La langue arabe se caractérise par la capacité d'inclure plusieurs constituants grammaticaux en un seul mot, ce qui exige des règles de transfert pour regrouper ces constituants dans un mot de la langue cible, à partir des mots indépendants. D'autres cas partageant le même principe n'ont pas été cités pour éviter la répétition.

Le chapitre suivant s'intéresse à la phrase et détaille les différentes transformations syntaxiques et l'ordre des mots lors du passage de la langue source à la langue cible. L'ordre des mots dans la phrase a une importance cruciale dans la détermination du sens de la phrase anglaise. Ainsi les langues diffèrent dans leurs structures syntaxiques ce qui nécessite des règles de transfert, ou règles métataxiques.

# Chapitre 6. Règles syntaxiques pour le transfert de l'anglais vers l'arabe

## *Introduction*

La génération de l'arabe suppose le texte anglais analysé et compris ; un des problèmes majeurs en anglais est celui du genre, trouvé parfois indirectement, non pas au niveau du nom, mais au niveau des pronoms qui s'y réfèrent ; un autre problème est celui des « that » sous-entendus ; un autre, l'ambivalence de « but » ; et, toujours, celui de trouver le verbe principal, notamment en cas de « verbification » (utilisation verbale d'un nom).

La langue arabe est caractérisé par un ordre libre de mots dans la phrase, il n'est pas étrange de trouver plusieurs ordres possibles comme : VSO, SVO, VOS et parfois même OVS comme dans les phrases suivantes :

VSO → (كتب الولد الدرس)

SVO → (الولد كتب الدرس)

VOS → (كتب الدرس الولد)

OVS → (الدرس كتب الولد)

Toutes ces phrases sont des phrases grammaticalement correctes et ont le même sens.

L'ordre le plus fréquent en langue arabe est l'ordre VSO, les autres sont rarement utilisés, ils servent à des situations pour accentuer des constituants de la phrase[60].

**Règles de transfert syntaxique**

La langue arabe est plus parataxique que l'anglais, plus hypotaxique. Nida et Taber [116] affirment que la traduction devrait minimiser la parataxe ou la transformer en hypotaxe.

L'arabe aussi est fortement flexionnelle, alors qu'on trouve des morphèmes avec des caractéristiques syntaxiques mais attachés à d'autres mots comme les conjonctions de coordi-nation, l'article défini, quelques prépositions ainsi les pronoms, ce qui engendre une diminution de nombre de constituants syntaxiques dans la langue cible.

---

[60] Aspect emphatique typique des langues flexionnelles, présentant un intérêt rhétorique.

**Exemple**

La figure 6.1 montre un exemple de transfert stucturel entre l'anglais et l'arabe.

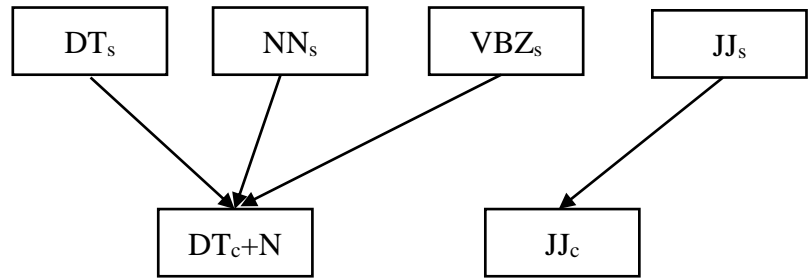

**Figure 6.1 Diminution de nombre de constituants lors d'un transfert syntaxique**

Quand on parle de phrase verbale ou nominale, on parle de la structure dans la langue cible, car parfois des phrases verbales en anglais deviennent des phrases nominales en arabe.

## *1. La phrase nominale*

La phrase nominale arabe a comme équivalent en anglais une phrase verbale. C'est une phrase qui comme son nom l'indique, ne contient pas de verbe. La surcharge du verbe « to be » en anglais, employé comme auxiliaire, le rend inévitable même dans les formulations les plus simples. Comme le verbe « to be » est implicite en arabe, il sera possible d'exprimer des énoncés simples, équivalents aux énoncés les plus simples d'anglais, sans employer de verbe.

La phrase nominale comporte un sujet et un attribut, ou plus précisément un groupe sujet et un groupe attribut.

### 1.1. La phrase nominale affirmative

#### 1.1.1.   Le présent

Si la phrase source comporte le verbe (*to be*) conjugué au **présent**, la phrase cible aura la structure donnée : (figure 6.2).

Si $VBZ_s = \{is, are\}$ : $[DT_s\ NN_s\ VBZ_s\ JJ_s] \rightarrow [DT_c + NN_c\ JJ_c]$

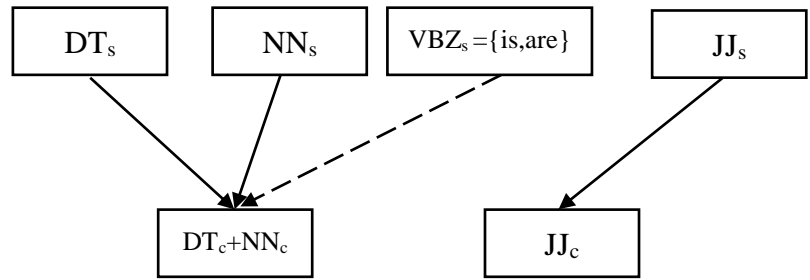

**Figure 6.2 Transfert d'une phrase verbale au présent vers une phrase nominale**

La phrase source est une phrase verbale qui a comme phrase cible une phrase nominale. Les phrases verbales sont toujours traduites par des phrases verbales sauf si le verbe de la phrase est le verbe (to be) : {VBZ="is", VBP="are"}, auquel cas elles sont traduites par des phrases nominales.

**Exemple**

(The book is useful) → (الكتاب مفيد)

Si le verbe (to be) est au présent, la phrase cible garde la forme sujet-attribut.

S'il est conjugué à un autre temps, la phrase cible reçoit des changements morphologiques et des changements structurels par l'ajout de pseudo-verbes, bien que la phrase soit qualifiée de nominale.

### 1.1.2. Le passé

Si le verbe (to be) est conjugué au **passé** (VBD ="was" ou "were"), la phrase sera précédée de "كان" ou "كانت" selon le genre du sujet (figure 6.3) et l'attribut sera suffixé (voir chapitre 5).

Si $VBD_s$ ={was, were}: [$DT_s$ $NN_s$ $VBD_s$ $JJ_s$] → ["كان" $DT_c$ + $NN_c$ $JJ_c$]

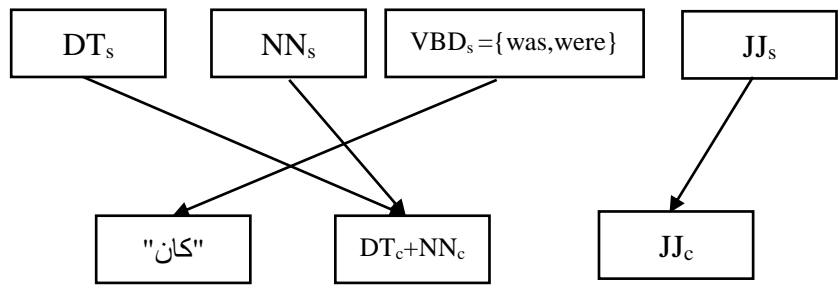

**Figure 6.3 Transfert d'une phrase verbale au passé vers une phrase nominale**

**Exemple**

(The book was useful) → (كان الكتاب مفيدا)

### 1.1.3. Le futur

Si le verbe (to be) est conjugué au **futur** (MD="will" & VB ="be"), la phrase sera précédée de "سيصبح" ou "ستصبح" selon le genre du sujet (figure 6.4) et l'attribut sera suffixé.

Si MD$_s$ ="will" & VB$_s$ ="be": [DT$_s$ NN$_s$ MD$_s$ VB$_s$ JJ$_s$] → ["سيصبح " DT$_c$+NN$_c$ JJ$_c$]

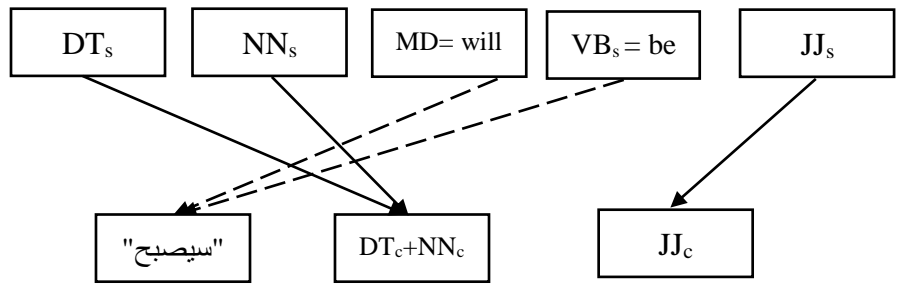

**Figure 6.4 Transfert d'une phrase verbale au futur vers une phrase nominale**

**Exemple**

(The book will be useful) → (سيصبح الكتاب مفيدا)

## 1.2. La négation dans les phrases nominales (sujet_ attribut)

### 1.2.1. Le présent

Si le verbe (to be) dans la phrase source est conjugué au **présent**, la phrase cible aura la structure donnée en figure 6.5. La phrase cible sera précédée de "ليس" ou "ليست" selon le genre du sujet et l'attribut sera suffixé.

Si VBZ$_s$={is, are}& RB$_s$ ={not}: [DT$_s$ NN$_s$ VBZ$_s$ RB$_s$ JJ$_s$] → ["ليس" DT$_c$+NN$_c$ JJ$_c$]

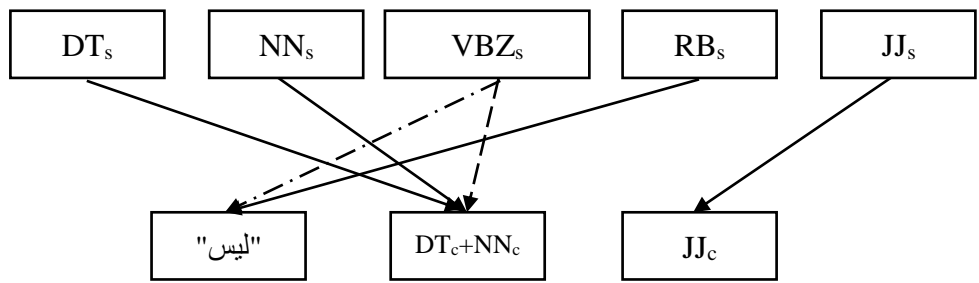

**Figure 6.5 Transfert de la négation dans les phrases comprenant (to be) au présent**

**Exemple**

(The book is not useful) → (ليس الكتاب مفيدا)

### 1.2.2. Le passé

Si le verbe (to be) est conjugué au **passé** (VBD ="was" ou "were"), la phrase négative sera précédée de "لم يكن" ou "لم تكن" selon le genre du sujet (figure 6.6) et l'attribut sera suffixé.

Si VBD$_s$ ={was, were}: [DT$_s$ NN$_s$ VBD$_s$ RB$_s$ JJ$_s$] → ["لم يكن" DT$_c$+NN$_c$ JJ$_c$]

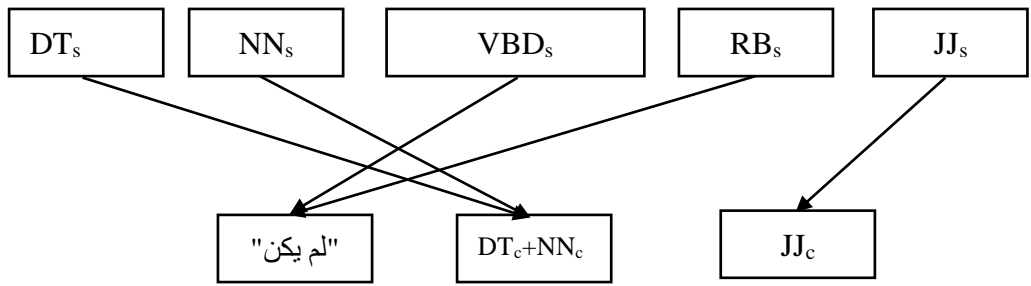

**Figure 6.6 Transfert de la négation dans les phrases comprenant (to be) au passé**

### Exemple

(The book was not useful) → (لم يكن الكتاب مفيدا)

### 1.2.3. Le futur

Si le verbe (to be) est conjugué au **futur** (MD="will" & VB ="be"), la phrase négative sera précédée de "لن يصبح" ou "لن تصبح" selon le genre du sujet (figure 6.7) et l'attribut sera suffixé.

Si $MD_s$ ="will" & $VB_s$ ="be": [$DT_s$ $NN_s$ $MD_s$ $VB_s$ $JJ_s$] → ["لن يصبح" $DT_c$+$NN_c$ $JJ_c$]

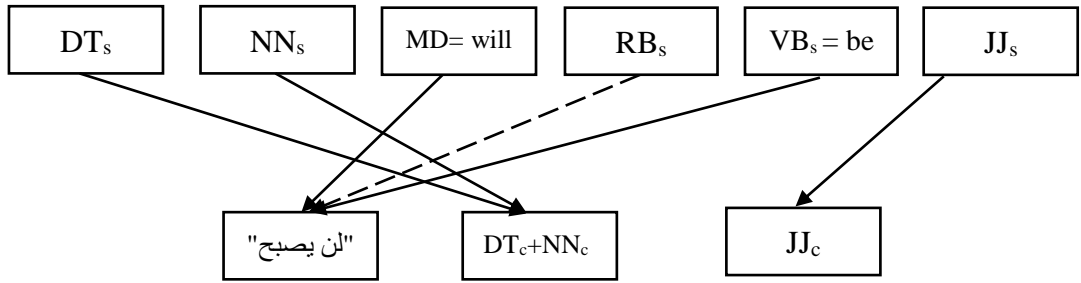

**Figure 6.7 Transfert de la négation dans les phrases comprenant (to be) au futur**

### Exemple

(The book will not be useful) → (لن يصبح الكتاب مفيدا)

## 1.3. L'interrogation dans les phrases nominales (sujet_ attribut)

### 1.3.1. Le présent

Si la phrase source comporte le verbe (to be) conjugué au **présent**, la phrase cible aura la structure de la figure 6.8. La phrase cible sera précédée de "هل" et l'attribut sera suffixé par les marques du nominatif.

Si $VBZ_s$= {is, are}: [$VBZ_s$ $DT_s$ $NN_s$ $JJ_s$] → ["هل" $DT_c$+$NN_c$ $JJ_c$]

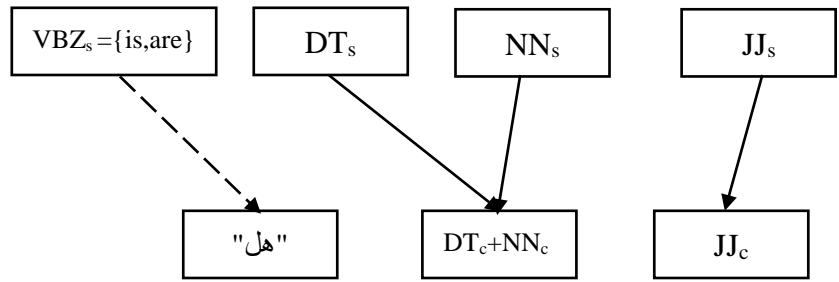

**Figure 6.8 Transfert de l'interrogation dans les phrases comprenant (to be) au présent**

**Exemple**

(Is the book useful?) → (هل الكتاب مفيد ؟)

### 1.3.2. Le passé

Si le verbe (to be) est conjugué au **passé** (VBD ="was" ou "were"), la phrase interrogative sera précédée de "هل كان" ou "هل كانت" selon le genre du sujet (figure 6.9) et l'attribut sera suffixé.

Si $VBD_s$ ={was, were}: $[VBD_s\ DT_s\ NN_s\ JJ_s]$ → ["هل كان" $DT_c+NN_c\ JJ_c$]

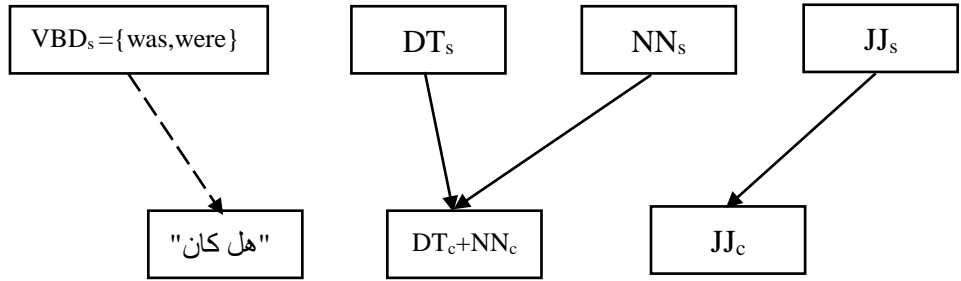

**Figure 6.9 Transfert de l'interrogation dans les phrases comprenant (to be) au passé**

**Exemple**

(Was the book useful?) → (هل كان الكتاب مفيدا ؟)

### 1.3.3. Le futur

Si le verbe (to be) est conjugué au **futur** (MD="will" & VB ="be"), la phrase interrogative sera précédée de "هل ستصبح" ou "هل سيصبح" selon le genre du sujet (figure 6.10) et l'attribut sera suffixé.

Si $MD_s$ ="will" & $VB_s$ ="be": $[MD_s\ DT_s\ NN_s\ VB_s\ JJ_s]$ → ["هل سيصبح" $DT_c+NN_c\ JJ_c$]

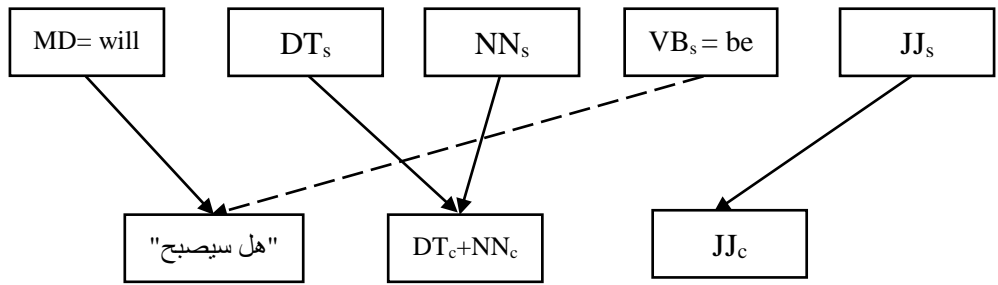

**Figure 6.10 Transfert de l'interrogation dans les phrases comprenant (to be) au futur**

**Exemple**

(Will the book be useful?) → (هل سيصبح الكتاب مفيدا)

## 1.4. Les adjectifs

Un adjectif lié à un nom, est traduit par un nom suivi d'un adjectif (figure 6.11).

$[DT_s\ JJ_s\ NN_s] \rightarrow [DT_c+NN_c\ DT_c+JJ_c]$

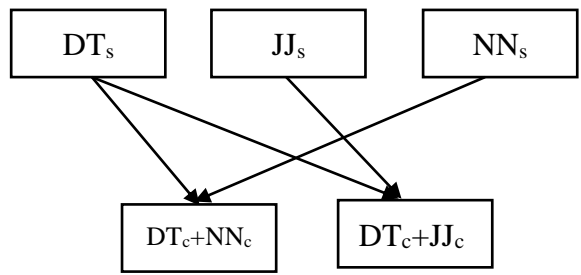

**Figure 6.11 Transfert des adjectifs**

Si le nombre d'adjectifs est supérieur à 1, les adjectifs seront inversés, le premier adjectif sera le dernier, le deuxième sera l'avant dernier, et le dernier sera le premier (figure 6.12).

$[DT_s\ JJ_{s1}\ JJ_{s2}\ JJ_{s3}\ \ldots JJ_{sn}\ NN_s] \rightarrow [DT_c+NN_c\ DT_c+JJ_n\ldots DT_c+JJ_{c3}\ DT_c+JJ_{c2}\ DT_c+JJ_{c1}]$

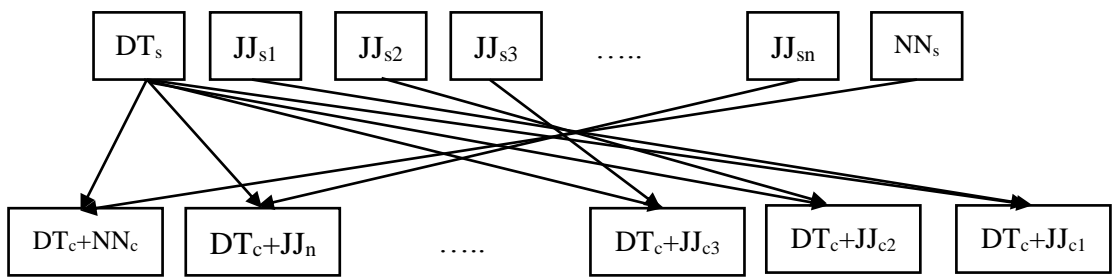

**Figure 6.12 Transfert des adjectifs multiples**

### 1.5. L'annexion

#### 1.5.1. L'annexion simple

L'annexion peut être exprimée par deux noms liés par (of) ou ('s).

[NN$_{s1}$ 's NN$_{s2}$] ou [NN$_{s2}$ of NN$_{s1}$]

Si POS= ( 's): l'ordre des noms est inversé (figure 6.13).

[DT$_s$ NN$_{s1}$ POS$_s$ NN$_{s2}$] → [NN$_{c2}$ DT$_c$+ NN$_{c1}$]

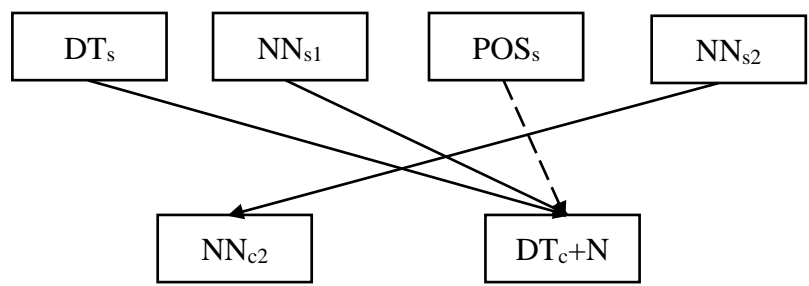

**Figure 6.13 Transfert d'annexion 1**

**Exemple**

(Student's book) → (كتاب التلميذ)

Si IN= (of): les noms dans la langue cible maintiennent le même ordre (figure 6.14).

[DT$_{s1}$ NN$_{s1}$ IN$_s$ DT$_{s2}$ NN$_{s2}$] → [NN$_{c1}$ DT$_c$+NN$_{c2}$]

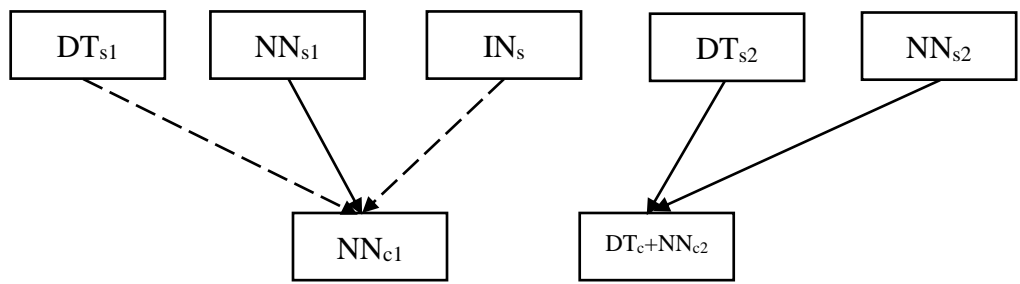

**Figure 6.14 Transfert d'annexion 2**

**Exemple**

(The book of the student) → (كتاب التلميذ)

#### 1.5.2. L'annexion multiple

Si le nombre de mots dans l'annexion est supérieur à 1, l'ordre des mots dans la langue cible reste invariable (figure 6.15).

**Exemple**

(The key of the door of the house) → (مفتاح باب البيت)

[DT$_{s1}$ NN$_{s1}$ IN$_s$ DT$_{s2}$ NN$_{s2}$ IN$_s$ DT$_{s3}$ NN$_{s3}$] →[NN$_{c1}$ NN$_{c2}$ DT$_c$+NN$_{c2}$]

Donc : [$DT_{s1}$ $NN_{s1}$ $IN_s$ $DT_{s2}$ $NN_{s2}$ … $IN_s$ $DT_{sn}$ $NN_{sn}$] →[$NN_{c1}$ $NN_{c2}$ … $DT_c$+$NN_{cn}$]

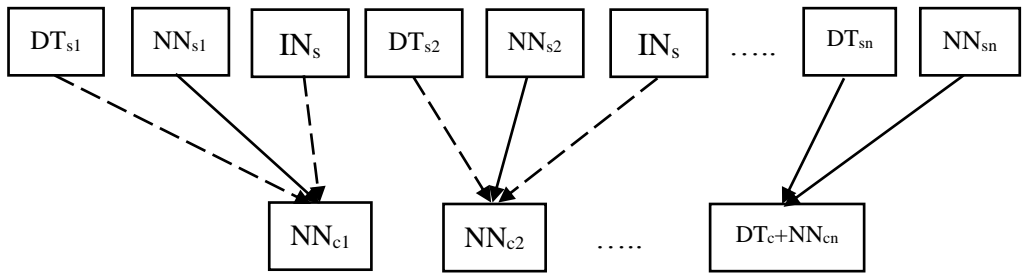

**Figure 6.15 Transfert d'annexion multiple**

### 1.5.3. Les adjectifs dans l'annexion

En anglais, il est possible d'avoir l'annexion sous forme d'adjectif : au lieu de dire the (key of the door), on dit (the key door) – or cette dernière formulation est considérée par exception comme un nom suivi d'un adjectif.

La phrase (The key door) peut être traduite par (مفتاح بابي) ou par (مفتاح الباب).

Par contre dans une phrase où l'annexion est précédée par l'adjectivation, il est possible de générer une ambigüité dans le rattachement des noms avec l'adjectif correspondant.

**Exemple**

(The key door of the house) → (مفتاح باب البيت) ou (مفتاح البيت البابي)

(The new key of the house) → (جديد مفتاح البيت) ou (مفتاح البيت الجديد)

Les phrases ont la même structure (DT JJ NN IN DT NN)

Pour la première phrase les deux propositions de traduction sont correctes, pour la deuxième phrase, seule la deuxième possibilité est correcte. Pour éviter de fausses traductions et sans savoir a priori si la structure est adjectivale ou d'annexion, on choisit le deuxième type de traduction pour ce genre de phrases.

L'adjectif sera reculé à la fin de la phrase et il sera toujours défini puisque le nom est toujours défini par annexion (figure 6.16).

[$DT_{s1}$ $JJ_{s1}$ $JJ_{s2}$ … $JJ_{sn}$ $NN_{s1}$ $IN_s$ $DT_{s2}$ $NN_{s2}$] → [$NN_{c1}$ $DT_c$+$NN_{c2}$ $DT_c$+$JJ_{cn}$ … $DT_c$+$JJ_{c2}$ $DT_c$+$JJ_{c1}$]

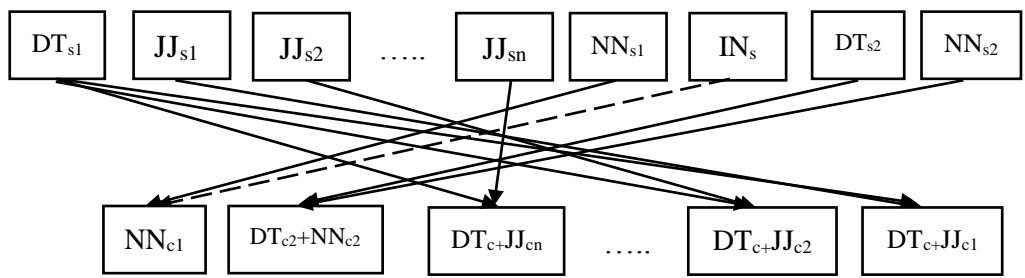

**Figure 6.16 Transfert dans l'annexion adjectivale**

### 1.6. Les conjonctions de coordination

Les noms liés par des conjonctions de coordination maintiennent le même ordre dans la langue cible (figure 6.17).

[$DT_{s1}$ $NN_{s1}$ $CC_s$ $DT_{s2}$ $NN_{s2}$ … $CC_n$ $DT_{sn}$ $NN_{sn}$] → [$DT_{c1}$+$NN_{c1}$ $CC_c$ $DT_{c2}$+$NN_{c2}$ … $CC_{cn}$ $DT_{cn}$ $NN_{cn}$]

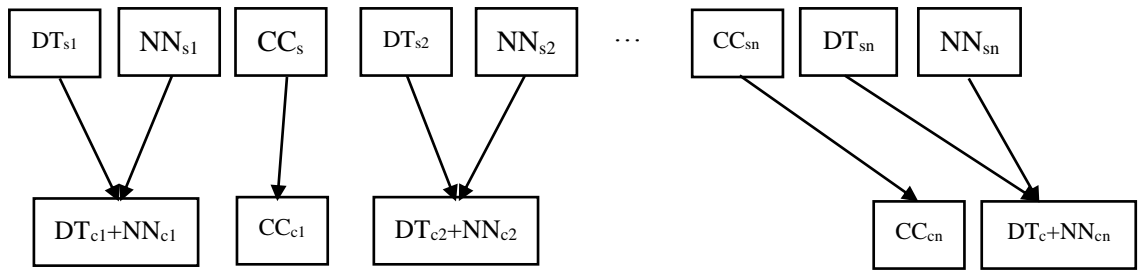

**Figure 6.17 Transfert des conjonctions de coordination**

### 1.7. Les prépositions

L'ordre de la préposition et de son objet reste inchangé de la langue source à la langue cible (figure 6.18).

[$IN_s$ $DT_s$ $NN_s$] → [$IN_c$ $DT_c$+ $NN_c$]

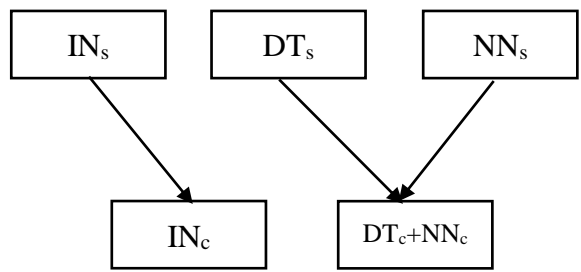

**Figure 6.18 Transfert des prépositions**

### 1.8. Les nombres

L'ordre des nombres entre l'anglais et l'arabe est différent, une méthode simple est de passer de la forme alphabétique anglaise du nombre vers la forme numérique puis vers la forme alphabétique en arabe (figure 6.19). Des changements morphologiques sont traités selon le cas du mot (nominatif, accusatif, génitif).

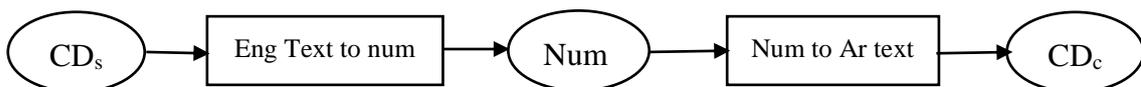

**Figure 6.19 Transfert des nombres**

**Exemple**

(One thousand four hundred thirty five) → (1435) → (ألف و أربع مائة و خمس و ثلاثون)

### 1.9. Le nom de nombre

Le nom du nombre est considéré comme un adjectif, donc les règles d'accord nom-adjectif sont applicables (figure 6.20).

[DT$_s$ JJ$_s$ NN$_s$] → [DT$_c$+NN$_c$ DT$_c$+JJ$_c$]

**Exemple**

(The first book) → (الكتاب الأول)

Si CD$_c \in [11-19]$

[DT$_s$ CD$_s$ JJ$_s$ NN$_s$] → [DT$_c$+NN$_c$ DT$_c$+JJ$_c$ CD$_c$]

Si CD$_c \in [21-99]$

[DT$_s$ CD$_s$ JJ$_s$ NN$_s$] → [DT$_c$+NN$_c$ DT$_c$+JJ$_c$ "و" DT$_c$+CD$_c$]

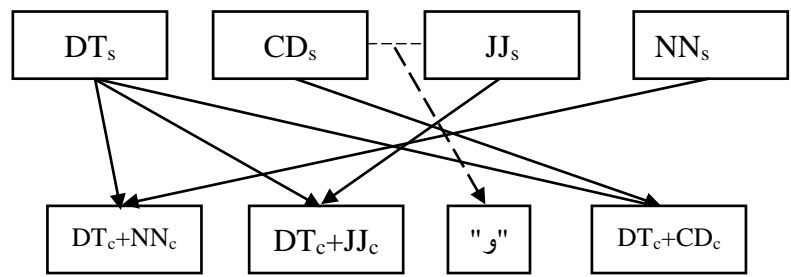

**Figure 6.20 Transfert des noms de nombres**

**Exemple**

(The twenty first book) → (الكتاب الواحد و العشرون)

## *2. La phrase verbale simple*

### 2.1. La phrase verbale affirmative

#### 2.1.1. Le verbe transitif

Nous avons mentionné dans l'introduction que la structure courante dans la langue arabe est VSO, alors la transformation de la phrase verbale comprenant un verbe transitif (figure 6.21) sera comme suit :

[DT$_{s1}$ NN$_{s1}$ VB$_s$ DT$_{s1}$ NN$_{s1}$] → [VB$_c$ DT$_{c1}$+NN$_{c1}$ DT$_{c2}$+NN$_{c2}$]

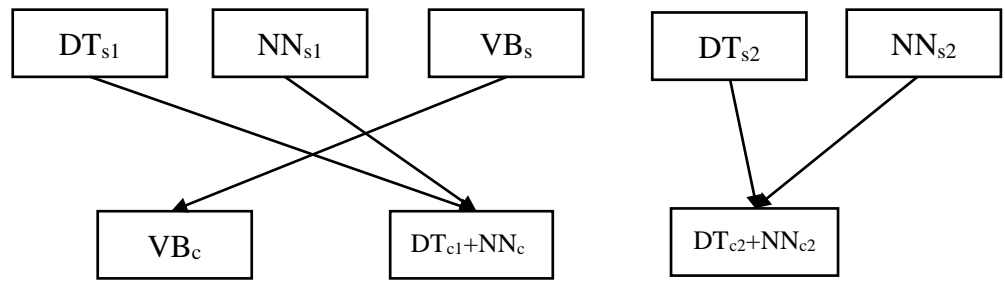

**Figure 6.21 Transfert de la phrase verbale simple, verbe transitif**

**Exemple**

(The boy writes the lesson) → (يكتب الولد الدرس)

Le temps du verbe n'a pas d'influence sur la structure syntaxique mais plutôt des changements morphologiques au niveau du verbe dans la structure VSO, où le verbe s'accorde uniquement en genre. (Voir section 3.1.1 chap.5), et des changements morphologiques au niveau du sujet et d'objet, le sujet est toujours au nominatif et l'objet est toujours à l'accusatif.

### 2.1.2. Le verbe intransitif

Si le verbe est intransitif, la présence du sujet est suffisante (figure 6.22)

$[DT_s\ NN_s\ VB_s] \rightarrow [VB_c\ DT_c+NN_c]$

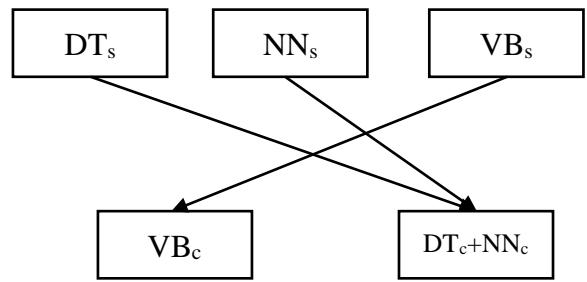

**Figure 6.22 Transfert de la phrase verbale simple, verbe intransitif**

**Exemple**

(The boy arrived) → (وصل الولد)

Il y a des verbes qui peuvent être transitifs et intransitifs en même temps, ils peuvent recevoir un objet ou non.

**Exemple**

(The boy ate) → (أكل الولد)

(The boy ate the apple) → (أكل الولد التفاحة)

### 2.2. La phrase verbale négative

Le transfert de la structure négative se fait par l'introduction de la particule "لا" ou "لن" ou "لم" avant la phrase. La phrase en négation maintient le même ordre que la phrase affirmative.

#### 2.2.1. Le présent

Le transfert de la négation dans les phrases verbales dont le verbe est conjugué au **présent** (figure 6.23) se fait comme suit :

[$DT_{s1}$ $NN_{s1}$ $VBZ_s$ $RB_s$ $VB_s$ $DT_{s2}$ $NN_{s2}$] → ["لا" $VB_c$ $DT_{c1}$+$NN_{c1}$ $DT_{c2}$+$NN_{c2}$]

Tel que $VBZ_s$ = "does", $RB_s$ = "not".

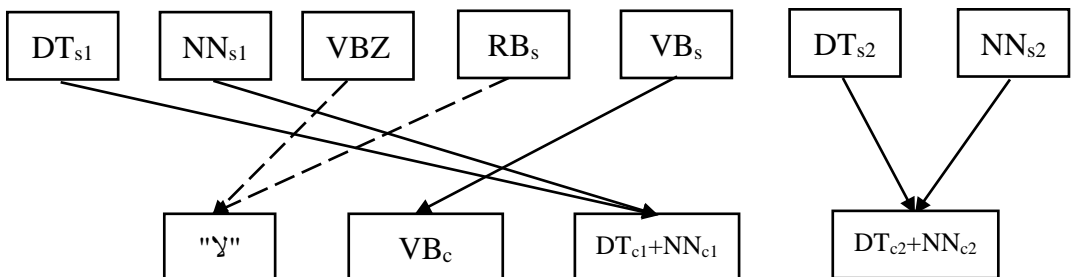

**Figure 6.23 Transfert de la négation dans la phrase verbale simple**

**Exemple**

(The boy does not write the lesson) → (لا يكتب الولد الدرس)

Le temps du verbe n'a pas d'influence sur la structure syntaxique mais entraîne plutôt des changements morphologiques au niveau du verbe.

#### 2.2.2. Le passé

Le transfert de la négation dans les phrases verbales dont le verbe est conjugué au **passé** se fait comme suit :

[$DT_{s1}$ $NN_{s1}$ $VBD_s$ $RB_s$ $VB_s$ $DT_{s2}$ $NN_{s2}$] → ["لم" $VB_c$ $DT_{c1}$+$NN_{c1}$ $DT_{c2}$+$NN_{c2}$]

Tel que $VBD_s$="did", $RB_s$="not".

**Exemple**

(The boy did not write the lesson) → (لم يكتب الولد الدرس)

#### 2.2.3. Le futur

Le transfert de la négation dans les phrases verbales dont le verbe est conjugué au **futur** se fait comme suit :

[$DT_s1$ $NN_{s1}$ $MD_s$ $RB_s$ $VB_s$ $DT_{s2}$ $NN_{s2}$] → ["لن" $VB_c$ $DT_{c1}$+$NN_{c1}$ $DT_{c2}$+$NN_{c2}$]

Tel que $MD_s$="will" ou "shall", $RB_s$="not".

**Exemple**

(The boy will not write the lesson) → (لن يكتب الولد الدرس)

### 2.3. La phrase verbale interrogative

Le transfert de la structure interrogative se fait par l'introduction de la particule "هل" avant la phrase, cette dernière entretient le même ordre dans la phrase affirmative.

#### 2.3.1. Le présent

Le transfert de l'interrogation dans les phrases verbales dont le verbe est conjugué au **présent** (figure 6.24) se fait comme suit :

[$VBZ_s$ $DT_{s1}$ $NN_{s1}$ $VB_s$ $DT_{s2}$ $NN_{s2}$?] → ["هل" $VB_c$ $DT_{c1}$+$NN_{c1}$ $DT_{c2}$+$NN_{c2}$ ؟]

Tel que $VBZ_s$="Does".

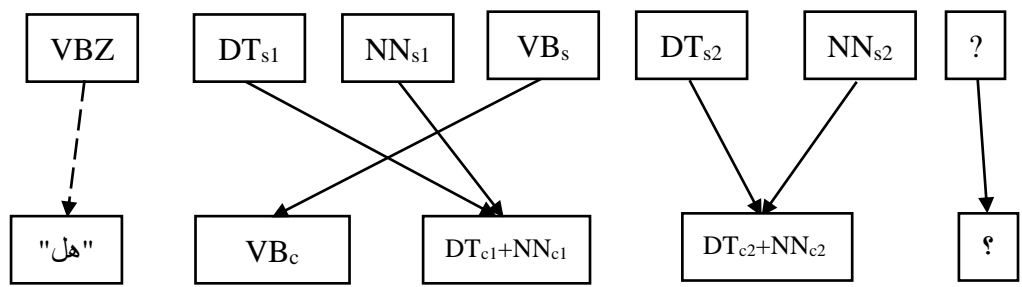

**Figure 6.24 Transfert de l'interrogation dans la phrase verbale simple**

**Exemple**

(Does the boy write the lesson?) → (هل يكتب الولد الدرس ؟)

#### 2.3.2. Le passé

Le transfert de l'interrogation dans les phrases verbales dont le verbe est conjugué au **passé** se fait comme suit :

[$VBD_s$ $DT_{s1}$ $NN_{s1}$ $VB_s$ $DT_{s2}$ $NN_{s2}$] → ["هل" $VB_c$ $DT_{c1}$+$NN_{c1}$ $DT_{c2}$+$NN_{c2}$]

Tel que $VBD_s$ = "did"

**Exemple**

(Did the boy write the lesson?) → (هل كتب الولد الدرس ؟)

#### 2.3.3. Le futur

Le transfert de l'interrogation dans les phrases verbales dont le verbe est conjugué au **futur** se fait comme suit :

[$MD_s$ $DT_{s1}$ $NN_{s1}$ $VB_s$ $DT_{s2}$ $NN_{s2}$] → ["هل" $VB_c$ $DT_{c1}$+$NN_{c1}$ $DT_{c2}$+$NN_{c2}$]

Tel que $MD_s$="will"ou "shall".

**Exemple**

(Will the boy write the lesson?) → (هل سيكتب الولد الدرس ؟)

### 2.4. Le complément d'agent

Les mêmes règles de sujet sont applicables sur le complément d'agent (figure 6.25). Le complément d'agent a toujours le cas nominatif.

[$DT_s$ $NN_s$ $VBD_s$ $VBN_s$] → [$VB_c$ $DT_c$+$NN_c$], tel que $VBD_s$ ="was" ou "were".

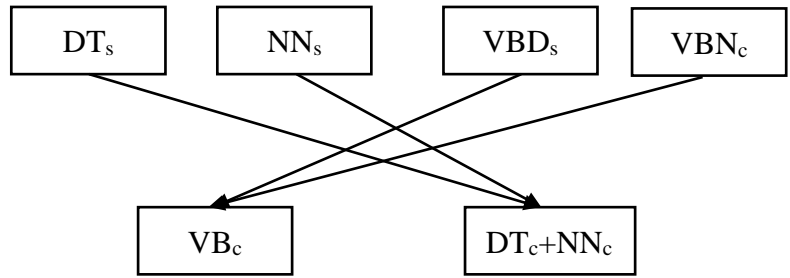

**Figure 6.25 Le transfert du complément d'agent**

**Exemple**

(The teachers were honored) → (كُرِّم المعلمون)

### 2.5. L'adverbe

L'adverbe vient à la fin de la phrase verbale, le même ordre est préservé dans la phrase cible (figure 6.26).

[$DT_{s1}$ $NN_{s1}$ $VB_s$ $DT_{s1}$ $NN_{s1}$ $RB_s$] → [$VB_c$ $DT_{c1}$+$NN_{c1}$ $DT_{c2}$+$NN_{c2}$ $RB_c$]

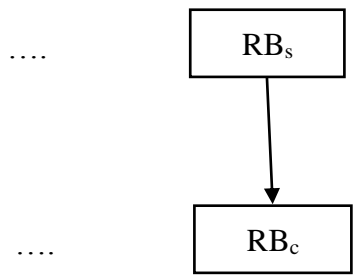

**Figure 6.26 Le transfert d'adverbe**

## 3. Les combinaisons

Les constituants essentiels seraient le groupe verbal, articulant sujet, verbe et complément, où sujet et complément peuvent être des groupes nominaux, dont les qualifiants sont des adjectifs, ou des propositions relatives, formées d'un pronom relatif, ….

Une phrase peut contenir plusieurs constituants en même temps, chaque type est traité comme un groupe, si le complément d'objet possède plusieurs adjectifs ces derniers avec l'objet constituent un groupe adjectival. Le groupe adjectival est considéré comme un seul mot vis-à-vis la phrase verbale représentant un complément d'objet. Cette méthode permet de décomposer la phrase en entités simples faciles à traduire.

- Le sujet peut être un groupe adjectival, un groupe prépositionnel.
- L'objet peut être un groupe adjectival, un groupe prépositionnel.
- L'attribut peut être un groupe prépositionnel.
- Le sujet et l'attribut peuvent être décrits par plusieurs adjectifs multiples.
- L'annexion peut être représentée par un sujet, un objet, un sujet de (sujet_attribut).
- ….

Dans les figures qui suivent, (IN) : veut dire préposition et (of) : annexion.

La figure 6.27 montre les cas possibles de combinaisons dans la phrase verbale affirmative.

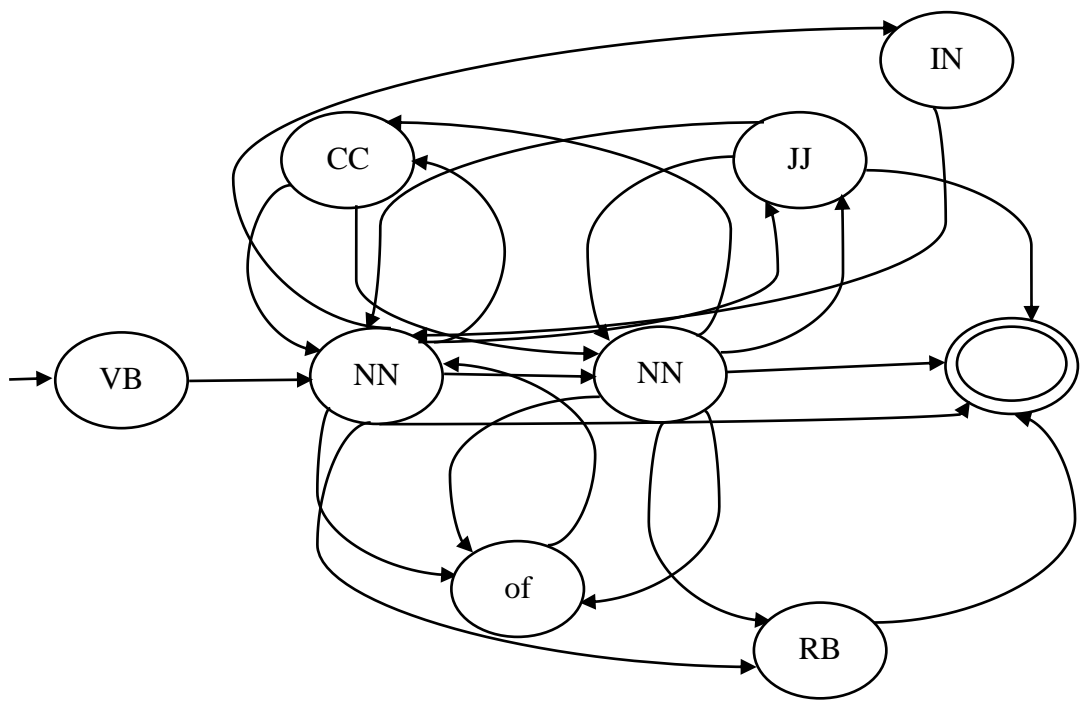

**Figure 6.27 Exemple de combinaisons dans les phrases verbales affirmatives**

La figure 6.28 détermine les cas possibles de début d'une phrase verbale.

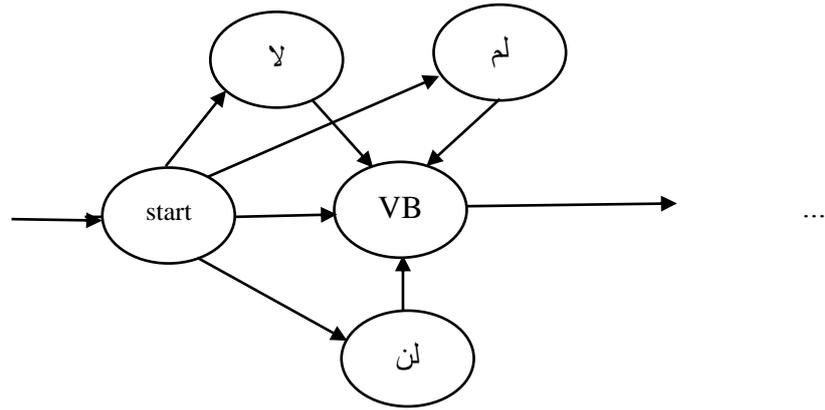

**Figure 6.28 Les cas possibles de début d'une phrase verbale**

La figure 6.29 montre les cas possibles de combinaisons dans la phrase nominale.

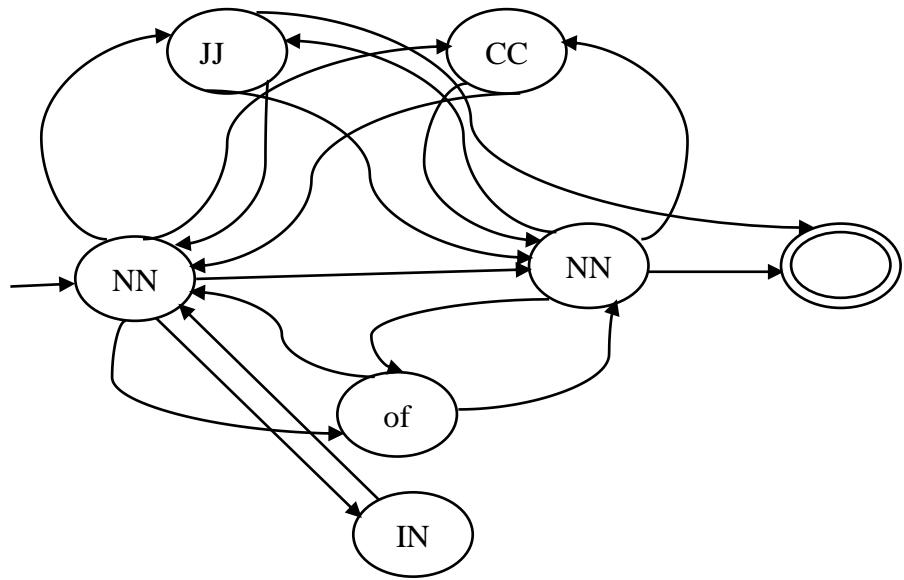

**Figure 6.29 Exemple de combinaisons dans les phrases nominales**

## *Conclusion*

Nous avons présenté dans ce dernier chapitre quelques règles syntaxiques pour le réordonnancement des mots en langue arabe à partir de l'anglais. La liste de règles n'est pas exhaustive, car d'autres exceptions peuvent se manifester. Des ambigüités peuvent être engendrées à cause des grandes différences structurelles entre les deux langues.

L'ordre entre génération morphologique et génération syntaxique n'est pas linéaire, des allez retour chaque fois peuvent se présenter entre les deux niveaux pour résoudre certaines ambiguïtés. La morphologie arabe est caractérisée par la présence de la syntaxe au niveau du mot, vu qu'un mot peut inclure toute une phrase.

# Chapitre 7. Expérimentations et Résultats

## *Introduction*

Dans cette section en présente des échantillons de résultats comprenant les transferts morphologiques et syntaxiques. On montre des phrases en anglais et leurs traductions en arabe par notre système (*Tordjman*). On a choisi les exemples où il y a des problèmes de traduction dans les autres systèmes comme *Google translate* et *Systran* et autres. On présente les exemples en mode ligne de commande pour faciliter la présentation et pour éviter d'encombrer les pages du manuscrit.

## *1. Langage de programmation*

Le langage de programmation utilisé est Python, on a opté pour la version Python 3 vue qu'elle utilise pour les chaines de caractères la norme *Unicode* compatible avec les caractères arabes quelle que soit la plate-forme informatique, contrairement aux versions antérieurs à la version Python 3.0.

## *2. Interfaces et résultats*

### 2.1. Interface de Tordjman

L'interface de *Tordjman* comporte un champ de saisie de la phrase source en anglais et un champ d'affichage de résultat de la phrase cible en arabe comme illustré dans les figures 7.1 et figure 7.2.

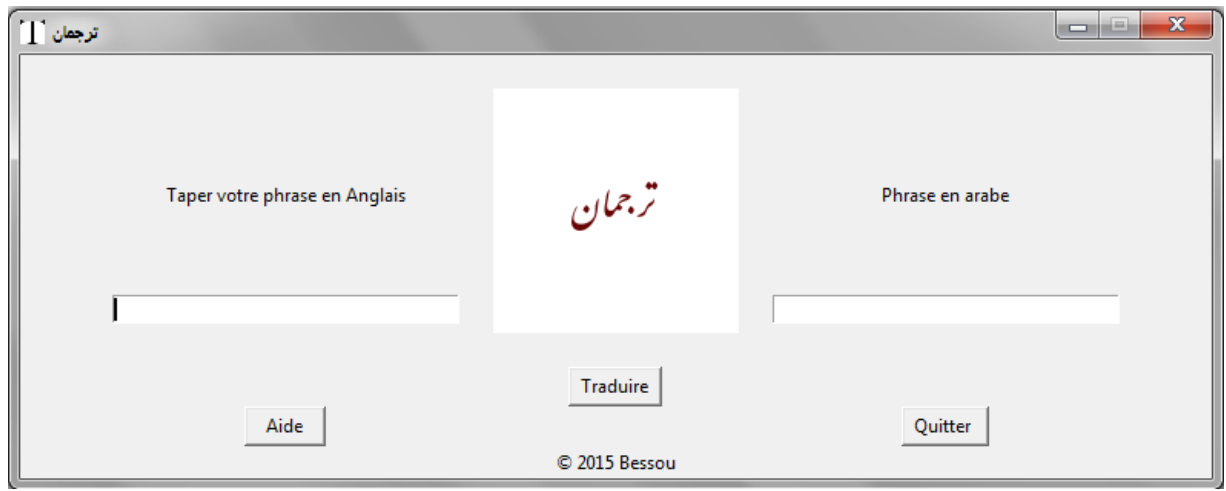

**Figure 7.1 Interface du système Tordjman**

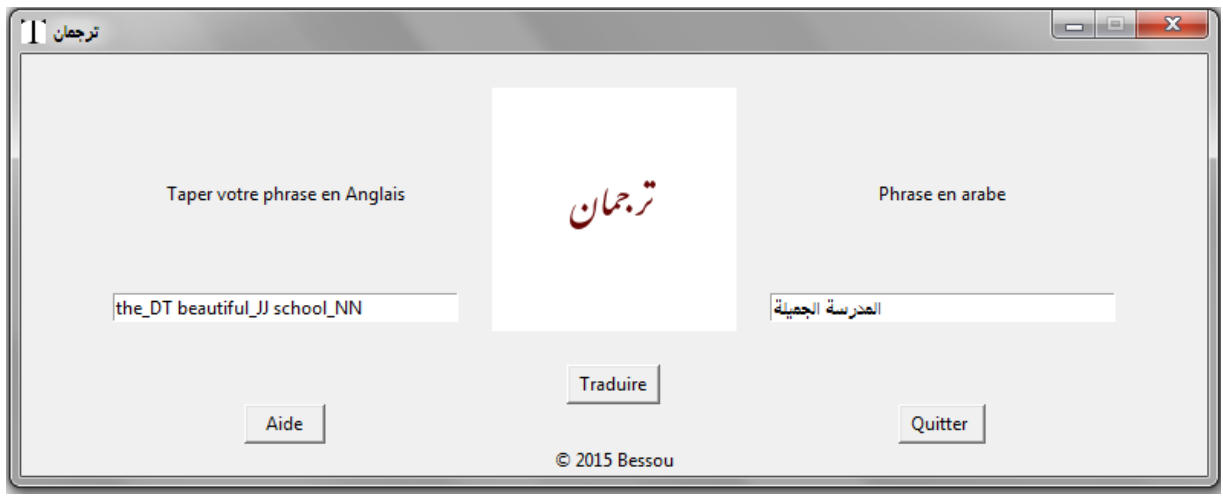

**Figure 7.2 Exemple de Traduction avec Tordjman**

La figure 7.3 montre un échantillon de code source de la fonction de traduction de la structure (DT JJ NN).

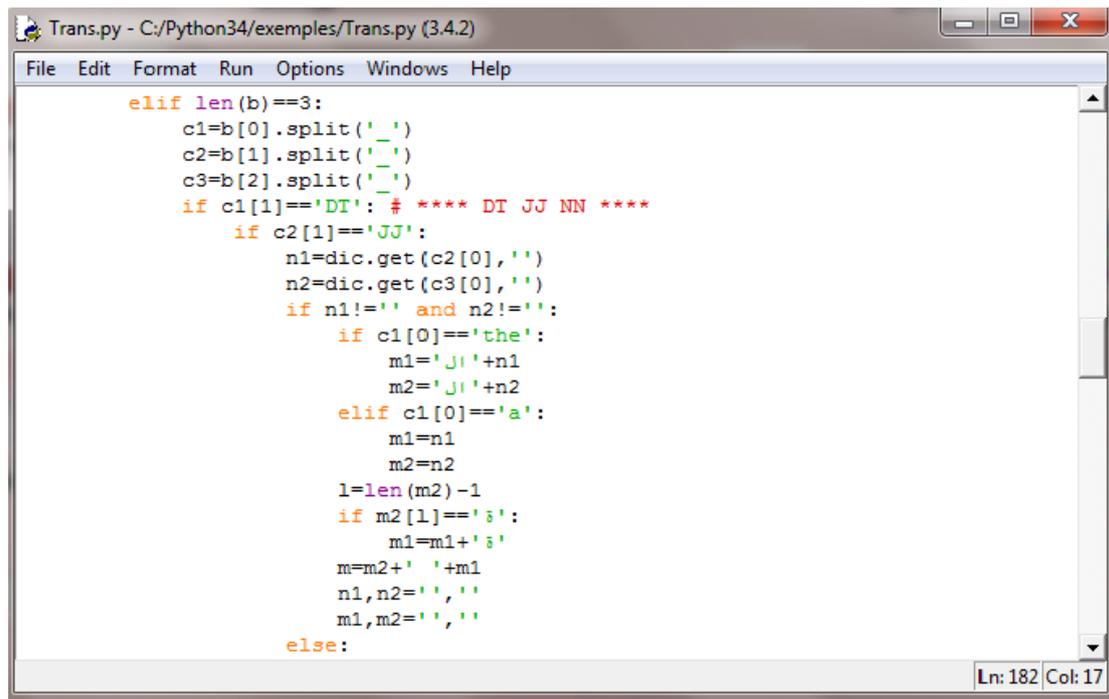

**Figure 7.3 Code source de la fonction de traduction de la structure DT JJ NN**

La figure 7.4 montre un échantillon de code source de la fonction de traduction de la structure (PRP VB DT NN).

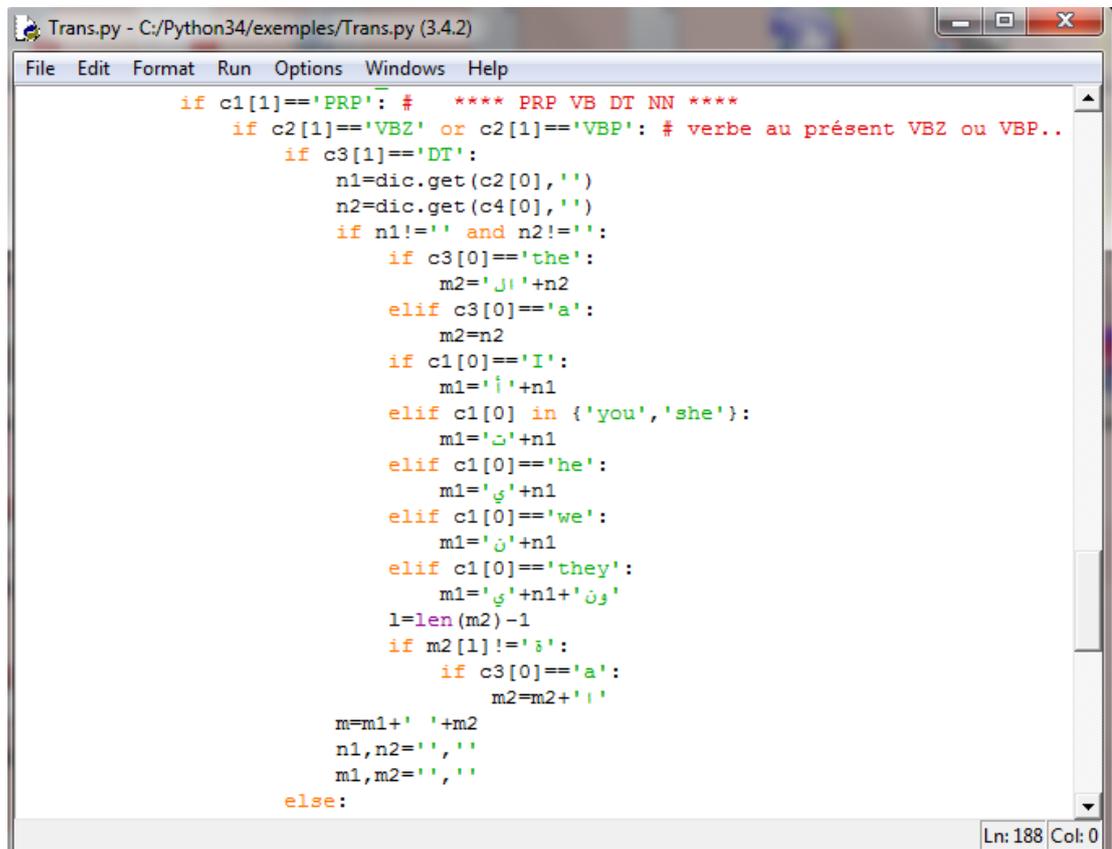

**Figure 7.4 Code source de la fonction de traduction de la structure PRP VB DT NN**

## 2.2. Test de quelques structures

La figure 7.5 montre l'interface de Stanford-Postagger pour avoir l'étiquetage des phrases en entrée.

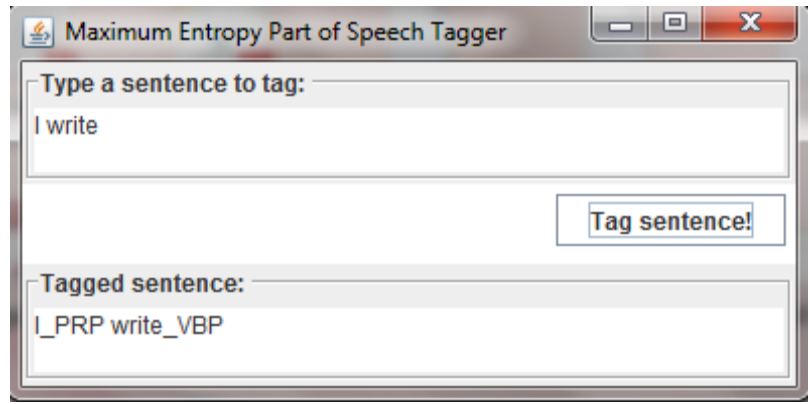

**Figure 7.5 Interface du PosTagger**

### 2.2.1. Les structures PRP VB

**Le présent**

La figure 7.6 montre le résultat de la traduction de la structure PRP VB au présent (VBP ou VBZ).

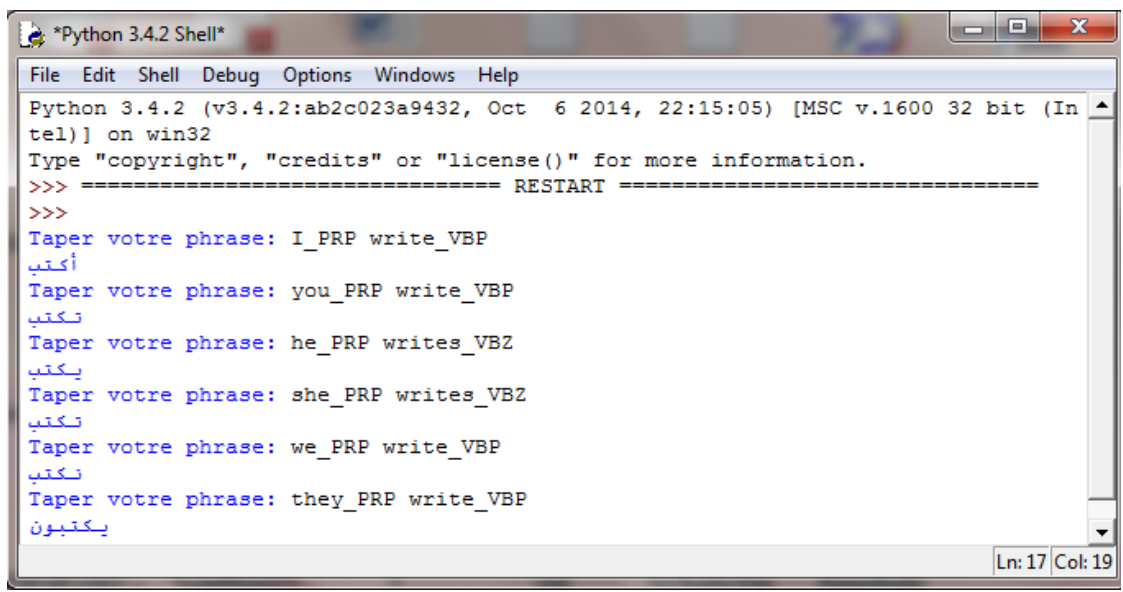

**Figure 7.6 Résultat de la traduction de la structure PRP VBP et PRP VBZ**

**Le passé**

La figure 7.7 montre le résultat de la traduction de la structure (PRP VBD).

**Figure 7.7 Résultat de la traduction de la structure PRP VBD**

### 2.2.2. La structure PRP MD VB

La figure 7.8 montre le résultat de la traduction de la structure (PRP MD VB).

**Figure 7.8 Résultat de la traduction de la structure PRP MD VB**

### 2.2.3. La structure PRP VB DT NN

**Le présent**

La figure 7.9 montre le résultat de la traduction de la structure (PRP VB DT NN) au présent (VBP ou VBZ), avec des noms masculins définis et indéfinis.

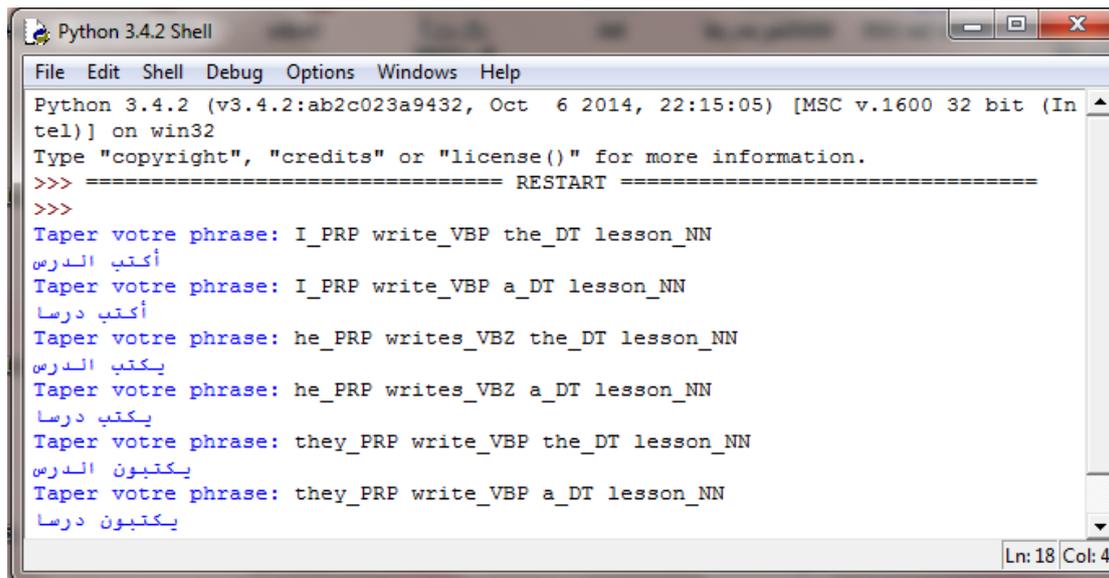

**Figure 7.9 Résultat de la traduction de la structure PRP VB DT NN au présent**

**Le passé**

La figure 7.10 montre le résultat de la traduction de la structure (PRP VBD DT NN) Avec des noms masculins, définis et indéfinis.

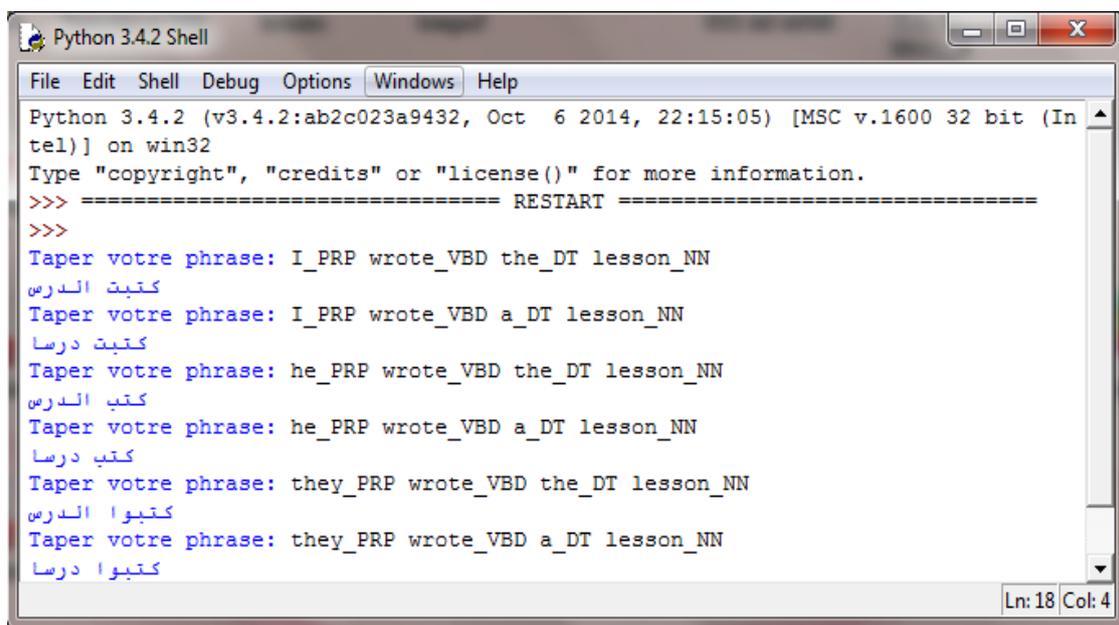

**Figure 7.10 Résultat de la traduction de la structure PRP VBD DT NN**

**Noms masculins et féminins**

La figure 7.11 montre le résultat de la traduction de la structure (PRP VB DT NN) avec des noms masculins et féminins, définis et indéfinis.

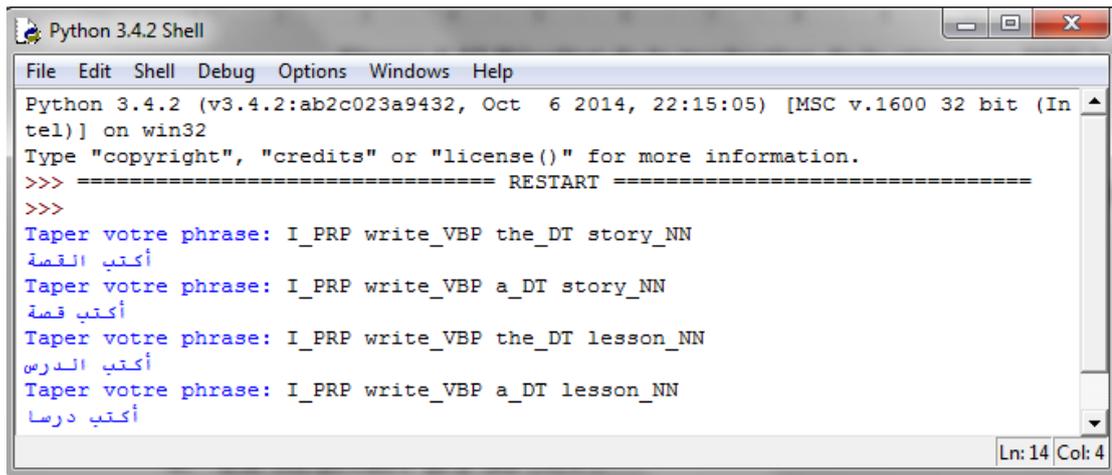

Figure 7.11 Résultat 2 de la traduction de la structure PRP VB DT NN : NN(F), NN(M)

### 2.2.4. La structure DT JJ NN

La figure 7.12 montre le résultat de la traduction de la structure (DT JJ NN) avec des noms masculins et féminins, définis et indéfinis.

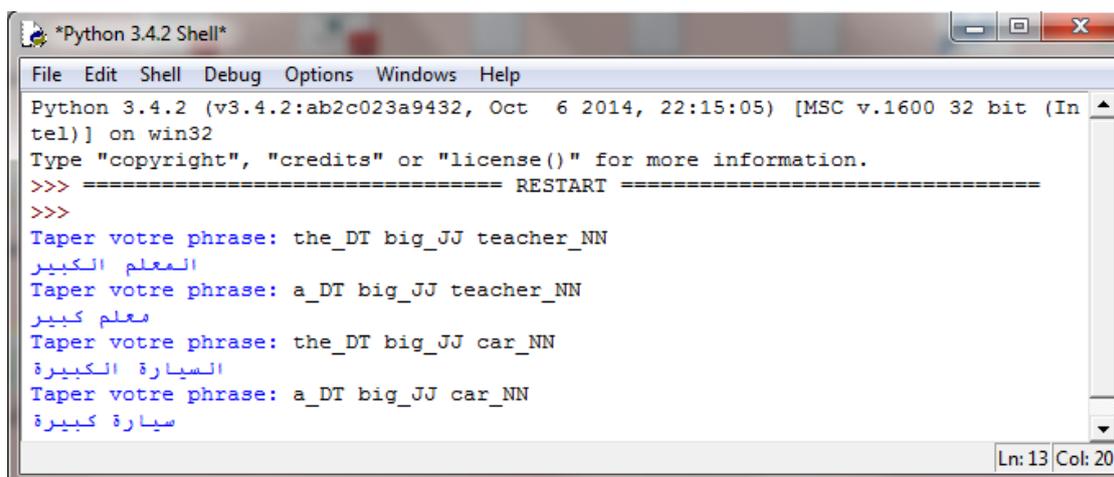

Figure 7.12 Résultat de la traduction de la structure DT JJ NN

## 3. Expérimentations et évaluation

Dans cette section, on présente les résultats obtenus lors de l'évaluation de l'approche sur des structures partielles de la langue. On doit mentionner ici que notre approche à base de règles est approuvable pour des domaines restreints avec des structures de phrases bien définis. Où elle peut donner de bons résultats par rapport à d'autres systèmes à base de méthodes statistiques. Or si on compare ces derniers comme système complet avec *Tordjman*, leurs résultats sont meilleurs.

Les expériences ont été réalisées on utilisant la métrique BLUE (voir chapitre 1, section 8.6) comme moyen d'évaluation automatique. On peut utiliser les codes fournis par Kenneth Heafield [117] ou Nitin Madnani [118]. La métrique BLUE exige un fichier test et un fichier référence (traduction humaine), pour cela on a utilisé un corpus parallèle ; les phrases sources en anglais peuvent être utilisé dans les échantillons de tests, les phrases cibles correspondantes en arabe représentent les phrases références pour la comparaison.

Parmi les corpus qui peuvent être utilisé comme traduction référence le corpus de Tatoeba tiré du projet OPUS (the open parallel corpus) de l'université d'Uppsala au suède [119]. OPUS propose une collection de corpus dédiée aux chercheurs et aux développeurs de traduction automatique. Pour la langue arabe, les corpus adéquats pour le test sont Tanzil et Tatoeba. Le premier corpus comprend des traductions de Coran, le deuxième comporte des phrases traduites entre différentes langues (plus de 132 langues). Le corpus le plus approprié à notre contribution est celui de Tatoeba. Ce corpus contient plus de 3469809 phrases, celui de la langue arabe plus de 15683 phrases.

Dans ces expériences, 76 phrases sont aléatoirement sélectionnées pour constituer le fichier de test de *Tordjman*. Le même fichier est utilisé pour comparer les résultats avec *Google translation* et *Systran*. La procédure d'évaluation est faite phrase par phrase dans le fichier de test. On calcule les scores BLUE (1-gram, 2-grams, 3-grams) pour toutes les phrases résultats. Puis on calcule la moyenne de chaque score n-gram. La figure 7.13 présente l'interface de calcul des scores pour *Tordjman* et *Google*.

**Figure 7.13 Calcul des scores pour Tordjman et Google**

La figure 7.14 présente l'interface de calcul des scores pour *Tordjman* et *Systran*

**Figure 7.14 Calcul des scores pour Tordjman et Systran**

La table 7.1 expose les scores BLUE des 27 segments traduits different de la traduction réference par *Google*.

**Table 7.1 Les scores BLUE du fichier test pour Google**

| Segment | 6    | 7    | 10   | 14   | 15   | 20   | 21   | 22   | 24   |
|---------|------|------|------|------|------|------|------|------|------|
| Score   | 0.50 | 0.71 | 0.38 | 0.84 | 0.84 | 0.50 | 0.71 | 0.16 | 0.23 |
| Segment | 29   | 32   | 36   | 37   | 41   | 42   | 44   | 47   | 48   |
| Score   | 0.64 | 0.32 | 0.64 | 0.71 | 0.64 | 0.38 | 0.32 | 0.71 | 0.64 |
| Segment | 49   | 53   | 56   | 57   | 61   | 63   | 65   | 66   | 68   |
| Score   | 0.71 | 0.64 | 0.32 | 0.71 | 0.71 | 0.71 | 0.64 | 0.64 | 0.32 |

La table 7.2 expose les scores BLUE des 69 segments traduits different de la traduction réference par *Systran*.

**Table 7.2 Les scores BLUE du fichier test pour Systran**

| Segment | 5    | 7    | 8    | 9    | 10   | 11   | 12   | 13   | 14   |
|---------|------|------|------|------|------|------|------|------|------|
| Score   | 0.71 | 0.71 | 0.71 | 0.71 | 0.71 | 0.71 | 0.71 | 0.71 | 0.71 |
| Segment | 15   | 16   | 17   | 18   | 19   | 20   | 21   | 22   | 23   |
| Score   | 0.71 | 0.50 | 0.71 | 0.71 | 0.71 | 0.71 | 0.71 | 0.71 | 0.71 |
| Segment | 24   | 25   | 26   | 27   | 29   | 30   | 31   | 32   | 33   |
| Score   | 0.71 | 0.71 | 0.71 | 0.71 | 0.64 | 0.64 | 0.64 | 0.64 | 0.64 |
| Segment | 34   | 35   | 36   | 37   | 38   | 39   | 40   | 41   | 42   |
| Score   | 0.64 | 0.64 | 0.64 | 0.64 | 0.38 | 0.64 | 0.64 | 0.38 | 0.38 |
| Segment | 43   | 44   | 45   | 46   | 47   | 48   | 49   | 50   | 51   |
| Score   | 0.38 | 0.38 | 0.38 | 0.38 | 0.38 | 0.38 | 0.38 | 0.23 | 0.38 |
| Segment | 52   | 53   | 54   | 55   | 56   | 57   | 58   | 59   | 60   |
| Score   | 0.38 | 0.64 | 0.64 | 0.64 | 0.64 | 0.64 | 0.64 | 0.64 | 0.64 |
| Segment | 61   | 62   | 63   | 64   | 65   | 66   | 67   | 68   | 69   |
| Score   | 0.64 | 0.38 | 0.64 | 0.64 | 0.64 | 0.64 | 0.64 | 0.64 | 0.64 |
| Segment | 70   | 71   | 72   | 73   | 74   | 75   | /    | /    | /    |
| Score   | 0.64 | 0.64 | 0.64 | 0.64 | 0.38 | 0.64 | /    | /    | /    |

Pour toutes les phrases du fichier test notre système a eu un score de 1.00. Selon la table 7.3, on peut conclure que les résultats de notre approche sont meilleurs que les résultats de *Google* et de *Systran*.

**Table 7.3 Les scores BLUE pour Tordjman, Google et Systran**

| Score *Tordjman* | Score *Google* | Score *Systran* |
|------------------|----------------|-----------------|
| 100.00           | 15.19          | 20.30           |

Alors notre système est capable de générer des traductions de phrases simples avec moins d'erreurs. Et ce grâce à l'aspect morphologique dans la génération de mots. *Google* et *Systran* optent pour une approche statistique qui considère le nombre d'apparition d'un mot dans le corpus d'apprentissage pour l'avantager à un autre mot semblable mais avec d'autres flexions ce qui rend le sens incohérent dans la phrase.

La figure 7.15 expose le résultat graphique de la métrique BLUE on comparant les traductions de *Tordjman* et celles de *Google*.

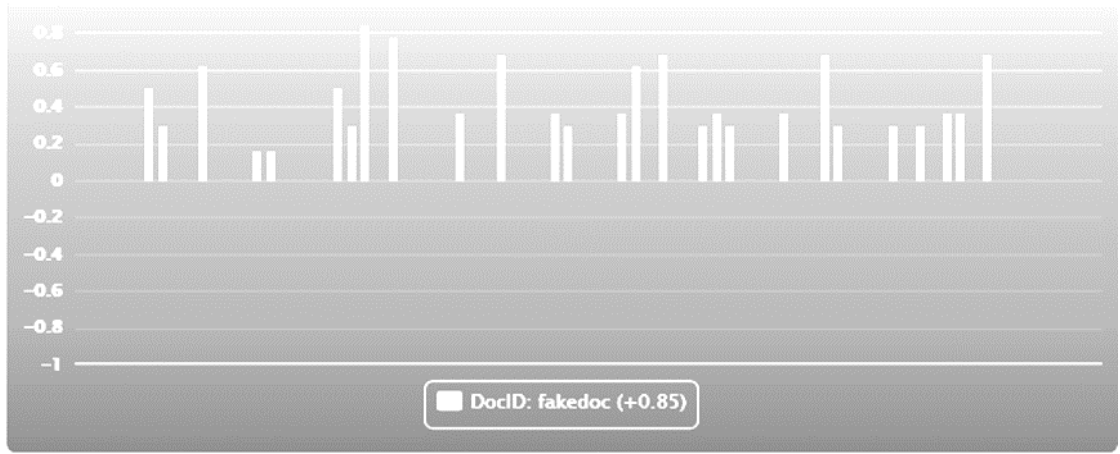

**Figure 7.15 Résultat visuel des scores BLUE du fichier test pour Google.**

Prenant par exemple le segment 6, la phrase source (big teacher) a eu un résultat similaire à la traduction référence, qui est (معلم كبير) alors que *Google* a donné (المعلم الكبير). La figure 7.16 présente les différences de N_gram par la métrique BLUE.

| ID | Segment 6, Document "fakedoc" [$\Delta_{BLEU}$=0.50] |
|---|---|
| Source | big teacher |
| Reference (fakeref) | معلم كبير |
| Hypothesis (*Tordjman*) | معلم كبير [1.00] |
| Hypothesis (*Google*) | المعلم الكبير [0.50] |

**Figure 7.16 Résultat BLUE du segment 6**

La figure 7.17 présente le résultat détaillé de la métrique BLUE du segment 15, la phrase source (he wrote) a eu par *Tordjman* (كتب) et par *Google* (وكتب).

La phrase Hypothesis (*Tordjman*) comporte 03 caractères en un seul mot avec un score de [1.00] alors que la phrase Hypothesis (*Google*) comporte 04 caractères en un seul mot avec un score de [0.84].

| ID | Segment 15, Document "fakedoc" [∆_BLEU=0.16] |
|---|---|
| Source | he wrote |
| Reference (fakeref) | كتب |
| Hypothesis (*Tordjman*) | كتب [1.00] |
| Hypothesis (*Google*) | وكتب [0.84] |

**Figure 7.17 Résultat BLUE du segment 15**

La figure 7.18 présente le résultat détaillé de la métrique BLUE du segment 22, la phrase source (she will write) a eu par *Tordjman* (ستكتب) et par *Google* (وقالت انها سوف أكتب).

La phrase Hypothesis (*Tordjman*) comporte 05 caractères en un seul mot avec un score de [1.00] alors que la phrase Hypothesis (*Google*) comporte 16 caractères en 4 mots avec un score de [0.16].

| ID | Segment 22, Document "fakedoc" [∆_BLEU=0.84] |
|---|---|
| Source | she will write |
| Reference (fakeref) | ستكتب |
| Hypothesis (*Tordjman*) | ستكتب [1.00] |
| Hypothesis (*Google*) | وقالت انها سوف أكتب [0.16] |

**Figure 7.18 Résultat BLUE du segment 22**

La figure 7.19 présente le résultat détaillé de la métrique BLUE du segment 57, la phrase source (we write the story) a eu par *Tordjman* (نكتب القصة) et par *Google* (نكتب قصة).

La phrase Hypothesis (*Tordjman*) comporte 09 caractères en 2 mots avec un score de [1.00] alors que la phrase Hypothesis (*Google*) comporte 07 caractères en 2 mots avec un score de [0.71].

| ID | Segment 57, Document "fakedoc" [∆_BLEU=0.29] |
|---|---|
| Source | we write the story |
| Reference (fakeref) | نكتب القصة |
| Hypothesis (*Tordjman*) | نكتب القصة [1.00] |
| Hypothesis (*Google*) | نكتب قصة [0.71] |

**Figure 7.19 Résultat BLUE du segment 57**

La figure 7.20 expose le résultat graphique de la métrique BLUE on comparant les traductions de *Tordjman* et celles de *Systran*.

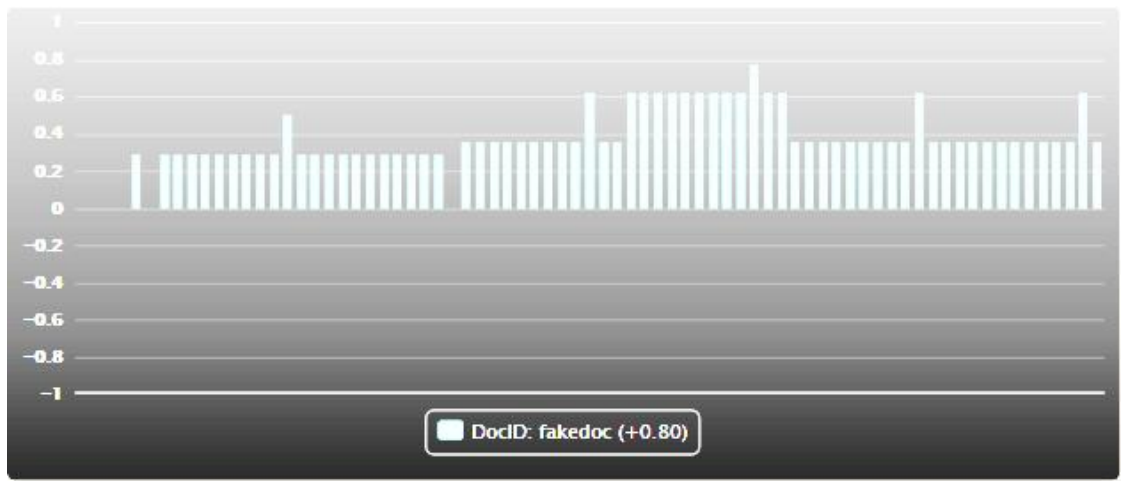

**Figure 7.20 Résultat visuel des scores BLUE du fichier test pour Systran.**

La figure 7.21 présente le résultat détaillé de la métrique BLUE du segment 5, la phrase source (big car) a eu par *Tordjman* (سيارة كبيرة) et par *Systran* (سيارة كبير).

La phrase Hypothesis (*Tordjman*) comporte 10 caractères en 2 mots avec un score de [1.00] alors que la phrase Hypothesis (*Systran*) comporte 09 caractères en 2 mots avec un score de [0.71].

| ID | Segment 5, Document "fakedoc" [$\Delta_{BLEU}$=0.29] |
|---|---|
| Source | big car |
| Reference (fakeref) | سيارة كبيرة |
| Hypothesis (*Tordjman*) | سيارة كبيرة [1.00] |
| Hypothesis (*Systran*) | سيارة كبير [0.71] |

**Figure 7.21 Résultat BLUE du segment 5**

La figure 7.22 présente le résultat détaillé de la métrique BLUE du segment 45, la phrase source (we write a lesson) a eu par *Tordjman* (نكتب درسا) et par *Systran* (نحن نكتب درس).

La phrase Hypothesis (*Tordjman*) comporte 8 caractères en 2 mots avec un score de [1.00] alors que la phrase Hypothesis (*Systran*) comporte 10 caractères en 03 mots avec un score de [0.38].

| ID | Segment 45, Document "fakedoc" [$\Delta_{BLEU}$=0.62] |
|---|---|
| Source | we write a lesson |
| Reference (fakeref) | نكتب درسا |
| Hypothesis (*Tordjman*) | **نكتب درسا** [1.00] |
| Hypothesis (*Systran*) | **نحن نكتب درس** [0.38] |

**Figure 7.22 Résultat BLUE du segment 45**

La figure 7.23 présente le résultat détaillé de la métrique BLUE du segment 50, la phrase source (she wrote a lesson) a eu par *Tordjman* (كتبت درسا) et par *Systran* (هو كتب درس).

La phrase Hypothesis (*Tordjman*) comporte 8 caractères en 2 mots avec un score de [1.00] alors que la phrase Hypothesis (*Systran*) comporte 8 caractères en 03 mots avec un score de [0.23].

| ID | Segment 50, Document "fakedoc" [$\Delta_{BLEU}$=0.77] |
|---|---|
| Source | she wrote a lesson |
| Reference (fakeref) | كتبت درسا |
| Hypothesis (*Tordjman*) | **كتبت درسا** [1.00] |
| Hypothesis (*Systran*) | **هو كتب درس** [0.23] |

**Figure 7.23 Résultat BLUE du segment 50**

La figure 7.24 présente le résultat détaillé de la métrique BLUE du segment 74, la phrase source (she wrote a story) a eu par *Tordjman* (كتبت قصة) et par *Systran* (هو كتب قصة).

La phrase Hypothesis (*Tordjman*) comporte 7 caractères en 2 mots avec un score de [1.00] alors que la phrase Hypothesis (*Systran*) comporte 08 caractères en 03 mots avec un score de [0.38].

| ID | Segment 74, Document "fakedoc" [$\Delta_{BLEU}$=0.62] |
|---|---|
| Source | she wrote a story |
| Reference (fakeref) | كتبت قصة |
| Hypothesis (*Tordjman*) | **كتبت قصة** [1.00] |
| Hypothesis (*Systran*) | **هو كتب قصة** [0.38] |

**Figure 7.24 Résultat BLUE du segment 74**

On peut expliquer ces résultats par deux raisons :

- Notre approche adopte une génération morphologique ce qui diminue le nombre de mots erronés alors augmente les scores BLUE.
- Nous avons testé des phrases simples, ce qui avantage notre approche par rapport à *Google*.

## *Conclusion*

A l'issue de cette expérience, on peut conclure que notre approche est convenable à des domaines restreints où les structures grammaticales sont limitées et invariables. Les résultats montrent une corrélation avec les jugements humains. L'incrémentation de nombre de règles pour plus de couverture de la langue augmente les possibilités d'interférence. A notre avis pour avoir un système robuste et général il faut utiliser ces règles de transfert comme étape de post traitement pour un système purement statistique ce qui élimine les failles des modèles statistiques et généralise les domaines de la langue pour les modèles à base de règles.

# Conclusion et perspectives

Ce mémoire a porté sur la traduction automatique et plus particulièrement sur le module de transfert entre l'anglais et l'arabe qui assure des transformations syntaxiques et morphologiques entre les deux langues.

Nous avons d'abord présenté les notions de traduction humaine et automatique, les techniques linguistiques de traduction automatique avec essentiellement l'approche à base de règles. Nous avons proposé une approche morphologique pour l'analyse et la génération des mots arabes, appliquée à la recherche d'information bilingue, et à la traduction de l'anglais vers l'arabe. Une modélisation à base de règles a été proposée par le biais de règles de transfert morphologiques et syntaxiques.

Pour la reconnaissance des formes fléchies, nous avons proposé une approche basée sur la notion de schème. Cette approche a abouti à des taux de reconnaissance des mots de 96%. Dans le domaine de la recherche d'information, les résultats de la lemmatisation ont permis une diminution du silence et ont abouti à une précision de 0,5732. Les résultats de la recherche multilingue étaient très satisfaisants car les sorties du niveau morphologique étaient suffisants pour traduire correctement les mots de la requête en anglais constituée de mots indépendants et non de phrases structurellement correctes.

En ce qui concerne le transfert morphologique, une liste quasiment exhaustive de règles a été proposée pour traiter tous les cas possibles. Nous avons donné beaucoup d'importance à ce transfert vue l'importance de la morphologie dans la langue arabe qui peut inclure des informations d'ordre syntaxique voire sémantique.

Pour le transfert syntaxique on s'est limité aux phrases simples nominales et verbales. On a proposé une liste de règles de transfert entre l'anglais et l'arabe, deux langues structurellement différentes. Une représentation visuelle est également procurée afin de faciliter l'appréhension de la modélisation de transfert syntaxique.

Les combinaisons entre phrases simples et les itérations dans les adjectifs et les annexions peuvent aboutir à des phrases complexes.

L'approche proposée a été concrétisé par le codage de certaines règles de transfert qui a donné des résultats satisfaisants qui étaient l'objet d'une comparaison avec des systèmes qui adoptent des approches statistiques.

Les travaux développés dans le cadre de cette thèse ont donc abouti à l'élaboration d'une *approche morphosyntaxique* pour assurer le transfert des structures syntaxiques de l'anglais vers l'arabe tout en insistant sur l'aspect morphologique très prépondérant dans le traitement automatique de la langue arabe.

**Perspectives**

A l'issue de cette thèse, nous envisageons :

- La finalisation des règles syntaxiques de transfert tout en étudiant toutes les structures possibles de la langue.
- L'adaptation du module de génération pour qui 'il puisse travailler avec des structures abstraites indépendantes de la langue source.

Des travaux sont en cours pour finaliser le module de transfert avec son générateur.

Nous projetons également d'aborder l'aspect multilingue dans la recherche d'information de plusieurs langues vers la langue arabe.

Une autre perspective intéressante serait de définir les règles de transfert syntaxique en retour de l'arabe vers l'anglais.

Néanmoins, il reste encore de nombreuses pistes à explorer dans ce domaine. On peut citer en particulier :

- Le traitement sémantique dans la traduction et dans la recherche d'information.
- Le transfert des textes arabes de/vers une langue pivot.
- L'intégration des méthodes statistiques pour remédier aux lacunes des approches linguistiques.

D'un point de vue plus pratique, il serait aussi intéressant d'implémenter les solutions que nous avons proposées dans d'autres applications de traitement automatique du langage naturel.

# Références

# Publications de l'auteur

**Revues Internationales**

S. Bessou, M. Touahria, *Morphological Analysis and Generation for Machine Translation From and to Arabic*, International journal of computer applications, ISSN: 0975-8887, volume 18(2), published by Foundation of Computer Science, New York, USA, 2011.

S. Bessou, *Machine translation*, Alarabia, ISSN: 3575-1112, volume 28, Algérie, 2012.

S. Bessou, M. Touahria, *An Accuracy-Enhanced Stemming Algorithm for Arabic Information Retrieval,* Neural Network world journal, ISSN: 1210-0552, volume 24(2), Czech Republic, 2014.

**Conférences internationales**

S. Bessou, M. Louail, A. Refoufi, Z. Kadem, M. Touahria, *Un système de lemmatisation pour les applications de TALN*, Colloque international sur le traitement automatique de la langue arabe CITALA, ISBN : 9954-412-12-3, Rabat, Maroc, 2007.

S. Bessou S, A. Saadi, M. Touahria, *Un système de lemmatisation des mots arabes*, HCLA, ISBN : 978-9961-795-53-8, Alger, Algérie, 2007.

S. Bessou, A. Saadi, M. Touahria, *Un système d'indexation et de recherche d'information en langue arabe*, HCLA, ISBN : 978-9961-795-53-8, Alger, Algérie, 2007.

S. Bessou, A. Saadi, M. Touahria, *Vers une recherche d'information plus intelligente application à la langue arabe*, SIIE : 1ère Conférence Internationale "Systèmes d'Information et Intelligence Economique", ISBN : 9978-9973868-19-0, Hammamet, Tunisie, 2008.

S. Bessou, S. Kendri, W. Zegadi, A. Saadi, M. Touahria, *Analyse linguistique statistique des textes en langue arabe pour la classification automatique par la méthode SVM*, IESC, Alep, Syrie, 2008.

S. Bessou, S. Kendri, W. Zegadi, A. Saadi, M. Touahria, *Représentation Textuelle pour la Classification par l'algorithme SVM, Application à la Langue Arabe*, OISLAM, Tlemcen, Algérie, 2008.

S. Bessou, A. Abaoui, A. Saadi, M. Touahria, *Méthode inspirée des arbres de décision pour la classification des textes en langue arabe*, OISLAM, Tlemcen, Algérie, 2008.

S. Bessou, *Structures de traits des mots arabes*, LexiPraxi09, Paris, France, 2009.

S. Bessou, M. Touahria, *Analyse morphosyntaxique pour la traduction automatique*, ICAI, ISBN : 978-9947-0-2763-9, BBA, Algérie, 2009.

S. Bessou, *Morphological analysis for machine translation from English to Arabic*, Séminaire international sur la traductologie et TAL, Oran, Algérie, 2010.

S. Bessou, *Analyse automatique de la structure morphologique*, $2^{\text{ème}}$ Séminaire international sur l'analyse automatique de la langue arabe, Université de Mascara, Algérie, 2014.

**Conférences nationales**

S. Bessou, A. Saadi, M. Touahria, *Un Système d'Indexation et de Recherche des Textes en Arabe (SIRTA)*, Séminaire national sur le langage naturel et l'intelligence artificielle, LANIA, Chlef, Algérie, 2007.

S. Bessou, A. Saadi, M. Touahria, *Un système d'extraction des racines des mots arabes pour l'indexation et la recherche d'information*, Séminaire national sur l'informatique et la langue arabe, Mostaganem, Algérie, 2008.

A. Saadi, S. Bessou, M. Touahria, *Extraction des informations à partir des données textuelles, séminaire national sur l'informatique et la langue arabe*, Mostaganem, Algérie, 2008.

S. Bessou, *L'importance de TALN dans l'administration électronique*, HCLA, ISBN : 978-9947-821-62-6, Alger, Algérie, 2011.

S. Bessou, *Les réseaux sociaux: l'aspect social et technique*, Séminaire sur la presse électronique et les technologies de la communication, Sétif, Algérie, 2011.

S. Bessou, *La publication électronique et la langue arabe*, 2$^{ème}$ conférence sur la publication électronique, HCLA, Alger, Algérie, 2012.

S. Bessou, *le langage SMS sur le web*, Université de Sétif-2-, Algérie, 2012.

S. Bessou, *L'analyse morphologique pour la traduction automatique de et vers la langue arabe*, Séminaire national sur le contenu numérique et les logiciels en langue arabe, Université de Tizi Ouzou, Algérie, 2013.

S. Bessou, M. Touahria, *Transfert Morphologique dans la Traduction automatique de l'anglais vers l'arabe*, 1$^{ère}$ journée d'informatique de l'université de Bordj Bou Arreridj (JIB-BA), Algérie, 2013.

S. Bessou, *La traduction automatique*, Journée d'étude sur la traduction et l'interdisciplinarité, Université de Sétif-2-, Algérie, 2014.

S. Bessou, *La terminologie informatique et le néologisme*, Séminaire national sur le Terme et la Terminologie, Université de Tizi Ouzou, Algérie, 2014.

**Écoles d'été**

*Les systèmes d'aides à la décision*, Oran, Algérie, 2007.
*La Bio-informatique Inspirée*, USTO, Oran, Algérie, 2008.

*Les systèmes d'aides à la décision*, Annaba, Algérie, 2009.

**Formations**

*Conception, développement et utilisation d'un cours en ligne*, Cellule de télé enseignement, université de Sétif, Agence Universitaire de la Francophonie, 2011.

*Prise en main de la plateforme d'apprentissage en ligne Moodle*, Cellule de télé enseignement, Université de Sétif, Agence Universitaire de la Francophonie, 2011.

*Workshop on research paper authorship and English technical writing*, Université de Sétif -1-, Algérie, 2012.

**Visites scientifiques**

Syrie : 2010 : Université d'Alep, 30 jours.

France : 2011 : INSA de Lyon, 30 jours.

Turquie : 2012 : LREC, 7 jours.

Canada : 2012 : Université de Québec à Montréal, 30 jours.

Suisse : 2013 : Université de Zürich, 30 jours.

Suède : 2014 : Université d'Uppsala, 30 jours.

**Projets de recherche**

Projet CNEPRU (2010)

Thème : *Systèmes de recherche d'information et de traduction automatique des langues naturelles : application à la langue arabe.*

Projet CNEPRU (2014)

Thème : *Système de traduction automatique de l'anglais vers la langue arabe par l'approche à base de transfert structurel.*